\documentclass[runningheads]{llncs}

\usepackage{eccv}

\usepackage{eccvabbrv}

\usepackage{graphicx}
\usepackage{booktabs}
\usepackage{multirow}
\usepackage{adjustbox}

\usepackage[accsupp]{axessibility}  %

\usepackage{hyperref}

\usepackage{orcidlink}

\usepackage{mwe} %

\begin{document}

\title{World Reconstruction From Inconsistent Views}

\author{Lukas H\"ollein \and Matthias Nie{\ss}ner}

\authorrunning{L.~H\"ollein, M. Nie{\ss}ner}

\institute{Technical University of Munich, Germany\\
\url{https://lukashoel.github.io/video_to_world}}

\maketitle

\vspace{-0.1in}
\begin{center}
\includegraphics[width=1\textwidth]{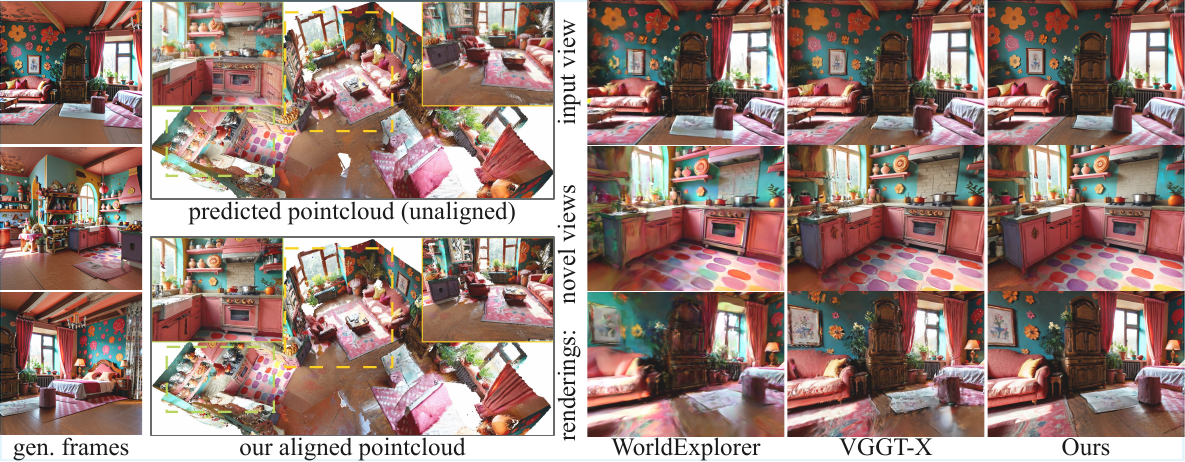}
\vspace{-0.3in}
\captionof{figure}{
\textbf{Our method reconstructs 3D worlds from video diffusion models.}
We propose a tailored non-rigid deformation of predicted pointcloud geometry (mid) that resolves the 3D inconsistencies inherent in generated video sequences.
Then, we utilize this improved alignment to optimize a Gaussian Splatting \cite{kerbl20233d} scene.
Our worlds can be explored from novel views at high visual fidelity (right).
}
\label{fig:teaser}
\end{center}

\begin{abstract}
Video diffusion models generate high-quality and diverse worlds; however, individual frames often lack 3D consistency across the output sequence, which makes the reconstruction of 3D worlds difficult.
To this end, we propose a new method that handles these inconsistencies by non-rigidly aligning the video frames into a globally-consistent coordinate frame that produces sharp and detailed pointcloud reconstructions.
First, a geometric foundation model lifts each frame into a pixel-wise 3D pointcloud, which contains unaligned surfaces due to these inconsistencies. %
We then propose a tailored non-rigid iterative frame-to-model ICP to obtain an initial alignment across all frames, followed by a global optimization that further sharpens the pointcloud. 
Finally, we leverage this pointcloud as initialization for 3D reconstruction and propose a novel inverse deformation rendering loss to create high quality and explorable 3D environments from inconsistent views. 
We demonstrate that our 3D scenes achieve higher quality than baselines, effectively turning video models into 3D-consistent world generators.
\end{abstract}

\section{Introduction}
\label{sec:intro}

World models build and simulate photorealistic environments.
The diversity and high fidelity of video diffusion models \cite{genie3, yang2024cogvideox, wan2025wanopenadvancedlargescale} has made the generation of 3D worlds a highly anticipated research goal.
Automating this creation process has many applications such as more diverse robotics training or faster ideation and production in movies, gaming, and VR.
Their vast prior knowledge makes video models a compelling foundation for world generation, where the goal is to lift these models into 3D generators.
A common approach employs a generation-reconstruction loop, that autoregressively generates multiple video sequences and reconstructs 3D scenes from these frames \cite{schneider_hoellein_2025_worldexplorer, chen2025flexworld, sun2024dimensionx}.

The core challenge in leveraging video diffusion models for world generation is their lack of 3D consistency.
While generated frame sequences look compelling, using them directly for 3D reconstruction leads to misaligned geometry and rendering artifacts (\Cref{fig:teaser}).
To this end, existing approaches add explicit camera trajectory control to the frame generation.
This is achieved by conditioning the network on extrinsic parameters \cite{gao2024cat3d, liang2024wonderland, zhou2025stable}, or by progressively building up a 3D cache from all previous frames and rendering it into novel perspectives \cite{yu2024viewcrafter, ren2025gen3c}.
More recent methods generate aligned RGB-D sequences \cite{huang2025voyager, bai2025geovideointroducinggeometricregularization} or finetune with consistency constraints \cite{kupyn2025epipolar, danier2025viewconsistentdiffusionrepresentations3dconsistent}.
Despite being pretrained and finetuned on huge amounts of data, the models still suffer from \textit{generative drift}: generated frames inconsistently warp object geometry to novel viewpoints and do not follow input cameras \cite{yu2024viewcrafter}.
We argue these generative models cannot be perfectly consistent and instead propose a lightweight alternative on the reconstruction side.

To this end, we propose a tailored non-rigid alignment of the scene geometry, that creates thin and sharp surfaces from an inconsistent initialization (\Cref{fig:teaser} mid).
We then leverage this geometry as initialization in a \textit{non-rigid aware} reconstruction, which yields high-quality, explorable 3D worlds  (\Cref{fig:teaser} right).
First, we utilize geometric foundation models \cite{lin2025depth} to obtain a dense pointcloud from all generated videos.
This reveals the inconsistencies of the input frames as multiple unaligned surfaces, which we now can correct.
Our tailored iterative frame-to-model ICP \cite{izadi2011kinectfusion, besl1992method} non-rigidly deforms these surfaces to align in a canonical space.
We additionally employ a sparse correspondence term to resolve larger misalignments, followed by a global optimization that further sharpens the pointcloud.
We use the aligned pointcloud as initialization for a 
\textit{non-rigid aware} Gaussian Splatting \cite{kerbl20233d, huang20242d} optimization.
Crucially, it factors out the misalignments by backward deforming the scene into the input frame spaces before rendering, which allows us to perform photometric optimization from an inconsistent image set.
Our resulting 3D worlds are consistent and can be rendered from novel perspectives at high visual fidelity.

\noindent To summarize, our contributions are:
\begin{itemize}
    \item We introduce a tailored non-rigid alignment of a scene pointcloud based on iterative frame-to-model ICP, that resolves the inconsistencies in the corresponding generated videos.
    \item We devise a \textit{non-rigid aware} 3D reconstruction that leverages this alignment to optimize consistent 3D worlds from inconsistent views.
    \item We demonstrate that our lightweight reconstruction method can be employed with many state-of-the-art video diffusion models, effectively turning them into 3D world generators.
\end{itemize}

\section{Related Work}
\label{sec:rw}

\subsection{Diffusion Models For Scene Generation}
Image and video diffusion models (VDM) generate diverse, high-quality 2D content \cite{SaharCSLWDGAMLSHFN2022, ramesh2022hierarchical, rombach2022highresolutionimagesynthesislatent, flux2023, yang2024cogvideox, wan2025wanopenadvancedlargescale}.
A highly anticipated goal is to leverage these models to generate 3D (e.g., NeRF \cite{mildenhall2021nerf}, 3DGS \cite{kerbl20233d}, or meshes \cite{siddiqui2024meshgpt, chen2024meshanything}).
Early works leverage score distillation to create assets \cite{DreamFusion, wang2023prolificdreamer, AssetGen} or scene chunks \cite{bahmani2025lyra, SceneTex, wu2024reconfusion}, which remains a costly per-scene optimization.
Recently, 2D models are finetuned with camera-control \cite{hollein2024viewdiff, gao2024cat3d, zhou2025stable, Tang2023mvdiffusion, liu2023syncdreamer, yu2024viewcrafter, ren2025gen3c} to jointly generate multi-view images and then reconstruct them in 3D \cite{szymanowicz2025bolt3d, hollein2023text2room, schneider_hoellein_2025_worldexplorer, chen2025flexworld, liang2024wonderland, schult2023controlroom3droomgenerationusing}.
However, the generated frames remain geometrically inconsistent \cite{yu2024viewcrafter}, which leads to floating artifacts when rendering.
To this end, recent methods remove such areas, generate more images, or fix artifacts at render-time \cite{wei2025gsfix3d, fischer2025flowrflowingsparsedense, delutio2026artifixerenhancingextending3d, warburg2023nerfbustersremovingghostlyartifacts, wu2025difix3dimproving3dreconstructions}.
They primarily focus on real-world scenarios with reliable (but sparse) camera observations, whereas we focus on the generative setting, improving from dense (but inconsistent) generated views.
Recent approaches finetune models on consistency objectives \cite{li2025flashworld, kupyn2025epipolar, bai2025geovideointroducinggeometricregularization, danier2025viewconsistentdiffusionrepresentations3dconsistent}, which is a costly per-model training.
In contrast, we propose a lightweight per-scene reconstruction that can turn any VDM into a 3D-consistent world generator.
Instead of explicit 3D, recent works explore minute-long rollout of consistent videos \cite{genie3, hyworld2025, song2025history}, which needs vast compute at inference and can hit context limits.
Our approach reconstructs consistent 3D worlds with high quality that can be rendered in real-time.

\subsection{Geometric Foundation Models}
Geometric foundation models (GFMs) predict 3D geometry from multiple images as input \cite{wang2024dust3r, leroy2024groundingimagematching3d, liu2025vggt, keetha2025mapanything, lin2025depth}, enabling online reconstruction \cite{smart2024splatt3r, maggio2025vggt, antsfeld2026smust3rslidingmultiview3d}.
When combined with rigid alignment akin to bundle adjustment, recent methods obtain high-quality 3D scenes from uncalibrated images \cite{liu2025vggt, huang20253rgsbestpracticeoptimizing, duisterhof2024mast3rsfmfullyintegratedsolutionunconstrained}.
However, since GFMs are trained on real-world or synthetic images, they cannot correct for the generative drift inherent in VDMs.
Instead, inconsistent images lead to non-overlapping surfaces, that reveal the geometric inaccuracy of generations.
We exploit this insight to correct for these artifacts in a 3D alignment stage.

\subsection{Pointcloud Alignment}
The alignment of 3D geometry has been extensively studied.
Foundational methods like ICP \cite{besl1992method, rusinkiewicz2001efficient, segal2009generalized} or volumetric fusion \cite{curless1996volumetric} enable large-scale integration of depth measurements \cite{chen1992object, choi2015robust, izadi2011kinectfusion, niessner2013real, zhou2016fast}.
They were further extended to reconstruct textures \cite{park2017colored, huang20173dlite, zhou2014color} and dynamic reconstruction by optimizing for non-rigid deformations \cite{newcombe2015dynamicfusion, innmann2016volumedeform, bozic2020neural, yunus2024recent, zollhofer2014real}.
Inspired by these approaches we propose a tailored pointcloud alignment, that is suitable for inconsistent video generations.
Specifically, we devise a non-rigid alignment energy akin to these approaches and utilize the reconstruction along with the optimized deformations in a \textit{non-rigid aware} Gaussian Splatting \cite{kerbl20233d} optimization to produce consistent 3D worlds.

\section{Method}
\label{sec:method}

\begin{figure*}[t]
    \centering
    \includegraphics[width=1\textwidth]{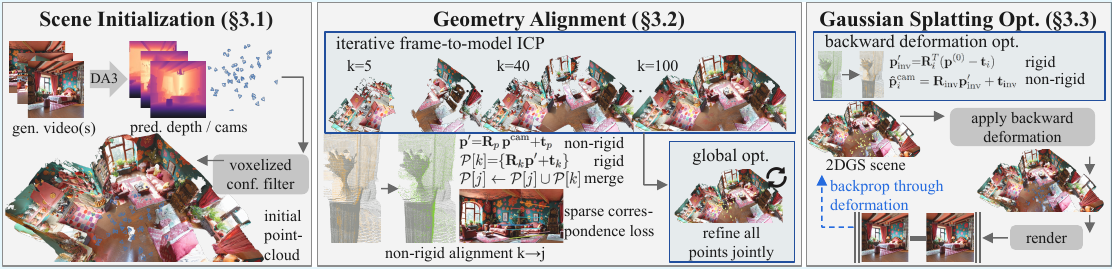}
    \caption{
        \textbf{Method overview.}
        We propose a three stage method that reconstructs a 2DGS \cite{huang20242d} scene from generated videos.
        First, we estimate multi-view depth and cameras with a geometric foundation model \cite{lin2025depth}.
        The resulting dense scene initialization is unaligned (multiple non-overlapping surfaces) due to the inconsistent input frames.
        We propose a tailored non-rigid geometry alignment that leverages iterative frame-to-model ICP \cite{izadi2011kinectfusion, besl1992method} and sparse correspondences, followed by global optimization, to create thin surfaces with detailed textures.
        Then, we leverage the alignment in a novel \textit{non-rigid aware} 2DGS \cite{huang20242d} optimization to obtain high-quality, consistent 3D worlds.
    }
    \label{fig:method}
\end{figure*}

Our method creates a 3D scene reconstruction from generated video sequences (see \Cref{fig:method}).
First, we initialize the scene by estimating dense per-pixel depth and poses and unproject them into a pointcloud (\Cref{subsec:init}).
Since the input frames are not 3D consistent, this geometry is unaligned and distorted.
To this end, we improve the geometry in our non-rigid alignment stage (\Cref{subsec:align}).
The output geometry accurately depicts unified surfaces and textures.
Finally, we use this refined pointcloud as initialization for our \textit{non-rigid aware} Gaussian Splatting optimization (\Cref{subsec:3dgs-baking}).
We obtain consistent and detailed 3D worlds that can be explored in real-time from arbitrary positions.

\subsection{Scene Initialization}
\label{subsec:init}

We leverage video diffusion models (VDMs) \cite{wan2025wanopenadvancedlargescale, yang2024cogvideox, blattmann2023stable} to generate input frames that depict static scenes from various perspectives.
VDMs model the conditional probability distribution $p_{\theta}(\mathbf{V} | \mathbf{c})$, where $\mathbf{V} {=} \{ \mathbf{I}_i \}_{i=0}^N$ is a continuous sequence of individual images $\mathbf{I} {\in} \mathbb{R}^{3 \times H \times W}$.
The condition $\textbf{c}$ can be a text prompt and/or previous frames $\mathbf{V}_\text{src}$.
This enables the generation of large-scale worlds via autoregressive roll-out of multiple sequences $\{ \mathbf{V} \}_{i=0}^N$ \cite{schneider_hoellein_2025_worldexplorer, genie3, hyworld2025}.
VDMs can be conditioned on camera trajectories $\{ \mathbf{R}_i {\in} \mathbb{R}^{3 \times 3}, \mathbf{t}_i {\in} \mathbb{R}^{3}, \mathbf{K}_i {\in} \mathbb{R}^{3 \times 3} \}_{i=1}^N$ to allow fine-grained per-frame spatial control.
This is typically achieved through dense Plücker embeddings \cite{zhou2025stable, liang2024wonderland} or 3D cache reprojection \cite{ren2025gen3c, yu2024viewcrafter}.

Despite explicitly modeling multi-view control, the frames suffer from \textit{generative drift}, making them inconsistent in 3D.
The generated sequences are thus hard to reconstruct, with floating artifacts or distortions when rendering (\Cref{fig:teaser} right).
We observe that geometric foundation models (GFMs) \cite{wang2024dust3r, liu2025vggt, keetha2025mapanything, lin2025depth} reveal these shortcomings and exploit this in a non-rigid alignment to obtain sharp and consistent surfaces (\Cref{fig:teaser} mid).
Concretely, GFMs predict dense per-pixel depth $\mathbf{d} {\in} \mathbb{R}^{H \times W}$ and confidence $\mathbf{c} {\in} \mathbb{R}^{H \times W}$ as well as per-frame pinhole cameras $(\mathbf{R}, \mathbf{t}, \mathbf{K})$: 
\begin{equation}
\{ \mathbf{d}_i, \mathbf{c}_i, \mathbf{R}_i, \mathbf{t}_i, \mathbf{K}_i \}_{i=0}^N = \mathcal{F}_{\text{GFM}}(\mathbf{V}),
\end{equation}
where $\mathcal{F}_{\text{GFM}}$ denotes the pre-trained metric DepthAnything-3 \cite{lin2025depth} model. 
In the case of camera-controlled video diffusion models, we condition the GFM on the trajectory to obtain $\{ \mathbf{d}_i, \mathbf{c}_i \}_{i=0}^N = \mathcal{F}_{\text{GFM}}(\mathbf{V}, \mathbf{R}_i, \mathbf{t}, \mathbf{K})$.

We utilize this information to construct a scene initialization as colored pointcloud via unprojection: 
\begin{equation}
\mathbf{p}_i^\text{cam}(u,v) {=} \mathbf{K}_i^{-1} [u, v, 1]^T \cdot \mathbf{d}_i(u, v),
\mathcal{\bar{P}} = \{ \mathbf{R}_i \mathbf{p}_i^\text{cam} + \mathbf{t}_i \}_{i=0}^N,
\mathcal{C} {=} \{ \mathbf{I}_i(u, v) \}_{i=0}^N,
\end{equation}
where $(u,v)$ are pixel coordinates.
We devise a filtering scheme to retain reliable points per scene chunk.
This ensures a dense initialization that does not require hallucination, but only alignment, to create detailed surfaces of entire scenes.
Each 3D point $\mathbf{p}_k {\in} \mathcal{\bar{P}}$ is assigned to a voxel $\mathbf{v}_k {=} \lfloor \mathbf{p}_k \,/\, s_{\text{vox}} \rfloor$ and then we compute the per-voxel confidence threshold: $\tau_{\text{loc}}[v] {=} \mathrm{perc}\bigl(\{\mathbf{c}_k {:} \mathbf{v}_k {=} v\},\; \theta_{\text{loc}}\bigr)$, where $\mathbf{c}_k$ is the point confidence.
We additionally compute the per-voxel occupancy threshold as $\tau_{\text{cnt}} {=} \mathrm{perc}(\{\mathrm{count}[v]\},\; \theta_{\text{cnt}})$ and use the two thresholds to obtain a filtered pointcloud: $\mathcal{P} {=} \{ \mathbf{p}_k {:} \mathbf{c}_k  \ge \tau_{\text{loc}}[\mathbf{v}_k]\bigr) {\;\wedge\;} \mathrm{count}[\mathbf{v}_k] \ge \tau_{\text{cnt}} \}$.

The resulting scene initialization shows misaligned points, i.e., corresponding surfaces across frames to not perfectly overlap (\Cref{fig:teaser} mid-top).
Next, we correct these artifacts in a 3D geometric alignment stage.

\subsection{Geometry Alignment}
\label{subsec:align}

We propose a two-stage non-rigid alignment of the pointcloud $\mathcal{P}$.
First, a tailored non-rigid iterative frame-to-model ICP \cite{besl1992method, amberg2007optimal, izadi2011kinectfusion} obtains an initial alignment across all frames.
Second, a global optimization further sharpens this alignment.

\vspace{-5mm}
\subsubsection{Non-rigid iterative frame-to-model ICP}
\label{subsubsec:non-rigid-icp}

The goal of this stage is to iteratively align two point subsets $\mathcal{P}[j]$, $\mathcal{P}[k]$ where $j,k$ denote image indices.
In other words, the subsets refer to the confidence-filtered per-frame point predictions of the GFM.
We optimize the camera extrinsics $(\mathbf{R}_k, \mathbf{t}_k)$ and the weights $\theta_k$ of a deformation neural field, implemented as a hashgrid MLP \cite{M_ller_2022}.
Given a 3D position $\mathbf{p} {\in} \mathcal{P}[k]$, the network predicts exponential coordinates $\xi {=} \mathcal{F}_\theta(\mathbf{p}) {\in} \mathbb{R}^6$, which are interpreted as the coordinate representation of a twist $\boldsymbol{\xi} {\in} \mathfrak{se}(3)$.
The corresponding per-point rigid transformation $(\mathbf{R}_p, \mathbf{t}_p)$ is obtained via the exponential map $T(\mathbf{p}) {=} \exp(\boldsymbol{\xi}^\wedge) {\in} \mathrm{SE}(3)$.
Then, we non-rigidly transform $\mathcal{P}[k]$ as
\begin{equation}
\label{eq:fwd-def}
    \mathbf{p}' {=} \mathbf{R}_p \, \mathbf{p}^{\text{cam}} + \mathbf{t}_p, \quad
    \mathcal{\bar{P}}[k] = \{ \mathbf{R}_k \mathbf{p}' + \mathbf{t}_k | \mathbf{p} {\in} \mathcal{P}[k]\}.
\end{equation}
This aligns the per-frame pointcloud with $\mathcal{P}[j]$ by first transforming each point independently (i.e., non-rigidly) in the $k$-th camera space, followed by a rigid transformation from $k$ to $j$.
We apply this principle on all frames $\{ \mathbf{I}_i \}_{i=1}^N$ by fixing $j {=} 0$ and iterating on $k {\in} \{1,...,N\}$.
Each iteration optimizes $\mathbf{R}_k, \mathbf{t}_k, \theta_k$ with a point-to-plane dense data term:
\begin{equation}
    \mathcal{L}_{\text{data}} = \frac{1}{\sum_{\mathbf{p} \in \mathcal{P}[k]} w(\mathbf{p})} \sum_{\mathbf{p} \in \mathcal{P}[k]} w(\mathbf{p}) \bigl((\mathbf{p}' - \mathbf{q}_{\mathrm{nn}(j)}) \cdot \mathbf{n}_{\mathrm{nn}(j)}\bigr)^2,
\end{equation}
where $\mathbf{q}_{\mathrm{nn}(j)}$ is the nearest neighbor of $\mathbf{p}'$ in $\mathcal{P}[j]$, $\mathbf{n}_{\mathrm{nn}(j)}$ its normal vector and we classify inliers via 
\begin{equation}
w(\mathbf{p}) =
\begin{cases}
1, & \text{if } \|\mathbf{p}' - \mathbf{q}_{\mathrm{nn}(i)}\|^2 < d_{\max}^2, \\
0, & \text{otherwise.}
\end{cases}
\end{equation}
We also adopt the color ICP term $\mathcal{L}_{\text{color}}$ of Park \etal \cite{park2017colored}, which aligns points by minimizing differences in local tangent-plane color gradients.
Additionally, we obtain 2D correspondence pairs via image matching of $\mathbf{I}_k$ and a subset of previous frames $\{\mathbf{I}_i\}_{i<k}$.
We utilize the RoMa matcher \cite{edstedt2024roma, edstedt2025roma} to obtain pairs $(s_i, t_i)$ and a certainty weight $w_i$, where $s_i$ is the pixel-coordinate in a source image and $t_i$ in $\mathbf{I}_k$.
Then, we compute the weighted sparse correspondence loss 
\begin{equation}
\label{eq:corr-loss}
    \mathcal{L}_{\text{corr}} = \frac{\sum_i w_i \|\mathcal{P}_j[\mathbf{s}_i] - \mathcal{P}_k[\mathbf{t}_i] \|^2}{\sum_i w_i}.
\end{equation}
In practice we found $\mathcal{L}_{\text{corr}}$ is highly beneficial, as it allows to align points that exceed the nearest neighbor distance $d_{\max}$.
We further regularize the non-rigid deformations to be small and locally similar, akin to an ARAP constraint \cite{arap}:
\begin{equation}
    \mathcal{L}_{\text{tv}} = \frac{1}{|\mathcal{P}[k]|} \sum_{p \in {\mathcal{P}[k]}} \| \mathcal{F}_{\theta_k}(\mathbf{p}) - \mathcal{F}_{\theta_k}(\mathbf{p \pm s_\text{vox}}) \|^2,
\end{equation}
where we compare the twist predictions of each point with its 6 axis-aligned neighbors at distance $s_\text{vox}$.
The final objective then becomes $E(\mathbf{R}_k, \mathbf{t}_k, \theta_k) {=} \mathcal{L}_{\text{data}} + \lambda_{\text{color}}\,\mathcal{L}_{\text{color}} + \lambda_{\text{corr}}\,\mathcal{L}_{\text{corr}} + \lambda_{\text{tv}}\,\mathcal{L}_{\text{tv}}$.
We optimize this energy with ADAM \cite{kingma2014adam} in a coarse-to-fine fashion with progressively smaller $d_{\max}, s_\text{vox}$, where we predict non-rigid deformations with $\mathcal{F}_\theta$ only in the finest scale.

After each iteration we merge the aligned points into the current model as $\mathcal{P}[j] \leftarrow \mathcal{P}[j] \cup \mathcal{\bar{P}}[k]$ using an adaptive outlier removal scheme.
It discards points, whose ICP losses are larger than the median absolute deviation (MAD) across all previous frames.
We compute the $k$-th thresholds $g^d_k {=} \mathrm{perc}\bigl( \mathcal{L}_\text{data}, \theta_d \bigr)$ and $g^c_k {=} \mathrm{perc}\bigl( \mathcal{L}_\text{color}, \theta_c \bigr)$ and accumulate them in $\tau_{d} = \mathrm{median}(\{g^d\}_{i<k}) + \sigma_{\text{d}} \cdot 1.4826 \cdot \mathrm{MAD}(\{g^d\}_{i<k})$ and $\tau_{c}$, respectively.
A point is then only merged if both its data and color ICP terms are smaller than $\tau_{d}$ and $\tau_{c}$, respectively.

\vspace{-5mm}
\subsubsection{Global non-rigid optimization}
\label{subsubsec:nrba}

After the iterative ICP stage, the pointcloud $\mathcal{P}[0]$ now represents an aligned 3D scene, where different per-frame points make up a unified surface.
We further improve this alignment by \textit{jointly} optimizing all cameras and deformation networks.
We find that this refinement leads to thinner surfaces and sharpens the overall texture quality of the pointcloud (e.g., compare the fence in \Cref{fig:qual_pcl} bottom).
Concretely, the frame-to-model optimization of the previous stage can drift over time towards aligned but thicker surfaces, since only the $k$-th frame is optimized at a time.

To this end, we optimize the global energy 
\begin{equation}
\label{eq:global-opt}
E(\{\mathbf{R}_i, \mathbf{t}_i, \theta_i\}_{i=1}^N) {=} \mathcal{L}_{\text{data}}^{\text{global}} + \lambda_{\text{color}} \mathcal{L}_{\text{color}}^{\text{global}} + \lambda_{\text{anchor}} \mathcal{L}_{\text{anchor}},
\end{equation}
that modifies the correspondence search in the ICP losses.
For each point $\mathbf{p} {\in} \mathcal{P}[0]$ we find its $k {=} 5$ nearest neighbors in \textit{other} frames and use them to calculate the respective objectives.
As gradients now flow to both $\mathbf{p}'$ and $\mathbf{q}_{\mathrm{nn}(j)}$, this drives the non-rigid deformations towards a thinner surface across all frames.

Additionally, we regularize the updates to the cameras and deformation networks to be as small as possible, which avoids degenerate solutions:
\begin{equation}
    \mathcal{L}_{\text{anchor}} = \frac{1}{N}\sum_{i=1}^{N}\biggl[\frac{1}{M}\sum_{k=1}^{M}\|\mathcal{F}_{\theta_i}(\mathbf{a}_{k,i}) - \xi^{(0)}_{k,i}\|^2 + \|\xi_{g,i} - \xi^{(0)}_{g,i}\|^2\biggr],
\end{equation}
where we subsample points from each frame $\{\mathbf{a}_{k,i}\}_{k=1}^M {\subset} \mathcal{P}[i]$ and encourage similarity of their twists to the state before global optimization $\xi^{(0)}_{k,i}$.
We similarly regularize the cameras by comparing their twist parameterizations $\xi_{g,i}$ and $\xi^{(0)}_{g,i}$.

\subsection{Non-rigid Gaussian Splatting Optimization}
\label{subsec:3dgs-baking}

After the geometry alignment we have obtained $\mathcal{P}[0]$, which represents an aligned 3D scene, as well as the non-rigid \textit{forward} deformations $\{\mathbf{R}_i, \mathbf{t}_i, \mathcal{F}_{\theta_i}\}_{i=1}^N$ from each frame to the first frame.
This is similar to a 4D scene reconstruction that is obtained in the dynamic reconstruction literature \cite{yunus2024recent, park2021nerfiesdeformableneuralradiance, newcombe2015dynamicfusion, wu20244d}.
In contrast, our goal is not to render a dynamic scene from a time-dependent state, but to unify the $N$ inconsistent geometry predictions into a single canonical state.
Thus, we propose a novel \textit{non-rigid aware} Gaussian Splatting \cite{kerbl20233d} optimization.
The output 3D reconstruction then renders photorealistic novel views of consistent 3D worlds, while being trained from the inconsistent image set $\{\mathbf{I}_i\}_{i=0}^N$.

\vspace{-5mm}
\subsubsection{Backward deformation optimization}
First, we obtain the inverse of the per-frame deformations $\{\mathbf{R}_i, \mathbf{t}_i, \mathcal{F}_{\theta_i}\}_{i=1}^N$, which transforms $\mathcal{P}[0]$ into all other frames.
Obtaining $\{\mathbf{R}_i^{-1}, \mathbf{t}_i^{-1}\}_{i=1}^N$ is straightforward and we optimize a single deformation neural field $\mathcal{F}^{-1}_{\theta_\text{inv}}$, that predicts the inverse exponential coordinates $\xi^\text{inv} {=} \mathcal{F}^{-1}_{\theta_\text{inv}}(\mathbf{p}, i)$ for any point $\mathbf{p} {\in} \mathcal{P}[0]$ to any view $i$.
We similarly implement $\mathcal{F}^{-1}_{\theta_\text{inv}}$ as a hashgrid MLP \cite{M_ller_2022}, but additionally condition the network on a learnable view embedding.
We generate training pairs $\{ \mathbf{p}^\text{cam}_i, \mathbf{p}^{(0)} \}_{i=1}^M$ by sampling a subset of points from all per-frame predictions in their respective camera space and transforming them to $\mathcal{P}[0]$ via \Cref{eq:fwd-def}.
We apply the \textit{backward} deformation on these pairs to obtain the predictions $\mathbf{\hat{p}}^\text{cam}_i$:
\begin{equation}
\label{eq:bkwd-def}
    \mathbf{p}'_\text{inv} {=} \mathbf{R}_i^T ( \mathbf{p}^{(0)} - \mathbf{t}_i ), \quad
    \mathbf{\hat{p}}^\text{cam}_i = \mathbf{R}_\text{inv} \mathbf{p}'_\text{inv} + \mathbf{t}_\text{inv},
\end{equation}
where $(\mathbf{R}_\text{inv}, \mathbf{t}_\text{inv})$ is the corresponding per-point inverse deformation that we obtain from $\xi^\text{inv}$.
Then, we define the total objective as 
\begin{equation}
    \mathcal{L}_{\text{inverse}} = \frac{1}{M}\sum_{i=1}^{M} \|\hat{\mathbf{p}}^{\text{cam}}_i - \mathbf{p}^{\text{cam}}_i\|^2, \quad
    E(\theta_\text{inv}) = \mathcal{L}_{\text{inverse}} + \lambda_\text{tv} \mathcal{L}_\text{tv}.
\end{equation}

\vspace{-8mm}
\subsubsection{3D Scene Consolidation}
We leverage $\mathcal{F}^{-1}_{\theta_\text{inv}}$ to train a Gaussian Splatting \cite{kerbl20233d} scene representation from the inconsistent image set $\{\mathbf{I}_i\}_{i=0}^N$.
Concretely, we convert $(\mathcal{P}[0], \mathcal{C})$ into the attributes of 2DGS \cite{huang20242d}, which parameterizes 2D Gaussian disks with position, rotation, scales, opacity, and color attributes.
We directly assign $\mathcal{P}[0]$ as the positions and use their estimated normals to orient the disks along the surface.
We calculate the average euclidean distance of each point to its $k{=}10$ nearest neighbors to determine the Gaussian scale.
We uniformly initialize the opacities to $0.1$ and convert the pointcloud colors into the degree-0 coefficients of Spherical Harmonics.

Since the generated image set $\{\mathbf{I}_i\}_{i=0}^N$ is inconsistent, we cannot directly render our scene from the corresponding cameras to train the 2D Gaussians.
Doing so would effectively \textit{undo} the geometric alignment to explain the generative drift in the image observations (see \Cref{fig:qual_ablation}). 
To this end, we propose a \textit{non-rigid aware} rendering objective, that first transforms the Gaussians into a per-frame camera space.
Concretely, we apply \Cref{eq:bkwd-def} on the Gaussian positions and rotations, which non-rigidly deforms them into the inconsistent frame spaces, and transform the scales using $(\mathbf{R}_i^{-1}, \mathbf{t}_i^{-1})$.
Then, we follow 2DGS \cite{huang20242d} to rasterize the Gaussians and perform volumetric alpha blending to render the image set $\{\mathbf{\hat{I}}_i\}_{i=0}^N$.
We jointly optimize the 2D Gaussians $\mathcal{G}$ and cameras $\{\mathbf{R}_i, \mathbf{t}_i \}_{i=0}^N$ with standard rendering objectives:
\begin{equation}
    \mathcal{L}_\text{rend} = \frac{1}{N} \sum_{i=0}^{N} \bigl( \lambda_{1} \|\mathbf{\hat{I}}_i {-} \mathbf{I}_i\|^1 {+} \lambda_{2} \text{LPIPS}(\mathbf{\hat{I}}_i, \mathbf{I}_i) \bigr),
\end{equation}
where $\text{LPIPS}$ denotes the perceptual loss \cite{johnson2016perceptual}.
Since \Cref{eq:bkwd-def} is differentiable, this loss optimizes the 2D Gaussians in their canonical state to explain the image observations.
Crucially, the non-rigid deformation \textit{factors out} the geometric inconsistencies, which keeps the scene representation aligned while optimizing for sharp and detailed appearance.
We additionally adopt the depth and normal regularizers of 2DGS \cite{huang20242d} and optimize the final objective 
\begin{equation}
    E(\mathcal{G}, \{\mathbf{R}_i, \mathbf{t}_i \}_{i=0}^N) = \mathcal{L}_\text{rend} + \lambda_d \mathcal{L}_d + \lambda_n \mathcal{L}_n.
\end{equation}
Since $\mathcal{P}[0]$ is a dense and accurate surface initialization, we observe fast convergence in only a few thousand iterations, similar to EDGS \cite{kotovenko2025edgs}.
For specular scenes we additionally optimize higher degree Spherical Harmonics coefficients in a second training phase.
In practice, we find it beneficial to freeze the cameras and positions in this stage to avoid degenerate solutions that model the generative drift of the input images into the view-dependent effects.

\section{Results}
\label{sec:results}

\subsection{Implementation Details}

We select uniform hyperparameters for all results.
The coarse-to-fine ICP optimizes with $s_\text{vox} {=} [4\text{cm}, 2\text{cm}]$ and $d_{\max} {=} [5\text{cm}, 3\text{cm}]$ for 50 and 150 iterations, respectively, with a learning rate of $1\mathrm{e}{-3}$ and $\lambda_\text{color} {=} 0.05, \lambda_\text{corr} {=} 1.0, \lambda_\text{tv} {=} 10.0$.
We set the percentile thresholds as $\theta_\text{loc} {=} 15.0, \theta_\text{cnt} {=} 50.0, \theta_\text{d} {=} \theta_\text{g} {=} 75.0$ and the maximum standard deviations for outlier removal to $\sigma_d {=} 2.5, \sigma_c {=} 1.5$.
We detect up to $5,000$ correspondences in up to $20$ image pairs for $\mathcal{L}_\text{corr}$.
The global optimization runs for another 100 iterations with $\lambda_\text{anchor} {=} 50.0$.
Both stages take a total of 25 minutes / 20GB on average for $N {=} 50$ images on a single A6000 GPU.
We subsample $\mathcal{P}[0]$ to roughly 1.5M Gaussians to optimize a 3D scene in 10 minutes / 8GB for $5,000$ iterations following the setup of 2DGS \cite{huang20242d}, but without densification. We optionally optimize the SH for another $10,000$ iterations.

\subsection{Single Video 3D Reconstruction}

\begin{table}[t]
\centering
\small
\setlength{\tabcolsep}{5pt}
\begin{tabular}{l | cc | cc}
\toprule
        \multirow{2}{*}{Method} & \multicolumn{2}{c}{Consistency} & \multicolumn{2}{c}{Fidelity}\\
                        \cmidrule(l{2pt}r{2pt}){2-3} \cmidrule(l{2pt}r{2pt}){4-5}
 & 3D & Photometric & CLIP-IQA+ & CLIP Aesthetic \\
\midrule
DA3 \cite{lin2025depth} & 69.53 & 71.62 & 31.64 & 36.64 \\
3DGS-MCMC \cite{kheradmand20243d} & 67.34 & 69.58 & 35.56 & 35.06 \\
VGGT-X \cite{liu2025vggt} & 65.73 & 65.58 & 38.21 & 37.24 \\
VGGT-X$^\dagger$ & 69.66 & 67.05 & 41.44 & 37.53 \\
Ours & \textbf{79.29} & \textbf{86.59} & \textbf{46.56} & \textbf{37.61} \\
\bottomrule
\end{tabular}
\caption{\textbf{Quantitative comparison.}
We compare against baselines on the single video 3D reconstruction task and report averaged results across all source video models.
Our method achieves the highest 3D and photometric consistency, which underlines its ability to reconstruct worlds from inconsistent frames.
We also improve in terms of image fidelity, which signals our images are of higher sharpness and texture detail.
}
\label{tab:quant_single}
\end{table}

\begin{figure*}[t]
    \centering
    \setlength{\tabcolsep}{1pt}
    \renewcommand{\arraystretch}{1.1}

    \begin{tabular}{c | c | c c c c}
        {\fontsize{8}{9}\selectfont Video Frames} &
        &
        {\fontsize{8}{9}\selectfont DA3~\cite{lin2025depth}} &
        {\fontsize{8}{9}\selectfont 3DGS-MCMC~\cite{kheradmand20243d}} &
        {\fontsize{8}{9}\selectfont VGGT-X$^\dagger$~\cite{liu2025vggt}} &
        {\fontsize{8}{9}\selectfont Ours} \\

        \midrule

        \includegraphics[width=0.19\textwidth]{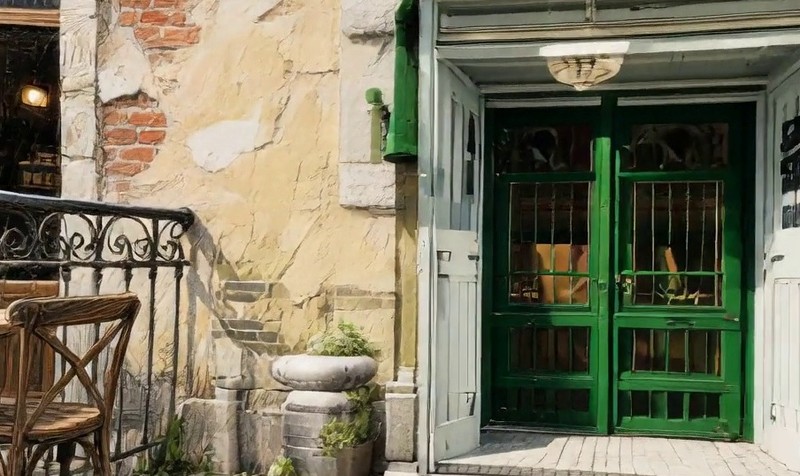}
        & \raisebox{0.0\height}{\rotatebox{90}{\fontsize{8}{9}\selectfont Input}}
        & \includegraphics[width=0.19\textwidth]{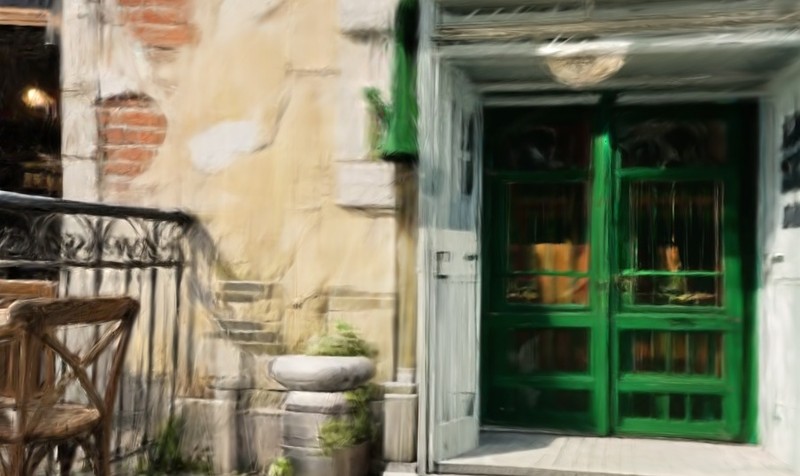}
        & \includegraphics[width=0.19\textwidth]{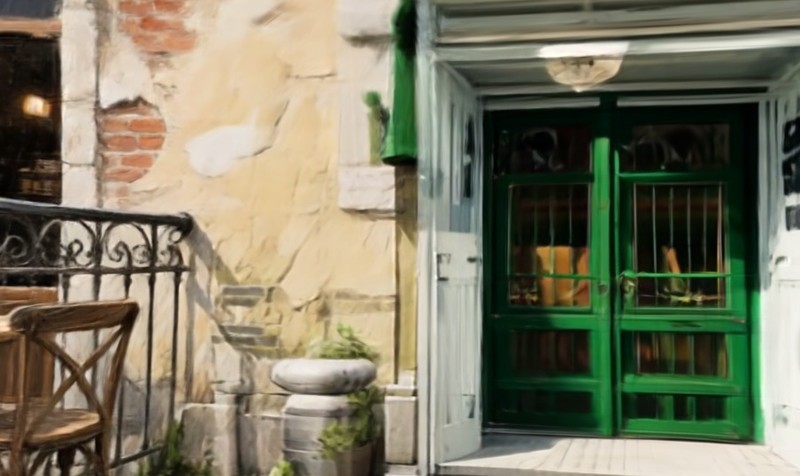}
        & \includegraphics[width=0.19\textwidth]{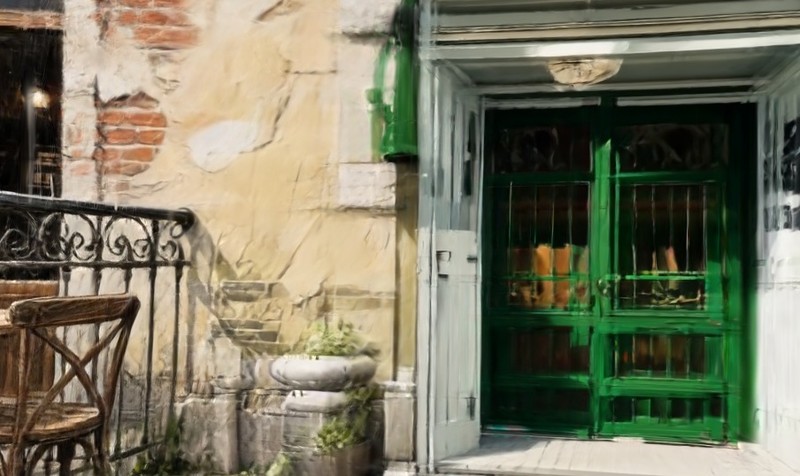}
        & \includegraphics[width=0.19\textwidth]{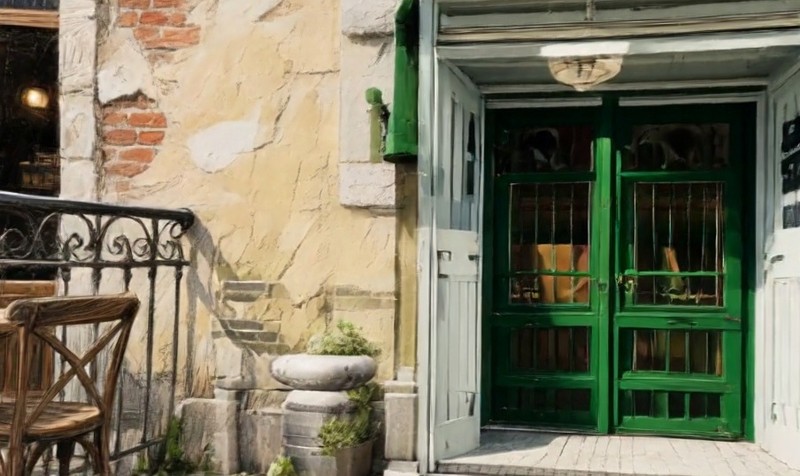} \\

        \includegraphics[width=0.19\textwidth]{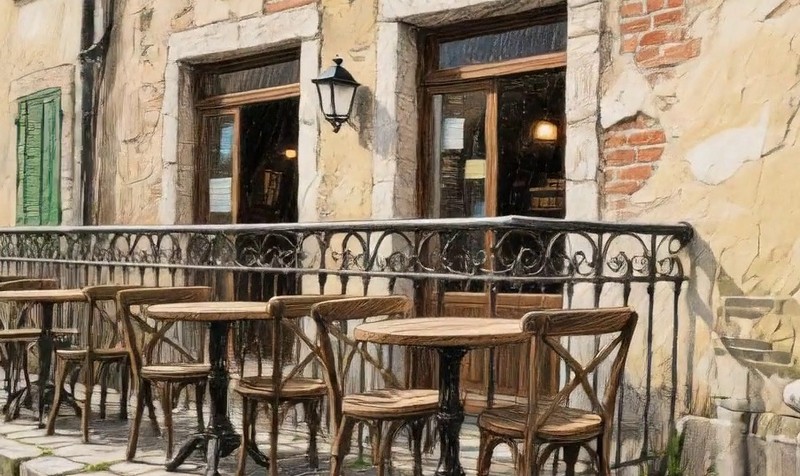}
        & \multirow{2}{*}{\rotatebox{90}{\fontsize{8}{9}\selectfont Novel Views}}
        & \includegraphics[width=0.19\textwidth]{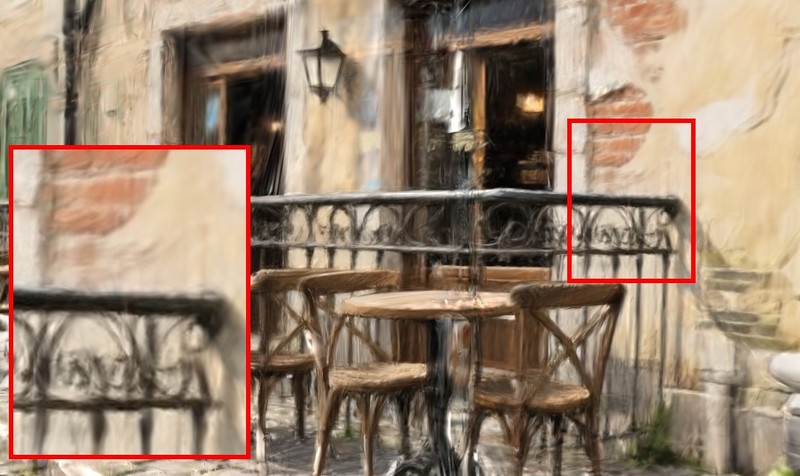}
        & \includegraphics[width=0.19\textwidth]{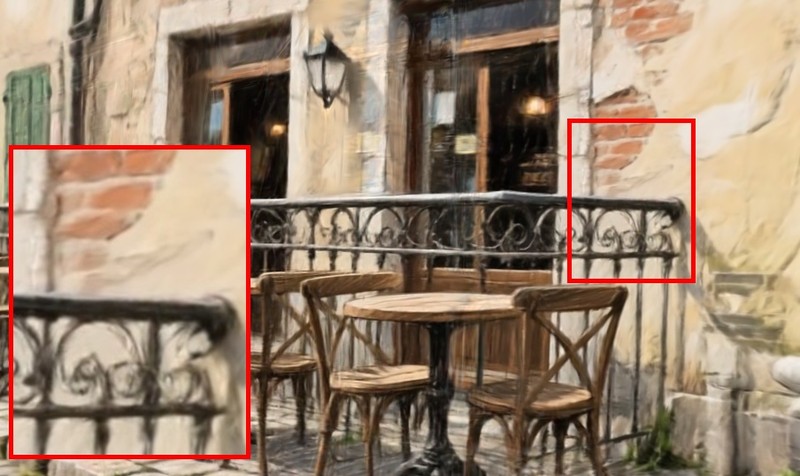}
        & \includegraphics[width=0.19\textwidth]{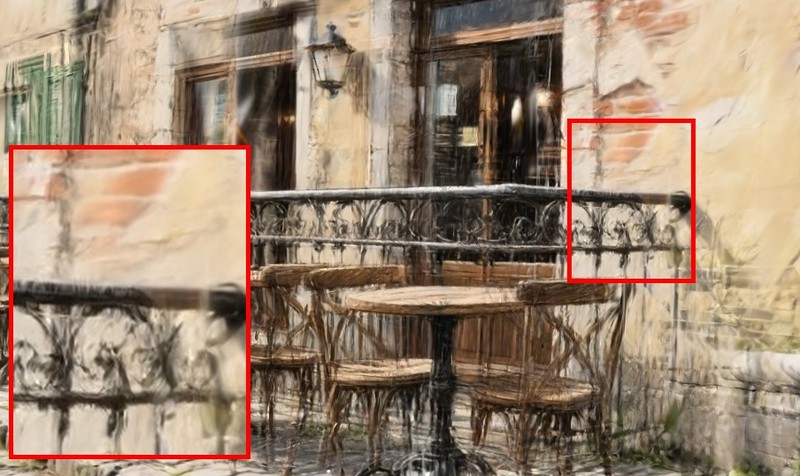}
        & \includegraphics[width=0.19\textwidth]{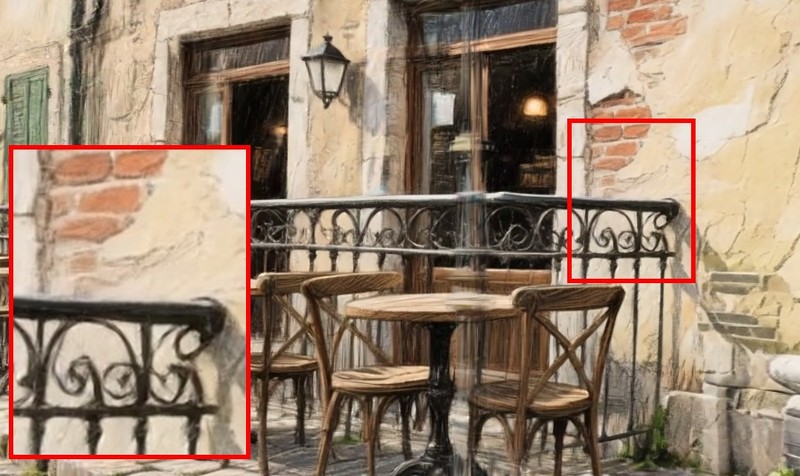} \\

        \includegraphics[width=0.19\textwidth]{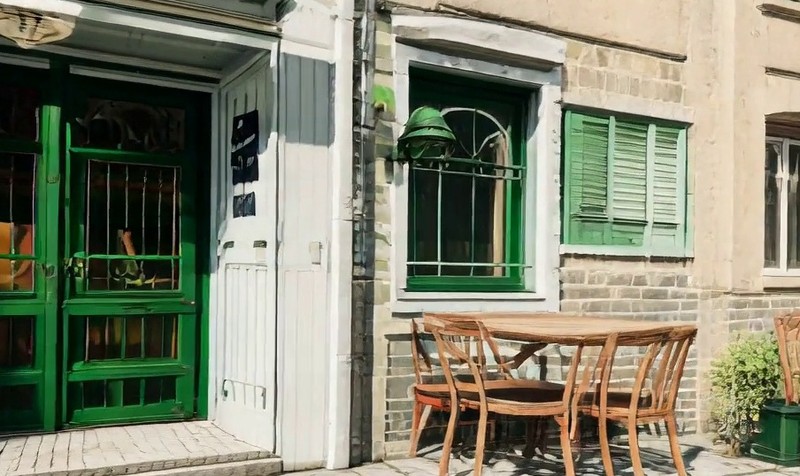}
        &
        & \includegraphics[width=0.19\textwidth]{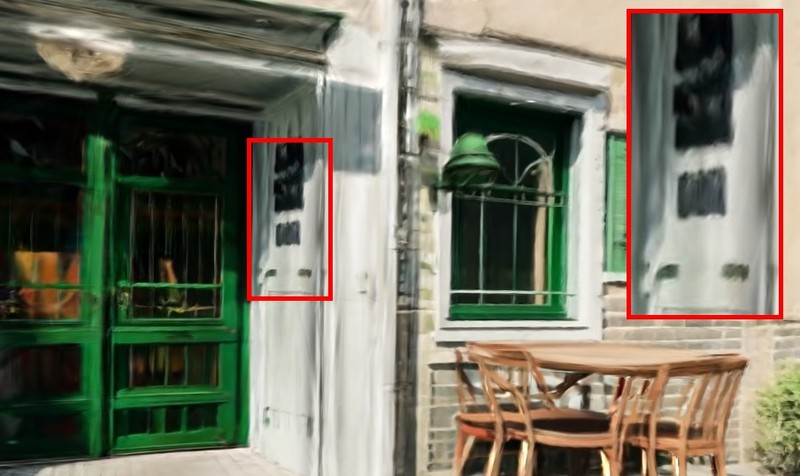}
        & \includegraphics[width=0.19\textwidth]{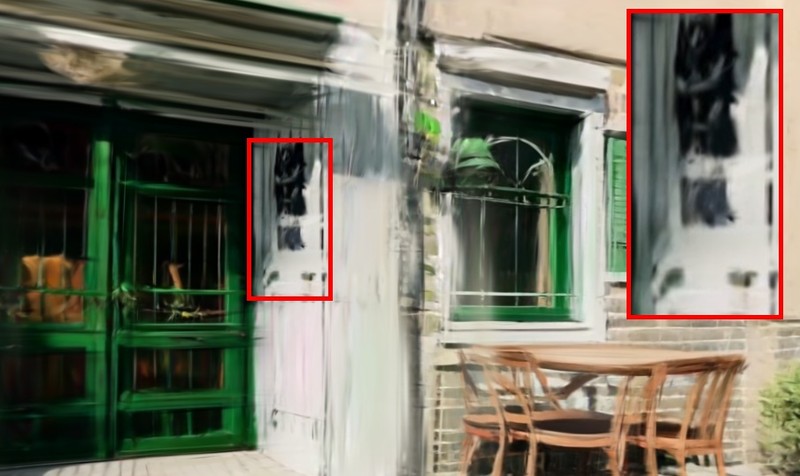}
        & \includegraphics[width=0.19\textwidth]{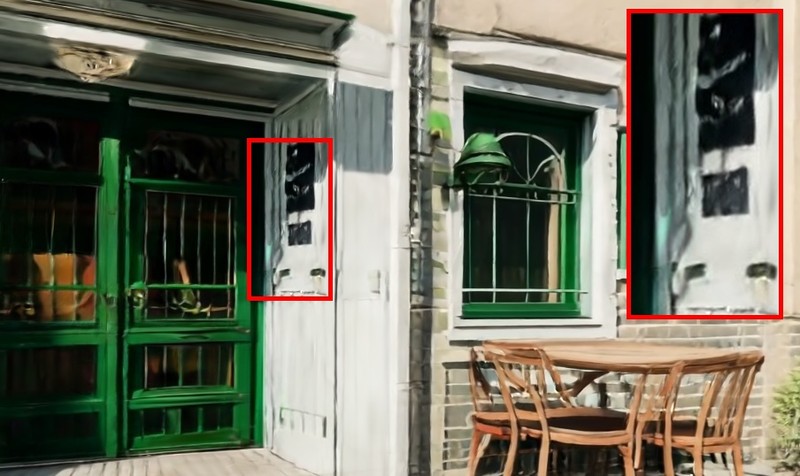}
        & \includegraphics[width=0.19\textwidth]{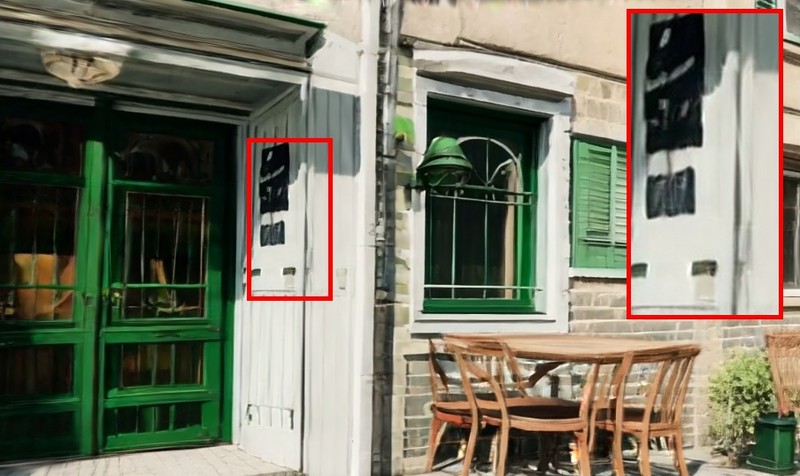} \\

        \midrule

        \includegraphics[width=0.19\textwidth]{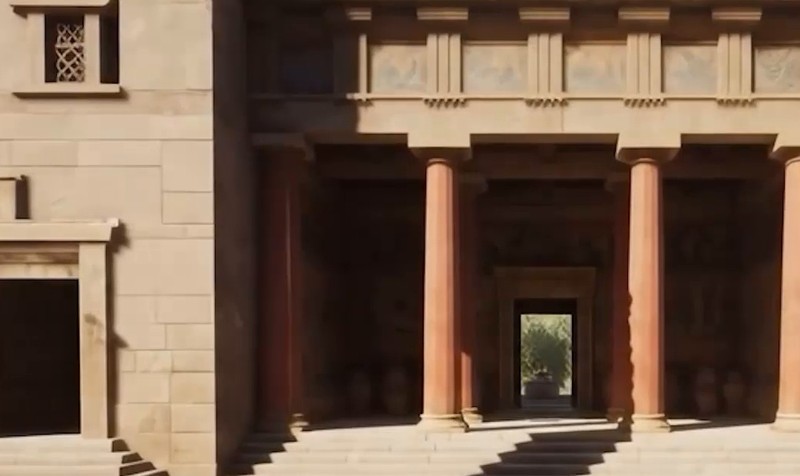}
        & \multirow{2}{*}{\rotatebox{90}{\fontsize{8}{9}\selectfont Novel Views}}
        
        & \includegraphics[width=0.19\textwidth]{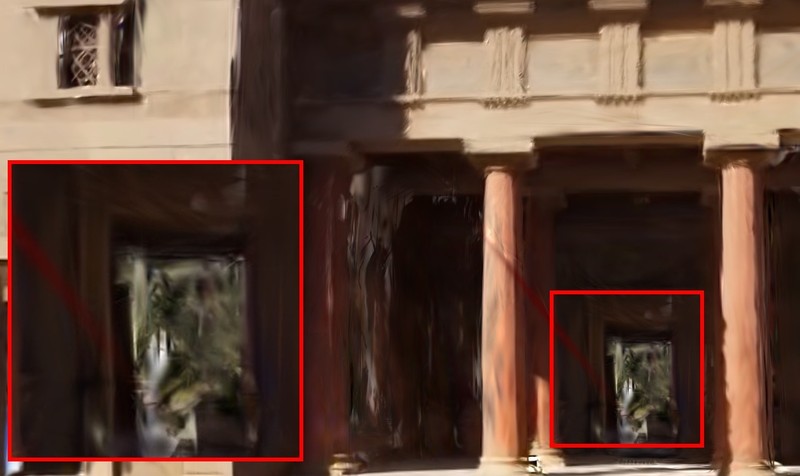}
        & \includegraphics[width=0.19\textwidth]{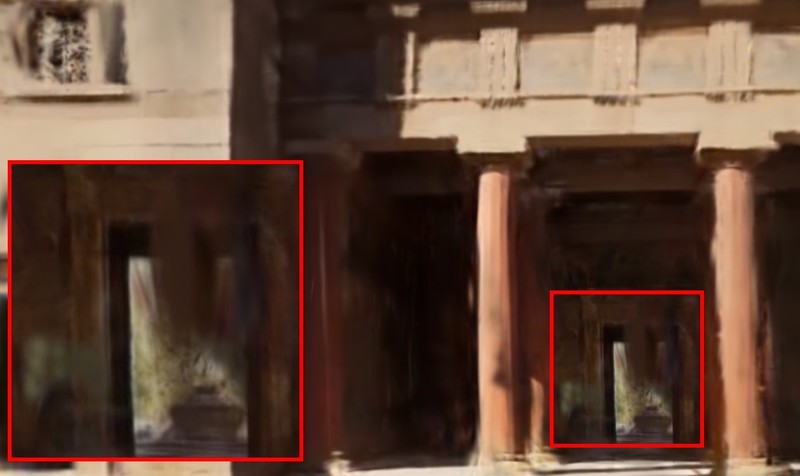}
        & \includegraphics[width=0.19\textwidth]{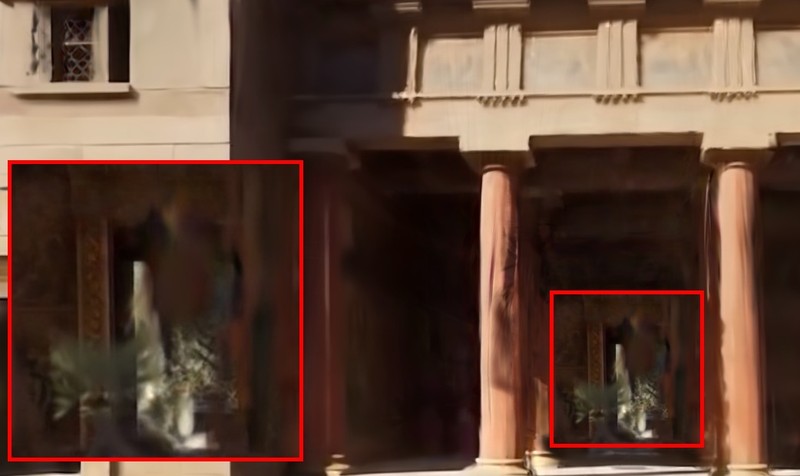}
        & \includegraphics[width=0.19\textwidth]{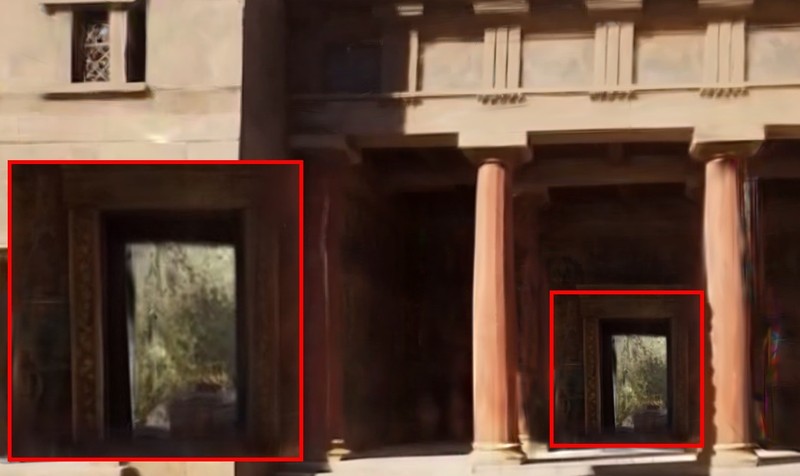} \\

        \includegraphics[width=0.19\textwidth]{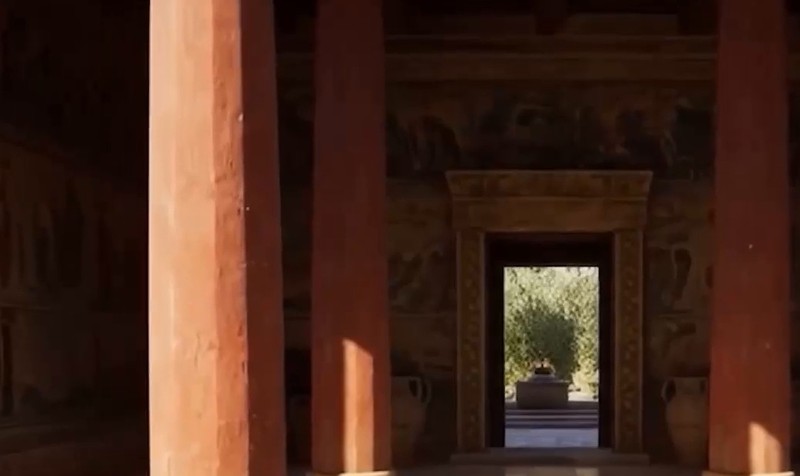}
        & 
        & \includegraphics[width=0.19\textwidth]{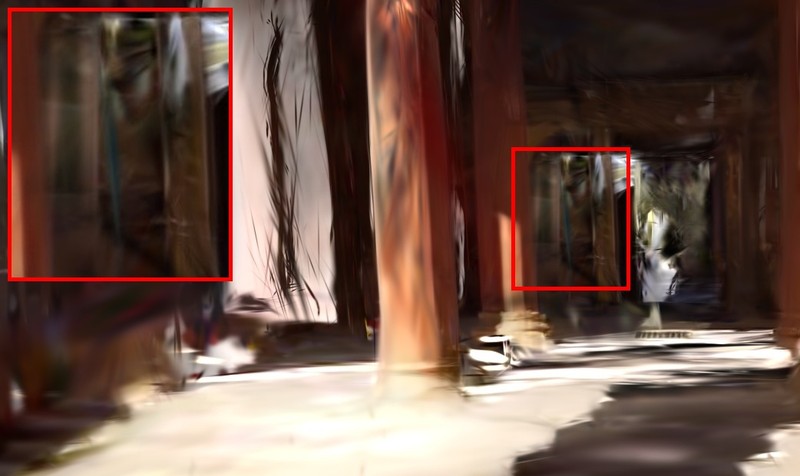}
        & \includegraphics[width=0.19\textwidth]{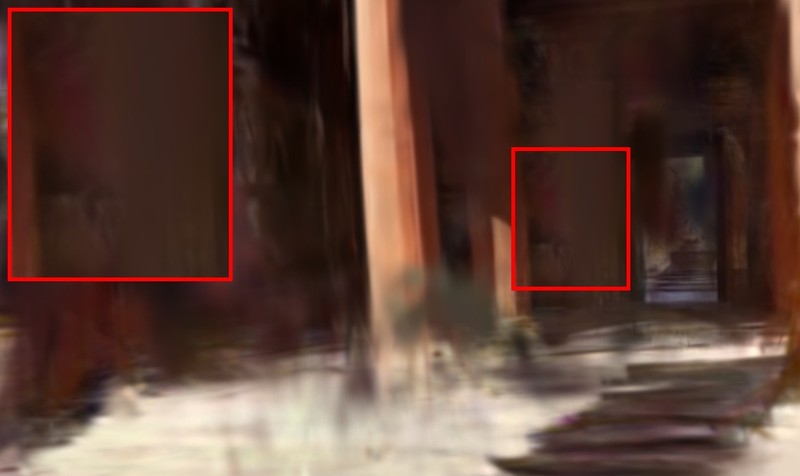}
        & \includegraphics[width=0.19\textwidth]{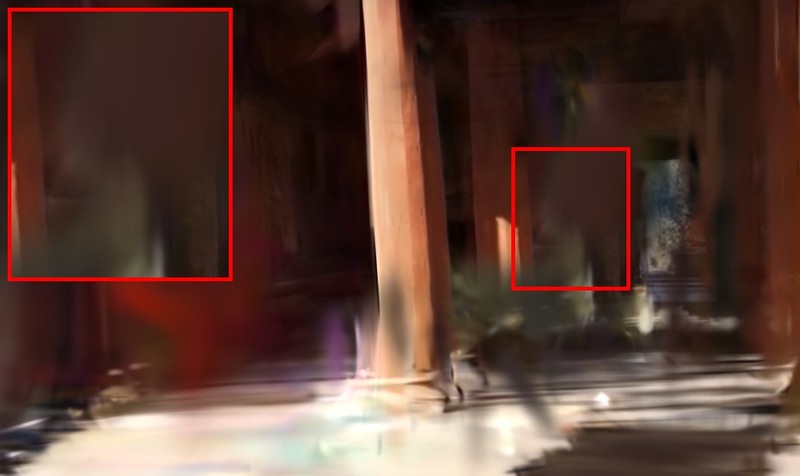}
        & \includegraphics[width=0.19\textwidth]{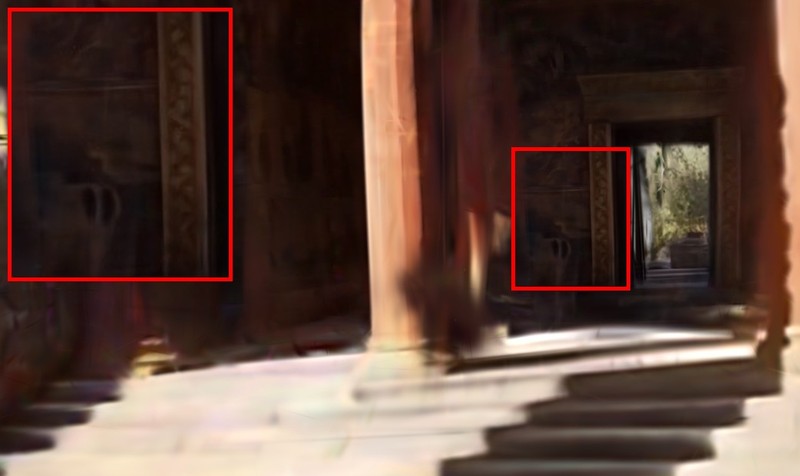} \\

        \midrule

        \includegraphics[width=0.19\textwidth]{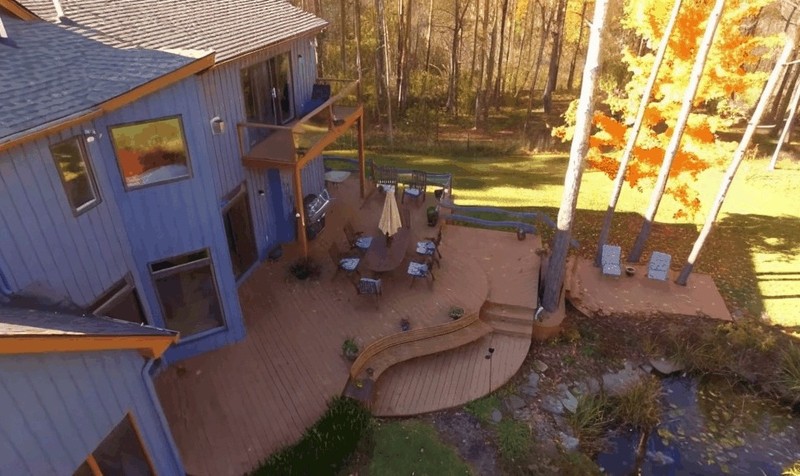}
        & \multirow{2}{*}{\rotatebox{90}{\fontsize{8}{9}\selectfont Novel Views}}
        
        & \includegraphics[width=0.19\textwidth]{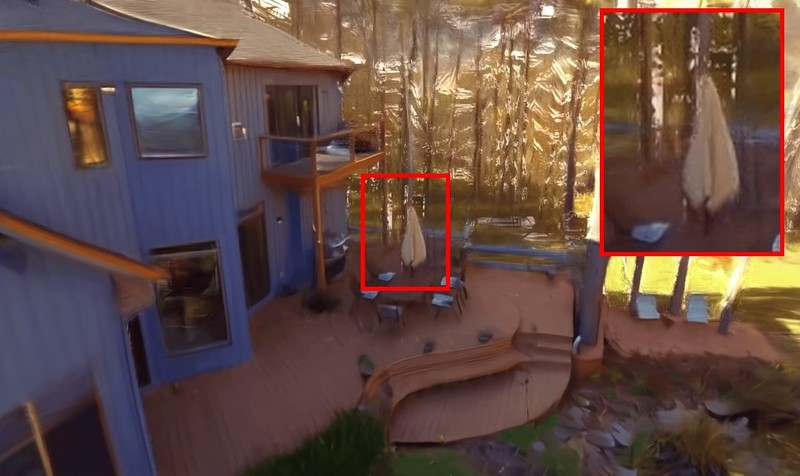}
        & \includegraphics[width=0.19\textwidth]{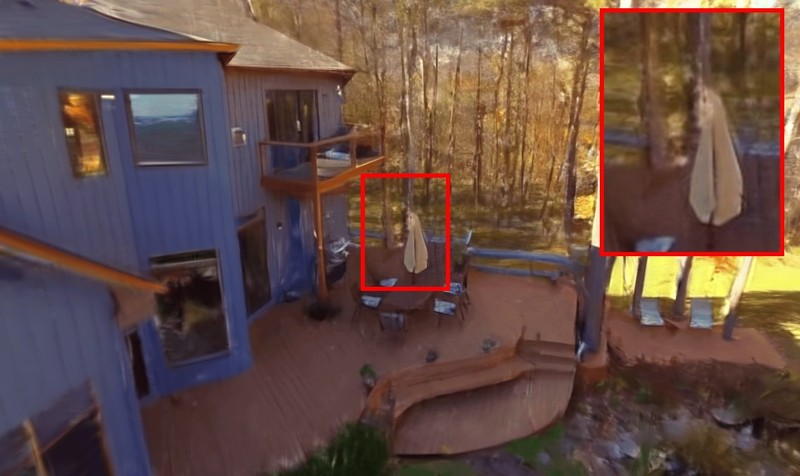}
        & \includegraphics[width=0.19\textwidth]{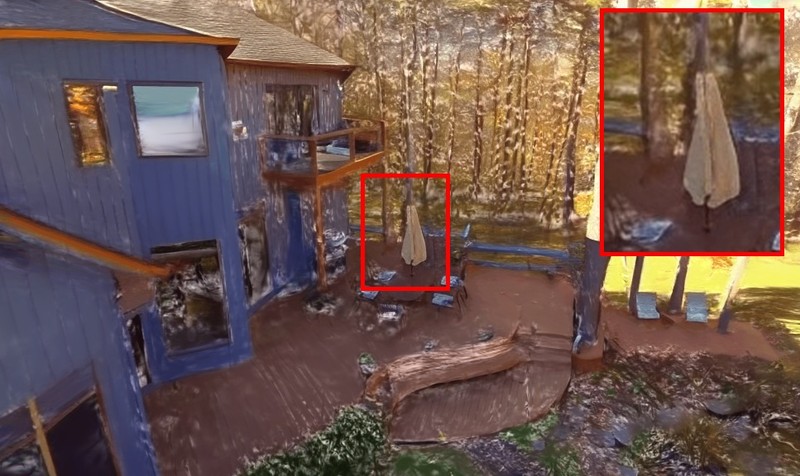}
        & \includegraphics[width=0.19\textwidth]{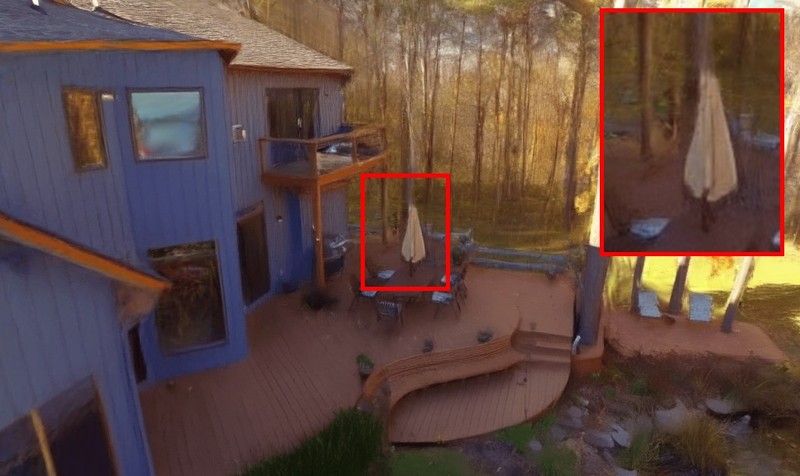} \\

        \includegraphics[width=0.19\textwidth]{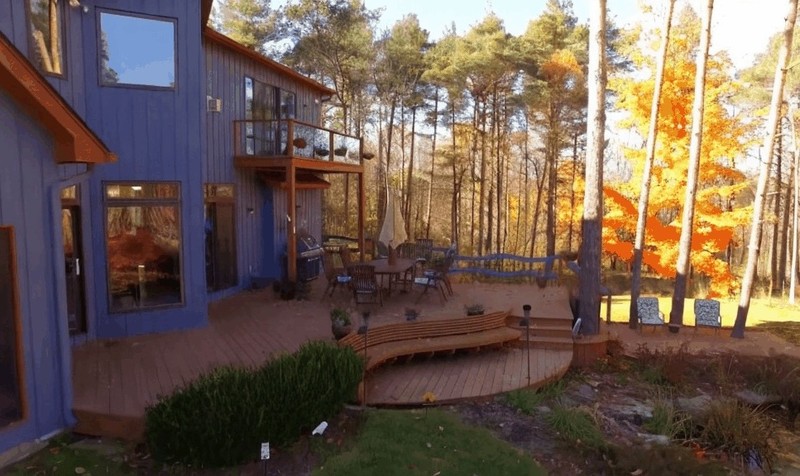}
        & 
        & \includegraphics[width=0.19\textwidth]{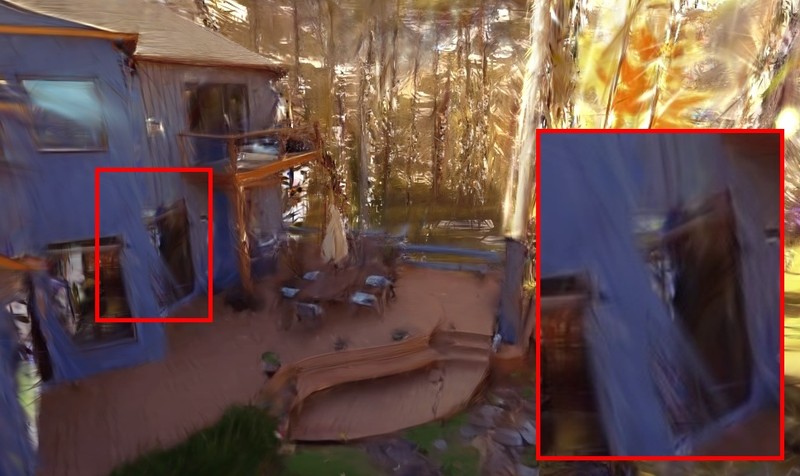}
        & \includegraphics[width=0.19\textwidth]{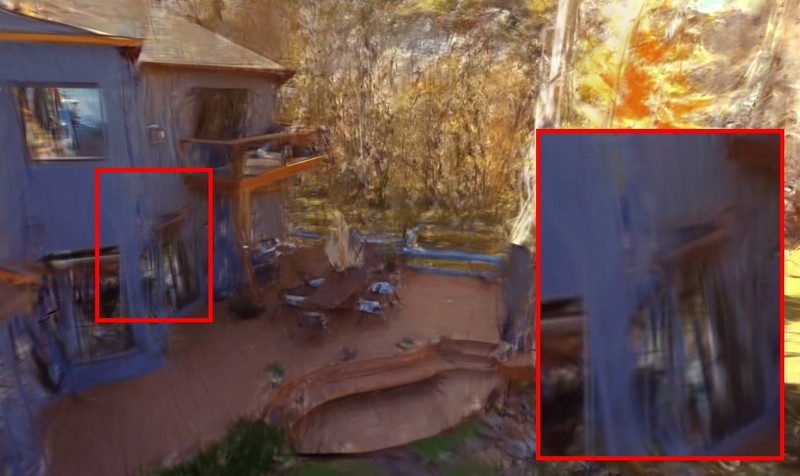}
        & \includegraphics[width=0.19\textwidth]{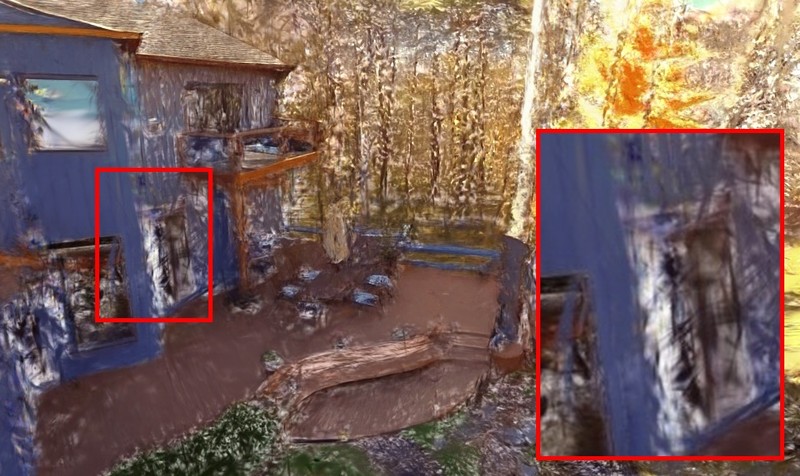}
        & \includegraphics[width=0.19\textwidth]{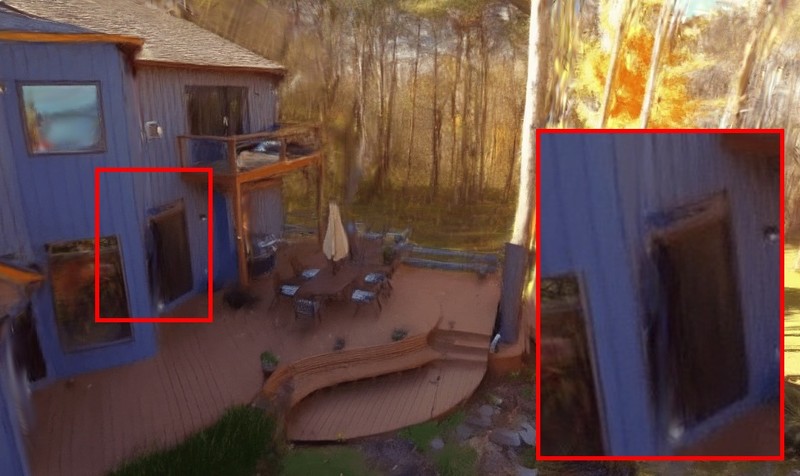} \\

    \end{tabular}
    \vspace{-4mm}
    \caption{\textbf{Single video 3D reconstruction.}
    We generate videos with HY-WorldPlay \cite{worldplay2025} (top), Genie3 \cite{genie3} (mid), ViewCrafter \cite{yu2024viewcrafter} (bottom) and 3D reconstruct them.
    Our method optimizes consistent worlds from inconsistent generated frames. 
    Compared to baselines, the renderings are of higher visual fidelity from both input and novel views.
    }
    \label{fig:qual_single1}
\end{figure*}

\begin{figure*}[t]
    \centering
    \setlength{\tabcolsep}{1pt}
    \renewcommand{\arraystretch}{1.1}

    \begin{tabular}{c | c | c c c c}
        {\fontsize{8}{9}\selectfont Video Frames} &
        &
        {\fontsize{8}{9}\selectfont DA3~\cite{lin2025depth}} &
        {\fontsize{8}{9}\selectfont 3DGS-MCMC~\cite{kheradmand20243d}} &
        {\fontsize{8}{9}\selectfont VGGT-X$^\dagger$~\cite{liu2025vggt}} &
        {\fontsize{8}{9}\selectfont Ours} \\

        \midrule

        \includegraphics[width=0.19\textwidth]{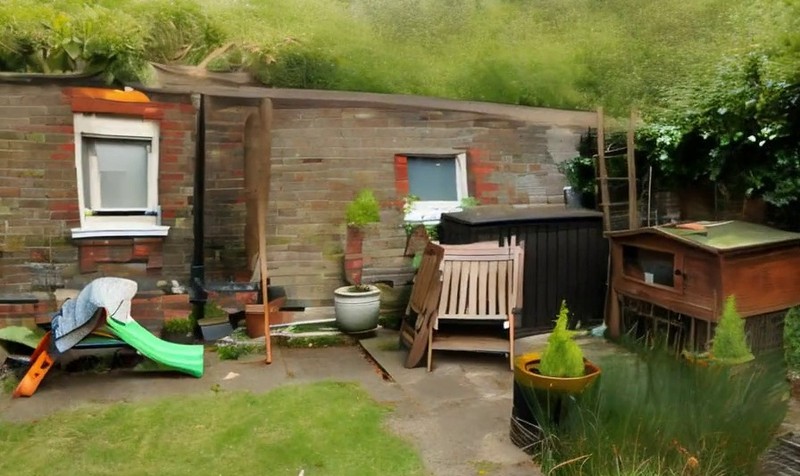}
        & \raisebox{0.0\height}{\rotatebox{90}{\fontsize{8}{9}\selectfont Input}}
        & \includegraphics[width=0.19\textwidth]{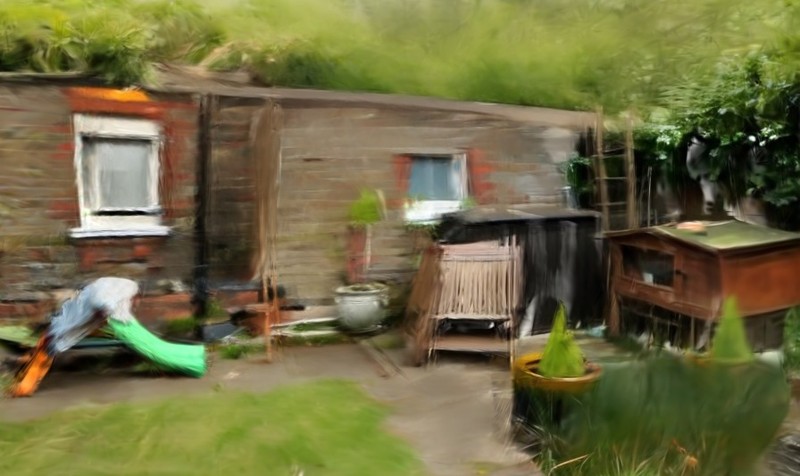}
        & \includegraphics[width=0.19\textwidth]{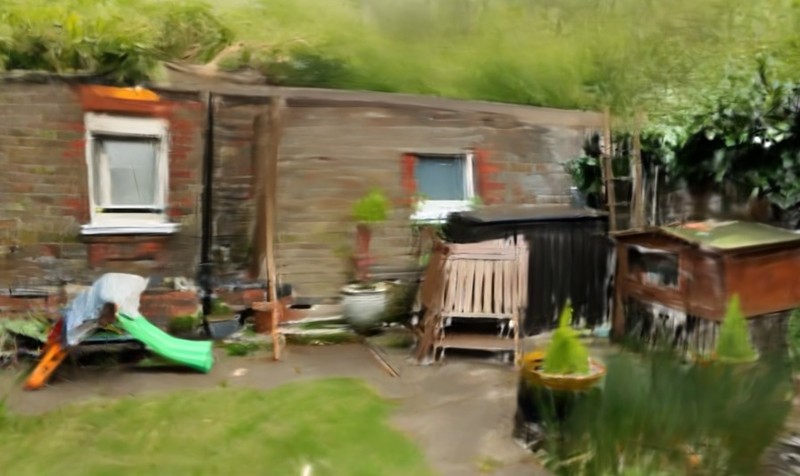}
        & \includegraphics[width=0.19\textwidth]{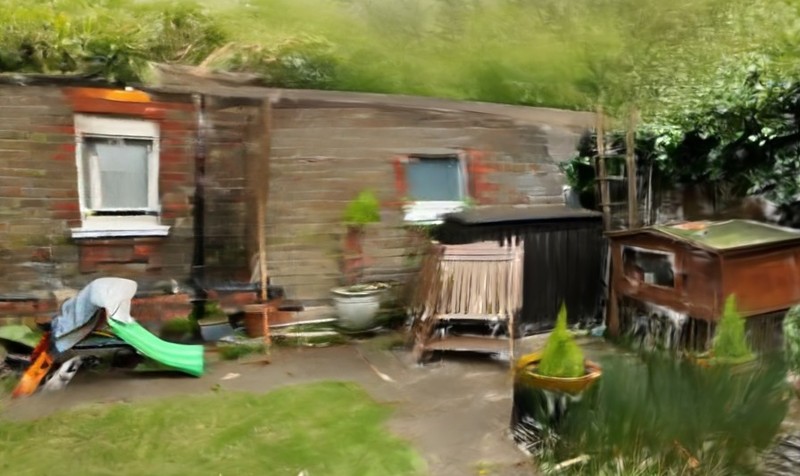}
        & \includegraphics[width=0.19\textwidth]{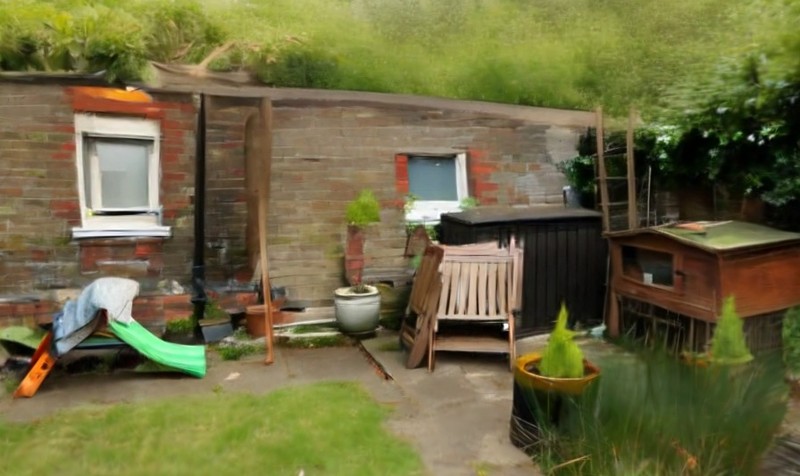} \\

        \includegraphics[width=0.19\textwidth]{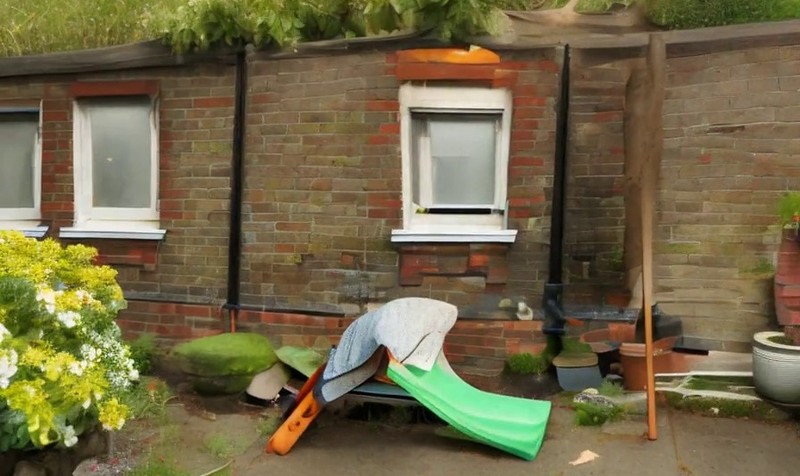}
        & \multirow{2}{*}{\rotatebox{90}{\fontsize{8}{9}\selectfont Novel Views}}
        & \includegraphics[width=0.19\textwidth]{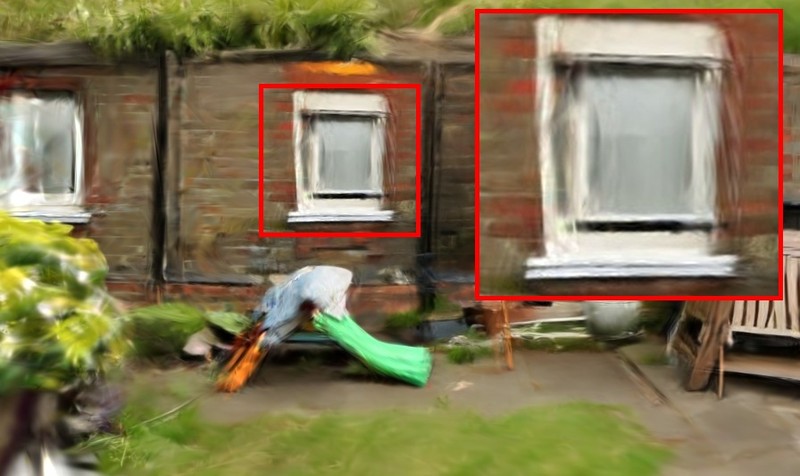}
        & \includegraphics[width=0.19\textwidth]{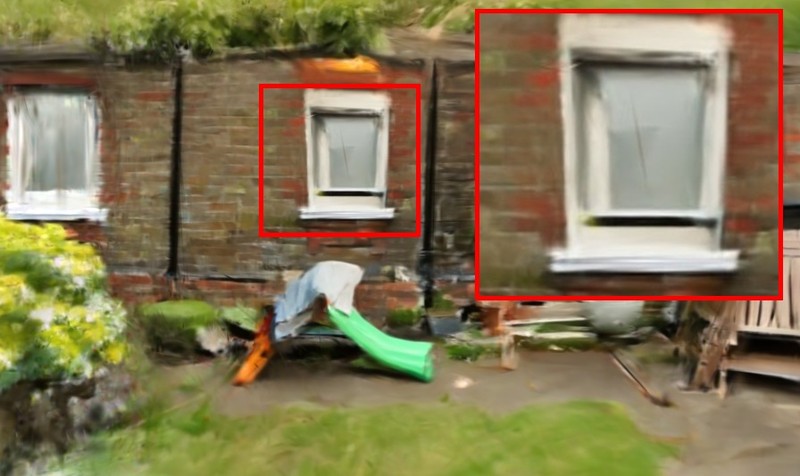}
        & \includegraphics[width=0.19\textwidth]{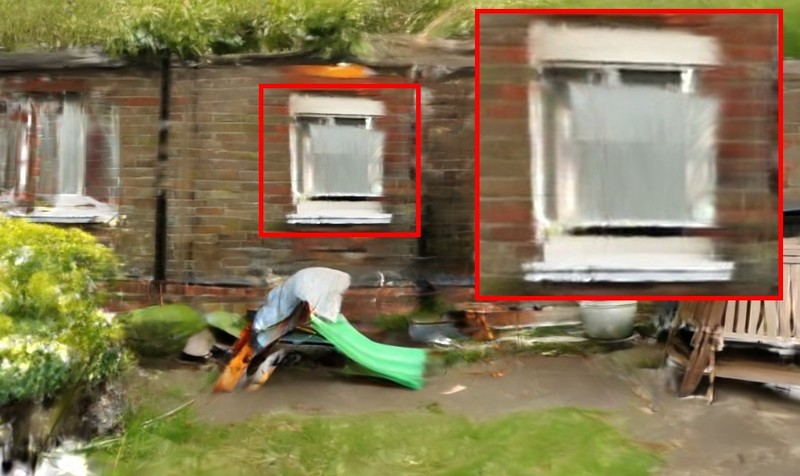}
        & \includegraphics[width=0.19\textwidth]{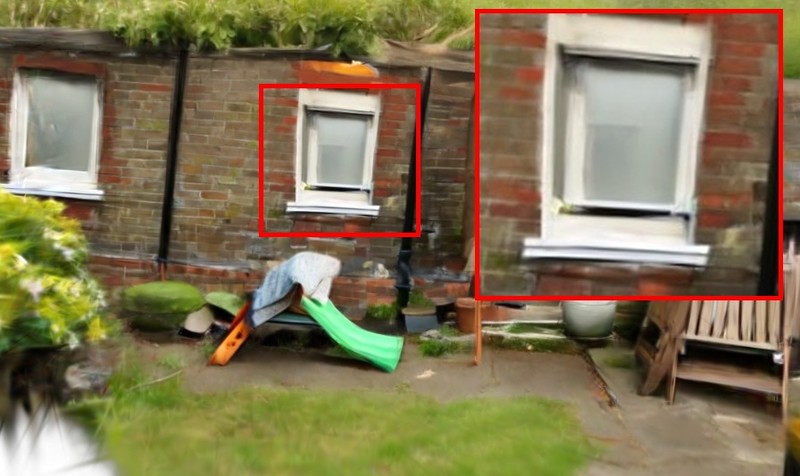} \\

        \includegraphics[width=0.19\textwidth]{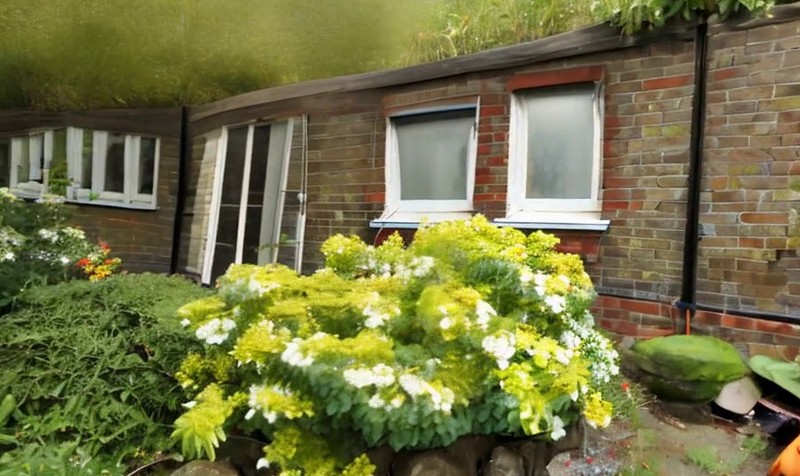}
        & 
        & \includegraphics[width=0.19\textwidth]{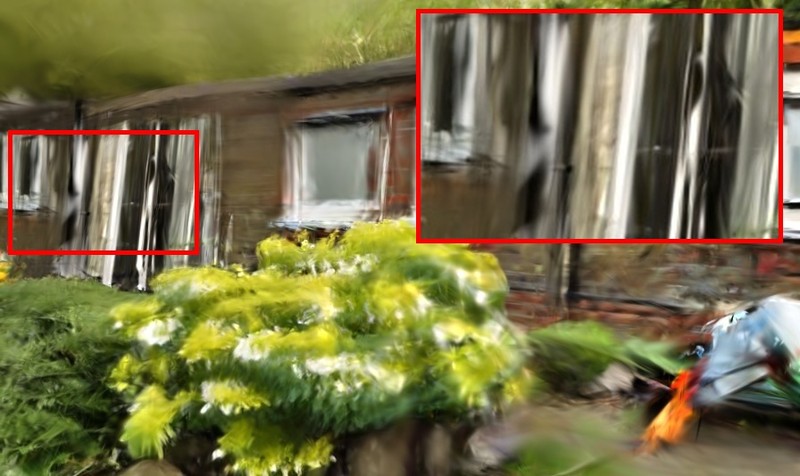}
        & \includegraphics[width=0.19\textwidth]{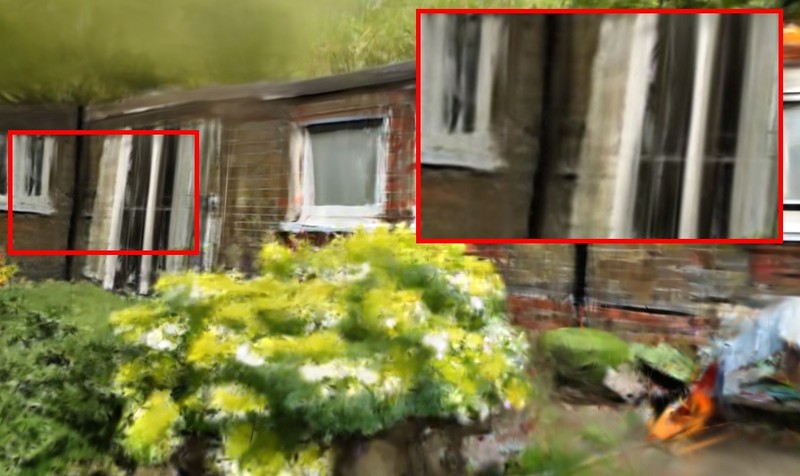}
        & \includegraphics[width=0.19\textwidth]{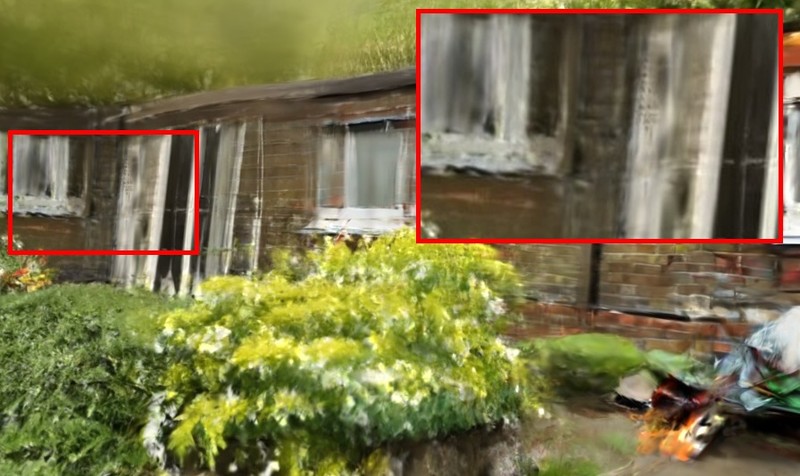}
        & \includegraphics[width=0.19\textwidth]{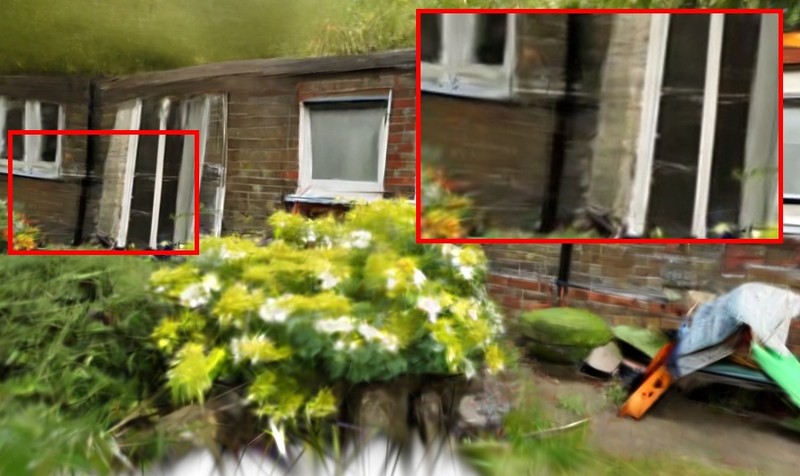} \\

        \midrule

        \includegraphics[width=0.19\textwidth]{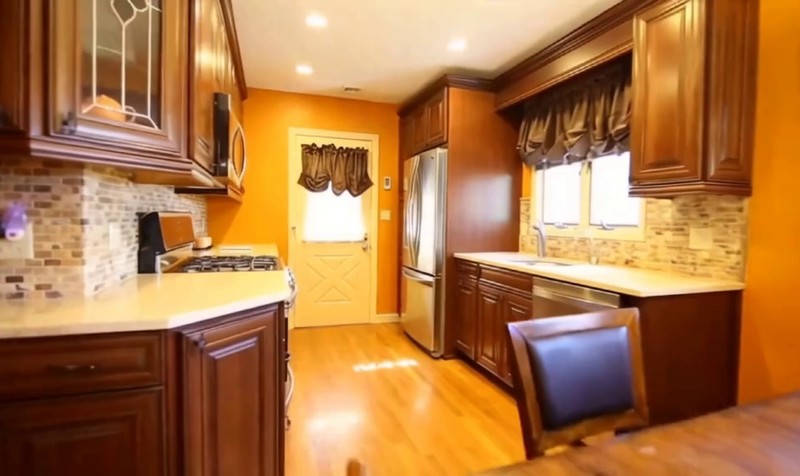}
        & \multirow{2}{*}{\rotatebox{90}{\fontsize{8}{9}\selectfont Novel Views}}
        
        & \includegraphics[width=0.19\textwidth]{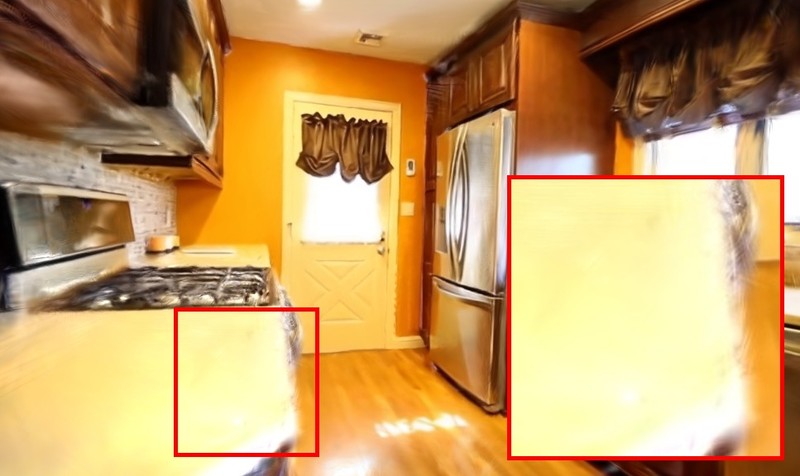}
        & \includegraphics[width=0.19\textwidth]{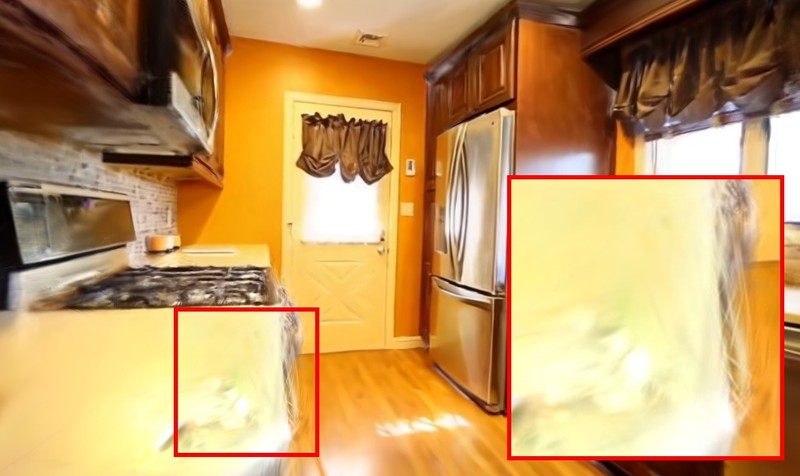}
        & \includegraphics[width=0.19\textwidth]{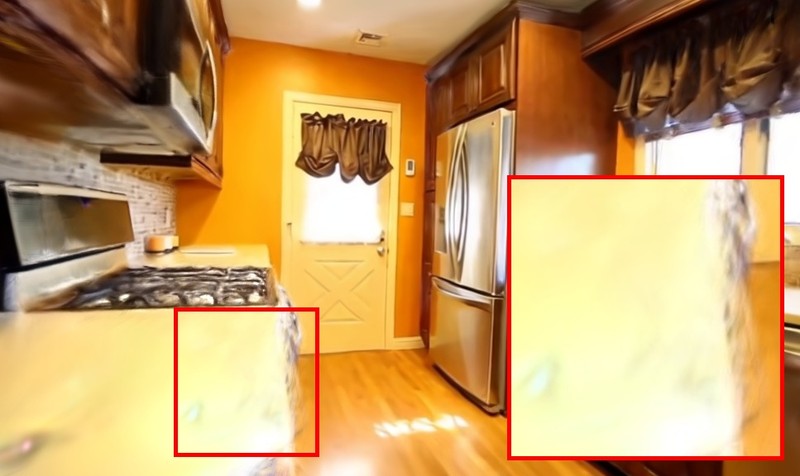}
        & \includegraphics[width=0.19\textwidth]{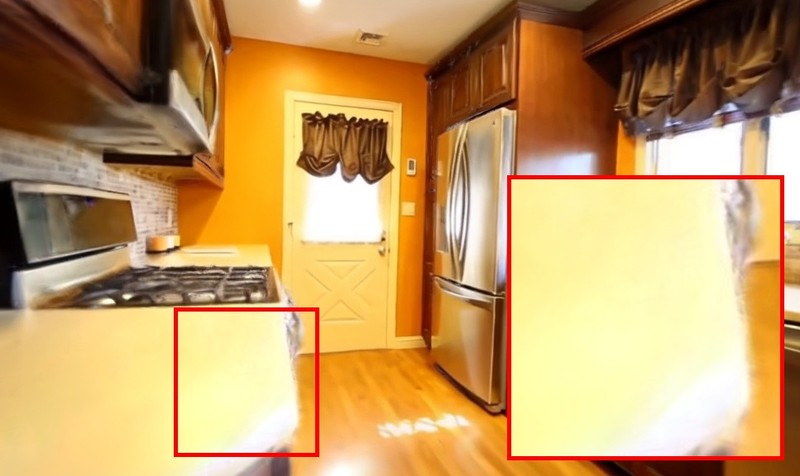} \\

        \includegraphics[width=0.19\textwidth]{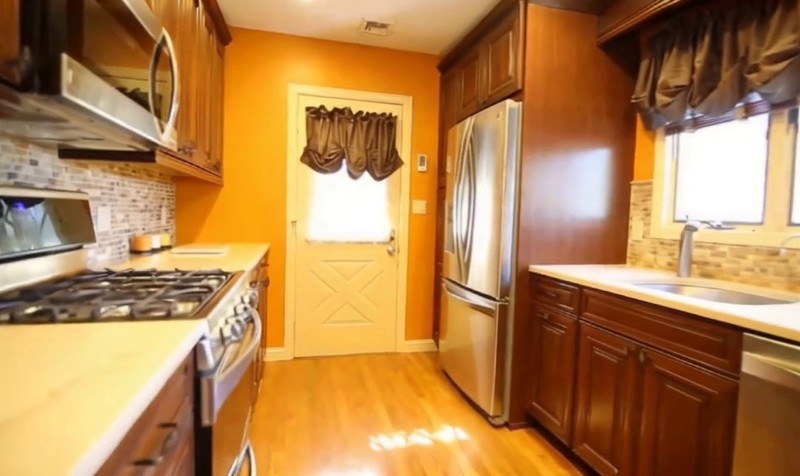}
        &
        & \includegraphics[width=0.19\textwidth]{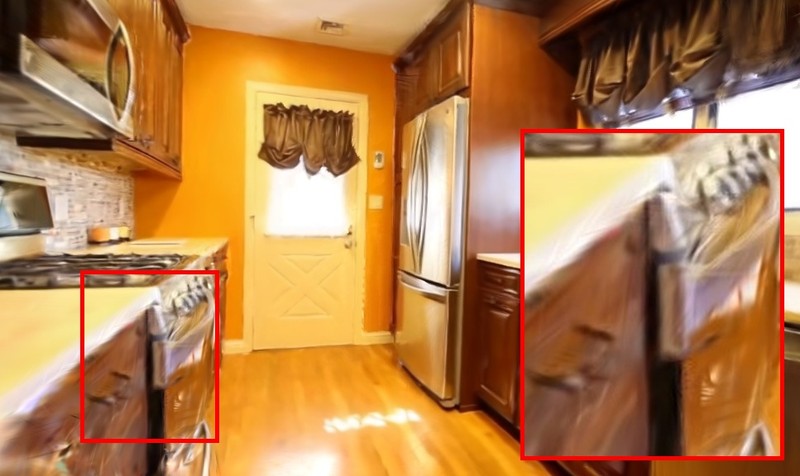}
        & \includegraphics[width=0.19\textwidth]{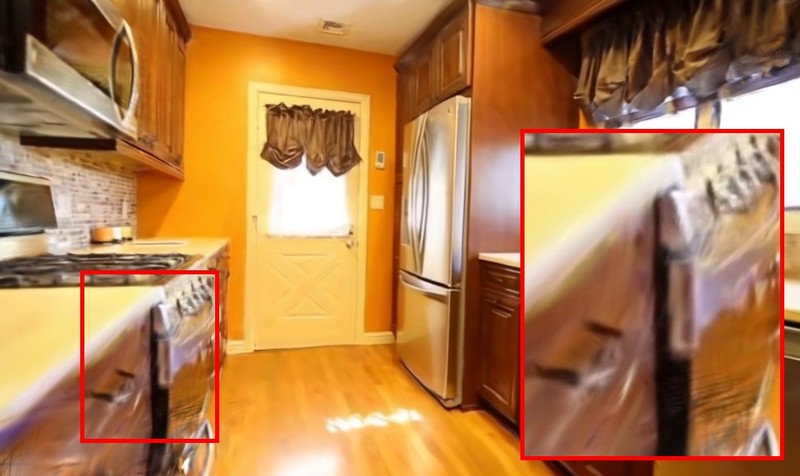}
        & \includegraphics[width=0.19\textwidth]{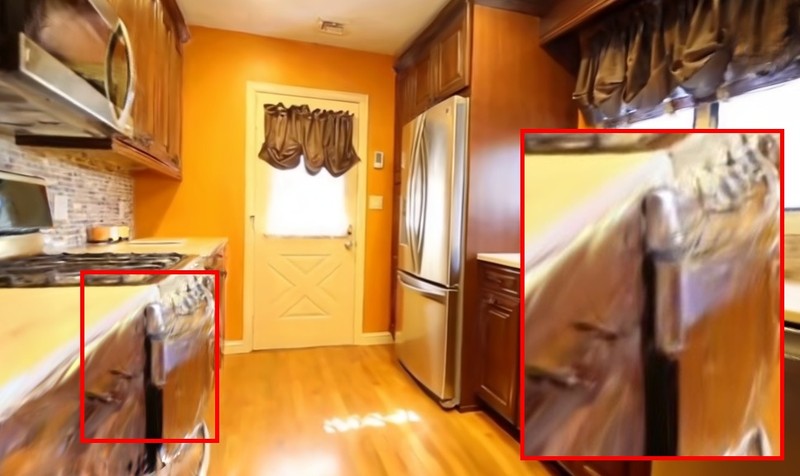}
        & \includegraphics[width=0.19\textwidth]{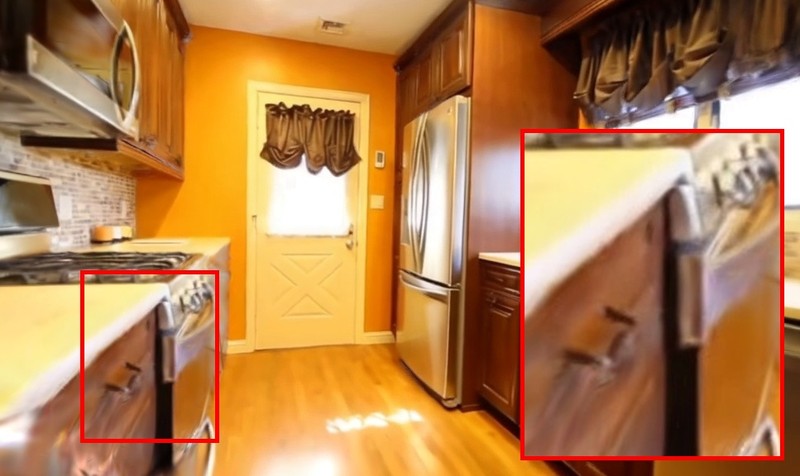} \\

        \midrule

        \includegraphics[width=0.19\textwidth]{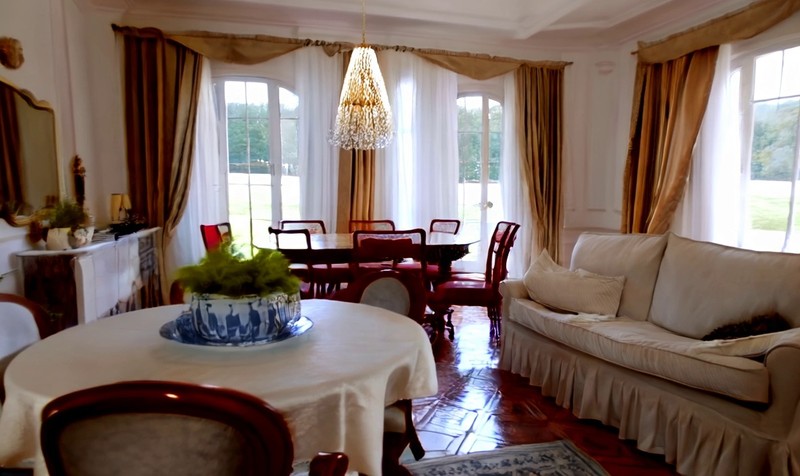}
        & \multirow{2}{*}{\rotatebox{90}{\fontsize{8}{9}\selectfont Novel Views}}
        
        & \includegraphics[width=0.19\textwidth]{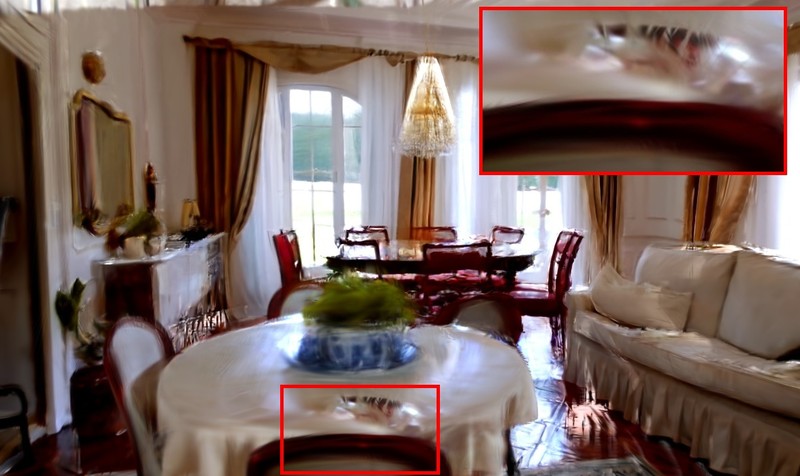}
        & \includegraphics[width=0.19\textwidth]{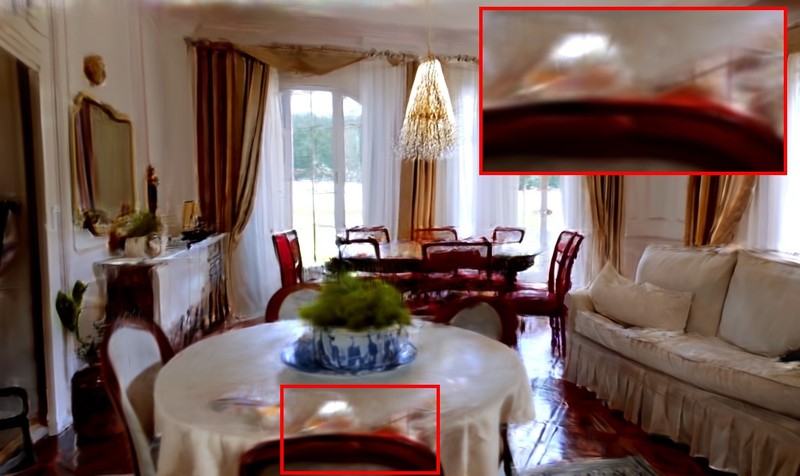}
        & \includegraphics[width=0.19\textwidth]{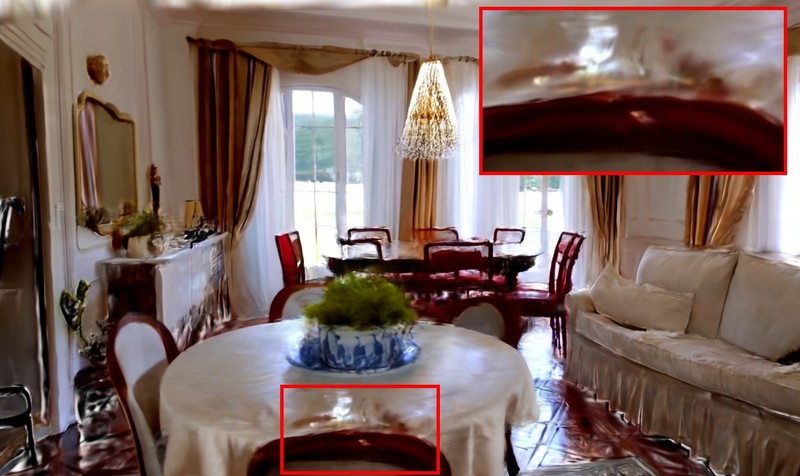}
        & \includegraphics[width=0.19\textwidth]{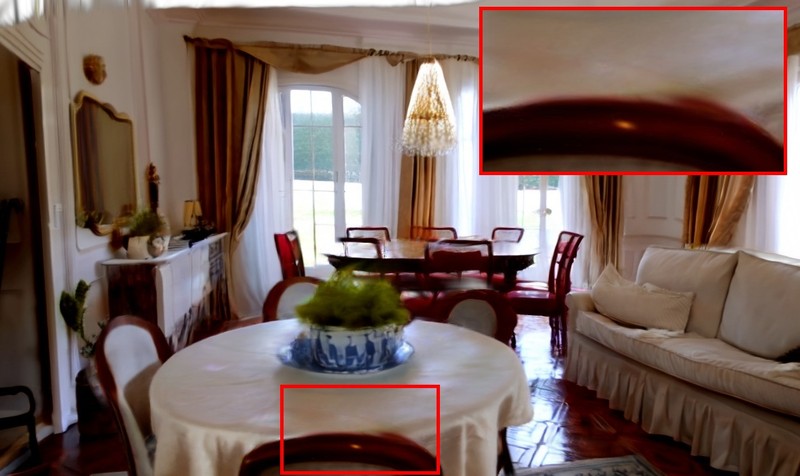} \\

        \includegraphics[width=0.19\textwidth]{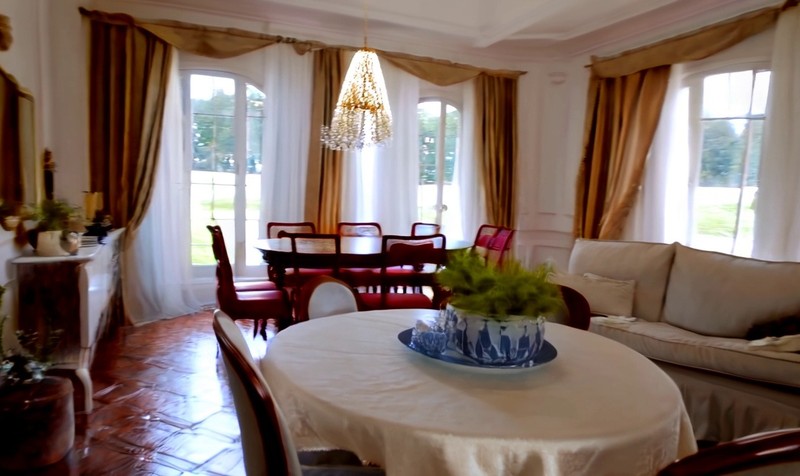}
        & 
        & \includegraphics[width=0.19\textwidth]{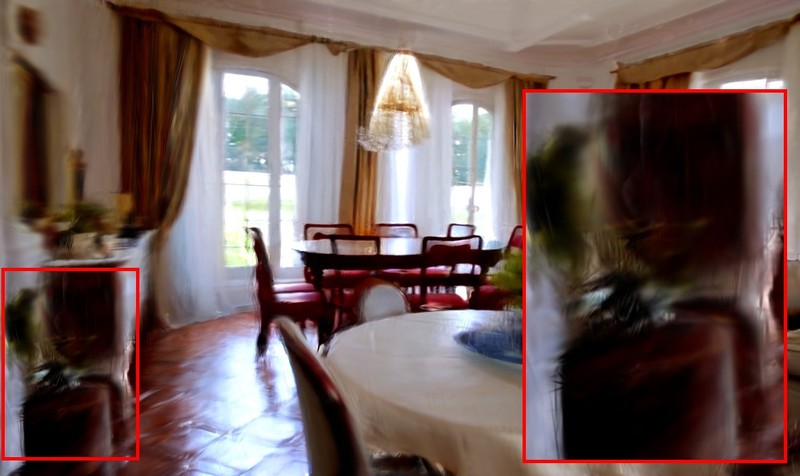}
        & \includegraphics[width=0.19\textwidth]{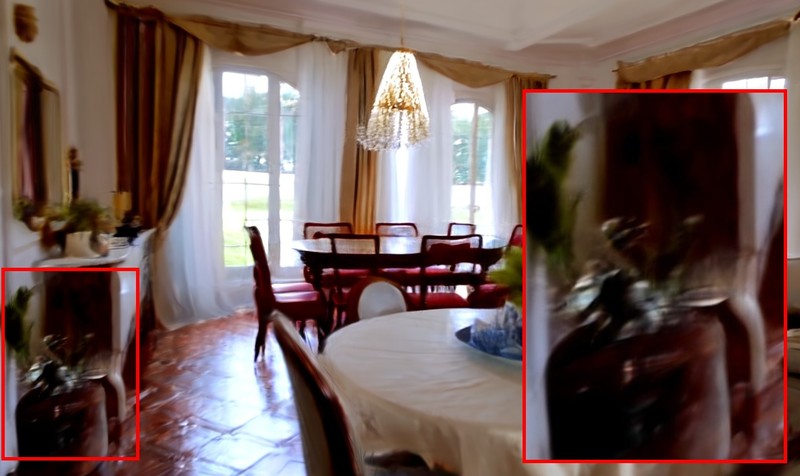}
        & \includegraphics[width=0.19\textwidth]{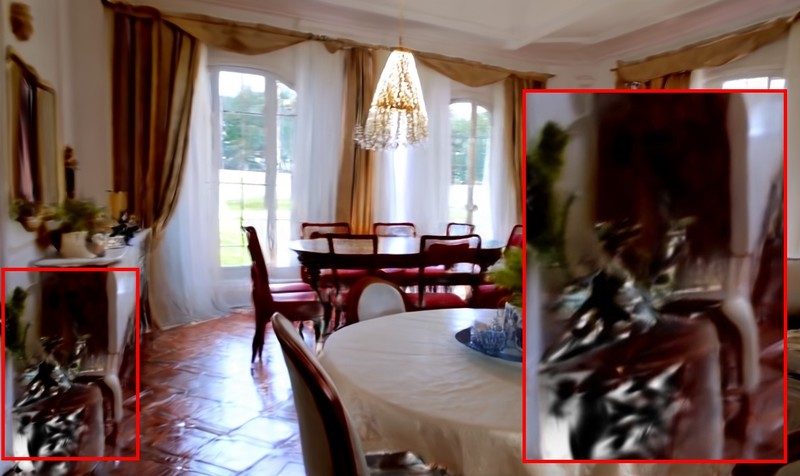}
        & \includegraphics[width=0.19\textwidth]{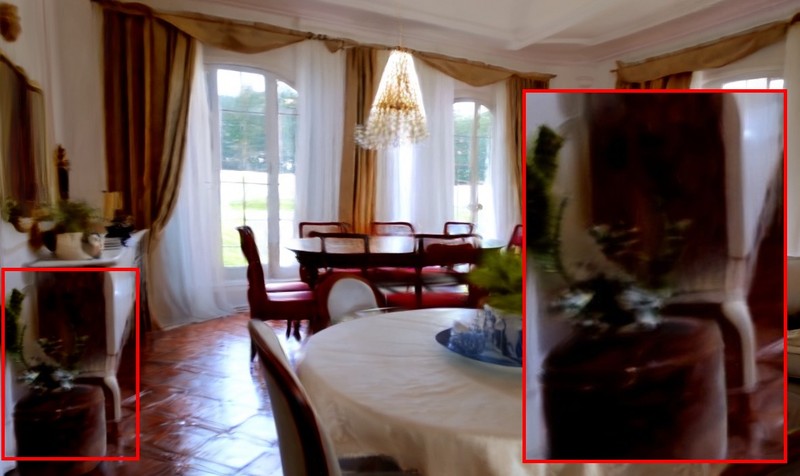} \\
        
    \end{tabular}
    \vspace{-4mm}
    \caption{\textbf{Single video 3D reconstruction.}
    We generate videos with SEVA \cite{zhou2025stable} (top), Gen3C \cite{ren2025gen3c} (mid), Wan \cite{wan2025wanopenadvancedlargescale} (bottom) and 3D reconstruct these frames.
    Inconsistencies in the generations lead to blurry textures for the baselines compared to the corresponding video, and to severe floating artifacts from novel views.
    In contrast, our method creates 3D consistent worlds with high fidelity beyond the generated perspectives.
    }
    \label{fig:qual_single2}
\end{figure*}

We compare our method against recent 3D reconstruction methods on the generated frames from multiple state-of-the-art video diffusion models.
Concretely, we generate single video sequences depicting various indoor/outdoor scenes and camera motions from text with Wan-2.2 \cite{wan2025wanopenadvancedlargescale} and with camera-control using ViewCrafter \cite{yu2024viewcrafter}, Gen3C \cite{ren2025gen3c}, Seva \cite{zhou2025stable}, and Voyager \cite{huang2025voyager}.
We additionally adopt the recent autoregressive world generators Genie3 \cite{genie3} and HY-WorldPlay \cite{worldplay2025}.

We reconstruct 3D scenes from $N {=} 50$ images sampled from these videos with various baselines. 
First, with 3DGS-MCMC \cite{kheradmand20243d} from a random initialization and optimizable cameras from DA3 \cite{lin2025depth}.
Second, we utilize the DA3 \cite{lin2025depth} points as 3DGS \cite{kerbl20233d} initialization and regularize their positions with a depth loss \cite{zhang2024rade} against the DA3 depth.
We refer to this as ``DA3'' \cite{lin2025depth} and refer to the suppl. material for more details.
We compare VGGT-X \cite{liu2025vggt}, which \textit{rigidly} aligns correspondences, akin to bundle adjustment, before optimizing 3DGS \cite{kerbl20233d}.
We refer to this as ``VGGT-X$^\dagger$'' when using DA3 \cite{lin2025depth} instead of VGGT \cite{liu2025vggt} predictions.

We showcase the reconstruction results in \Cref{fig:qual_single1,fig:qual_single2} and refer to the suppl. material for more samples and animated results.
Our method renders sharper and more detailed textures that are of comparable visual fidelity as the input video frames, just in a 3D consistent space.
In contrast, the baselines do not correct for the generative drift inherent in the generated frames.
This can lead to blurrier renderings from input poses (e.g., the brick textures in \Cref{fig:qual_single1}top).
It becomes especially noticeable from novel perspectives: the inconsistent image observations lead to floating artifacts in the 3D scenes that limit their explorability from arbitrary positions.
Our method optimizes the scenes in a \textit{non-rigid aware} fashion which leads to greatly improved viewpoint stability, i.e., we can explore the 3D worlds from novel perspectives while retaining the high visual fidelity of the video frames.
Our method effectively turns any video diffusion model into a 3D world generator that allows persistent, high quality, and real-time rendering.

\begin{figure*}
    \centering
    \setlength{\tabcolsep}{1pt}
    \renewcommand{\arraystretch}{0.5}

    \begin{tabular}{c c c c}
        {\fontsize{8}{9}\selectfont DA3~\cite{lin2025depth}} &
        {\fontsize{8}{9}\selectfont VGGT-X$^\dagger$~\cite{liu2025vggt}} &
        {\fontsize{8}{9}\selectfont Ours (before global)} &
        {\fontsize{8}{9}\selectfont Ours}\\

        \midrule

        \includegraphics[width=0.25\textwidth]{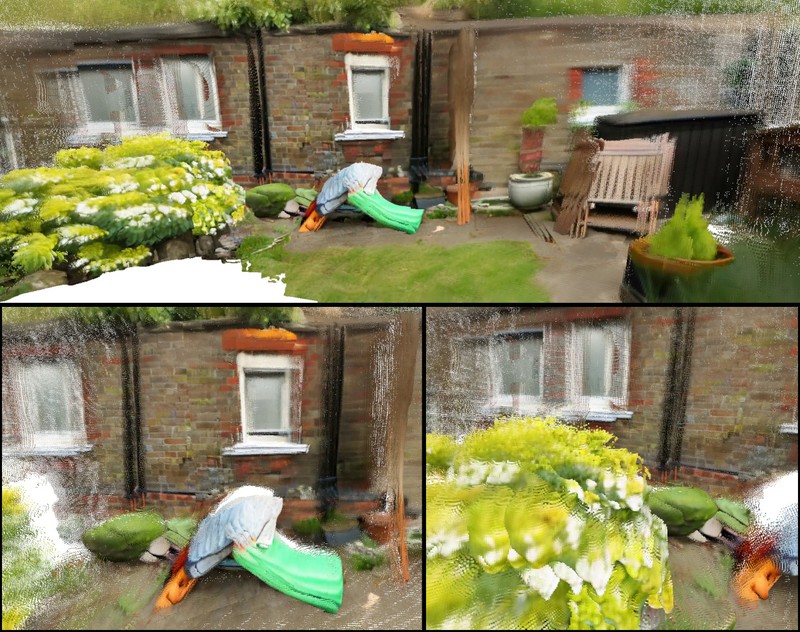}
        & \includegraphics[width=0.25\textwidth]{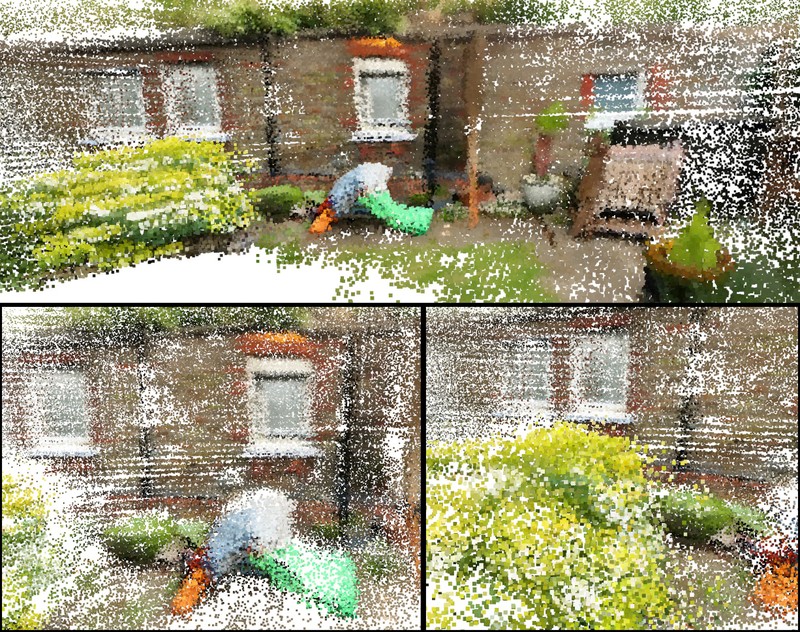}
        & \includegraphics[width=0.25\textwidth]{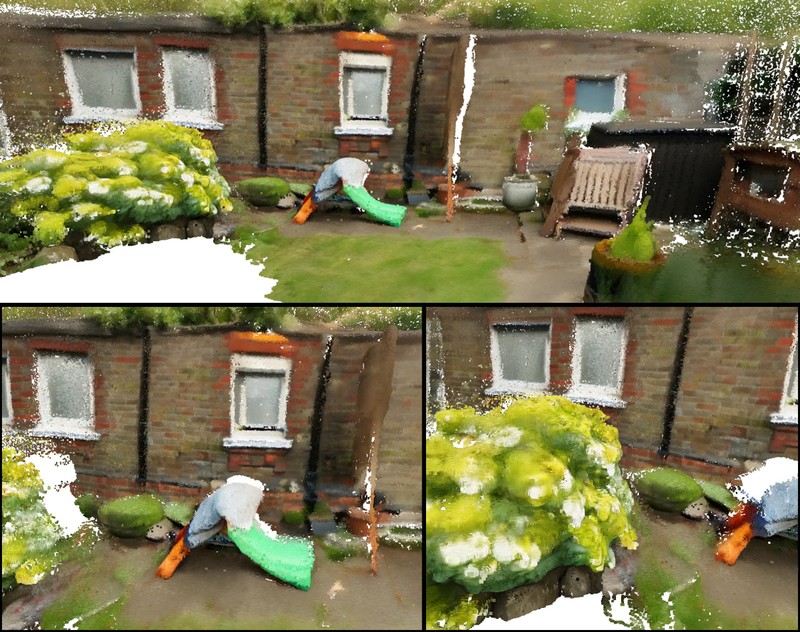}
        & \includegraphics[width=0.25\textwidth]{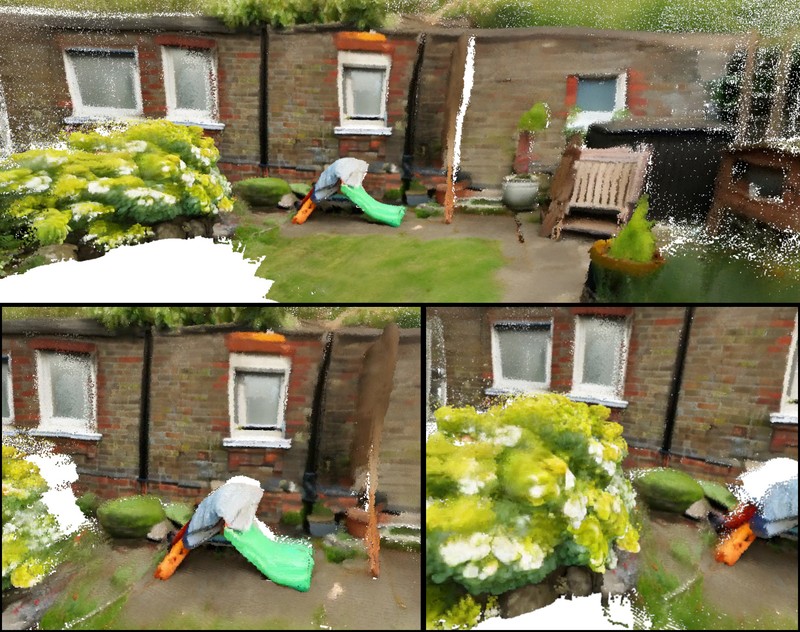} \\

        \includegraphics[width=0.25\textwidth]{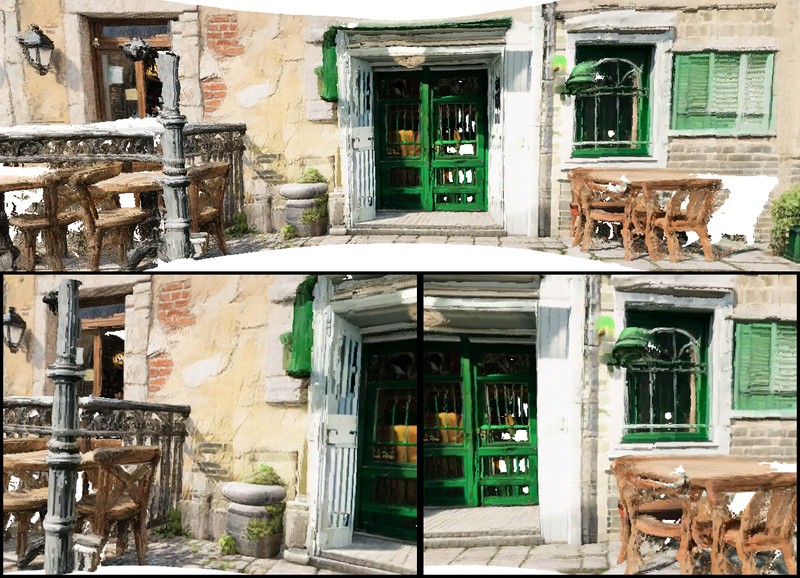}
        & \includegraphics[width=0.25\textwidth]{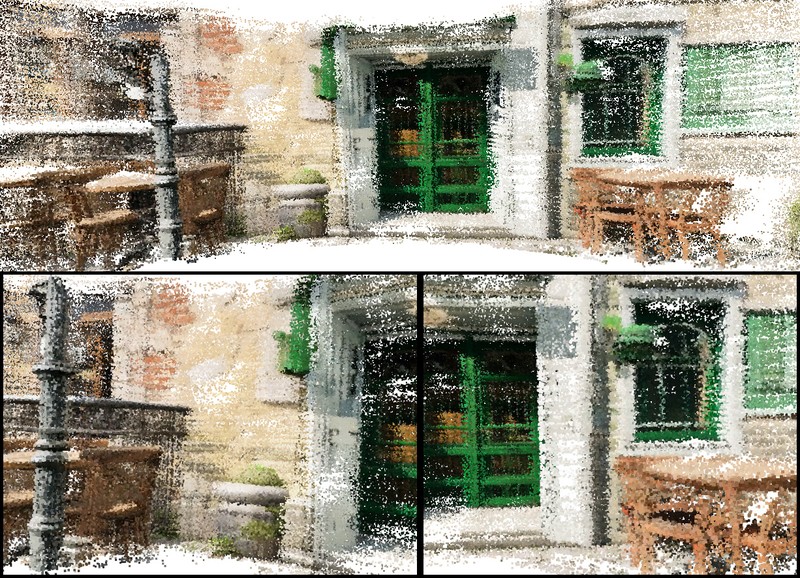}
        & \includegraphics[width=0.25\textwidth]{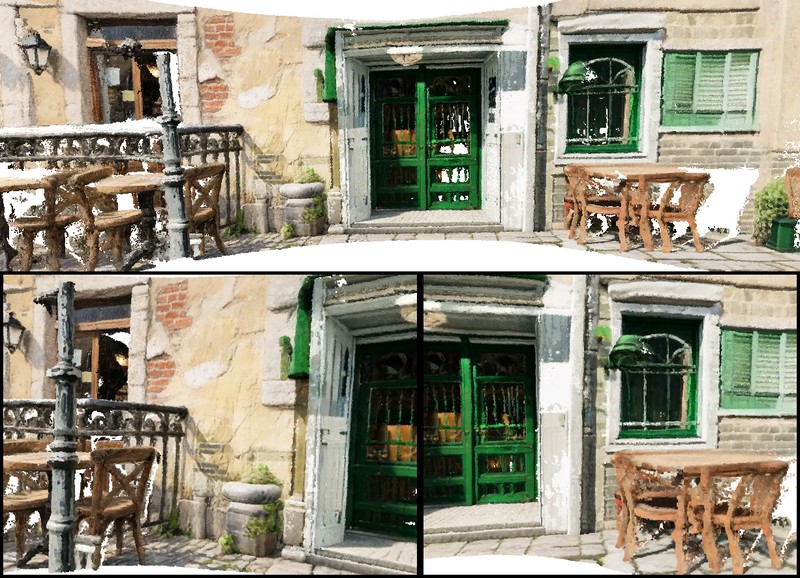}
        & \includegraphics[width=0.25\textwidth]{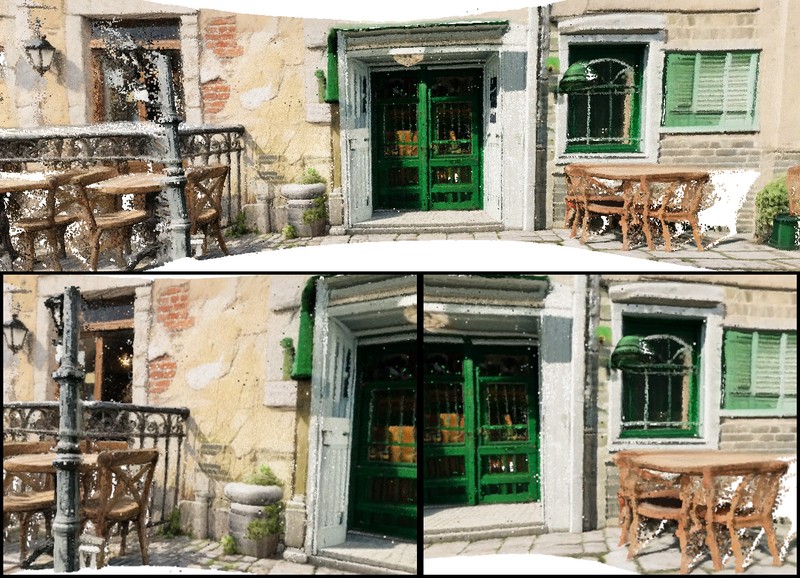} \\

    \end{tabular}
    \vspace{-4mm}
    \caption{\textbf{Pointcloud reconstructions.}
    We compare the quality of the reconstructed pointclouds that are used as initialization for Gaussian Splatting \cite{kerbl20233d} optimization in the subsequent stages for each method.
    Our approach achieves the highest alignment and compelling textures for individual objects with no overlap of multiple surfaces.
    }
    \label{fig:qual_pcl}
\end{figure*}

We confirm this in \Cref{tab:quant_single} by calculating consistency and fidelity metrics following the established WorldScore benchmark \cite{duan2025worldscore}, averaged across all VDMs per reconstruction method.
Our method obtains the highest consistency scores, which underlines its ability to reconstruct 3D worlds from inconsistent views.
The rendering quality is the highest among baselines and comparable to that of the input videos (CLIP-IQA+: 47.39, CLIP Aesthetic: 39.04).
Additionally, we compare the 3D pointclouds that are used as initialization for all applicable baselines in \Cref{fig:qual_pcl}.
In comparison to the DA3 \cite{lin2025depth} predictions, our results greatly improve the alignment of individual surfaces (e.g., unifies multiple windows and slides in the garden scene).
VGGT-X \cite{liu2025vggt} obtains sparser pointclouds and less-precise alignment since they do not model non-rigid deformations.

\subsection{Ablations}

The key components of our method are the non-rigid alignment (\cref{eq:fwd-def,eq:bkwd-def}) and its usage in the 3D reconstruction.
We ablate these decisions in \Cref{fig:qual_ablation} and \Cref{tab:quant_ablation} (averaged across four SEVA \cite{zhou2025stable} scenes) and additionally compare dropping the correspondence loss, global optimization, and point filtering.

This highlights that both only rigid alignment and vanilla 3DGS optimization (``no inv'') are not enough to obtain aligned surfaces and consistent renderings.
Dropping the correspondences fails to align surfaces beyond the matching distance $d_{\max}$, which depending on the amount of generative drift can be noticeable in the pointcloud and renderings.
No point filtering or global optimization adversely impact the texture quality of reconstructed and rendered surfaces.
In contrast, our full approach obtains the highest rendering quality and alignment.

\begin{table}[b]
\centering
\small
\setlength{\tabcolsep}{5pt}
\begin{tabular}{l | cc | cc}
\toprule
        \multirow{2}{*}{Method} & \multicolumn{2}{c}{Consistency} & \multicolumn{2}{c}{Fidelity}\\
                        \cmidrule(l{2pt}r{2pt}){2-3} \cmidrule(l{2pt}r{2pt}){4-5}
 & 3D & Photometric & CLIP-IQA+ & CLIP Aesthetic \\
\midrule
only rigid & 76.98 & 64.93 & 11.43 & 49.12 \\
no inv & 78.28 & 55.16 & 15.81 & 46.39 \\
no filt & 72.58 & 64.88 & 55.17 & 51.05 \\
no corr & 65.62 & 74.76 & 44.81 & 46.27 \\
no global & 63.90 & 77.85 & 49.69 & 51.61 \\
Ours & \textbf{79.79} & \textbf{86.88} & \textbf{55.30} & \textbf{52.40} \\
\bottomrule
\end{tabular}
\caption{\textbf{Ablation study.}
3D reconstruction from only rigid alignment, without correspondences, or without backward deformation (``no inv'') negatively impacts 3D consistency and image quality.
No point filter or global optimization decrease image fidelity and consistency.
Our full method achieves the highest consistency and fidelity.
}
\label{tab:quant_ablation}
\end{table}

\begin{figure*}
    \centering
    \setlength{\tabcolsep}{1pt}
    \renewcommand{\arraystretch}{1.1}

    \begin{tabular}{c | c c c c c c c}

        &
        {\fontsize{8}{9}\selectfont Video Frames} &
        {\fontsize{8}{9}\selectfont only rigid} &
        {\fontsize{8}{9}\selectfont no inv} &
        {\fontsize{8}{9}\selectfont no corr} &
        {\fontsize{8}{9}\selectfont no filt} &
        {\fontsize{8}{9}\selectfont no global} &
        {\fontsize{8}{9}\selectfont Ours} \\

        \midrule

        \rotatebox{90}{\fontsize{8}{9}\selectfont Input Pose}
        &\includegraphics[width=0.14\textwidth]{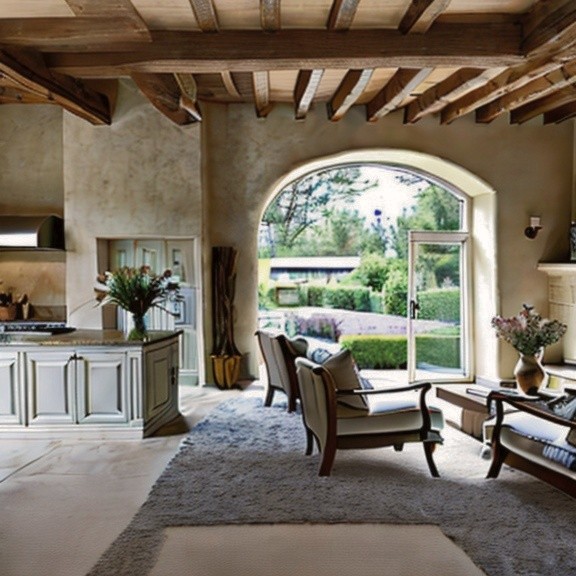}
        & \includegraphics[width=0.14\textwidth]{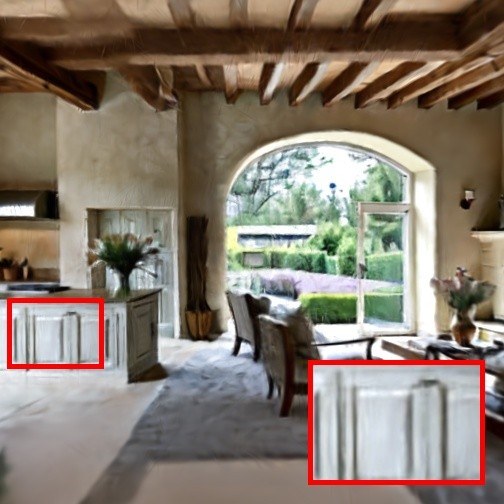}
        & \includegraphics[width=0.14\textwidth]{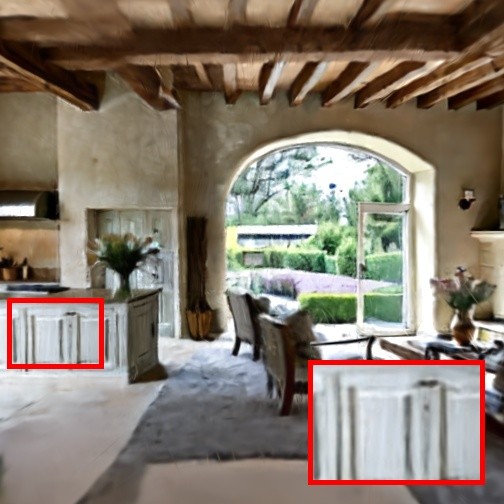}
        & \includegraphics[width=0.14\textwidth]{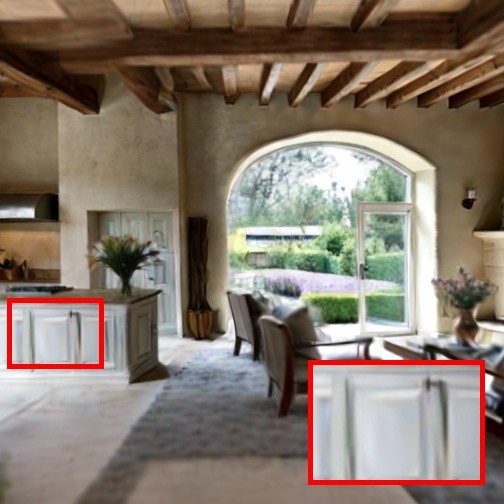}
        & \includegraphics[width=0.14\textwidth]{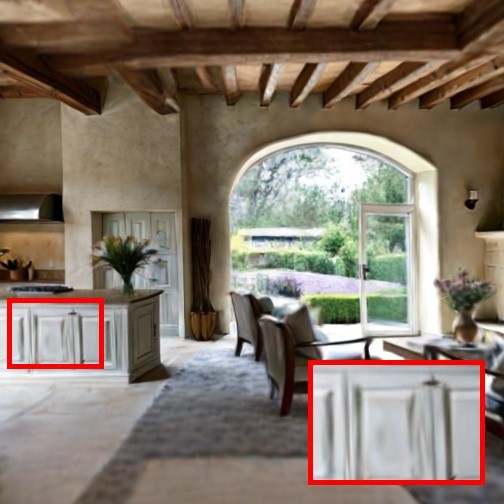}
        & \includegraphics[width=0.14\textwidth]{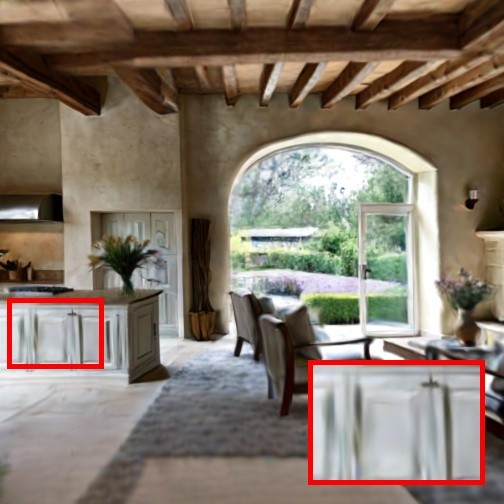}
        & \includegraphics[width=0.14\textwidth]{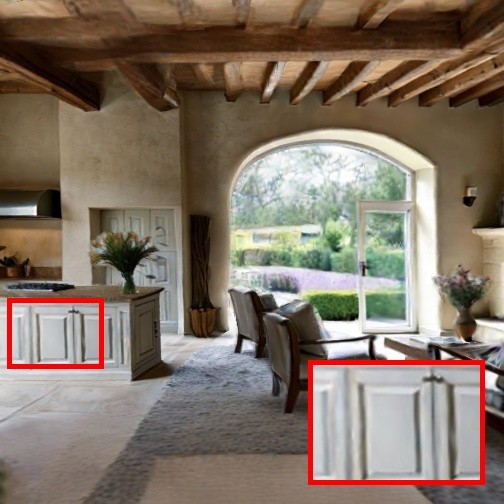} \\

        \rotatebox{90}{\fontsize{8}{9}\selectfont Novel Pose}
        & \includegraphics[width=0.14\textwidth]{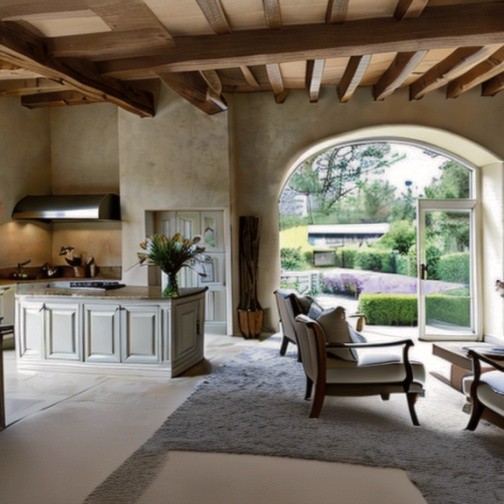}
        & \includegraphics[width=0.14\textwidth]{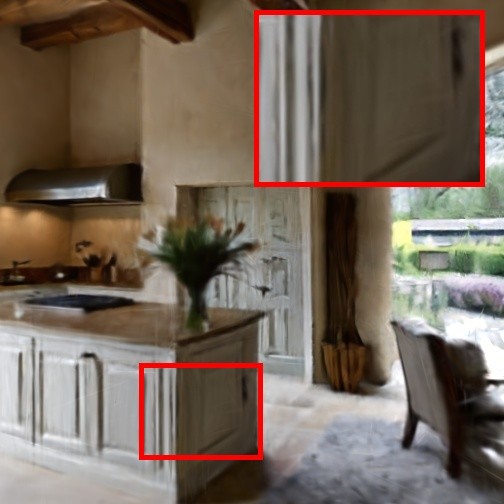}
        & \includegraphics[width=0.14\textwidth]{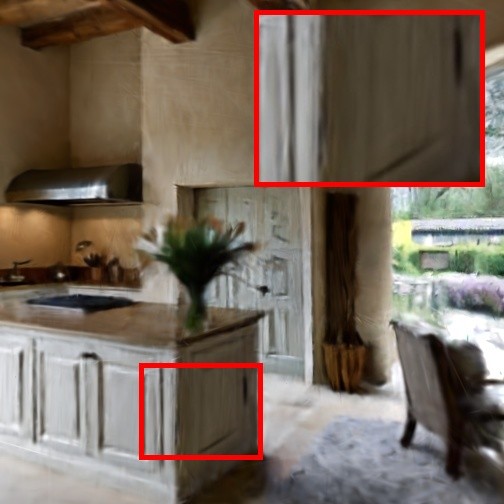}
        & \includegraphics[width=0.14\textwidth]{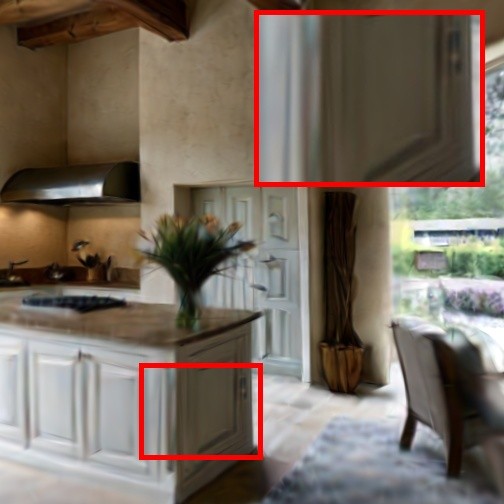}
        & \includegraphics[width=0.14\textwidth]{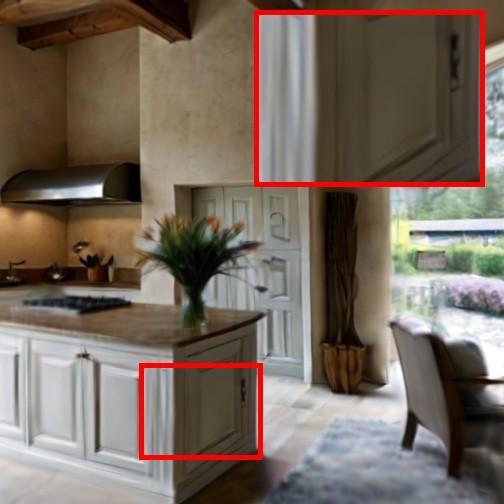}
        & \includegraphics[width=0.14\textwidth]{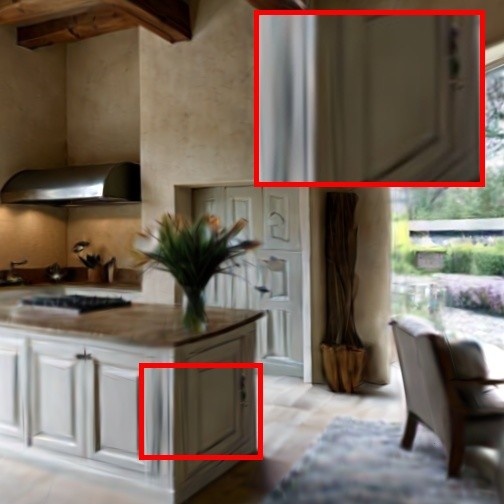}
        & \includegraphics[width=0.14\textwidth]{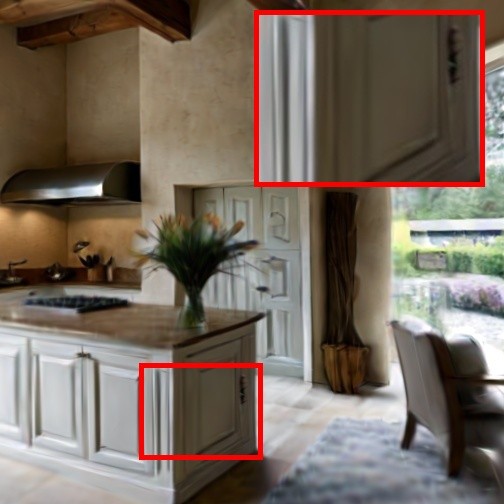} \\

    \end{tabular}

    \vspace{-1mm}

    \begin{tabular}{c | c c c c c c}

        \midrule
        \vspace{-0.5mm}

        &
        {\fontsize{8}{9}\selectfont DA3 \cite{lin2025depth}} &
        {\fontsize{8}{9}\selectfont only rigid} &
        {\fontsize{8}{9}\selectfont no corr} &
        {\fontsize{8}{9}\selectfont no filt} &
        {\fontsize{8}{9}\selectfont no global} &
        {\fontsize{8}{9}\selectfont Ours} \\

        \midrule

        \rotatebox{90}{\fontsize{8}{9}\selectfont Pointcloud}  
        & \includegraphics[width=0.163\textwidth]{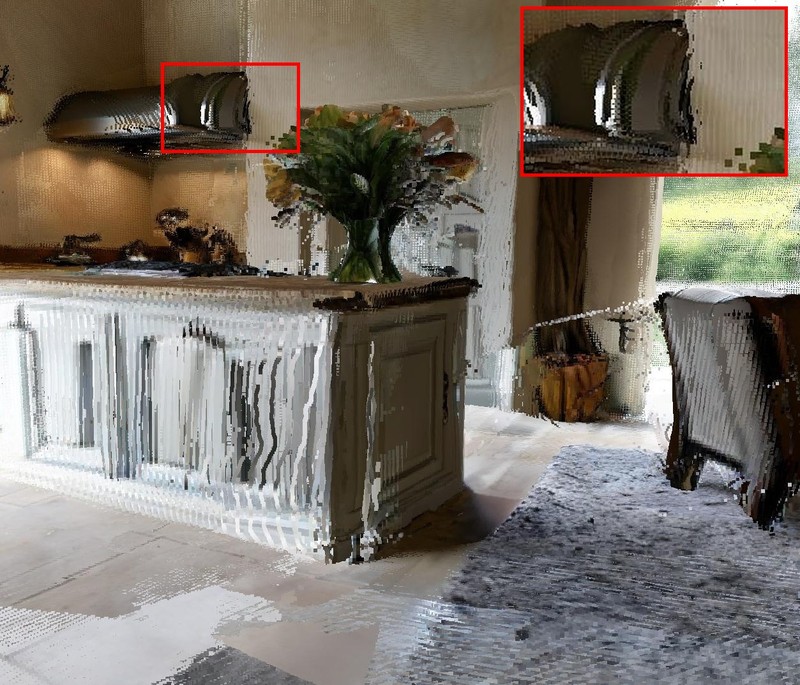}
        & \includegraphics[width=0.163\textwidth]{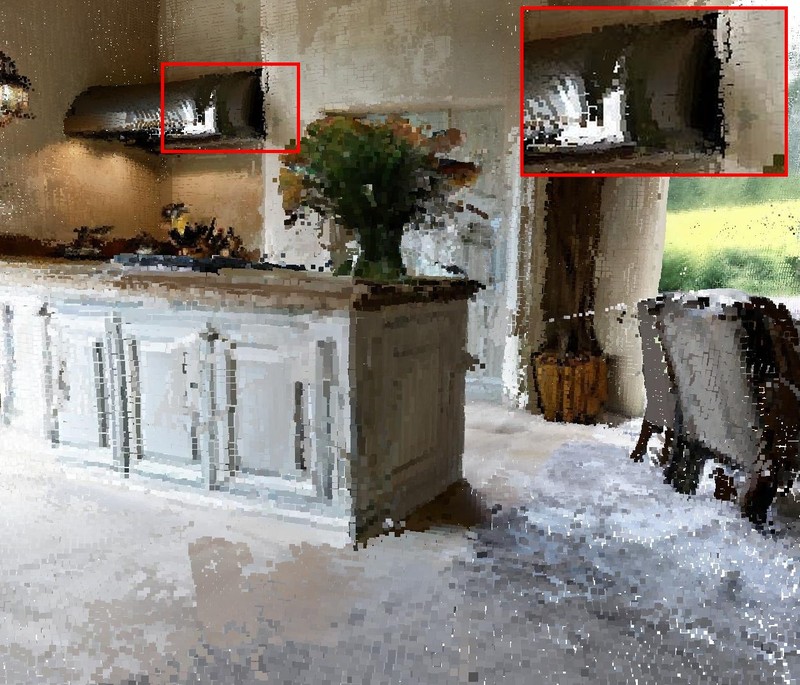}
        & \includegraphics[width=0.163\textwidth]{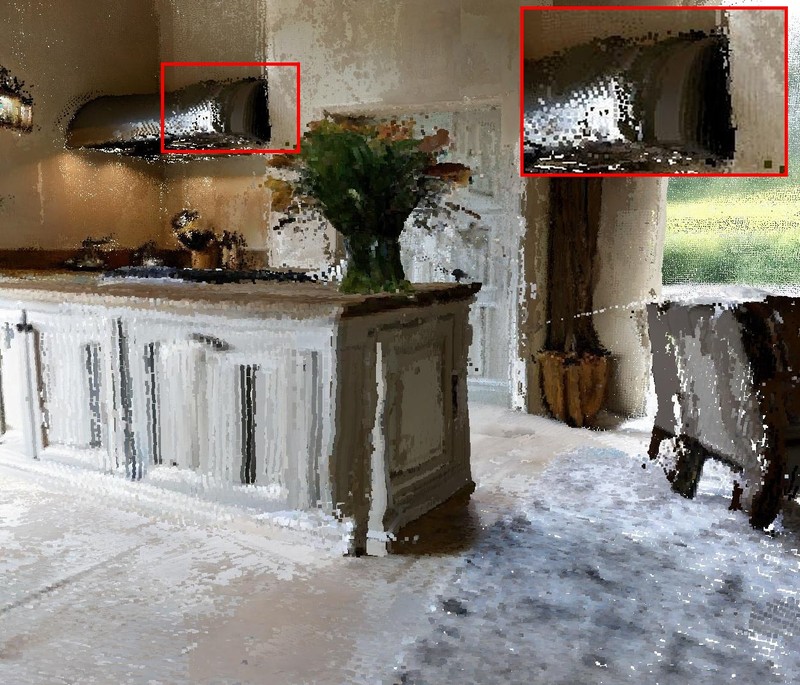}
        & \includegraphics[width=0.163\textwidth]{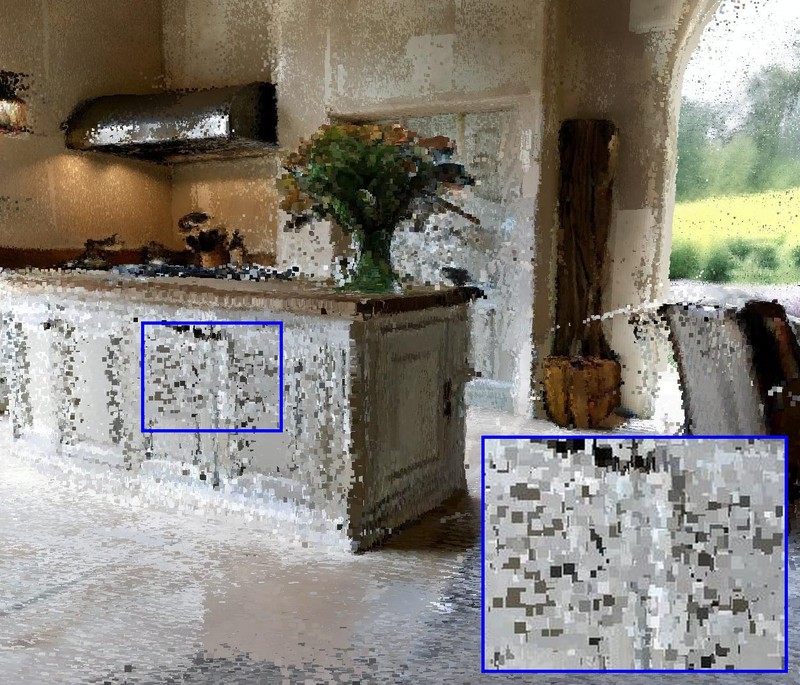}
        & \includegraphics[width=0.163\textwidth]{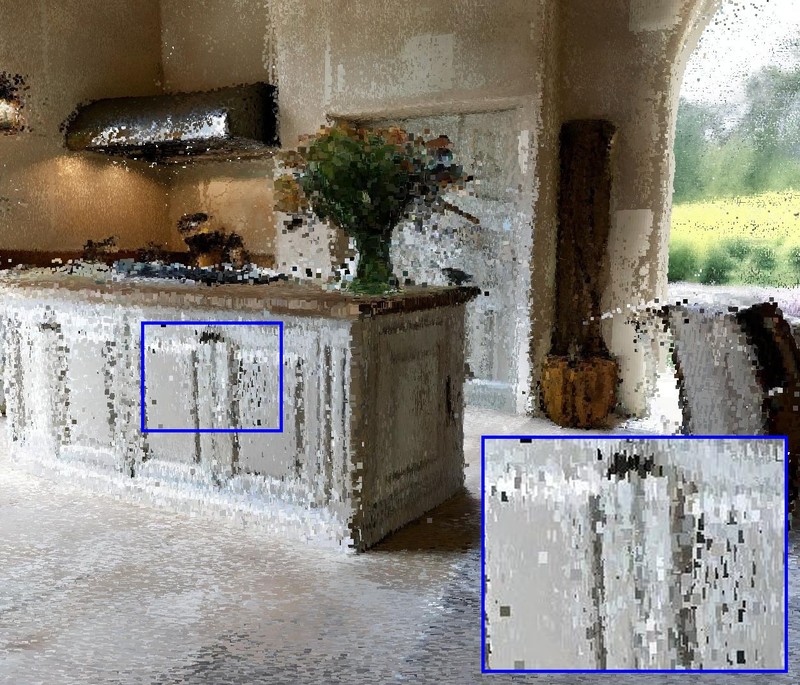}
        & \includegraphics[width=0.163\textwidth]{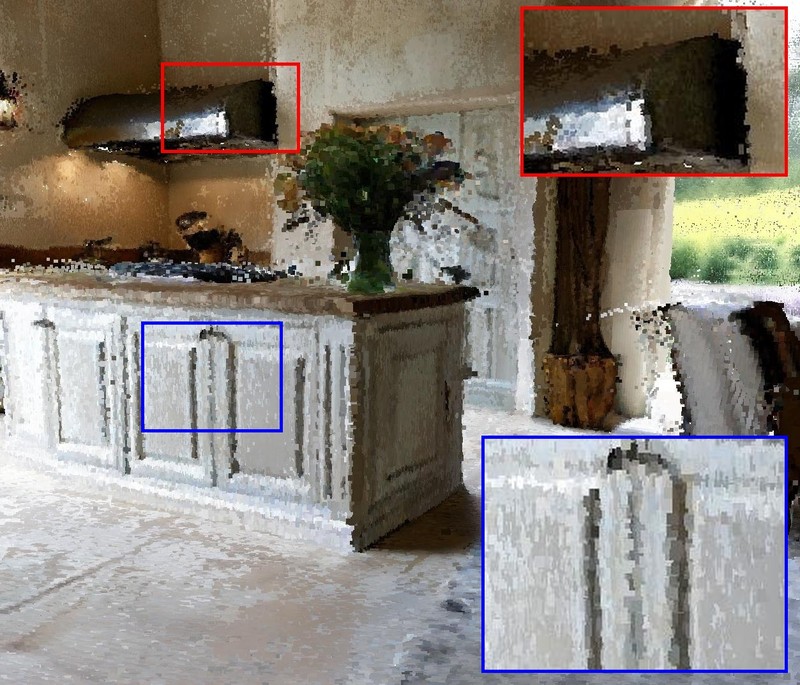} \\

    \end{tabular}

    \vspace{-4mm}
    \caption{\textbf{Ablation.} 
    We compare our full method on rendering quality of input views (top) and novel views (mid), as well as on pointcloud reconstruction quality (bottom).
    The DA3 \cite{lin2025depth} prediction contains misalignments that cannot be resolved by only rigid alignment or without correspondences.
    Removing the color filtering and global optimization steps adversely impact texture quality.
    Removing our novel inverse deformation loss reduces sharpness.
    In contrast, our method obtains flat/sharp surfaces which better constrain the 3D reconstruction, yielding high quality rendering results.
    }
    \label{fig:qual_ablation}
\end{figure*}

\subsection{Large-Scale World Generation}

\newcommand{\croppimg}[1]{%
  \adjustbox{
    width=0.245\textwidth,
    trim=0 {.1\height} 0 {.16\height},
    clip
  }{\includegraphics{#1}}%
}

\newcommand{\croppimgg}[1]{%
  \adjustbox{
    width=0.245\textwidth,
    trim=0 {.05\height} 0 {.21\height},
    clip
  }{\includegraphics{#1}}%
}

\newcommand{\croppimggg}[1]{%
  \adjustbox{
    width=0.33\textwidth,
    trim=0 {.1\height} 0 {.16\height},
    clip
  }{\includegraphics{#1}}%
}

\newcommand{\croppimgggg}[1]{%
  \adjustbox{
    width=0.33\textwidth,
    trim=0 {.13\height} 0 {.13\height},
    clip
  }{\includegraphics{#1}}%
}

\begin{figure*}
    \centering
    \setlength{\tabcolsep}{1pt}
    \renewcommand{\arraystretch}{1.1}

    \resizebox{0.97\textwidth}{!}{
    \begin{minipage}{\textwidth}
    \begin{tabular}{c | c | c c c}

        &
        {\fontsize{8}{9}\selectfont Video Frames} &
        {\fontsize{8}{9}\selectfont WorldExplorer \cite{schneider_hoellein_2025_worldexplorer}} &
        {\fontsize{8}{9}\selectfont VGGT-X$^\dagger$ \cite{liu2025vggt}} &
        {\fontsize{8}{9}\selectfont Ours} \\

        \midrule

        \rotatebox{90}{\fontsize{8}{9}\selectfont Input Pose}
        & \croppimg{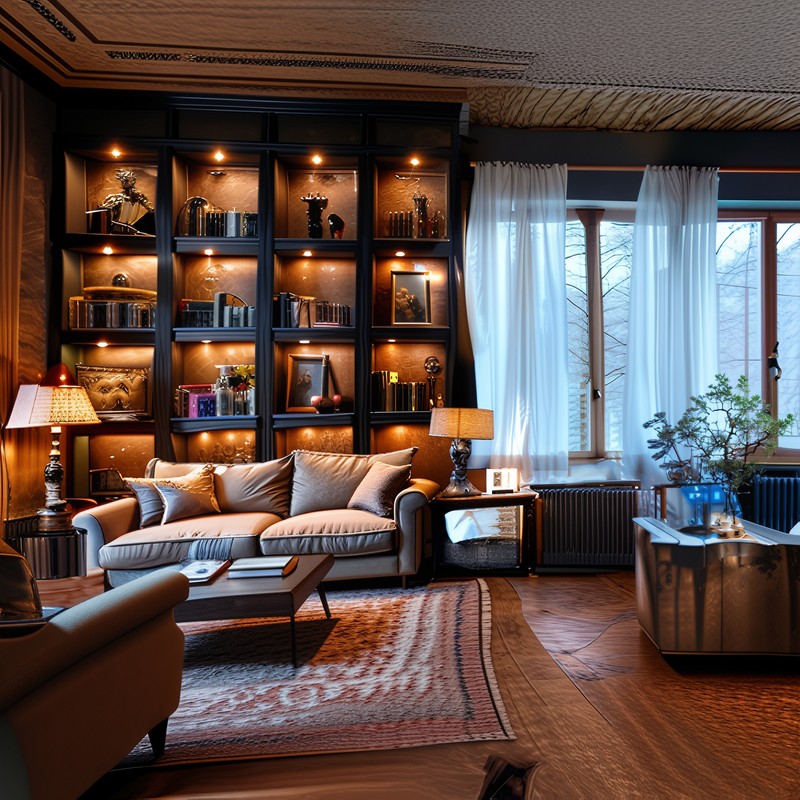}
        & \croppimg{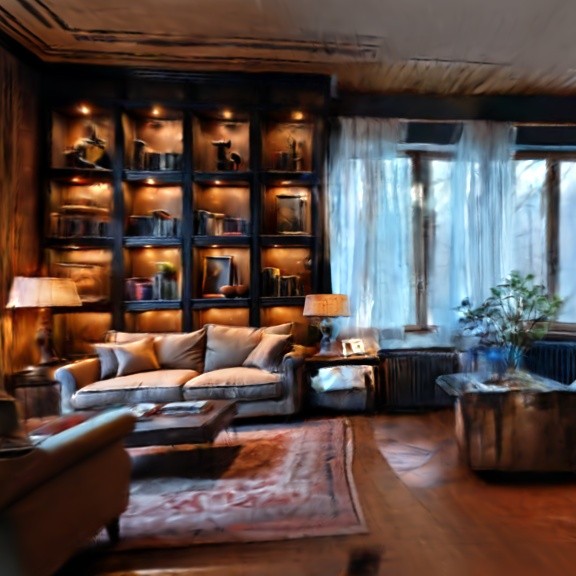}
        & \croppimg{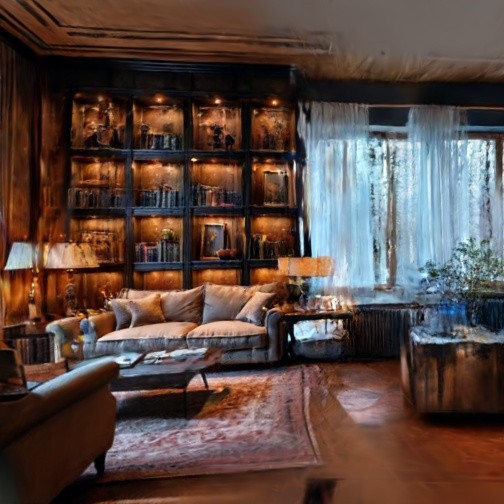}
        & \croppimg{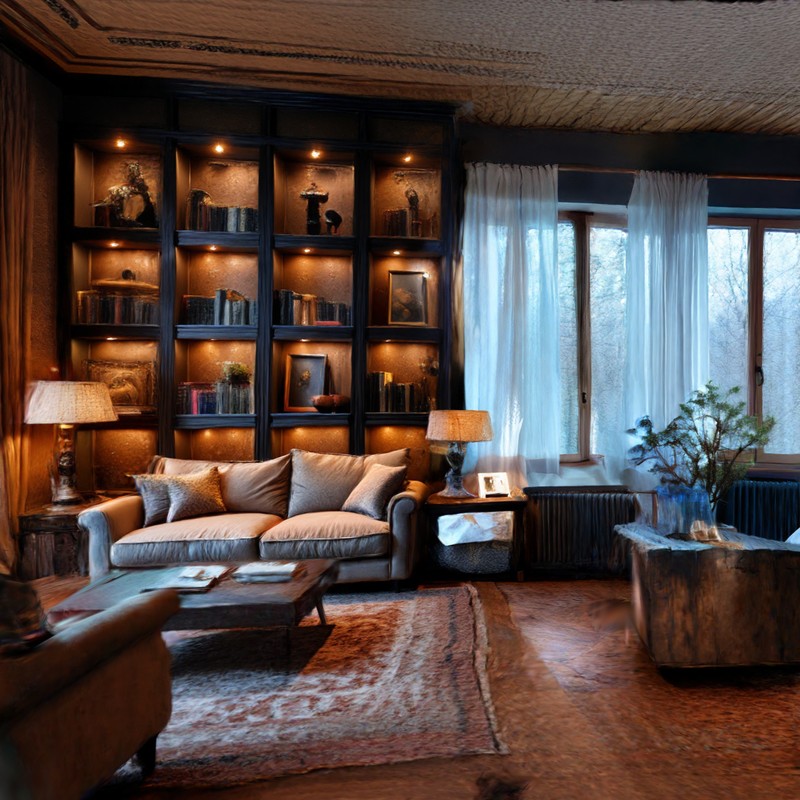} \\

        \rotatebox{90}{\fontsize{8}{9}\selectfont Input Pose}
        & \croppimgg{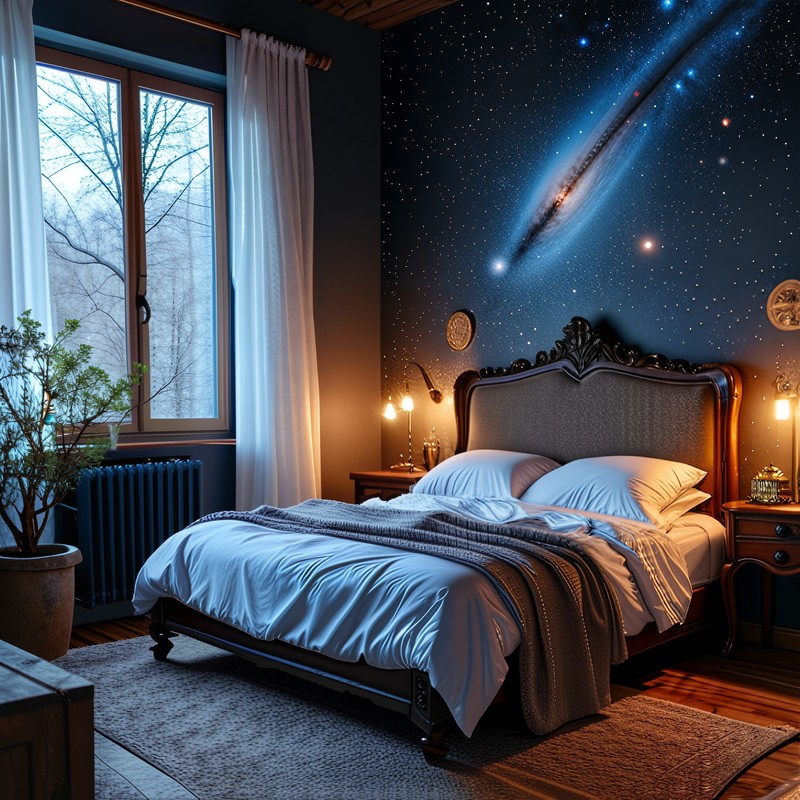}
        & \croppimgg{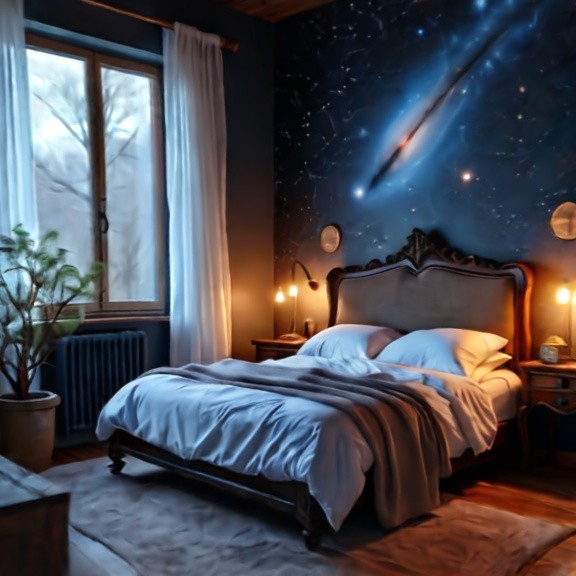}
        & \croppimgg{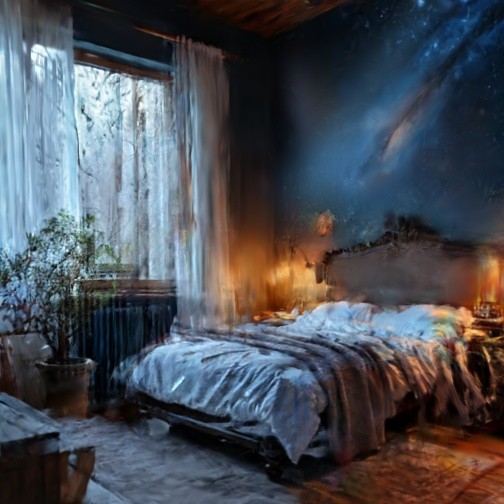}
        & \croppimgg{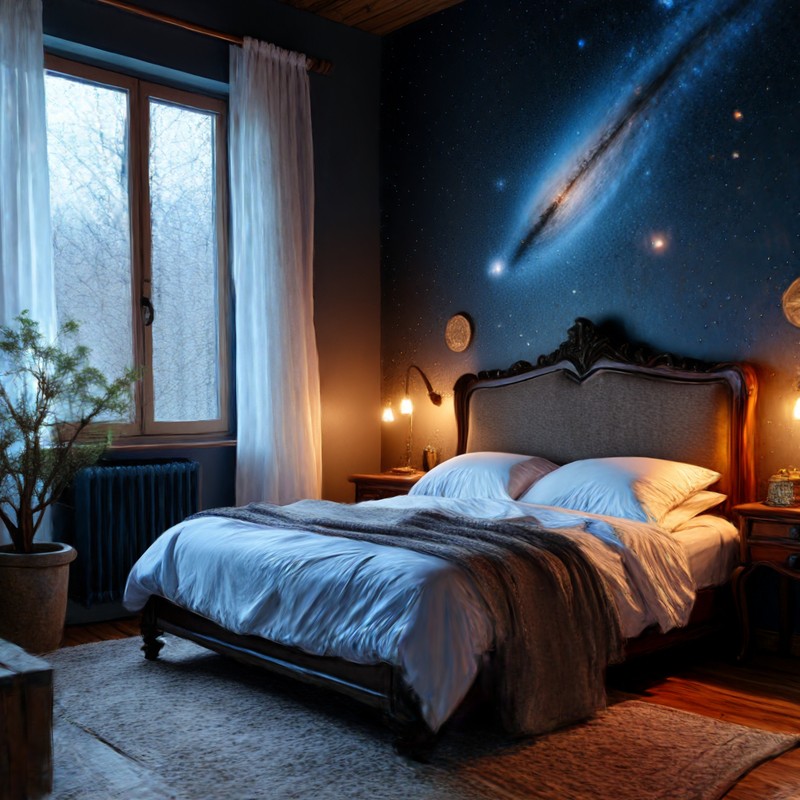} \\

    \end{tabular}

    \vspace{-1mm}

    \begin{tabular}{c | c c c}

        \midrule
        \vspace{-0.5mm}

        &
        {\fontsize{8}{9}\selectfont WorldExplorer \cite{schneider_hoellein_2025_worldexplorer}} &
        {\fontsize{8}{9}\selectfont VGGT-X$^\dagger$ \cite{liu2025vggt}} &
        {\fontsize{8}{9}\selectfont Ours} \\

        \midrule

        \rotatebox{90}{\fontsize{8}{9}\selectfont Novel Views}  
        & \croppimggg{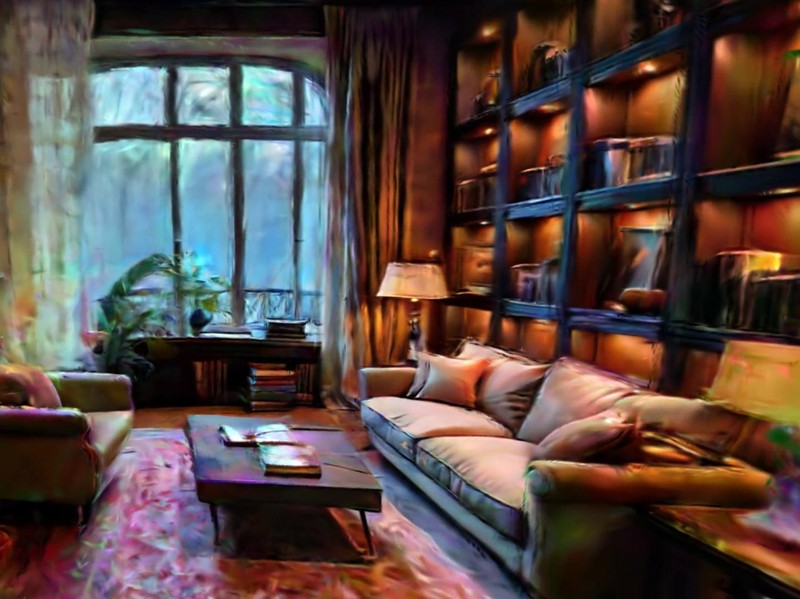}
        & \croppimggg{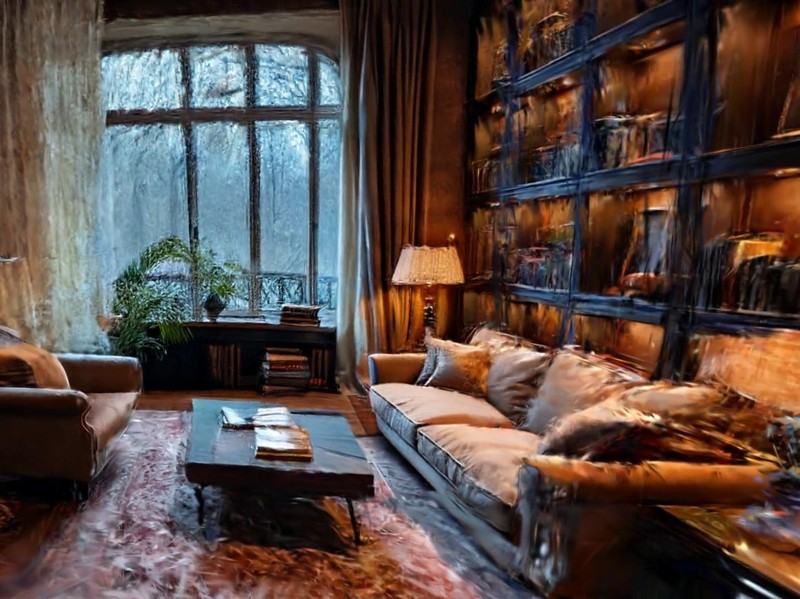}
        & \croppimggg{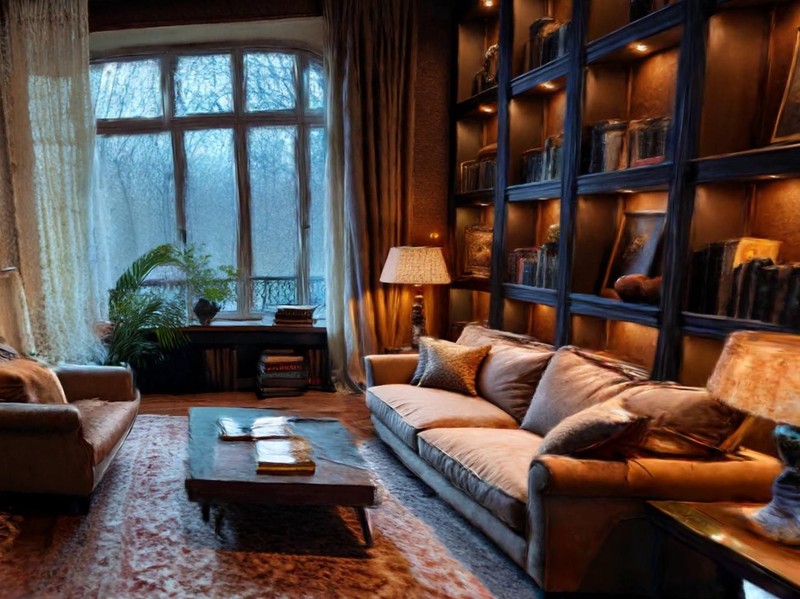} \\

    \end{tabular}
    \end{minipage}
    }

    \vspace{-4mm}
    \caption{\textbf{Large-scale 3D reconstructions.} 
    We compare rendering quality on input poses and novel views for entire 360 degree explorable scenes.
    Our method creates 3D consistent worlds with a high rendering fidelity far beyond the training views.
    Please see the supplementary material for animated flythroughs.
    }
    \label{fig:qual_multi2}
\end{figure*}

A single generated video is limited in the amount of scene exploration it can show.
Recent works exploit VDMs autoregressively to generate multiple sequences that depict entire 360 degree scenes \cite{chen2025flexworld, schneider_hoellein_2025_worldexplorer, genie3, hyworld2025}.
\Cref{fig:teaser,fig:qual_multi2} compare against WorldExplorer \cite{schneider_hoellein_2025_worldexplorer} and VGGT-X \cite{liu2025vggt} by generating and then reconstructing up to 32 video sequences with SEVA \cite{zhou2025stable} via the progressive scene expansion strategy of \cite{schneider_hoellein_2025_worldexplorer}.
The generative drift grows larger across multiple videos and thus the 3D reconstructions suffer from increased consistency problems.
While the baselines can depict complete and diverse worlds, the amount of exploration they enable is still limited (i.e., novel views far beyond the training poses suffer from extensive floating artifacts).
In contrast, our worlds remain consistent and retain a high rendering fidelity, even from extreme novel perspectives.

\subsection{Limitations}

Our method reconstructs 3D scenes from inconsistent generated views, however, some limitations remain (see supplementary material).
First, we tackle the problem of \textit{generative drift} in VDMs, i.e., existing objects move geometrically inconsistent.
However, VDMs can also suffer from hallucinations (e.g., revisiting previous scene areas creates novel or removes existing objects, changes their texture, etc.).
Adopting robust reconstruction mechanisms to detect such outlier frames could resolve these problems \cite{sabour2023robustnerf}.
Second, our method proposes a per-scene alignment stage for generated frames.
While this lightweight approach can turn any video diffusion model into a reliable world generator, it still entails additional computational cost before reconstruction.
One could explore finetuning VDMs with our alignments acting as gradient signals for the generators.

\section{Conclusion}
\label{sec:conclusion}

We have presented a method to reconstruct 3D worlds from the inconsistent generated views of video diffusion models \cite{genie3, wan2025wanopenadvancedlargescale, zhou2025stable, ren2025gen3c, yu2024viewcrafter, huang2025voyager}.
We exploit geometric foundation models \cite{lin2025depth} to reveal the generative drift inherent in the generations (\Cref{subsec:init}).
Our tailored non-rigid alignment creates sharp and flat surfaces that resolve these inconsistencies (\Cref{subsec:align}).
Finally, we propose a novel \textit{non-rigid aware} Gaussian Splatting \cite{kerbl20233d} optimization that optimizes a canonical 3D scene representation from the distorted frame observations (\Cref{subsec:3dgs-baking}).
Overall, this leads to sharp and high-quality scene renderings from novel poses, without floating artifacts that otherwise occur due to the inaccurate frames.
We believe this demonstrates the exciting ability to turn any video diffusion model into a reliable 3D world generator, which will open up further research avenues and make world generation more practical across many real-world applications.

\section*{Acknowledgements}
This project was funded by the ERC Consolidator Grant Gen3D (101171131).
We also thank Angela Dai for the video voice-over.

\bibliographystyle{splncs04}
\bibliography{main}

\appendix
\renewcommand{\thefigure}{A\arabic{figure}}
\setcounter{figure}{0}

\section{Supplemental Video}
Please watch our attached video for a comprehensive evaluation of the proposed method.
We include rendered videos of multiple generated scenes from novel trajectories, that showcase the quality of both single video 3D reconstructions and our large-scale scenes.
This highlights our methods abilities to create consistent 3D worlds from inconsistent views with high quality renderings and fewer floating artifacts than the baseline methods.

\section{Extended Baseline Discussion}

\subsection{Comparison Against Dynamic Reconstruction}

\newcommand{\cropimg}[1]{%
  \adjustbox{
    width=0.245\textwidth,
    trim=0 {.13\height} 0 {.13\height},
    clip
  }{\includegraphics{#1}}%
}

\newcommand{\cropimgg}[1]{%
  \adjustbox{
    width=0.245\textwidth,
    trim=0 {.05\height} 0 {.21\height},
    clip
  }{\includegraphics{#1}}%
}

\newcommand{\cropimggg}[1]{%
  \adjustbox{
    width=0.245\textwidth,
    trim=0 {.13\height} 0 {.13\height},
    clip
  }{\includegraphics{#1}}%
}

\newcommand{\cropimgggg}[1]{%
  \adjustbox{
    width=0.245\textwidth,
    trim=0 {.26\height} 0 0,
    clip
  }{\includegraphics{#1}}%
}

\begin{figure*}
    \centering
    \setlength{\tabcolsep}{1pt}
    \renewcommand{\arraystretch}{1.1}

    \begin{tabular}{c | c c c}

        {\fontsize{8}{9}\selectfont Video Frames} &
        {\fontsize{8}{9}\selectfont D-3DGS (dyn) \cite{yang2024deformable}} &
        {\fontsize{8}{9}\selectfont D-3DGS (static) \cite{yang2024deformable}} &
        {\fontsize{8}{9}\selectfont Ours} \\

        \midrule

        \cropimg{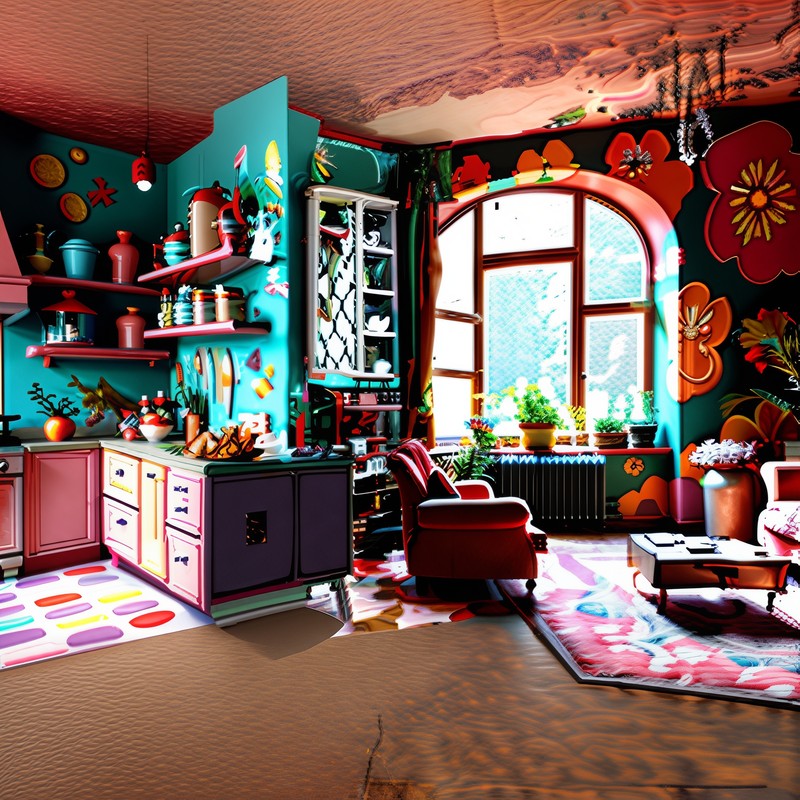}
        & \cropimg{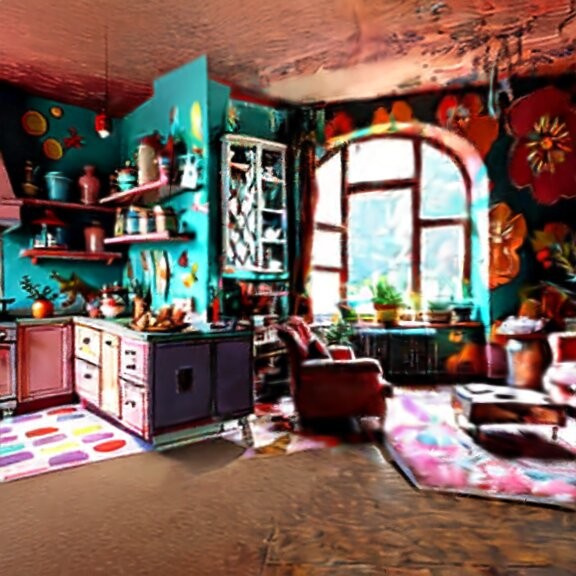}
        & \cropimg{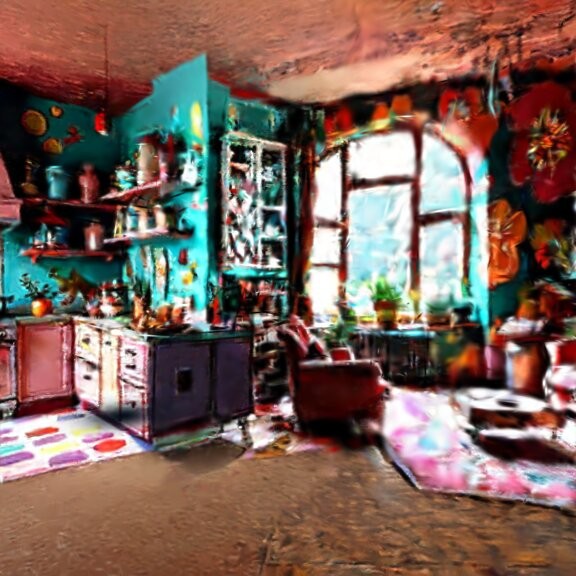}
        & \cropimg{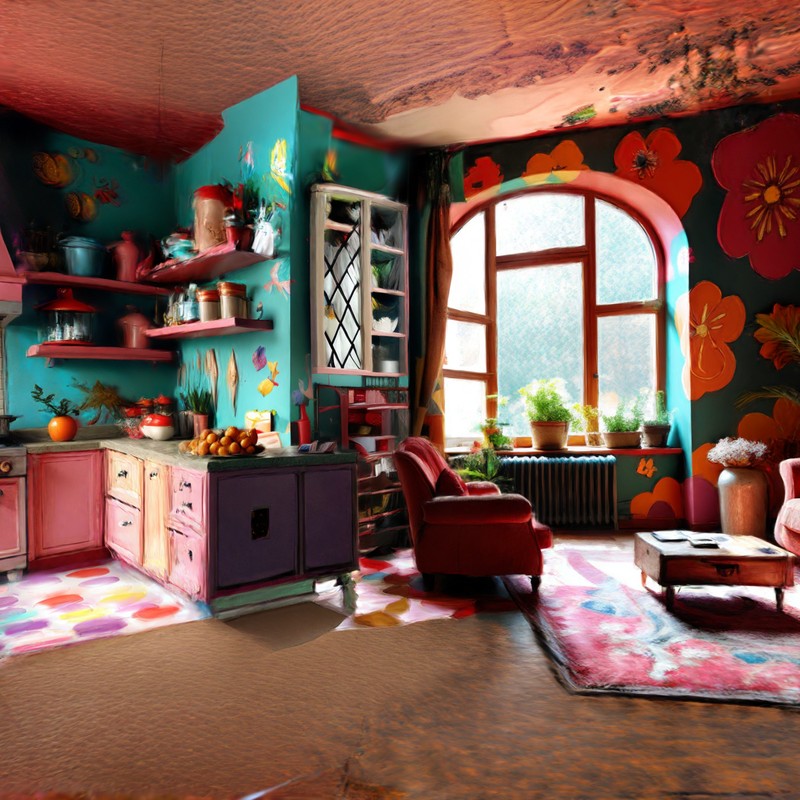} \\

        \cropimggg{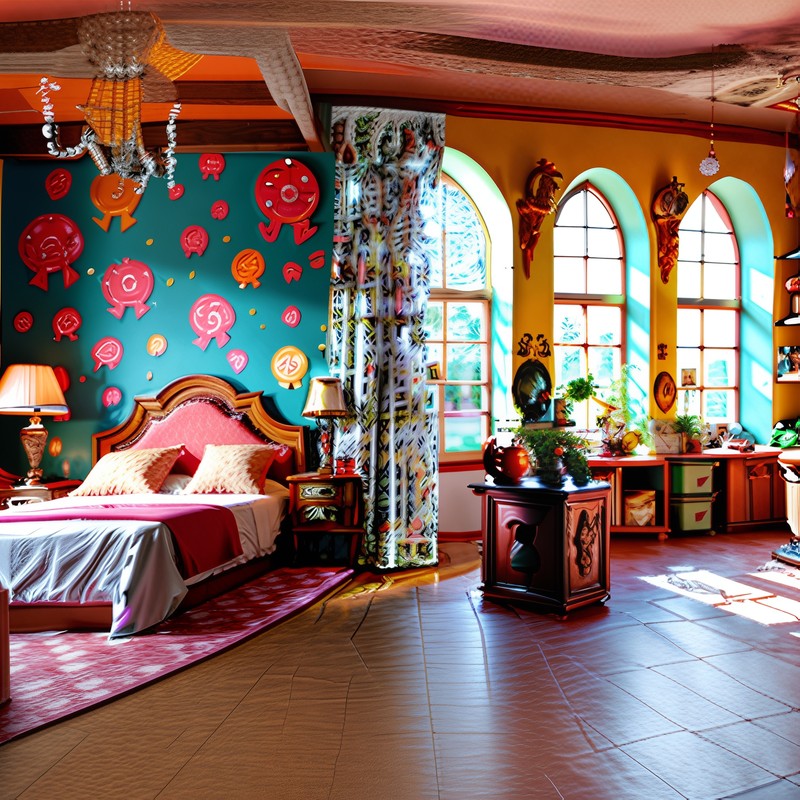}
        & \cropimggg{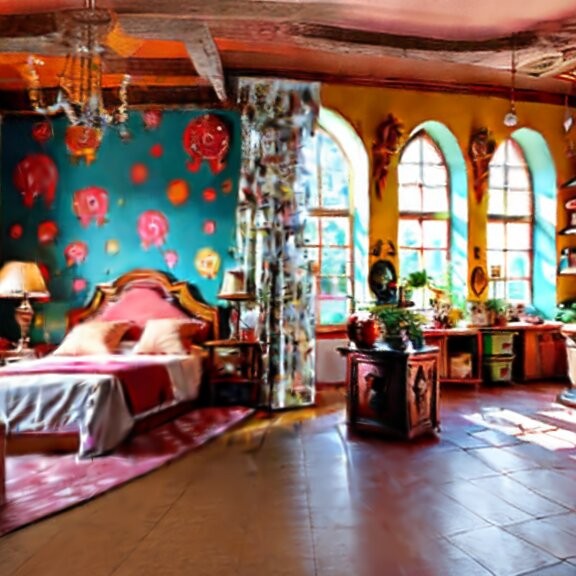}
        & \cropimggg{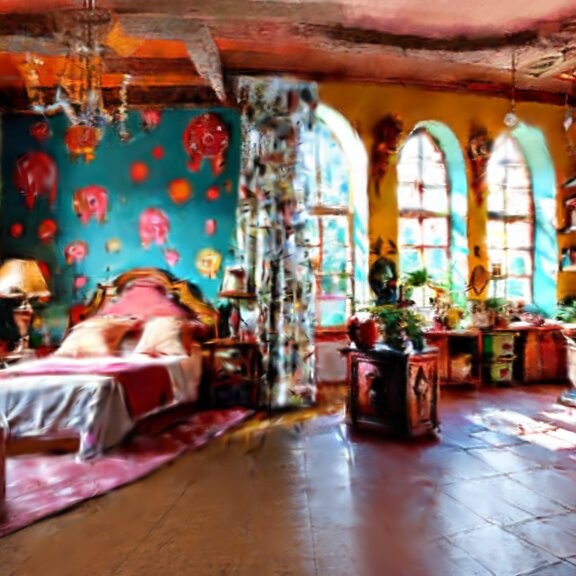}
        & \cropimggg{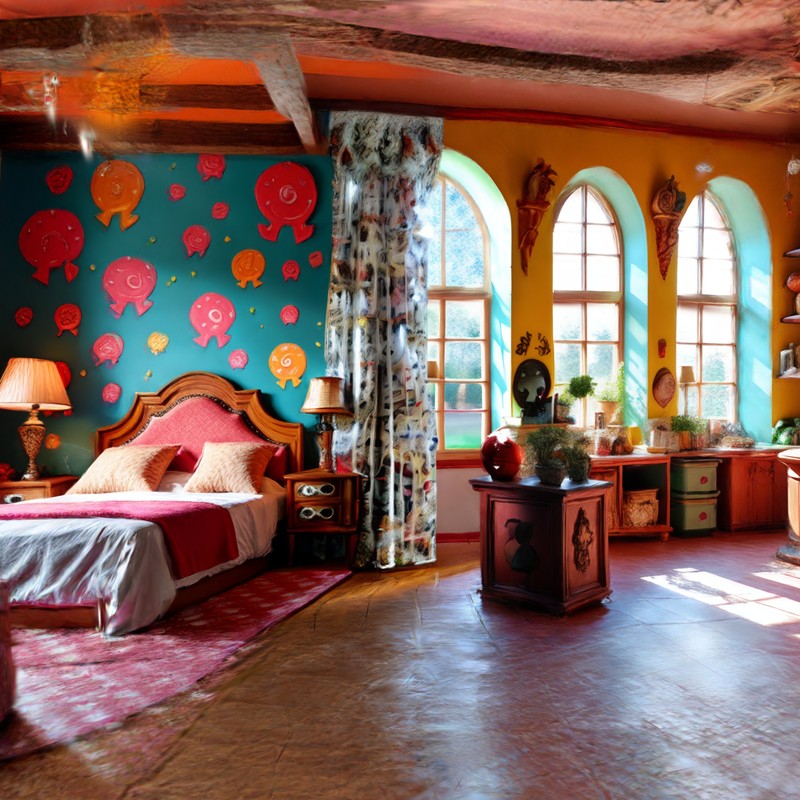} \\

        \cropimgggg{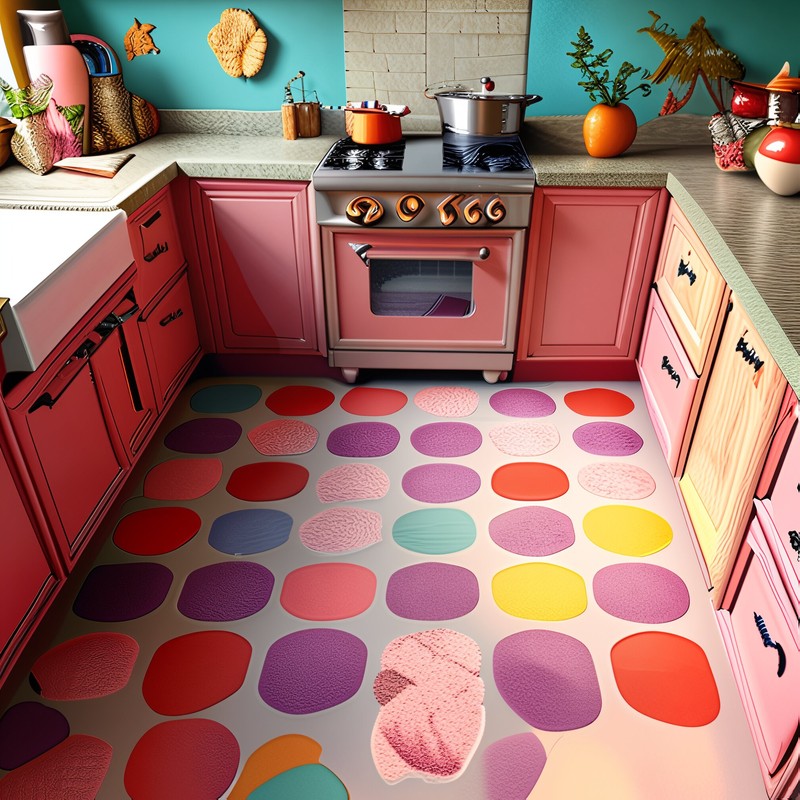}
        & \cropimgggg{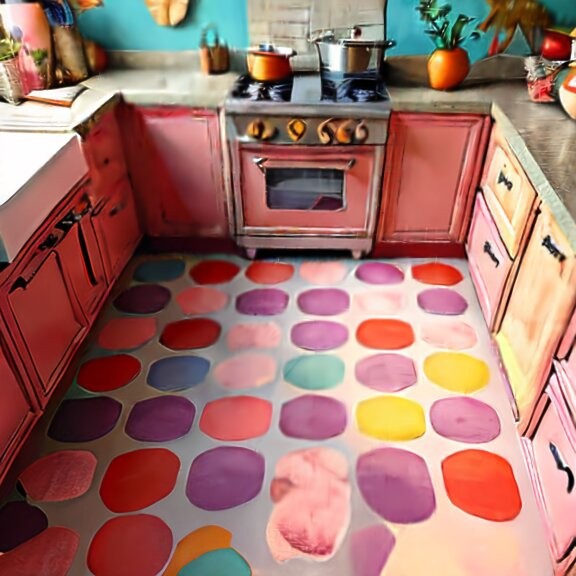}
        & \cropimgggg{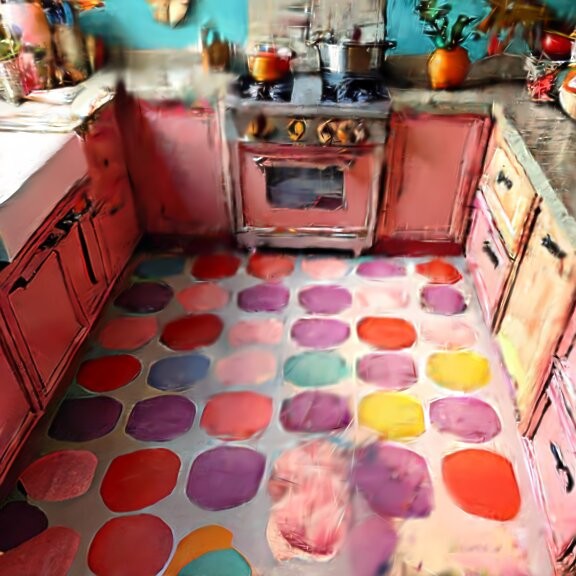}
        & \cropimgggg{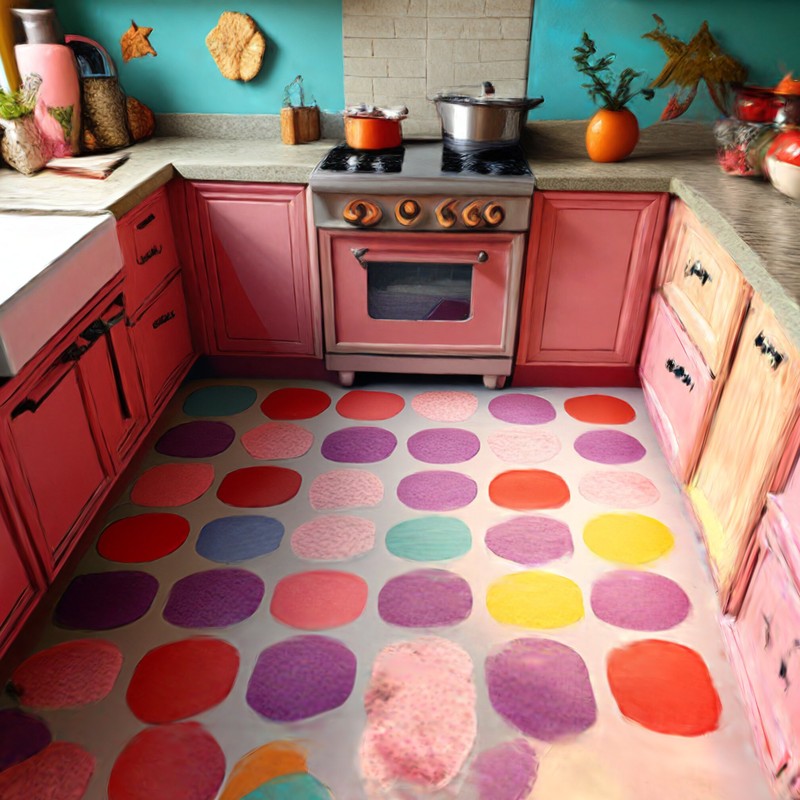} \\

    \end{tabular}

    \vspace{-4mm}
    \caption{\textbf{Comparison against dynamic reconstruction.}
    We compare against ``Deformable 3DGS'' \cite{yang2024deformable} on reconstructing inconsistent generated video sequences.
    Since they do not model an aligned canonical space, their renderings both with deformation (dyn) and without (static) suffer from floating artifacts and decreased image quality.
    }
    \label{fig:suppl_deformable}
\end{figure*}

After the geometry alignment we have obtained $\mathcal{P}[0]$, which represents an aligned 3D scene, as well as the non-rigid \textit{forward} deformations $\{\mathbf{R}_i, \mathbf{t}_i, \mathcal{F}_{\theta_i}\}_{i=1}^N$ from each frame to the first frame (see Section~3.2).
This is similar to a 4D scene reconstruction that is obtained in the dynamic reconstruction literature \cite{yunus2024recent, park2021nerfiesdeformableneuralradiance, newcombe2015dynamicfusion, wu20244d}.
We compare against ``Deformable 3DGS'' \cite{yang2024deformable} as an exemplary method of that line of work.
Concretely, they assign a timestep to each viewpoint and a learnable deformation network then transforms all canonical Gaussians to the dynamic scene state at that timestep.
At inference, this allows to replay the dynamically moving scene from any timestep at any observed (or novel) view.
These methods define a canonical scene state from which the Gaussians are warped either as the first timestep or as a learned embedding.
However, when applying this principle to the dynamic reconstruction of inconsistent generated video sequences, we observe severe limitations (see \Cref{fig:suppl_deformable}).
The key observation is that these methods \textit{do not model} an aligned canonical space.
Instead, the gradients from the rendering loss incentify a strong deformation network to explain the training views from \textit{any canonical scene constellation}.
This is sufficient to \textit{reproduce} the inconsistent views when applying the deformation network at inference time (marked as ``dyn'').
However, when rendering the scene at a fixed timestep (to simulate a static and unified 3D reconstruction from inconsistent views; marked as ``static''), severe degradation becomes visible through floating artifacts.
In contrast, our method first optimizes for a geometric alignment, that explicitly models the canonical space.
Using this as a basis for deformable Gaussian Splatting optimization leads to consistent and high-quality 3D reconstructions.

\subsection{Discussion Of DA3 Feedforward 3DGS Prediction}
\begin{figure*}[t]
    \centering
    \setlength{\tabcolsep}{1pt}
    \renewcommand{\arraystretch}{1.1}

    \begin{tabular}{c | c | c c c c}
        {\fontsize{8}{9}\selectfont Video Frames} &
        &
        {\fontsize{8}{9}\selectfont DA3~\cite{lin2025depth}} &
        {\fontsize{8}{9}\selectfont DA3 (FF)~\cite{lin2025depth}} &
        {\fontsize{8}{9}\selectfont SHARP~\cite{mescheder2026sharpmonocularviewsynthesis}} &
        {\fontsize{8}{9}\selectfont Ours} \\

        \midrule

        \includegraphics[width=0.19\textwidth]{img/single_video/gen3c/086/input_poses_cropped/input_video_frame_00001_center.jpeg}
        & \raisebox{0.0\height}{\rotatebox{90}{\fontsize{8}{9}\selectfont Input}}
        & \includegraphics[width=0.19\textwidth]{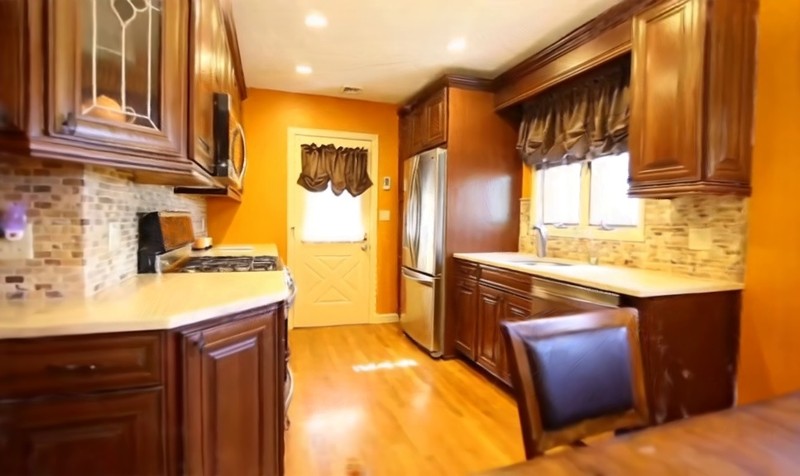}
        & \includegraphics[width=0.19\textwidth]{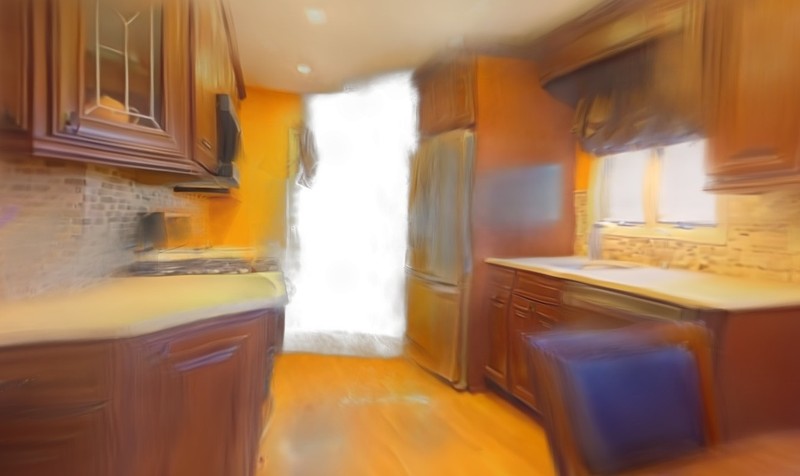}
        & \includegraphics[width=0.19\textwidth]{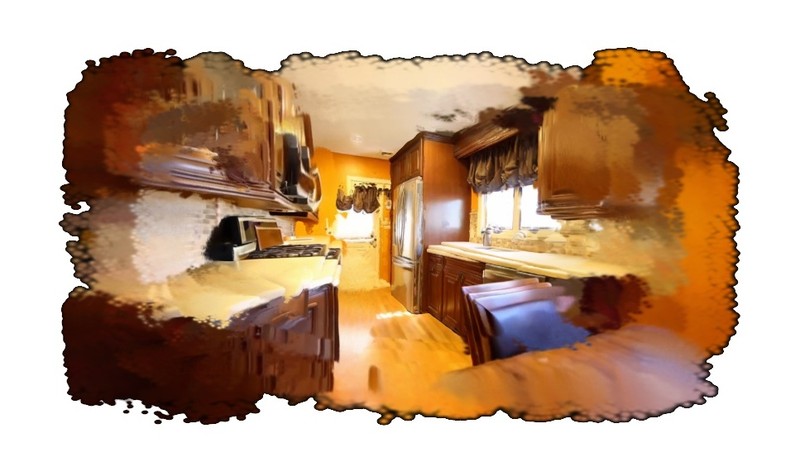}
        & \includegraphics[width=0.19\textwidth]{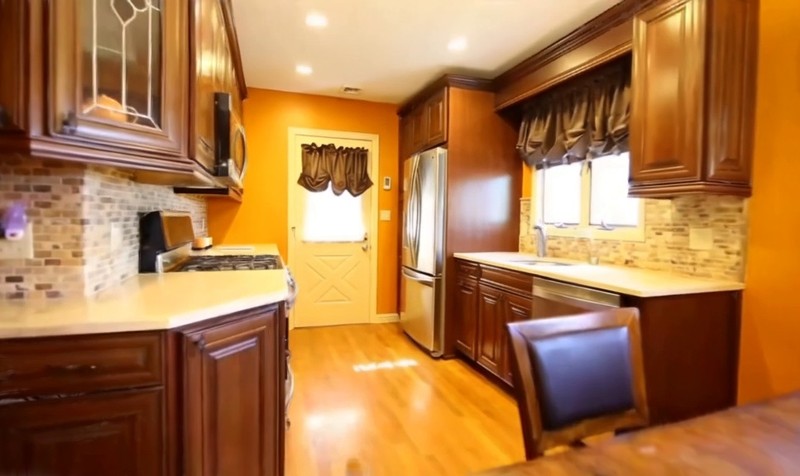} \\

        \includegraphics[width=0.19\textwidth]{img/single_video/gen3c/086/input_poses_cropped/input_video_frame_00001_center.jpeg}
        & \multirow{2}{*}{\rotatebox{90}{\fontsize{8}{9}\selectfont Novel Views}}
        
        & \includegraphics[width=0.19\textwidth]{img/single_video_zoom_insets/gen3c/086/00060_da3_3dgs_white_bg_left_zoom.jpeg}
        & \includegraphics[width=0.19\textwidth]{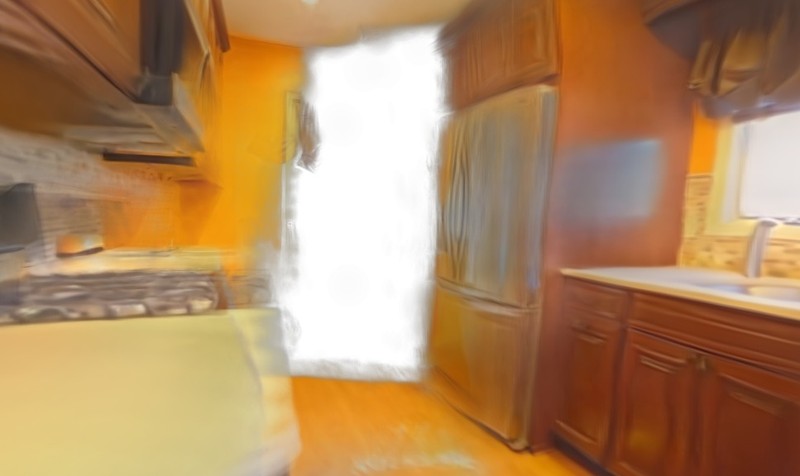}
        & \includegraphics[width=0.19\textwidth]{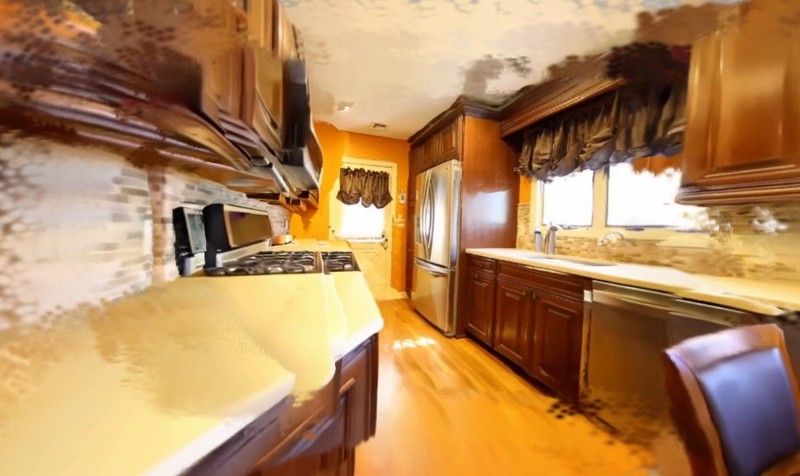}
        & \includegraphics[width=0.19\textwidth]{img/single_video_zoom_insets/gen3c/086/00060_nvs_white_bg_left_zoom.jpeg} \\

        \includegraphics[width=0.19\textwidth]{img/single_video/gen3c/086/input_poses_cropped/input_video_frame_00055_center.jpeg}
        &
        & \includegraphics[width=0.19\textwidth]{img/single_video_zoom_insets/gen3c/086/00071_da3_3dgs_white_bg_left_zoom.jpeg}
        & \includegraphics[width=0.19\textwidth]{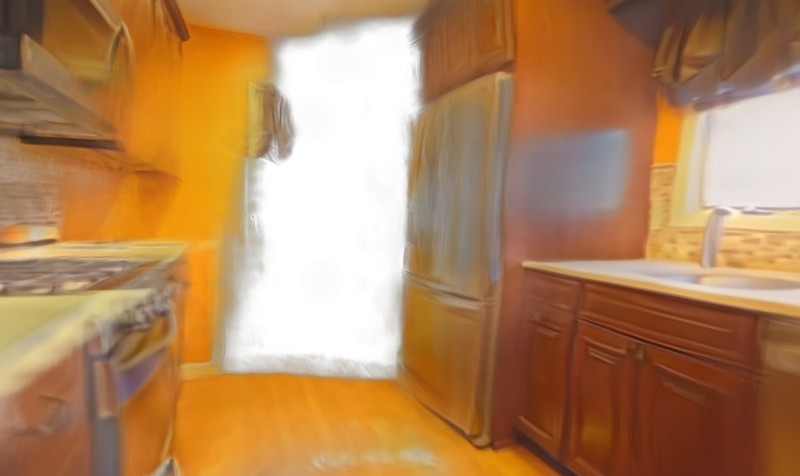}
        & \includegraphics[width=0.19\textwidth]{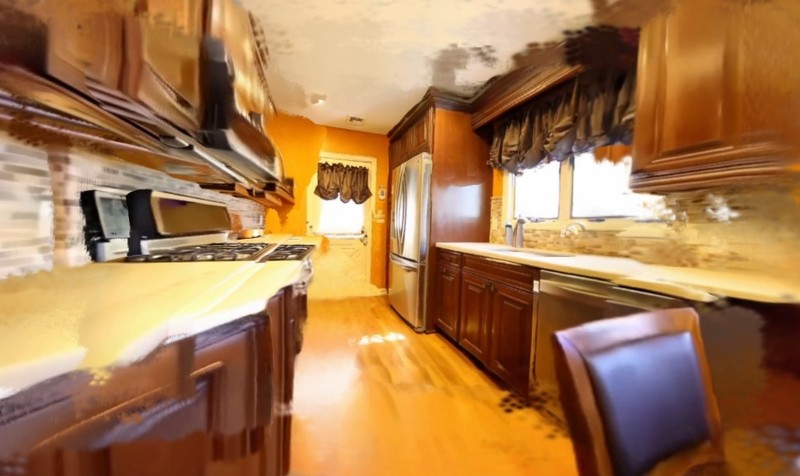}
        & \includegraphics[width=0.19\textwidth]{img/single_video_zoom_insets/gen3c/086/00071_nvs_white_bg_left_zoom.jpeg} \\
        
    \end{tabular}
    \vspace{-4mm}
    \caption{\textbf{Single video 3D reconstruction.}
    We generate a video with Gen3C \cite{ren2025gen3c} and 3D reconstruct the frames.
    The feedforward 3DGS prediction from DA3~\cite{lin2025depth} suffers from decreased quality in comparison to optimizing 3DGS from scratch from the DA3 predicted depth/cams.
    Combining SHARP~\cite{mescheder2026sharpmonocularviewsynthesis} and DA3 to obtain feedforward 3DGS predictions from multiple cameras suffers from the misaligned world space of the DA3 predictions (even though we utilize our optimized poses).
    In contrast, our method obtains high-quality renderings from inconsistent views by lifting the corresponding geometry into an aligned space.
    }
    \label{fig:qual_sharp_da3}
\end{figure*}

In our main results, we utilize the DA3 \cite{lin2025depth} points as 3DGS \cite{kerbl20233d} initialization and regularize their positions with a depth loss \cite{zhang2024rade} against the DA3 depth.
We refer to this as ``DA3'' \cite{lin2025depth} and additionally compare against DA3 by utilizing its feed-forward 3DGS prediction head in \Cref{fig:qual_sharp_da3} (marked as ``DA3 (FF)'').
We noticed severe quality degradation both in terms of color-balance and background density in comparison to re-optimizing the 3DGS scenes, across all available released model checkpoints.
To ensure a fair comparison, we thus opt for the ``DA3'' comparison described above.
It is conceptually similar to the capabilities of the feed-forward head, but without the quality degradations.
Concretely, the FF head predicts the remaining Gaussian attributes and places them at the positions obtain from unprojecting the predicted depth maps.
Similarly, the point-based initialization and depth loss in our ``DA3'' implementation ensures that the Gaussians are positioned in this way.

\subsection{Comparison Against SHARP}

Recently, the monocular Gaussian Splatting predictor SHARP~\cite{mescheder2026sharpmonocularviewsynthesis} demonstrated impressive results on novel view synthesis from a single image.
We propose a baseline comparison on our task by predicting and merging per-view 3DGS attributes from our generated video sequence with this method.
Since the per-view predictions do not yet reside in a unified world space, we utilize our cameras $\{\mathbf{R}_i, \mathbf{t}_i\}_{i=1}^N$ that we optimize from the DA3~\cite{lin2025depth} predictions to align them.
However, \Cref{fig:qual_sharp_da3} shows that this is insufficient to obtain aligned worlds, which highlights the need for our non-rigid alignment and reconstruction stages.
While the per-frame predictions render high-quality textures, they remain geometrically unaligned, since the input generated frames are inconsistent.
In contrast, our full approach is able to produce aligned and high-quality surfaces.

\section{Additional Implementation Details}

\subsection{Voxelized Confidence Filter}

\begin{figure*}
    \centering
    \setlength{\tabcolsep}{1pt}
    \renewcommand{\arraystretch}{1.1}

    \begin{tabular}{c c c c c}

        \includegraphics[width=0.19\textwidth]{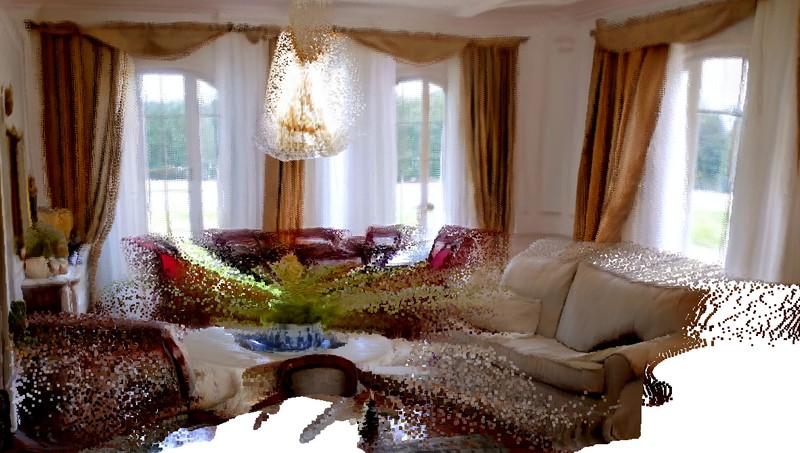}
        & \includegraphics[width=0.19\textwidth]{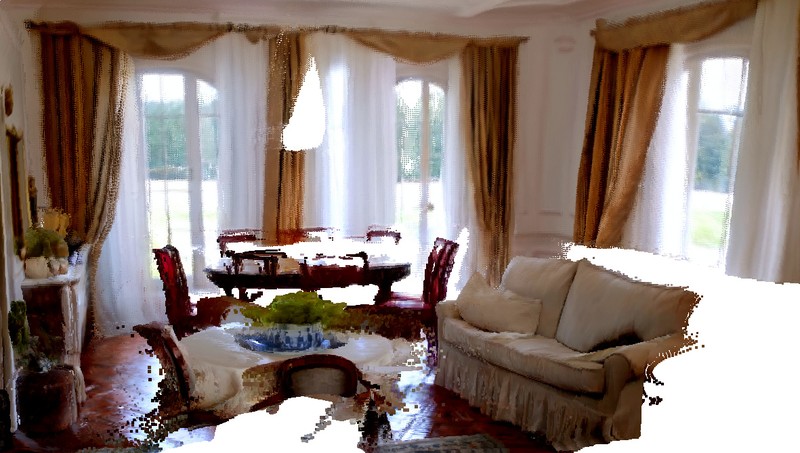}
        & \includegraphics[width=0.19\textwidth]{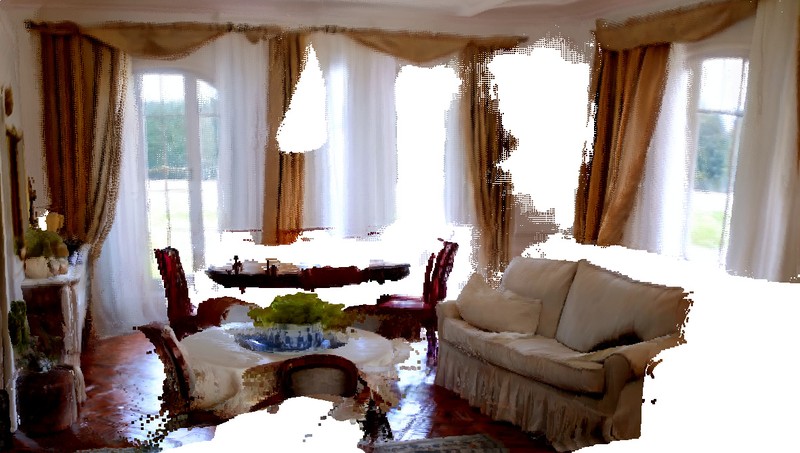}
        & \includegraphics[width=0.19\textwidth]{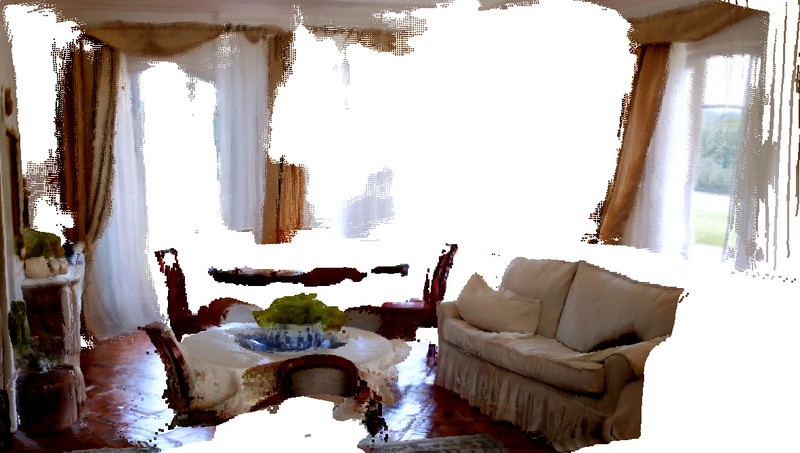}
        & \includegraphics[width=0.19\textwidth]{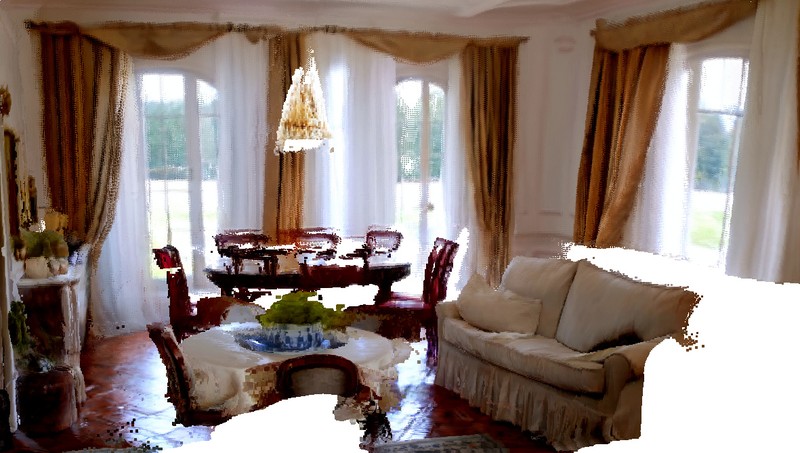} \\

        {\fontsize{8}{9}\selectfont global 0} &
        {\fontsize{8}{9}\selectfont global 10} &
        {\fontsize{8}{9}\selectfont global 20} &
        {\fontsize{8}{9}\selectfont global 40} &
        {\fontsize{8}{9}\selectfont ours voxel-based} \\
        
    \end{tabular}
    \vspace{-2mm}
    \caption{\textbf{Voxel-based confidence filtering.}     
    We propose a point pruning strategy of DA3~\cite{lin2025depth} predictions to retain reliable, but spatially dense geometry, based on the predicted confidence scores.
    Naive global filtering schemes can prune too few points (global 0; noticeable stretch artifacts), which makes it hard to obtain high quality surface alignments.
    Global filtering with too high thresholds (global 10-40) can remove important scene content (e.g., the lamp or windows), which requires additional densification in the reconstruction stage.
    In contrast, our voxel-based filtering obtains dense surfaces that can directly be aligned in the subsequent stages and require no further densification to model complete scenes.
    }
    \label{fig:qual_conf_thresh}
\end{figure*}

After unprojecting the initial DA3 \cite{lin2025depth} predictions into the pointcloud $\mathcal{\bar{P}}$, we propose a voxelized confidence filtering scheme to obtain the scene initialization $\mathcal{P}$ (see Section 3.1).
We compare this filtering scheme against common global percentile thresholds that are similarly based on the predicted confidence scores in \Cref{fig:qual_conf_thresh}.
The challenge with this global filtering scheme is to find reliable thresholds across scenes that do not prune too few or too much points.
Pruning too few points creates a scene initialization with many stretched-out artifacts between objects.
This makes our subsequent geometry alignment stage harder, as the nearest neighbor correspondence search in the ICP terms is less reliable and struggles to create unified surfaces.
In contrast, removing too much points creates hard to reconstruct background regions in the photometric Gaussian Splatting optimization.
While this can be tackled with densification, the introduction of additional Gaussian parameters undermines the previous efforts of obtaining an aligned pointcloud geometry: they are harder to regularize onto a single surface and could re-introduce floating artifacts if the corresponding background regions are inconsistently generated in the input sequences.
Our voxel-based strategy solves both problems by retaining only reliable points, but densely across the entire scenes.
This makes it possible to optimize our final scene representation without densification and thus we obtain 3D consistent worlds that can render complete and high-quality images also for the background regions.

\subsection{Details Of Quantitative Evaluation}

We quantitatively evaluate our main results by calculating consistency and fidelity metrics of the established WorldScore \cite{duan2025worldscore} benchmark.
Here, we provide additional details.
Since our input video sequences are generated with video diffusion models, the individual frames are not 3D consistent (i.e., they suffer from generative drift; see Figure 1).
Common reconstruction metrics like PSNR, SSIM \cite{wang2004image}, LPIPS \cite{johnson2016perceptual} compare rendered and observed images.
However, since the input frames are inconsistent, these metrics would only measure the degree to which the respective reconstruction methods can \textit{reproduce the inconsistencies}.
In contrast, we are interested to measure how geometrically consistent the 3D scenes are.
This entails that renderings from input camera poses do not precisely (i.e., pixel-wise) reproduce the input frames, but instead small movements of individual objects become noticeable, that solve the distortions inherent in the input frames.
To this end, we follow WorldScore \cite{duan2025worldscore} and calculate two types of consistency metrics.
The ``3D consistency'' measures reprojection error from one frame to another based on a reconstruction of the rendered frames with DROID-SLAM \cite{teed2022droidslamdeepvisualslam}.
Similarly, the ``photometric consistency'' estimates the optical flow between consecutive frames using RAFT \cite{teed2020raftrecurrentallpairsfield} and calculates the warping error.
We are also interested to reason about the methods abilities to render sharp textures without blurriness or floating artifacts.
To quantify this image quality, we additionally adopt the fidelity scores CLIP-IQA+ \cite{wang2022exploringclipassessinglook} and CLIP Aesthetic \cite{schuhmann2022improvedaestheticpredictor}.

We calculate these metrics on re-renderings of the input video sequence after 3D reconstruction that we uniformly downsample to 100 frames (between 3-5 second clips on average).
This ensures that image fidelity is not negatively impacted by rendering novel views that show unoptimized areas like backgrounds.
It also allows to compare the image fidelity against the video frames itself, which serve as the upper bound in this category.
Importantly, this still allows to quantify the 3D consistency, since the input frames themselves are inconsistent.
Naively reproducing them thus yields lower scores in these metrics.
We report average results across all source video diffusion models (VDM) for each reconstruction method, which amounts to 16 different scenes (2-3 per VDM).

\section{Limitations}

\begin{figure*}
    \centering
    \setlength{\tabcolsep}{1pt}
    \renewcommand{\arraystretch}{1.1}

    \begin{tabular}{c c c c c}
        {\fontsize{8}{9}\selectfont Video Frame 1} &
        {\fontsize{8}{9}\selectfont Video Frame 2} &
        {\fontsize{8}{9}\selectfont Ours Pointcloud} &
        {\fontsize{8}{9}\selectfont Ours Rendering 1} &
        {\fontsize{8}{9}\selectfont Ours Rendering 2} \\

        \midrule

        \includegraphics[width=0.19\textwidth]{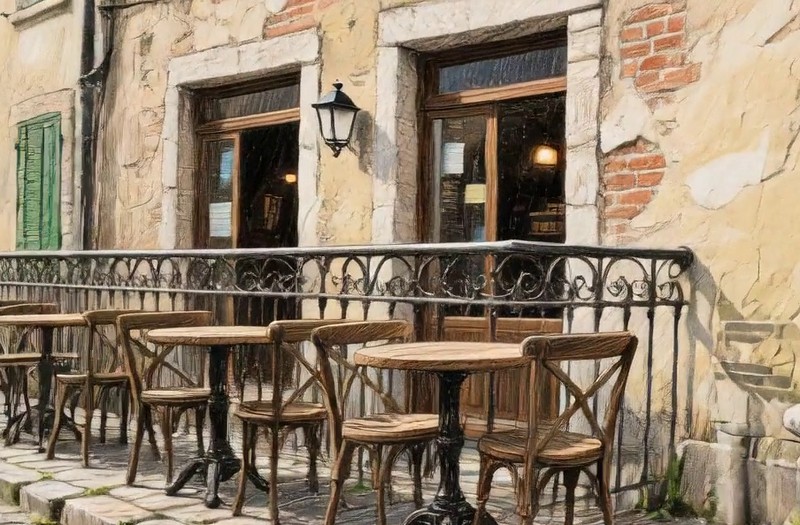}
        & \includegraphics[width=0.19\textwidth]{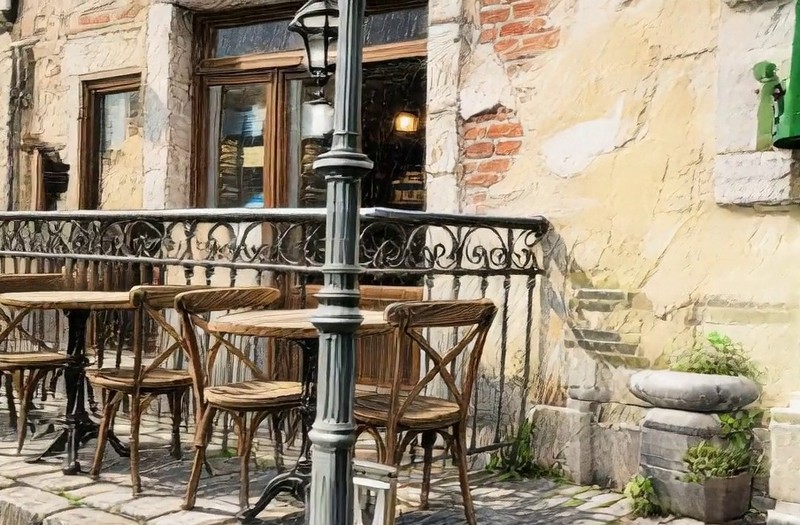} 
        & \includegraphics[width=0.19\textwidth]{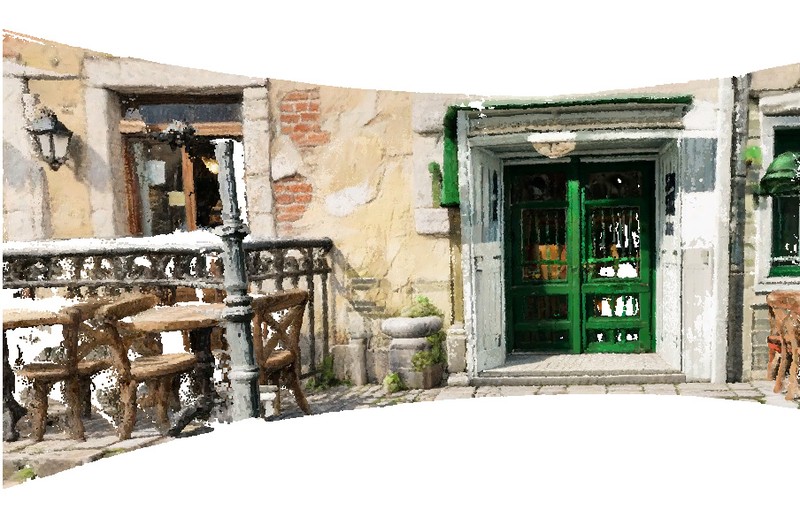}
        & \includegraphics[width=0.19\textwidth]{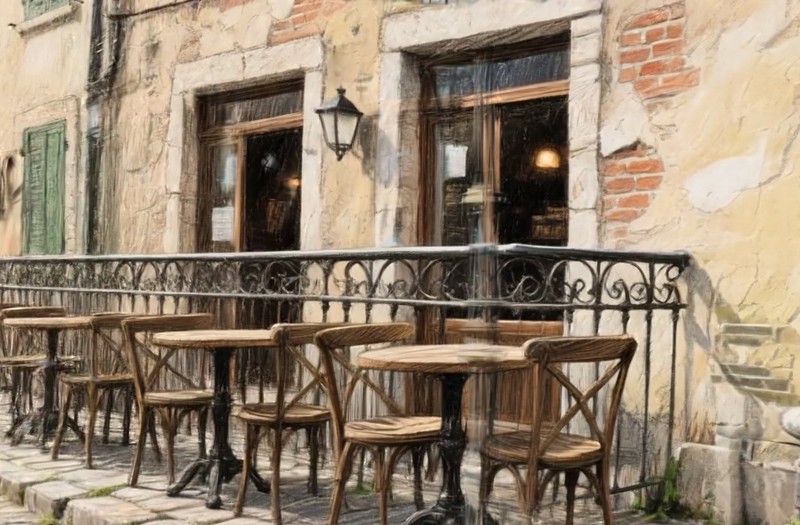}
        & \includegraphics[width=0.19\textwidth]{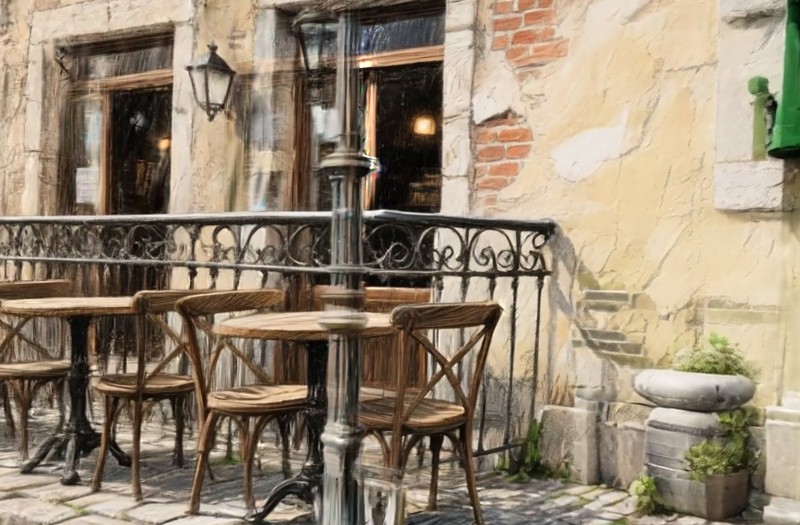} \\
        
    \end{tabular}
    \vspace{-2mm}
    \caption{\textbf{Limitations.}     
    Our method reconstructs 3D consistent worlds from inconsistent video sequences through geometry alignment and non-rigid Gaussian Splatting optimization.
    However, video diffusion models can also hallucinate novel content when revisiting areas (e.g., the street lamp appears suddenly).
    We can still align multiple frames that show such hallucinations, but they become a part of the 3D-consistent static geometry.
    Additionally, 3D reconstruction optimizes for the average explanation of all input views and as such, these inconsistencies stay visible (e.g., the street lamp is visible in both viewpoints when rendering from our 3D scene).
    }
    \label{fig:qual_limitations}
\end{figure*}

Our method reconstructs 3D scenes from inconsistent generated views, however, some limitations remain.
We tackle the problem of \textit{generative drift} in video diffusion models (VDM), i.e., existing objects move geometrically inconsistent.
However, VDMs can also suffer from hallucinations (e.g., revisiting previous scene areas creates novel or removes existing objects, changes their texture, etc.).
We visualize this phenomenon in \Cref{fig:qual_limitations}, where a street lamp suddenly appears in the input video sequence upon revisiting that area.
Our aligned pointcloud geometry thus includes this object (it is still aligned from multiple frames that show the object).
Our optimized 3D scene renders the street lamp in both input perspectives, since the optimization is jointly performed from all views on a single static 3D world and thus averages out inconsistent viewpoints.
Adopting robust reconstruction mechanisms to detect such outlier frames could resolve these problems \cite{sabour2023robustnerf}.

\section{Additional Results}

\subsection{Single Video 3D Reconstruction}

\begin{figure*}
    \centering
    \setlength{\tabcolsep}{1pt}
    \renewcommand{\arraystretch}{1.1}

    \begin{tabular}{c | c | c c c c}
        {\fontsize{8}{9}\selectfont Video Frames} &
        &
        {\fontsize{8}{9}\selectfont DA3~\cite{lin2025depth}} &
        {\fontsize{8}{9}\selectfont 3DGS-MCMC~\cite{kheradmand20243d}} &
        {\fontsize{8}{9}\selectfont VGGT-X$^\dagger$~\cite{liu2025vggt}} &
        {\fontsize{8}{9}\selectfont Ours} \\

        \midrule

        \includegraphics[width=0.19\textwidth]{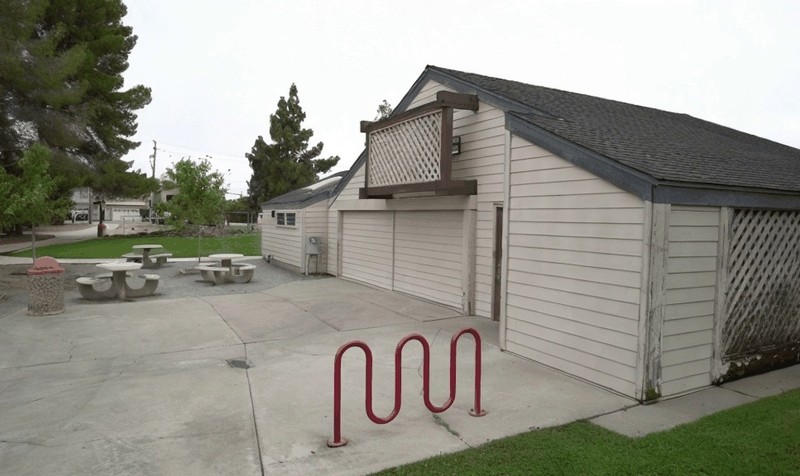}
        & \raisebox{0.0\height}{\rotatebox{90}{\fontsize{8}{9}\selectfont Input}}
        & \includegraphics[width=0.19\textwidth]{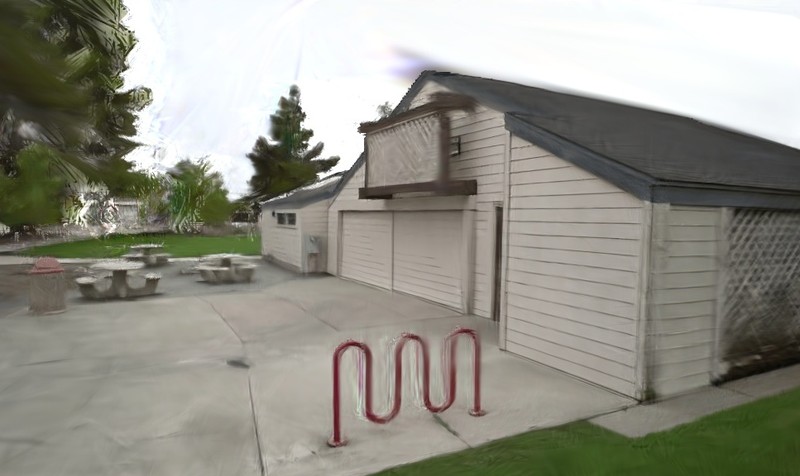}
        & \includegraphics[width=0.19\textwidth]{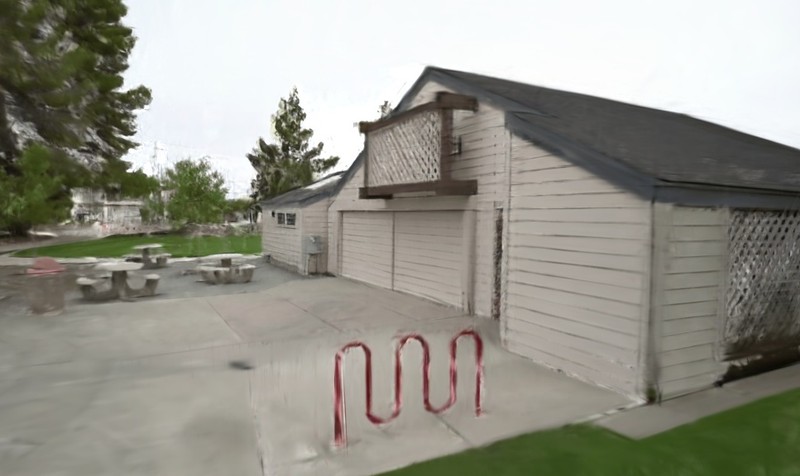}
        & \includegraphics[width=0.19\textwidth]{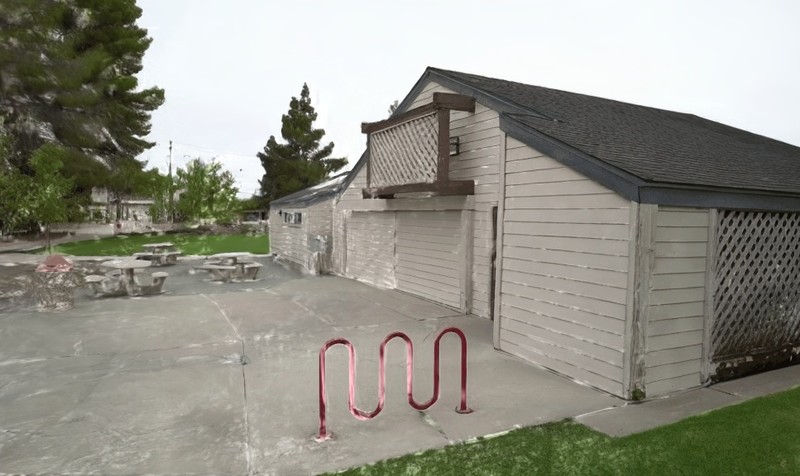}
        & \includegraphics[width=0.19\textwidth]{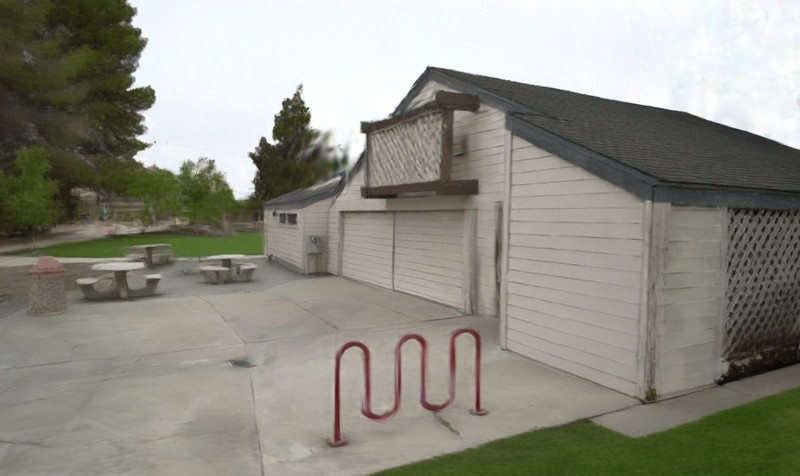} \\

        \includegraphics[width=0.19\textwidth]{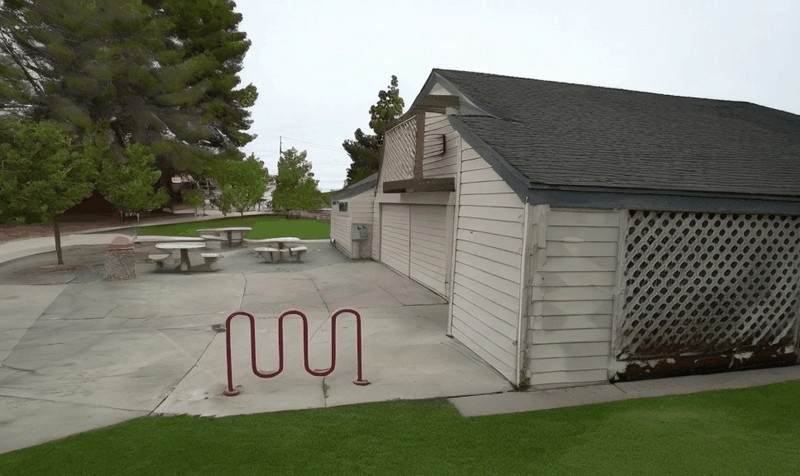}
        & \multirow{2}{*}{\rotatebox{90}{\fontsize{8}{9}\selectfont Novel Views}}
        & \includegraphics[width=0.19\textwidth]{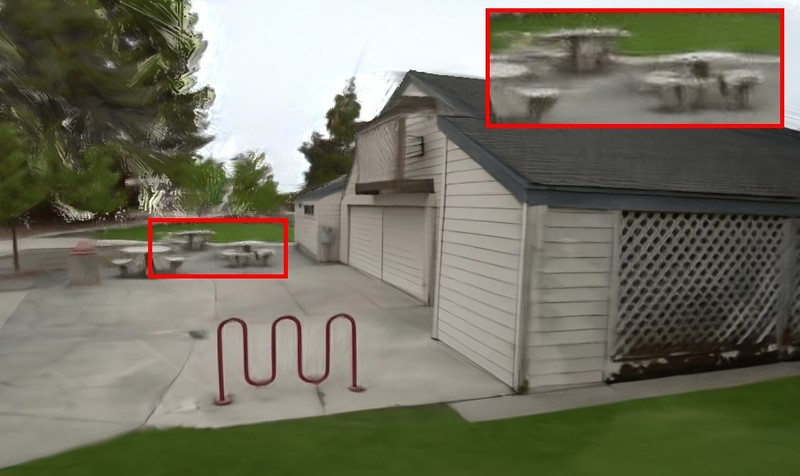}
        & \includegraphics[width=0.19\textwidth]{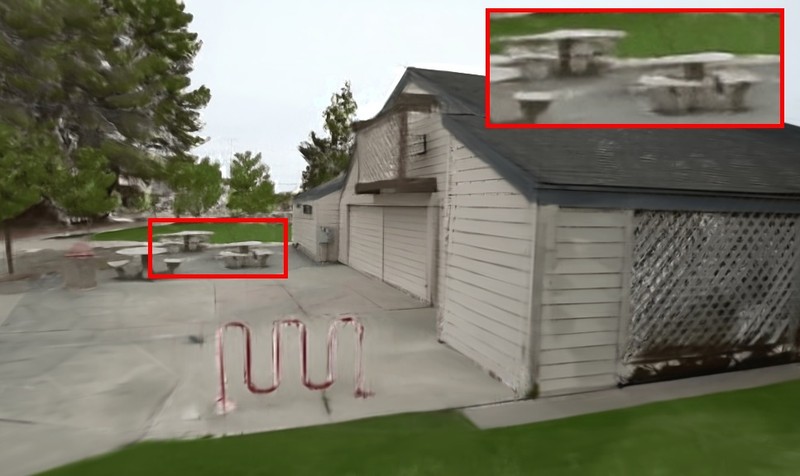}
        & \includegraphics[width=0.19\textwidth]{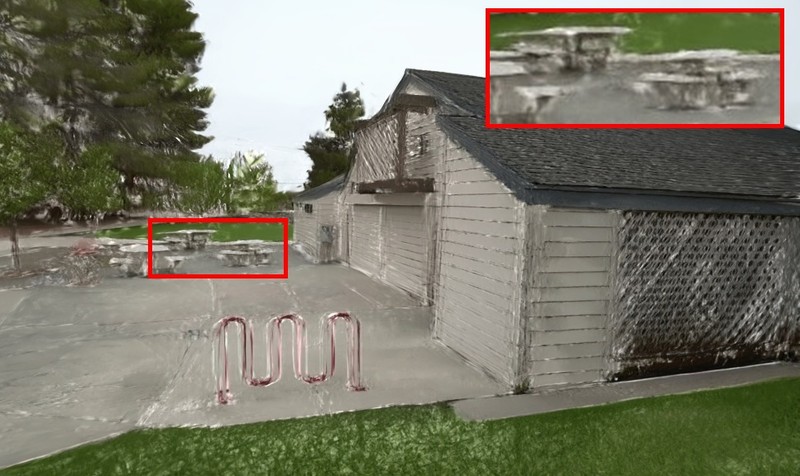}
        & \includegraphics[width=0.19\textwidth]{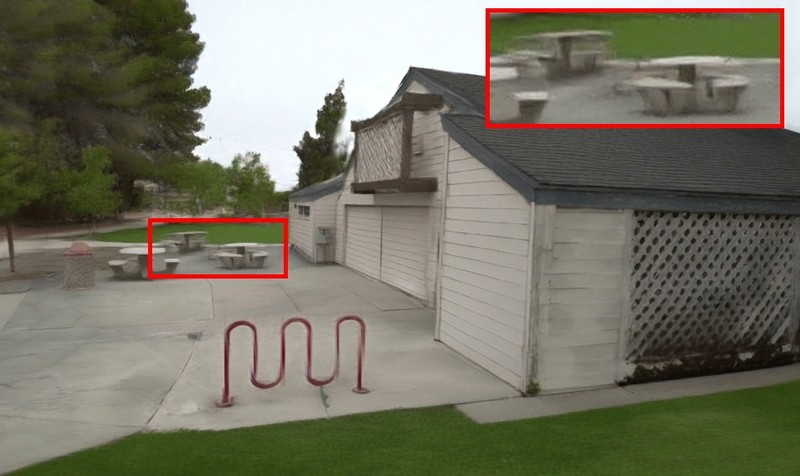} \\

        \includegraphics[width=0.19\textwidth]{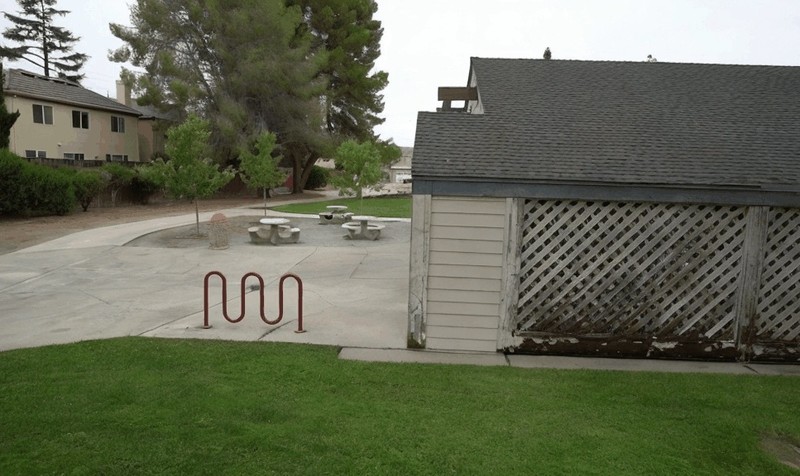}
        &
        & \includegraphics[width=0.19\textwidth]{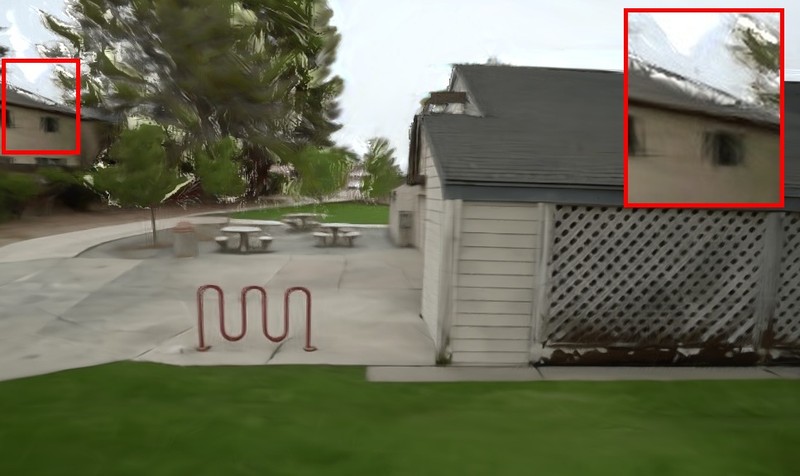}
        & \includegraphics[width=0.19\textwidth]{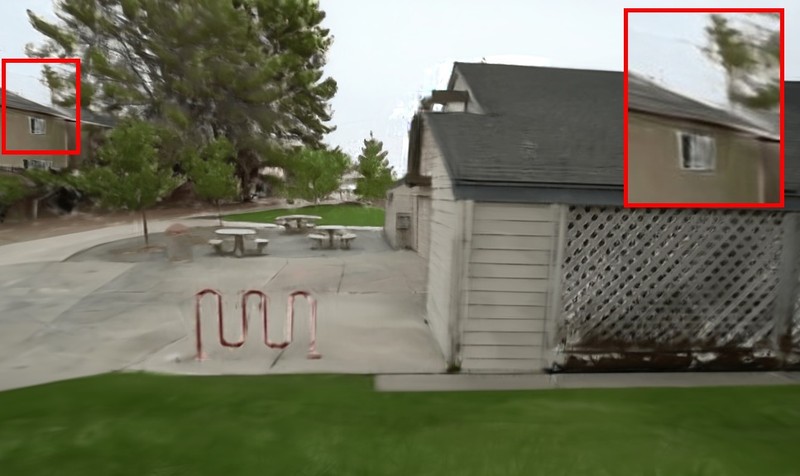}
        & \includegraphics[width=0.19\textwidth]{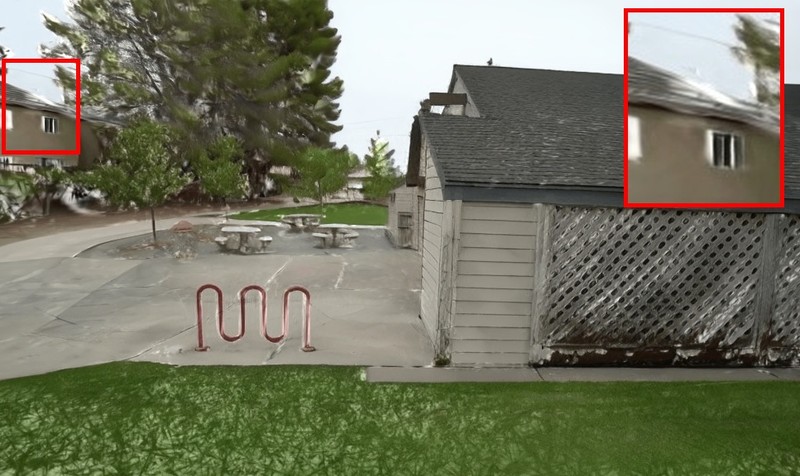}
        & \includegraphics[width=0.19\textwidth]{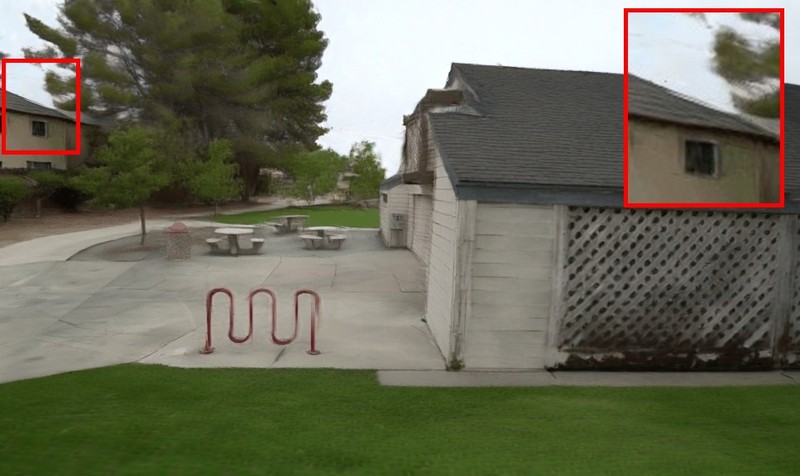} \\

        \midrule

        \includegraphics[width=0.19\textwidth]{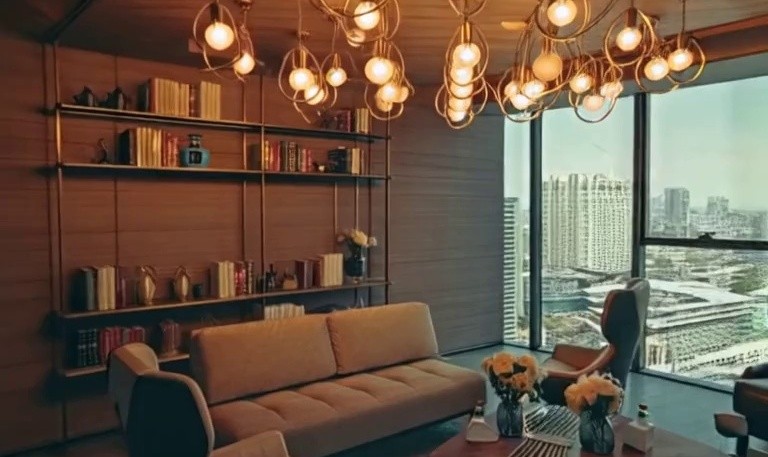}
        & \multirow{2}{*}{\rotatebox{90}{\fontsize{8}{9}\selectfont Novel Views}}
        & \includegraphics[width=0.19\textwidth]{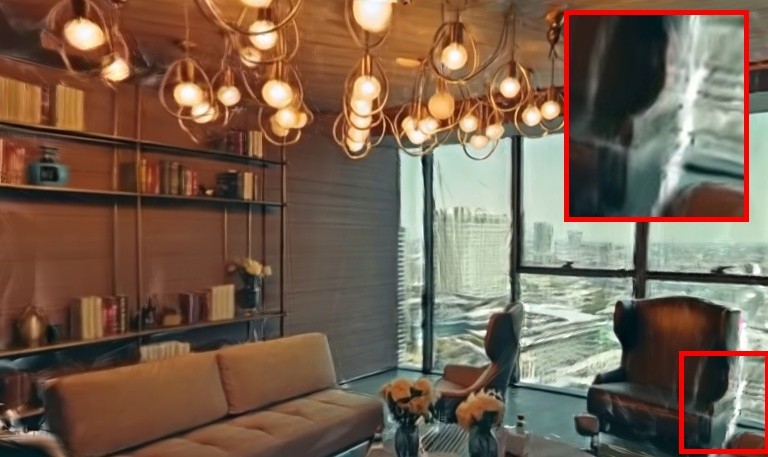}
        & \includegraphics[width=0.19\textwidth]{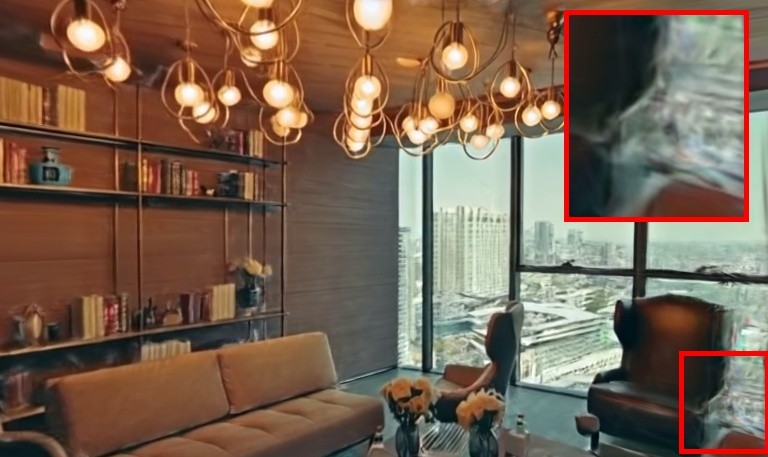}
        & \includegraphics[width=0.19\textwidth]{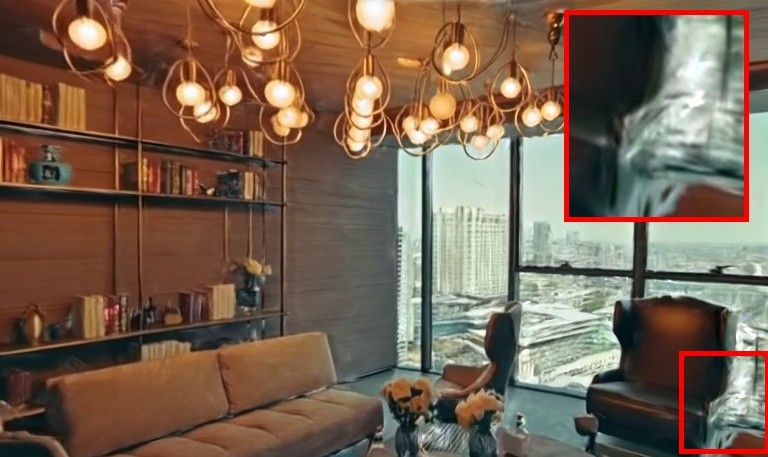}
        & \includegraphics[width=0.19\textwidth]{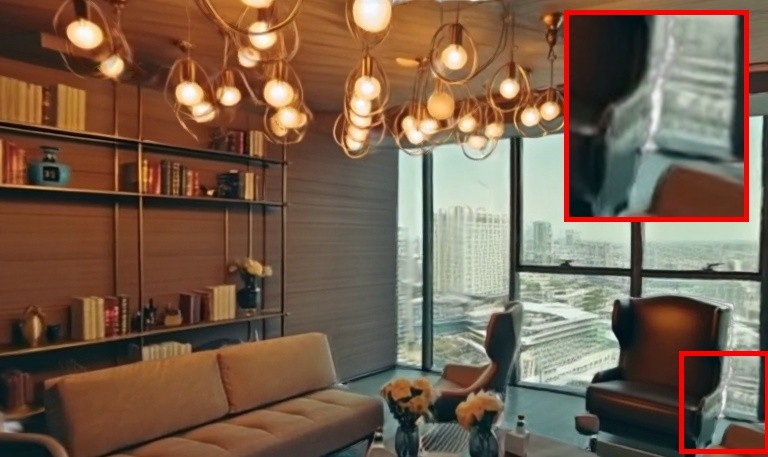} \\

        \includegraphics[width=0.19\textwidth]{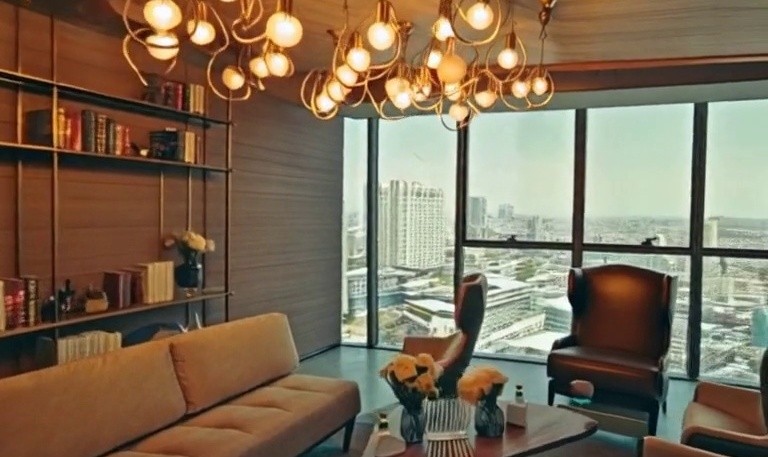}
        &
        & \includegraphics[width=0.19\textwidth]{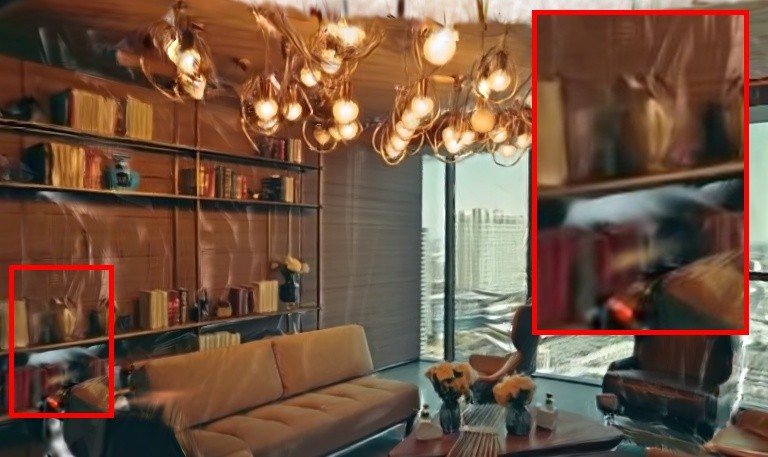}
        & \includegraphics[width=0.19\textwidth]{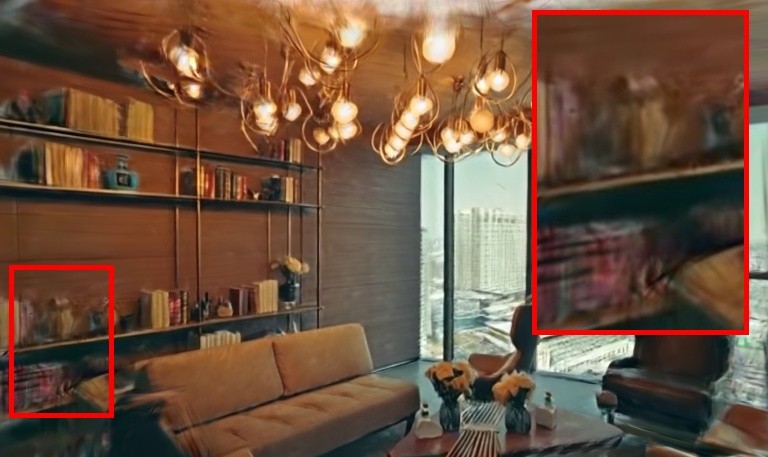}
        & \includegraphics[width=0.19\textwidth]{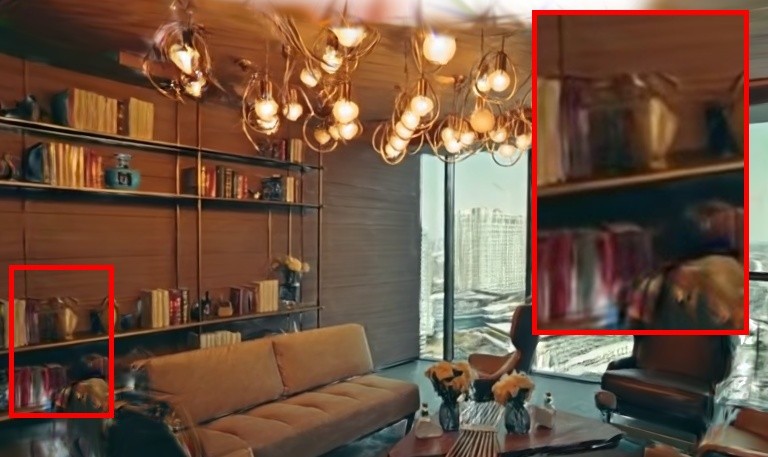}
        & \includegraphics[width=0.19\textwidth]{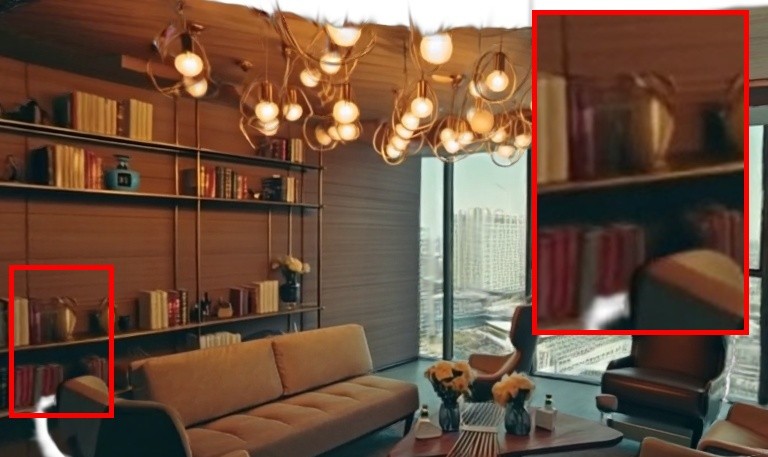} \\

        \midrule

        \includegraphics[width=0.19\textwidth]{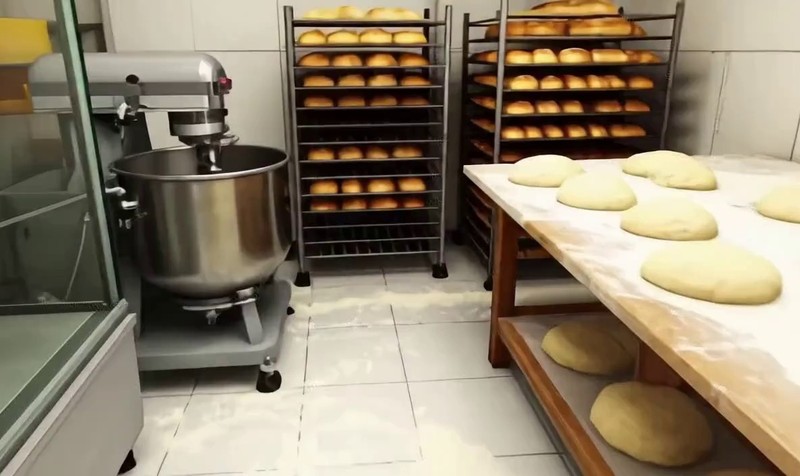}
        & \multirow{2}{*}{\rotatebox{90}{\fontsize{8}{9}\selectfont Novel Views}}
        & \includegraphics[width=0.19\textwidth]{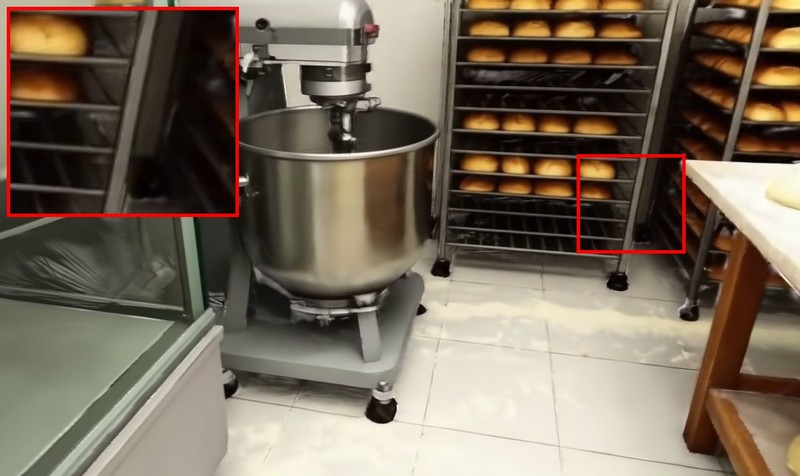}
        & \includegraphics[width=0.19\textwidth]{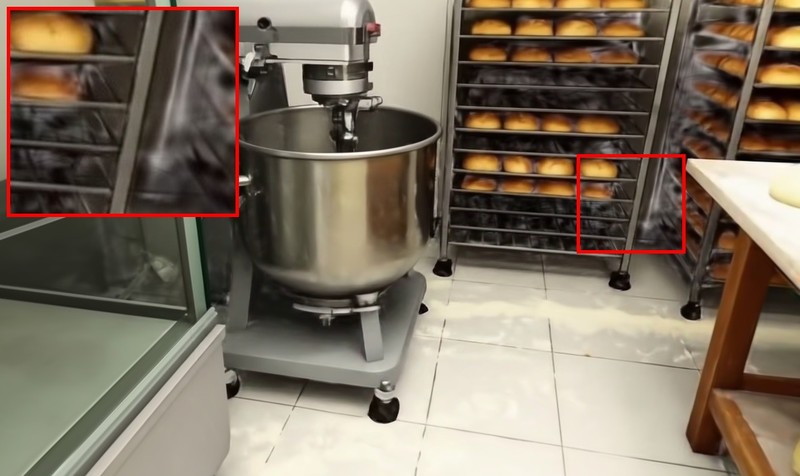}
        & \includegraphics[width=0.19\textwidth]{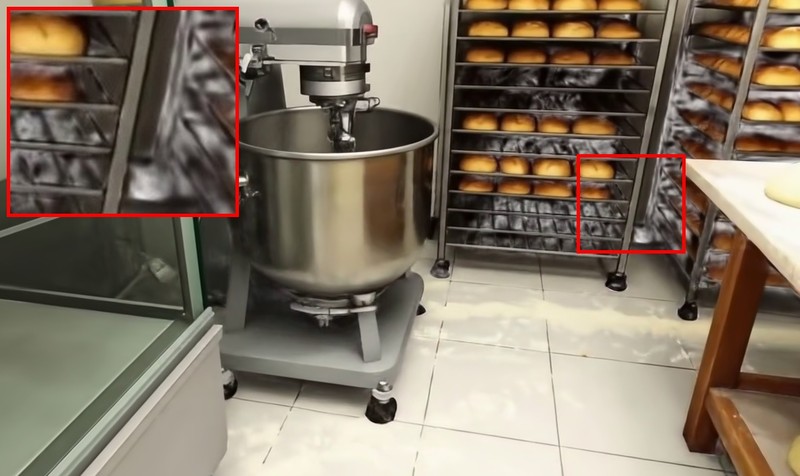}
        & \includegraphics[width=0.19\textwidth]{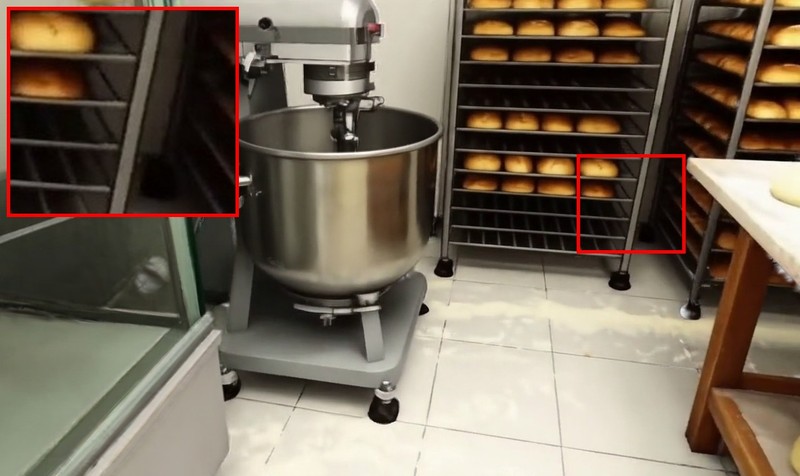} \\

        \includegraphics[width=0.19\textwidth]{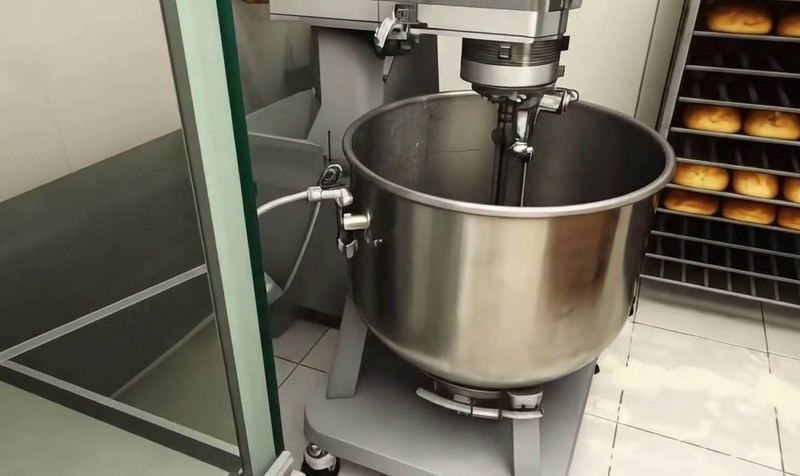}
        &
        & \includegraphics[width=0.19\textwidth]{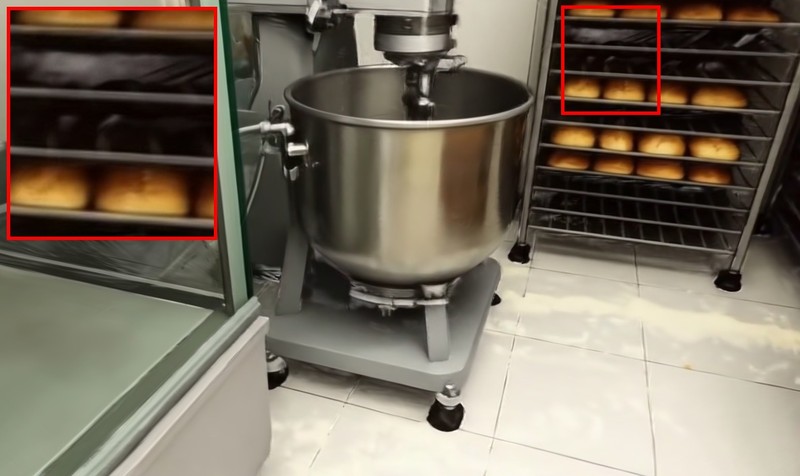}
        & \includegraphics[width=0.19\textwidth]{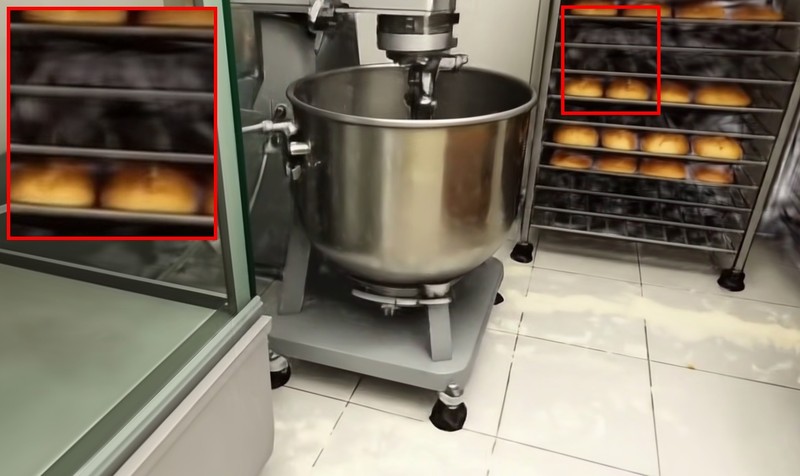}
        & \includegraphics[width=0.19\textwidth]{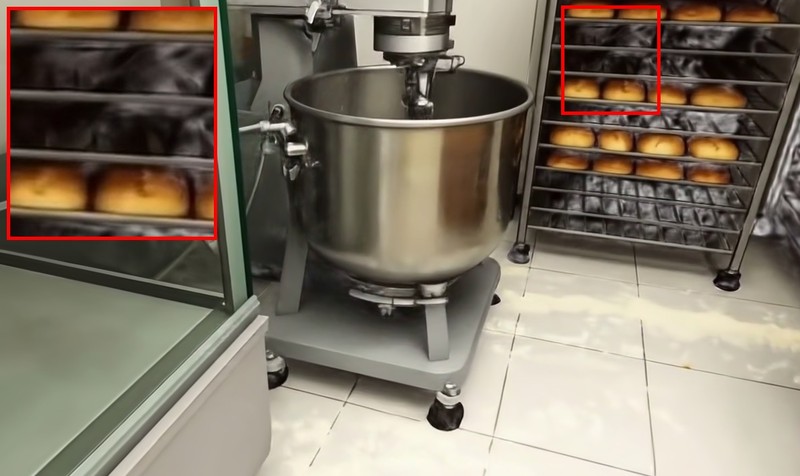}
        & \includegraphics[width=0.19\textwidth]{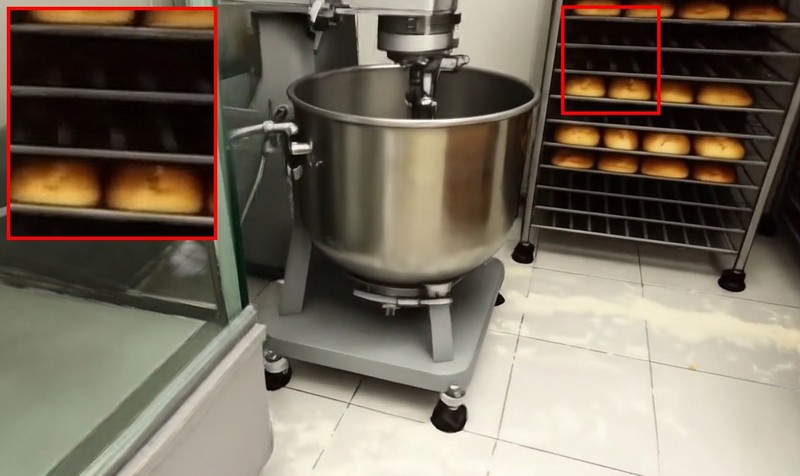} \\
        
    \end{tabular}
    \vspace{-4mm}
    \caption{\textbf{Single video 3D reconstruction.}     
    We generate videos with ViewCrafter \cite{yu2024viewcrafter} (top), Voyager \cite{huang2025voyager} (mid), Genie3 \cite{genie3} (bottom) and 3D reconstruct these frames.
    Inconsistencies in the generations lead to blurry textures for the baselines compared to the corresponding video, and to floating artifacts from novel views.
    In contrast, our method creates 3D consistent worlds with high fidelity beyond generated perspectives.
    }
    \label{fig:qual_single_suppl1}
\end{figure*}

\begin{figure*}
    \centering
    \setlength{\tabcolsep}{1pt}
    \renewcommand{\arraystretch}{1.1}

    \begin{tabular}{c | c | c c c c}
        {\fontsize{8}{9}\selectfont Video Frames} &
        &
        {\fontsize{8}{9}\selectfont DA3~\cite{lin2025depth}} &
        {\fontsize{8}{9}\selectfont 3DGS-MCMC~\cite{kheradmand20243d}} &
        {\fontsize{8}{9}\selectfont VGGT-X$^\dagger$~\cite{liu2025vggt}} &
        {\fontsize{8}{9}\selectfont Ours} \\

        \midrule

        \includegraphics[width=0.19\textwidth]{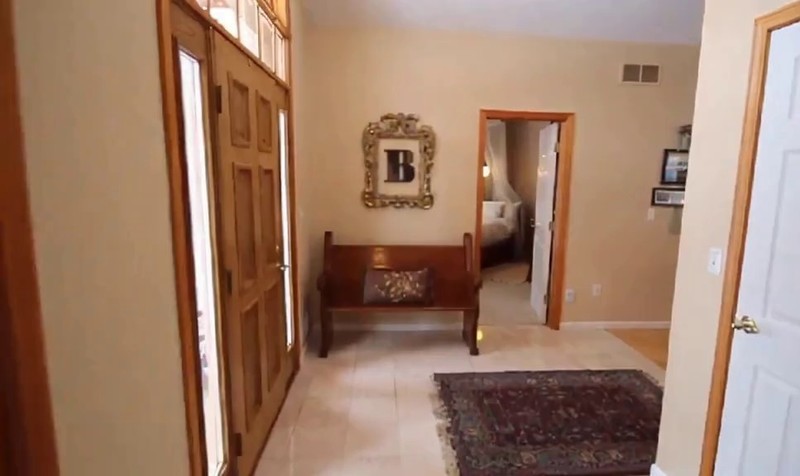}
        & \raisebox{0.0\height}{\rotatebox{90}{\fontsize{8}{9}\selectfont Input}}
        & \includegraphics[width=0.19\textwidth]{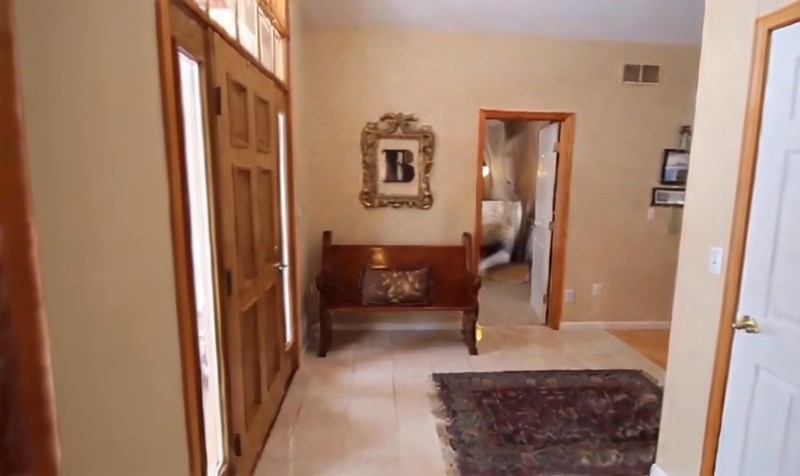}
        & \includegraphics[width=0.19\textwidth]{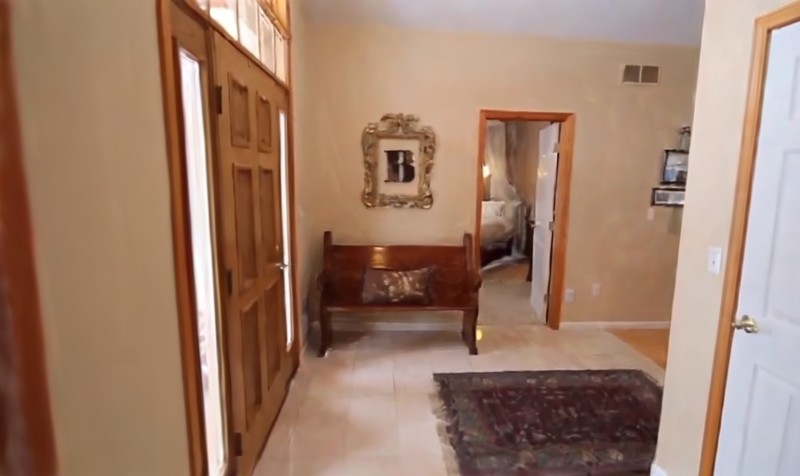}
        & \includegraphics[width=0.19\textwidth]{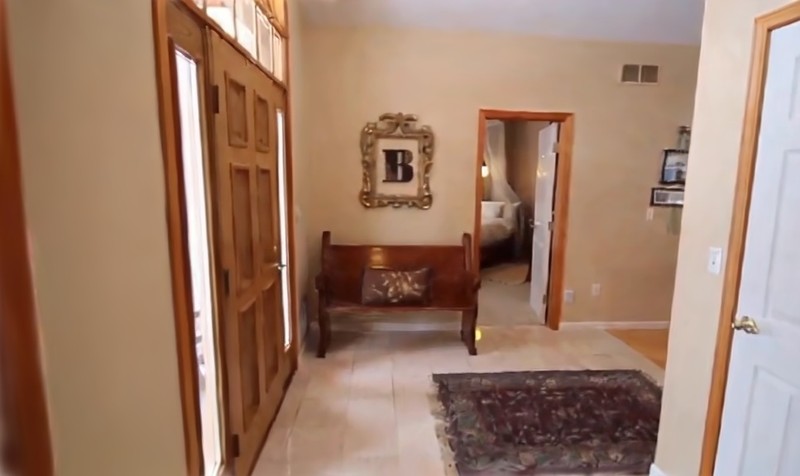}
        & \includegraphics[width=0.19\textwidth]{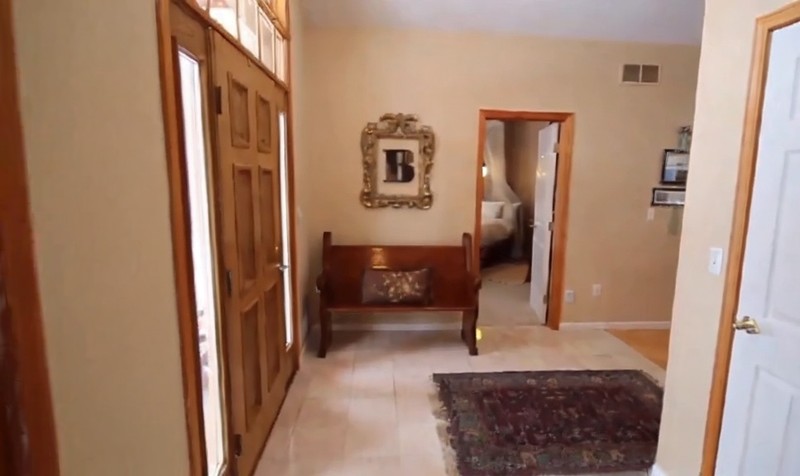} \\

        \includegraphics[width=0.19\textwidth]{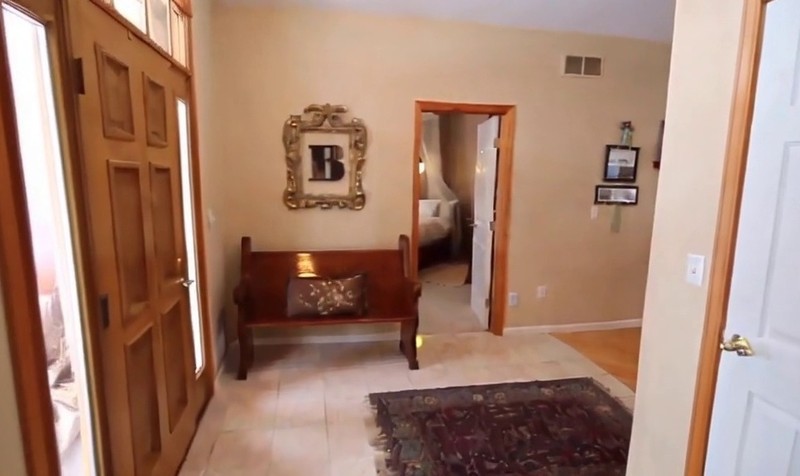}
        & \multirow{2}{*}{\rotatebox{90}{\fontsize{8}{9}\selectfont Novel Views}}
        & \includegraphics[width=0.19\textwidth]{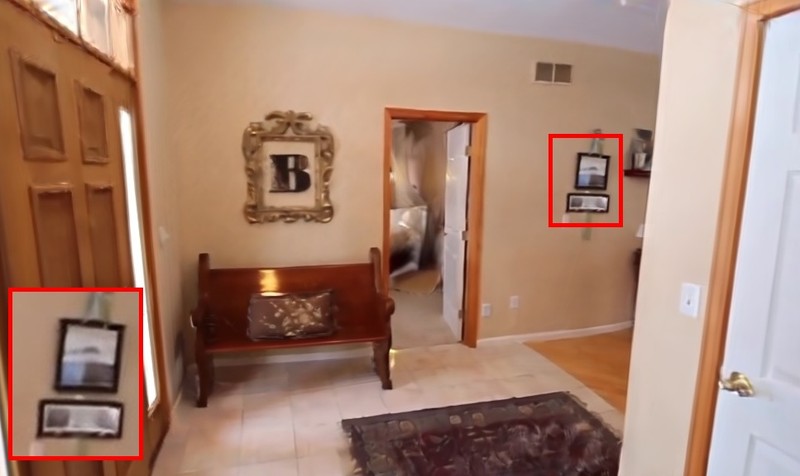}
        & \includegraphics[width=0.19\textwidth]{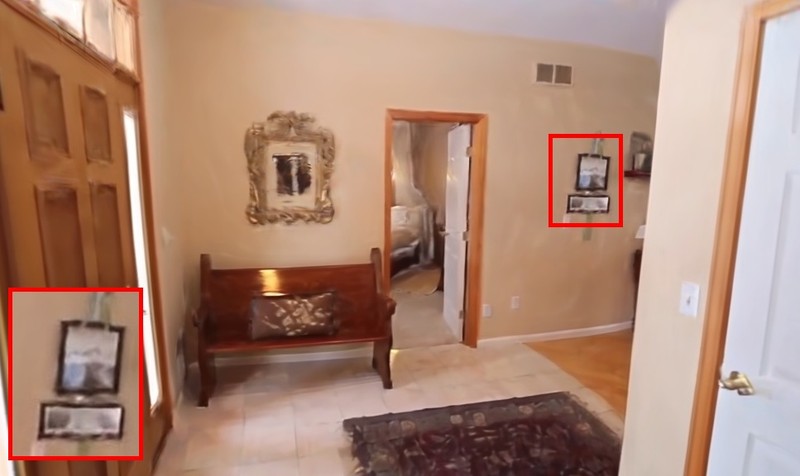}
        & \includegraphics[width=0.19\textwidth]{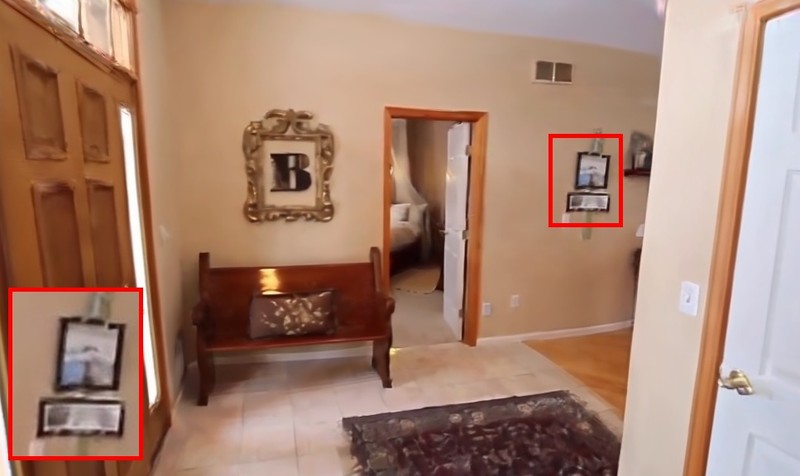}
        & \includegraphics[width=0.19\textwidth]{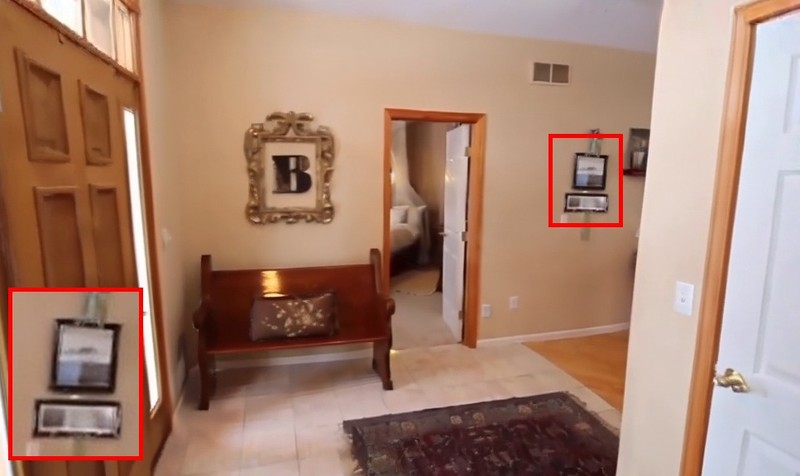} \\

        \includegraphics[width=0.19\textwidth]{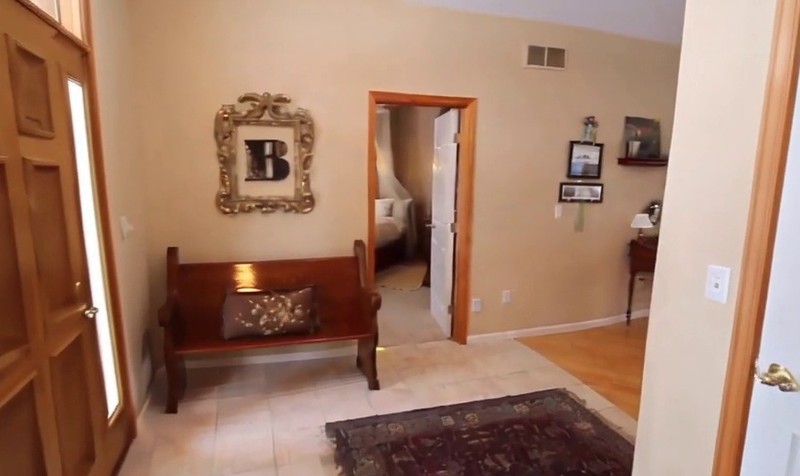}
        &
        & \includegraphics[width=0.19\textwidth]{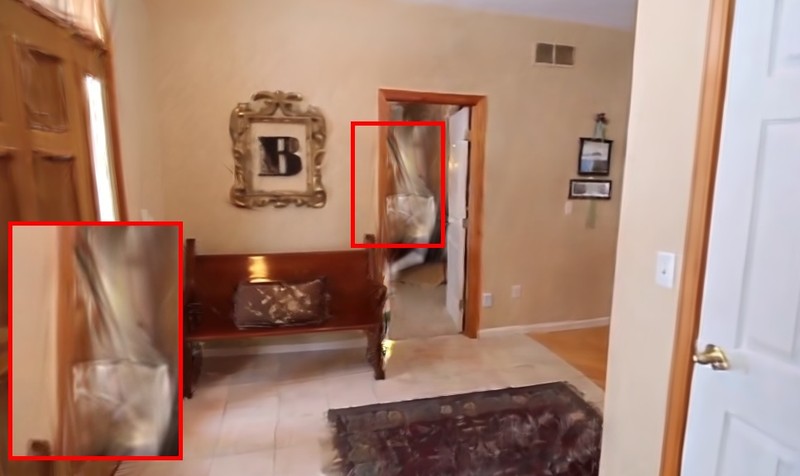}
        & \includegraphics[width=0.19\textwidth]{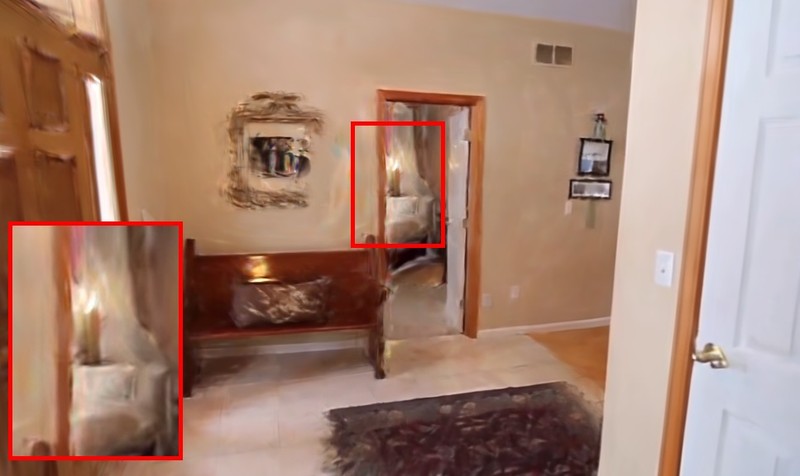}
        & \includegraphics[width=0.19\textwidth]{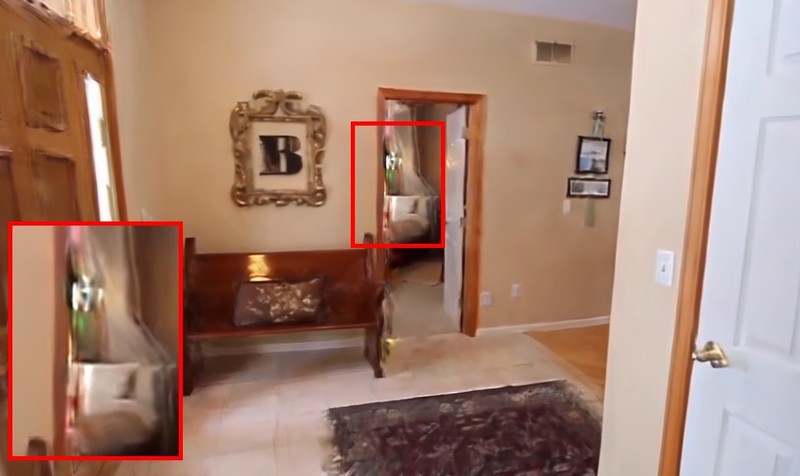}
        & \includegraphics[width=0.19\textwidth]{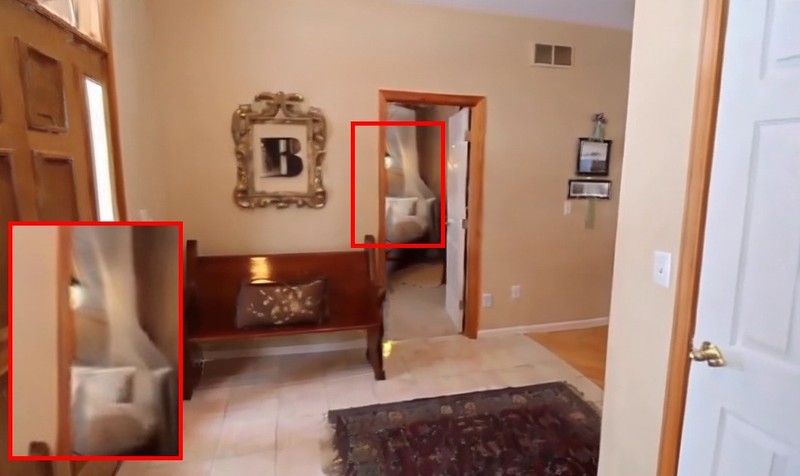} \\

        \midrule

        \includegraphics[width=0.19\textwidth]{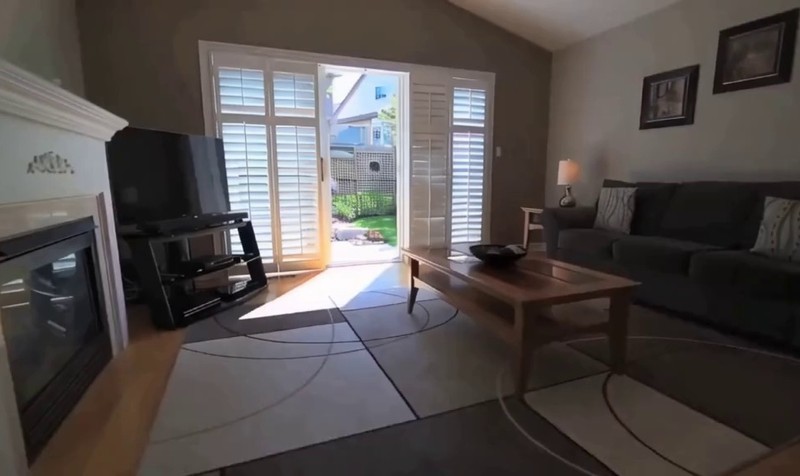}
        & \multirow{2}{*}{\rotatebox{90}{\fontsize{8}{9}\selectfont Novel Views}}
        & \includegraphics[width=0.19\textwidth]{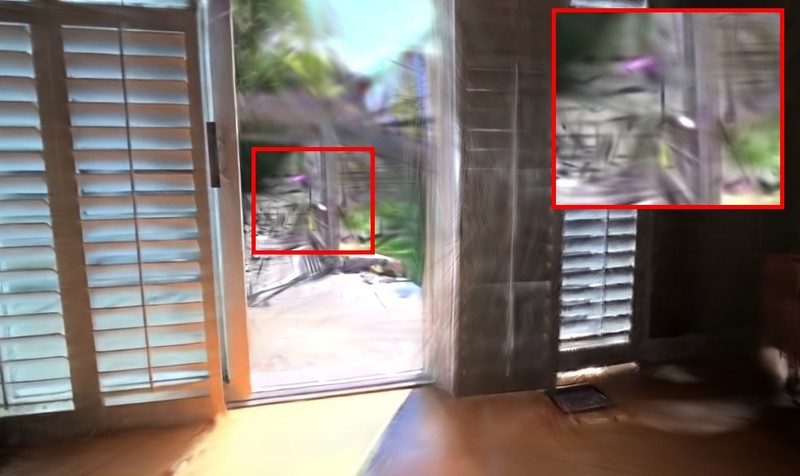}
        & \includegraphics[width=0.19\textwidth]{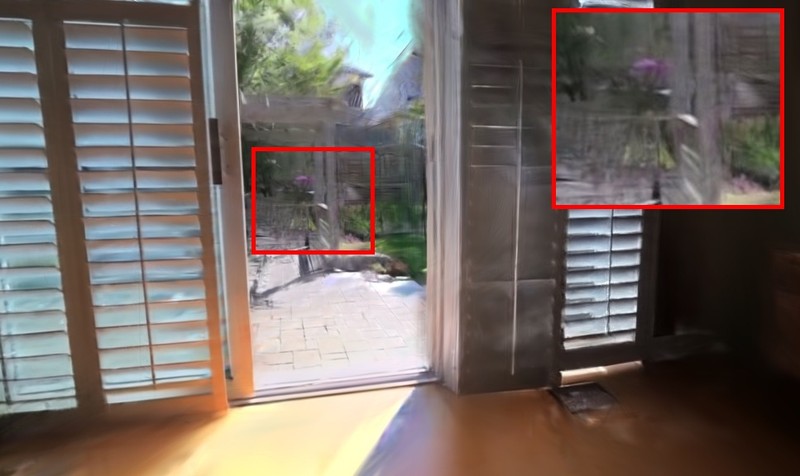}
        & \includegraphics[width=0.19\textwidth]{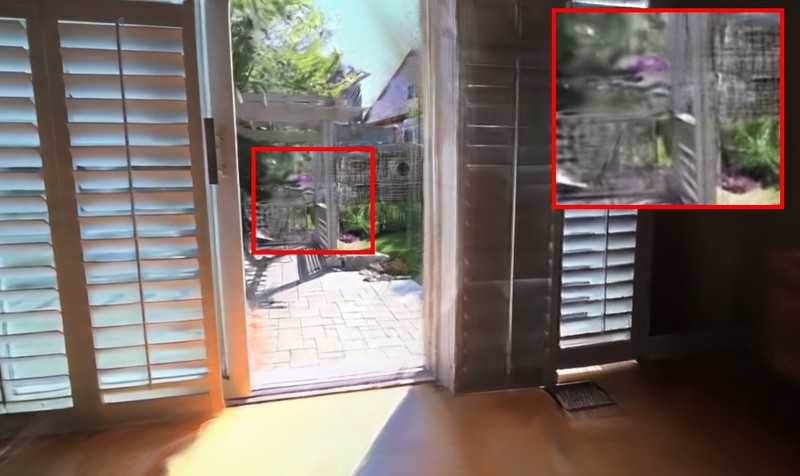}
        & \includegraphics[width=0.19\textwidth]{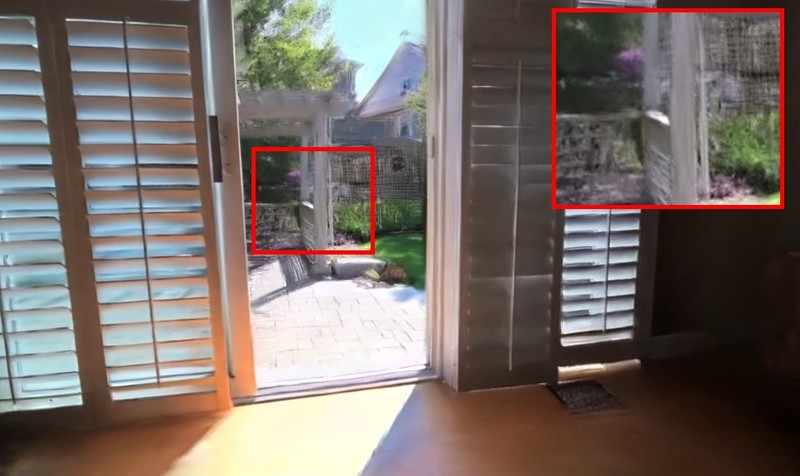} \\

        \includegraphics[width=0.19\textwidth]{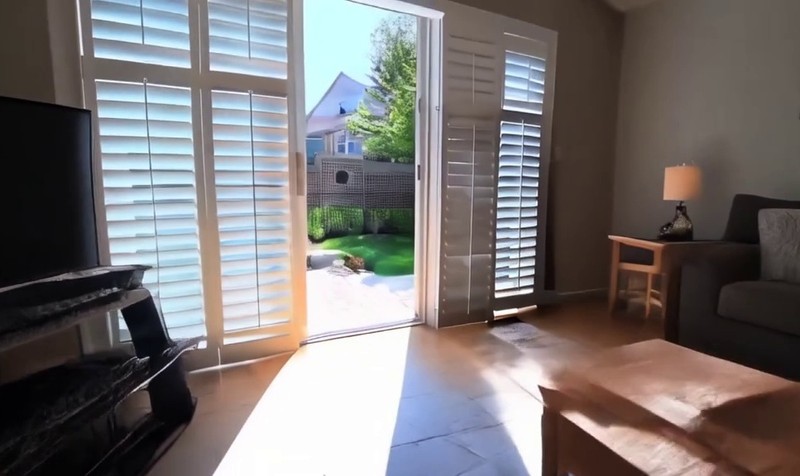}
        &
        & \includegraphics[width=0.19\textwidth]{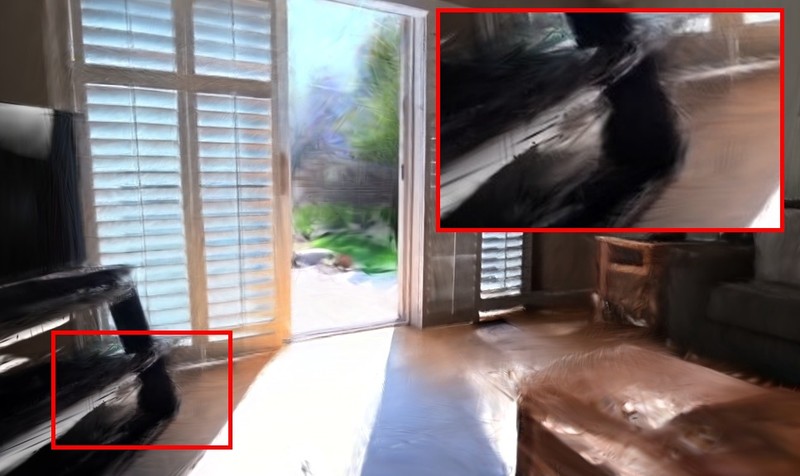}
        & \includegraphics[width=0.19\textwidth]{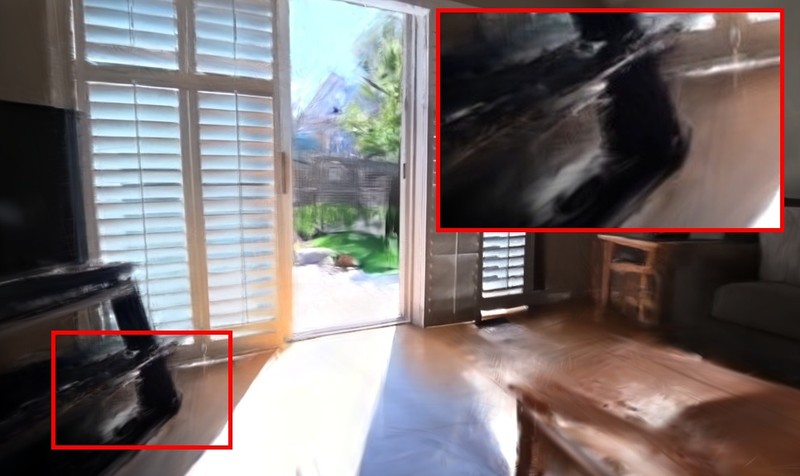}
        & \includegraphics[width=0.19\textwidth]{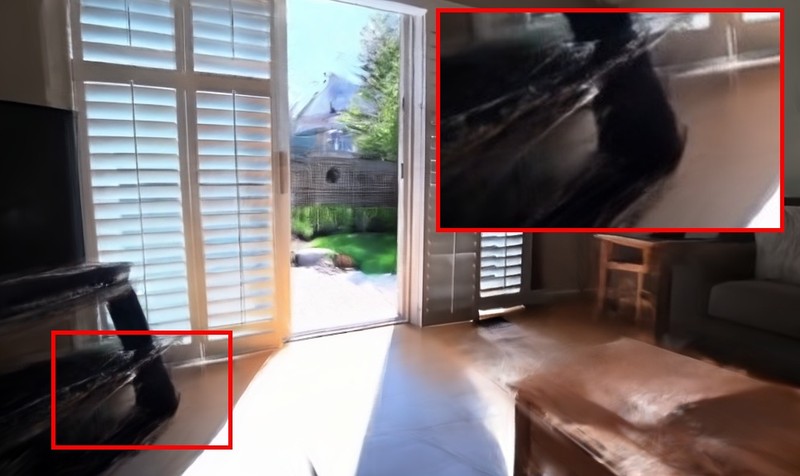}
        & \includegraphics[width=0.19\textwidth]{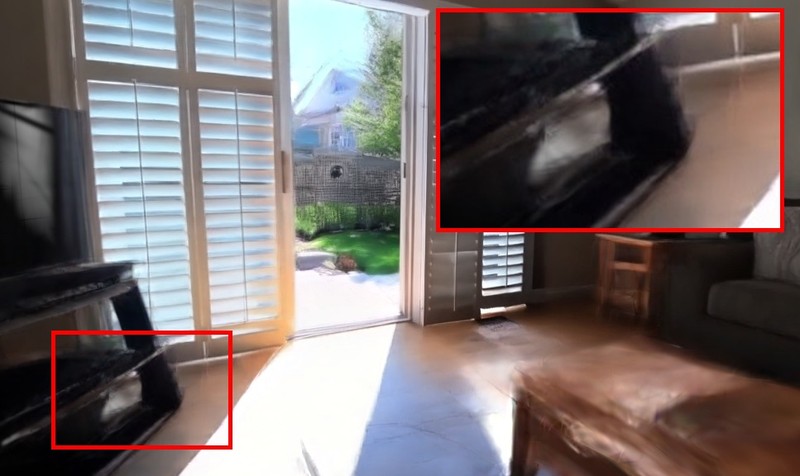} \\

        \midrule

        \includegraphics[width=0.19\textwidth]{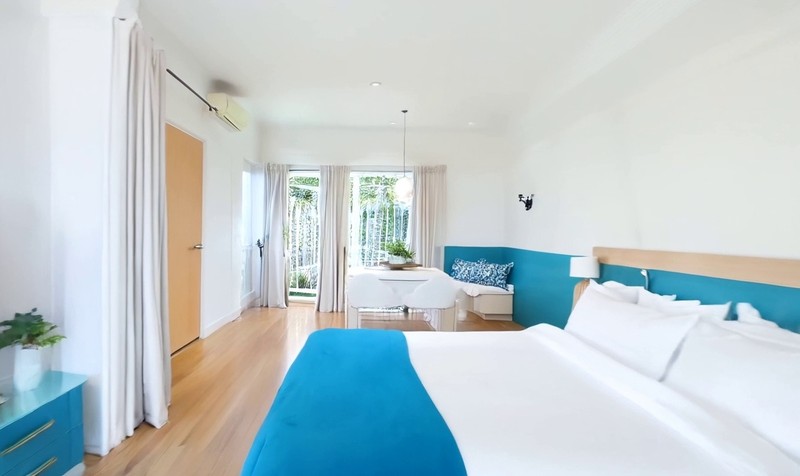}
        & \multirow{2}{*}{\rotatebox{90}{\fontsize{8}{9}\selectfont Novel Views}}
        & \includegraphics[width=0.19\textwidth]{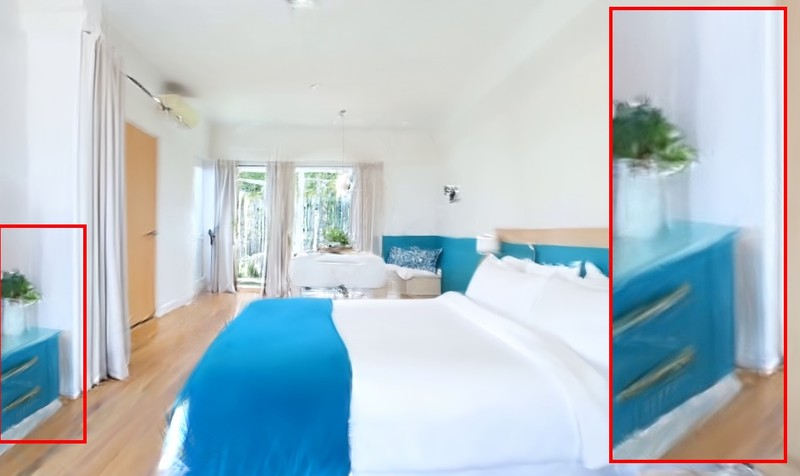}
        & \includegraphics[width=0.19\textwidth]{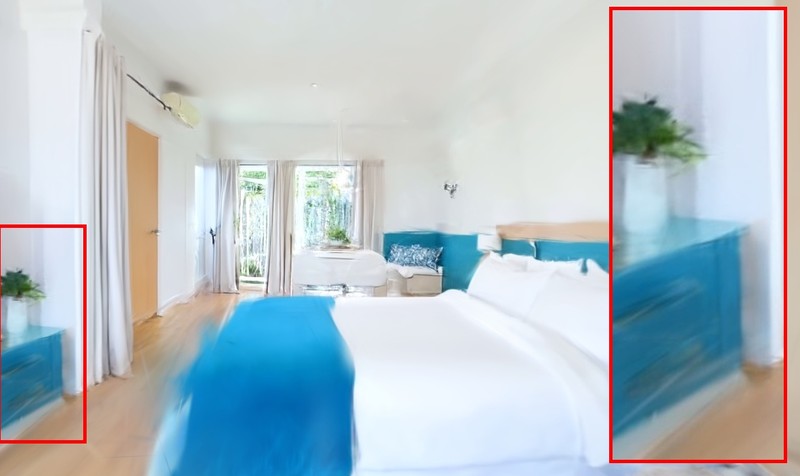}
        & \includegraphics[width=0.19\textwidth]{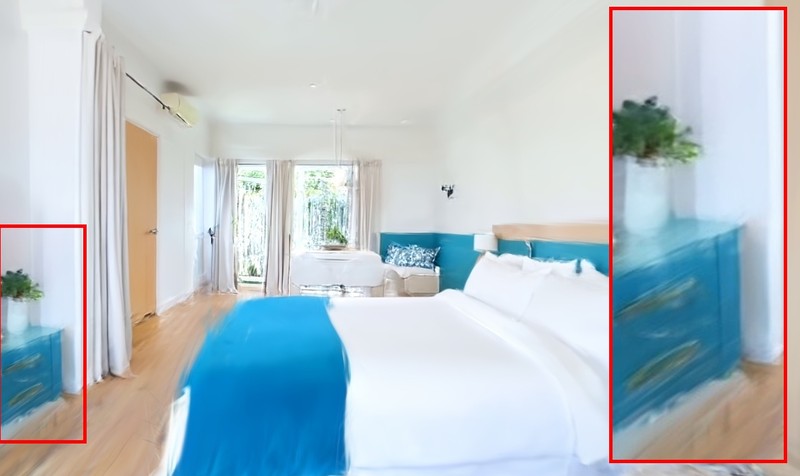}
        & \includegraphics[width=0.19\textwidth]{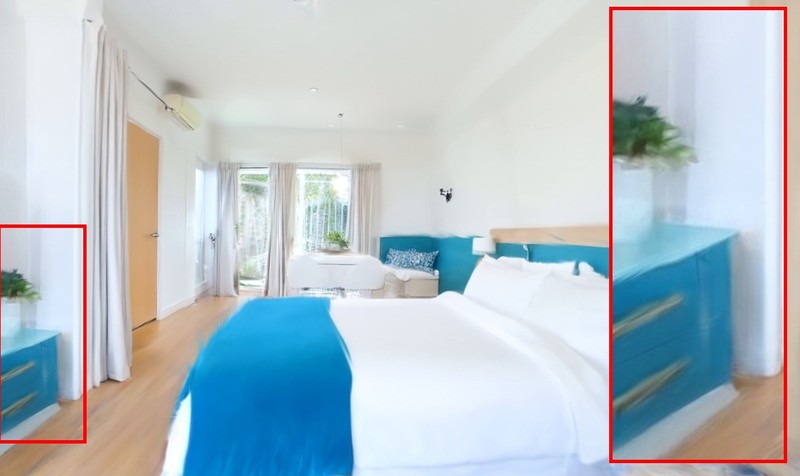} \\

        \includegraphics[width=0.19\textwidth]{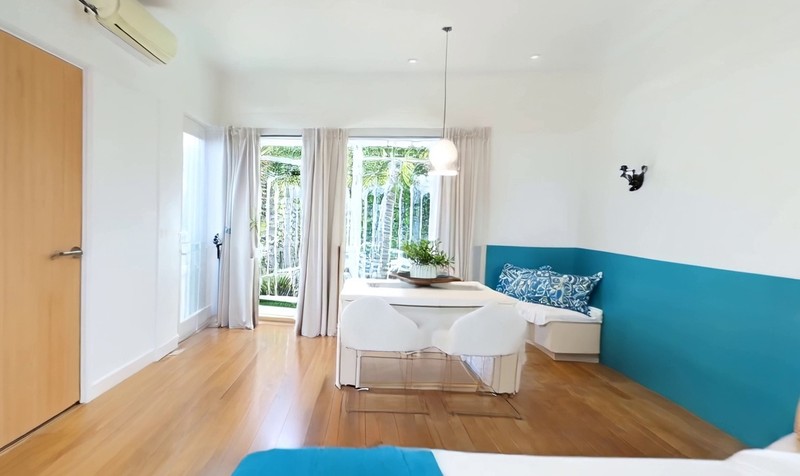}
        &
        & \includegraphics[width=0.19\textwidth]{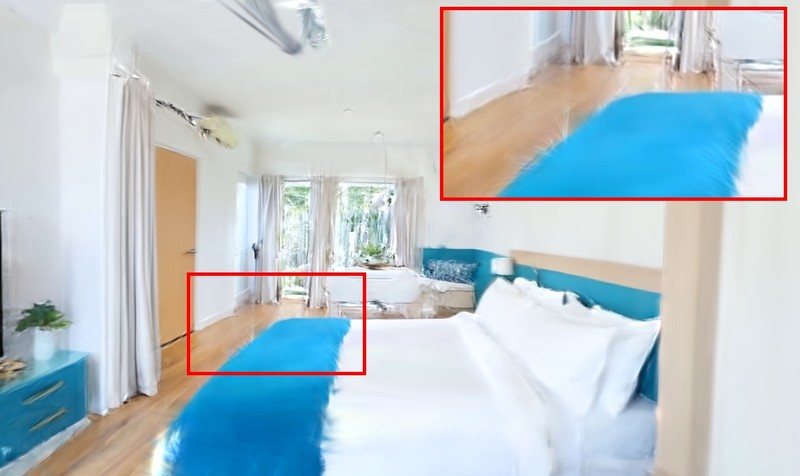}
        & \includegraphics[width=0.19\textwidth]{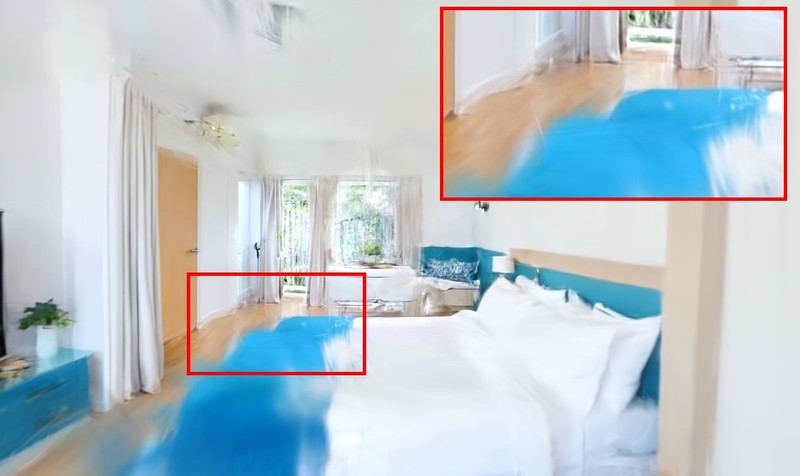}
        & \includegraphics[width=0.19\textwidth]{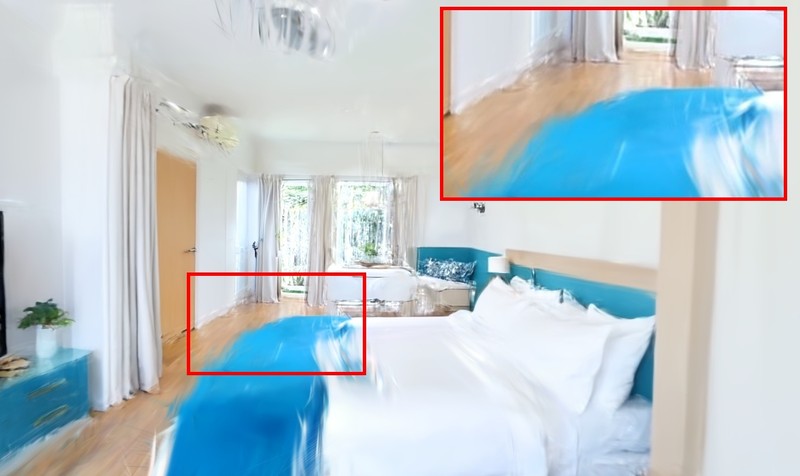}
        & \includegraphics[width=0.19\textwidth]{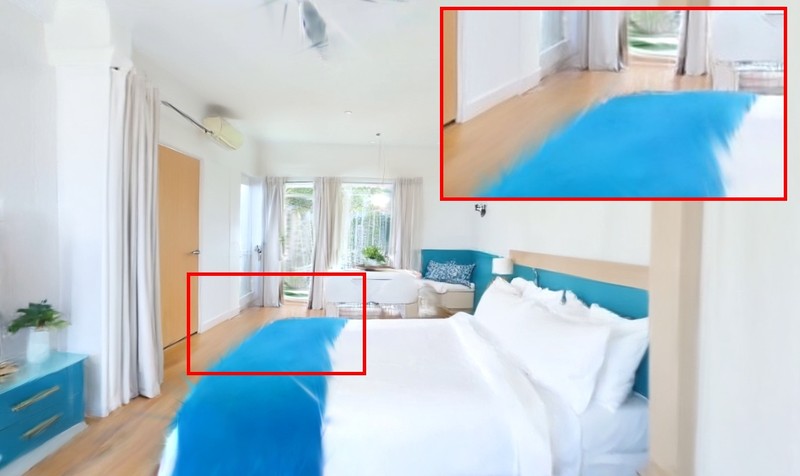} \\
        
    \end{tabular}
    \vspace{-4mm}
    \caption{\textbf{Single video 3D reconstruction.}     
    We generate videos with SEVA \cite{zhou2025stable} (top), Gen3C \cite{ren2025gen3c} (mid), Wan \cite{wan2025wanopenadvancedlargescale} (bottom) and 3D reconstruct these frames.
    Inconsistencies in the generations lead to blurry textures for the baselines compared to the corresponding video, and to floating artifacts from novel views.
    In contrast, our method creates 3D consistent worlds with high fidelity beyond generated perspectives.
    }
    \label{fig:qual_single_suppl2}
\end{figure*}

We show additional results on the single video 3D reconstruction task in \Cref{fig:qual_single_suppl1,fig:qual_single_suppl2}.
Our method renders sharper and more detailed textures that are of comparable visual fidelity as the input video frames, just in a 3D consistent space.
In contrast, the baselines do not correct for the generative drift inherent in the generated frames, which leads to blurrier renderings (especially noticeable from novel perspectives).
Our method optimizes the scenes in a \textit{non-rigid aware} fashion which leads to greatly improved viewpoint stability, i.e., we can explore the 3D worlds from novel perspectives while retaining the high visual fidelity of the video frames.
Our method effectively turns any video diffusion model into a 3D world generator that allows persistent, high quality, and real-time rendering.

\begin{figure*}
    \centering
    \setlength{\tabcolsep}{1pt}
    \renewcommand{\arraystretch}{1.1}

    \begin{tabular}{c | c | c c c c}
        {\fontsize{8}{9}\selectfont Video Frames} &
        &
        {\fontsize{8}{9}\selectfont DA3~\cite{lin2025depth}} &
        {\fontsize{8}{9}\selectfont 3DGS-MCMC~\cite{kheradmand20243d}} &
        {\fontsize{8}{9}\selectfont VGGT-X$^\dagger$~\cite{liu2025vggt}} &
        {\fontsize{8}{9}\selectfont Ours} \\

        \midrule

        \includegraphics[width=0.19\textwidth]{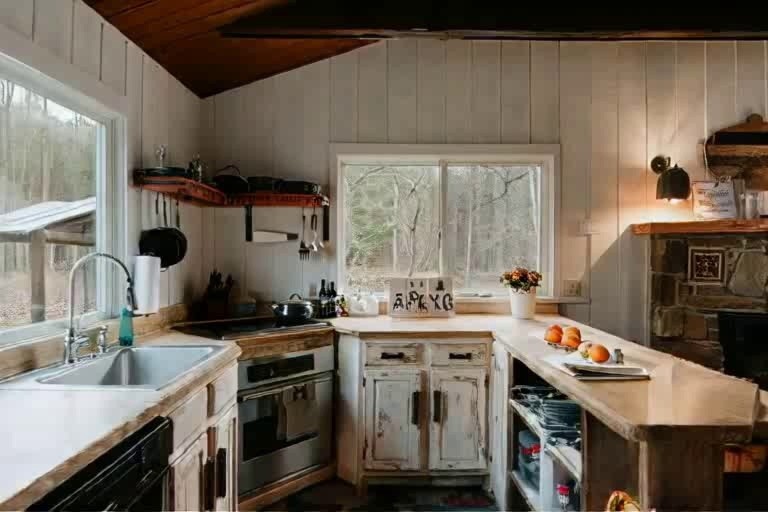}
        & \raisebox{0.0\height}{\rotatebox{90}{\fontsize{8}{9}\selectfont Input}}
        & \includegraphics[width=0.19\textwidth]{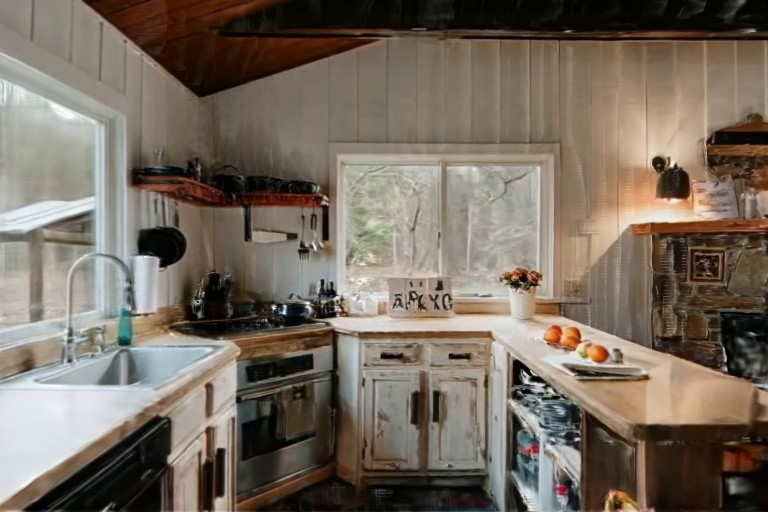}
        & \includegraphics[width=0.19\textwidth]{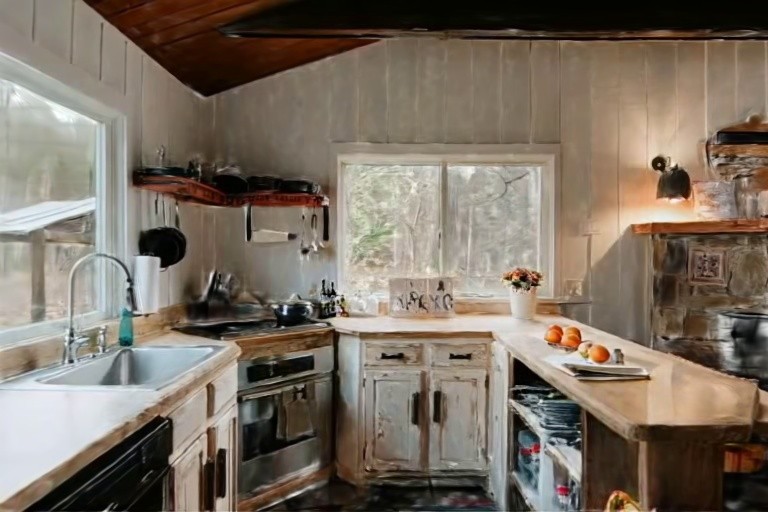}
        & \includegraphics[width=0.19\textwidth]{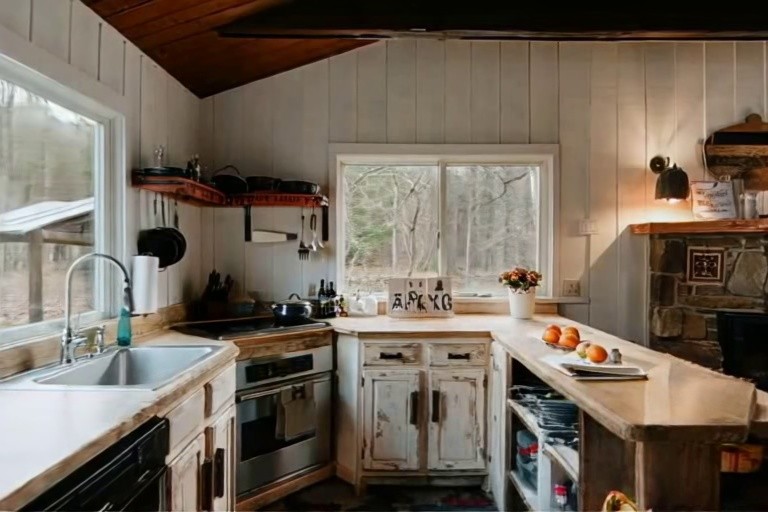}
        & \includegraphics[width=0.19\textwidth]{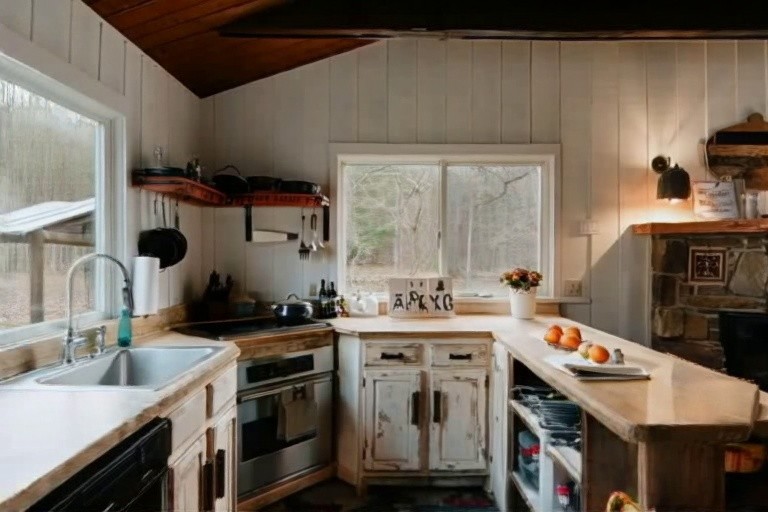} \\

        \includegraphics[width=0.19\textwidth]{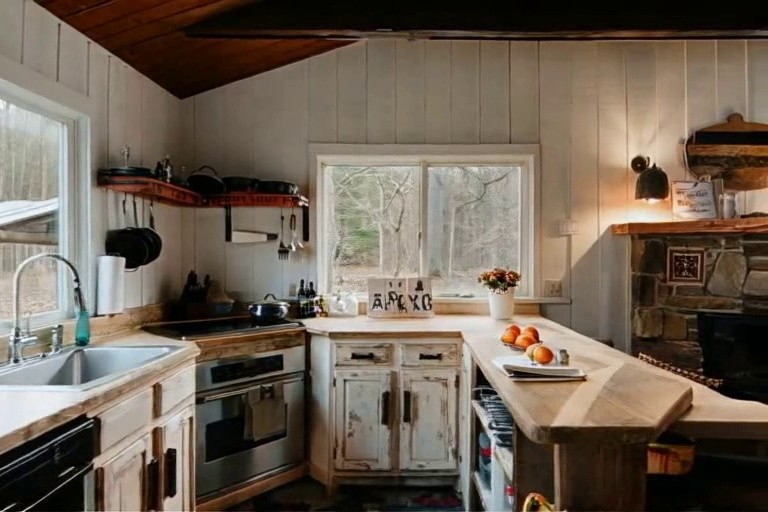}
        & \multirow{2}{*}{\rotatebox{90}{\fontsize{8}{9}\selectfont Novel Views}}
        & \includegraphics[width=0.19\textwidth]{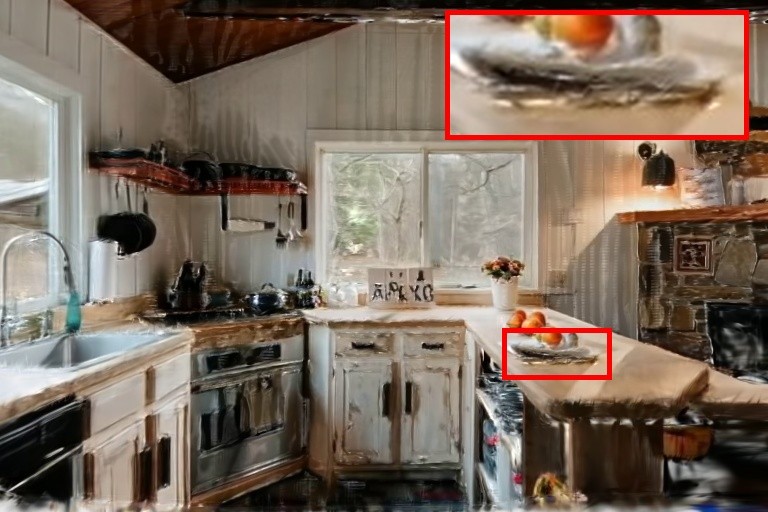}
        & \includegraphics[width=0.19\textwidth]{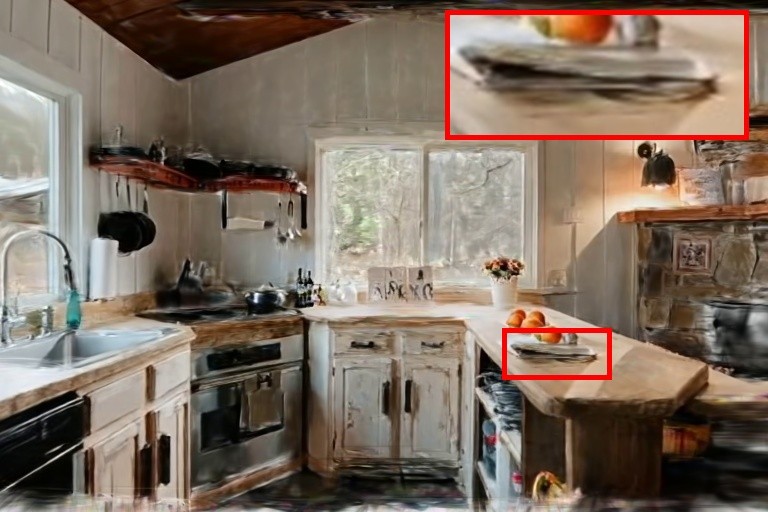}
        & \includegraphics[width=0.19\textwidth]{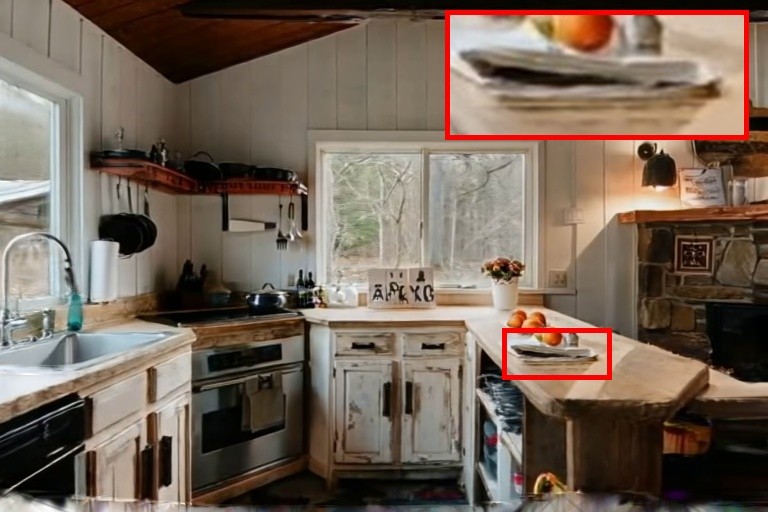}
        & \includegraphics[width=0.19\textwidth]{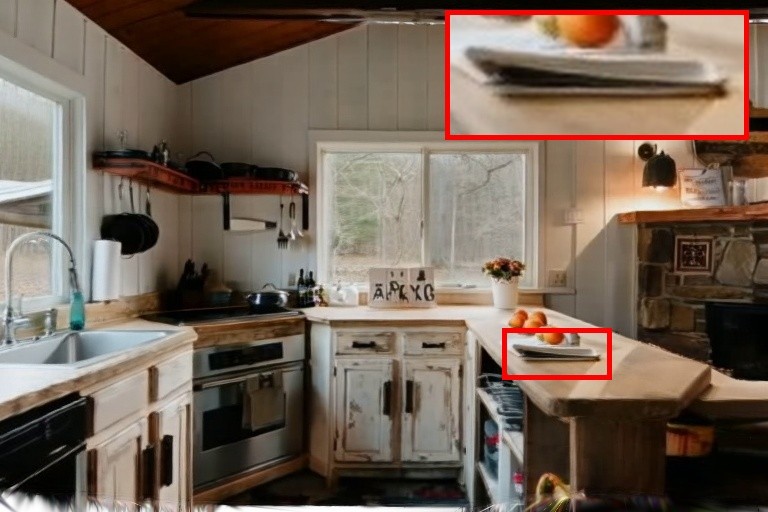} \\

        \includegraphics[width=0.19\textwidth]{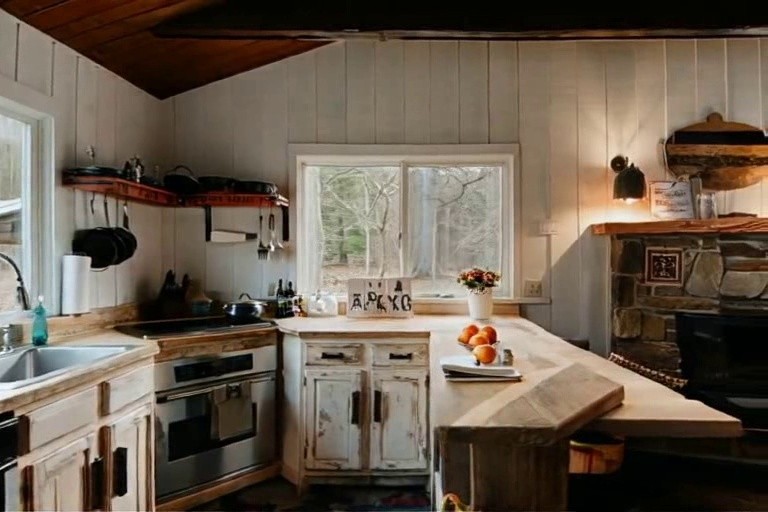}
        &
        & \includegraphics[width=0.19\textwidth]{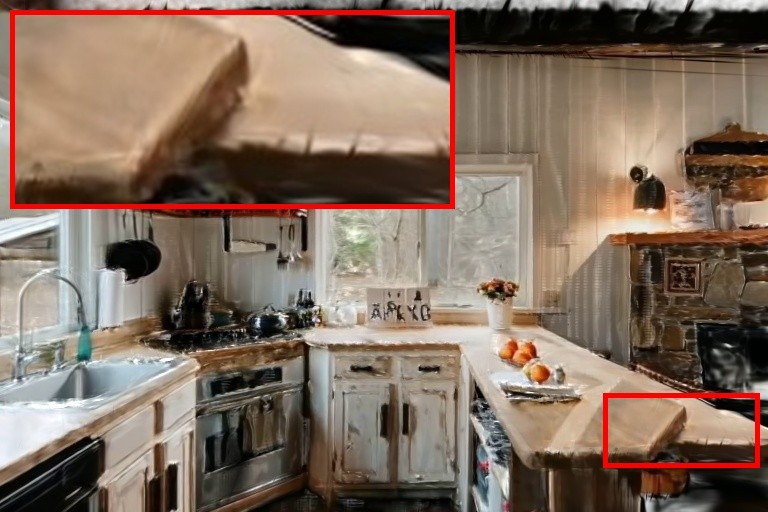}
        & \includegraphics[width=0.19\textwidth]{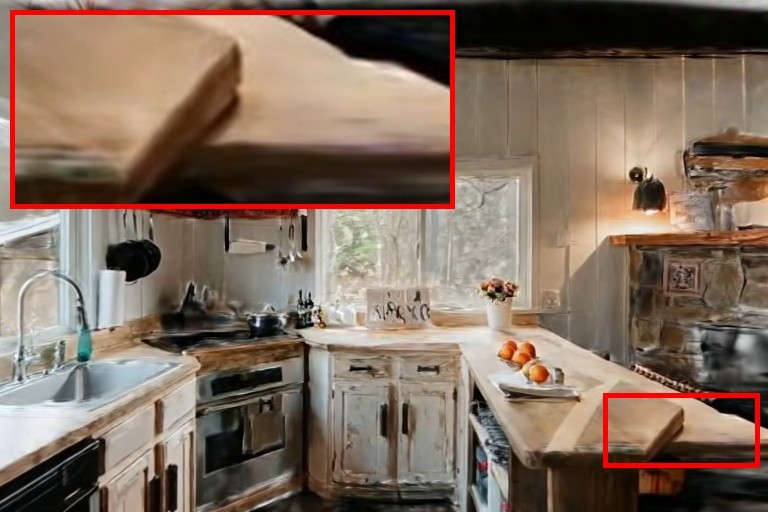}
        & \includegraphics[width=0.19\textwidth]{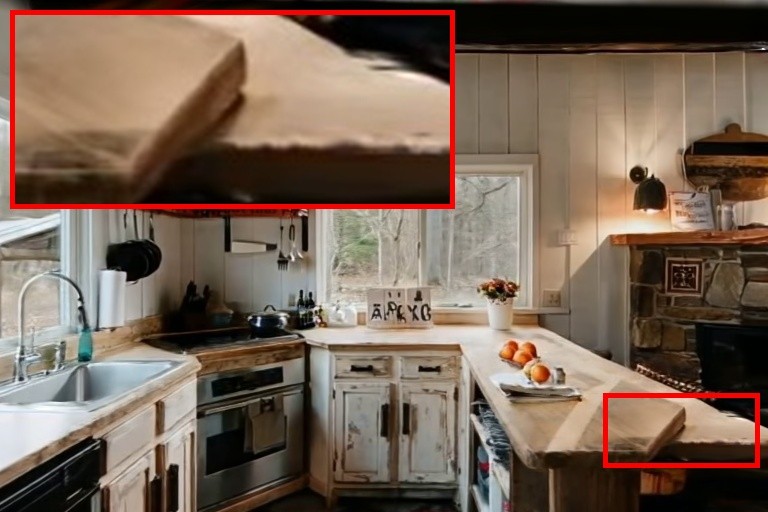}
        & \includegraphics[width=0.19\textwidth]{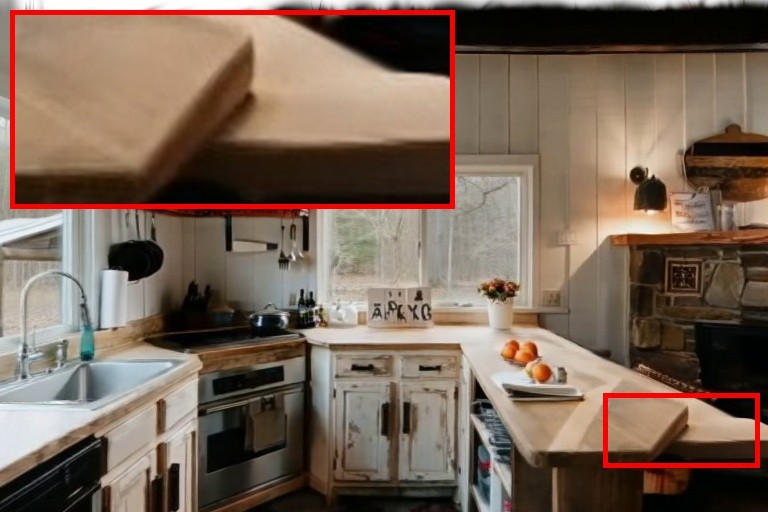} \\

        \midrule

        \includegraphics[width=0.19\textwidth]{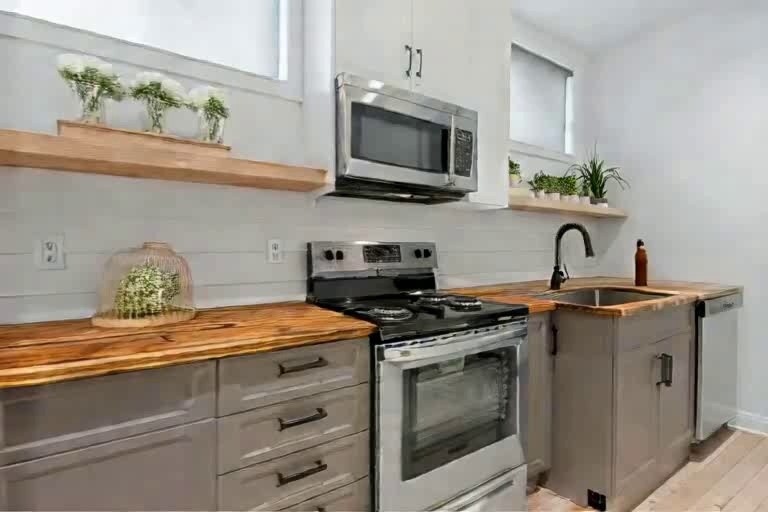}
        & \raisebox{0.0\height}{\rotatebox{90}{\fontsize{8}{9}\selectfont Input}}
        & \includegraphics[width=0.19\textwidth]{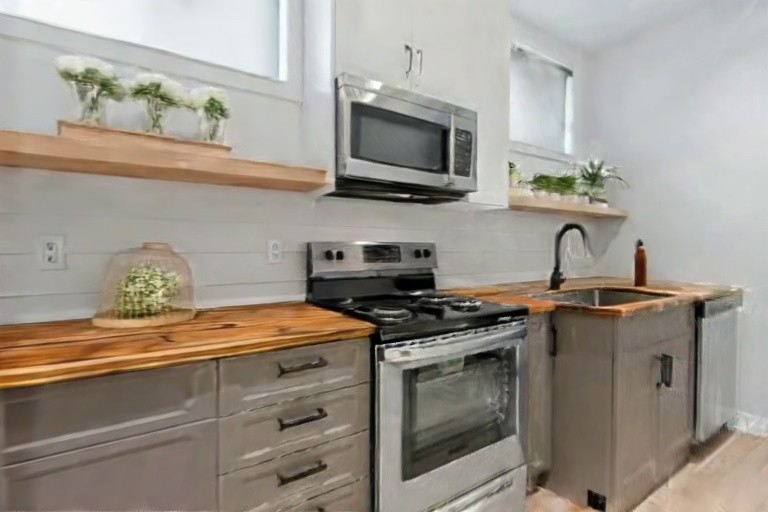}
        & \includegraphics[width=0.19\textwidth]{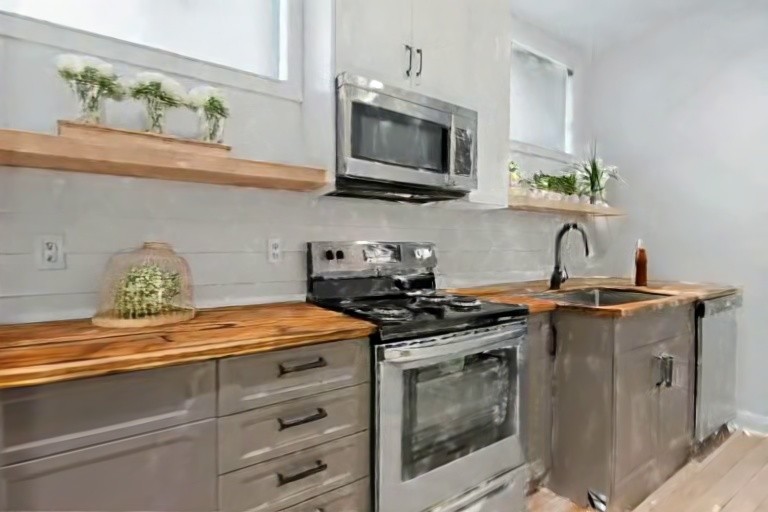}
        & \includegraphics[width=0.19\textwidth]{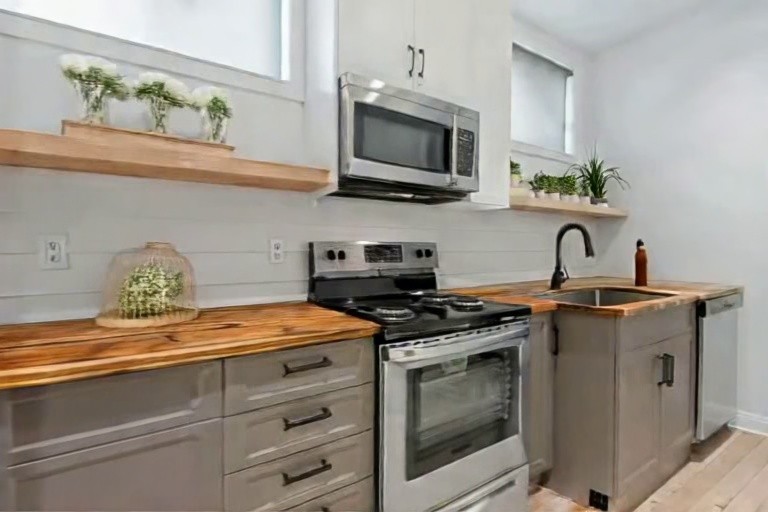}
        & \includegraphics[width=0.19\textwidth]{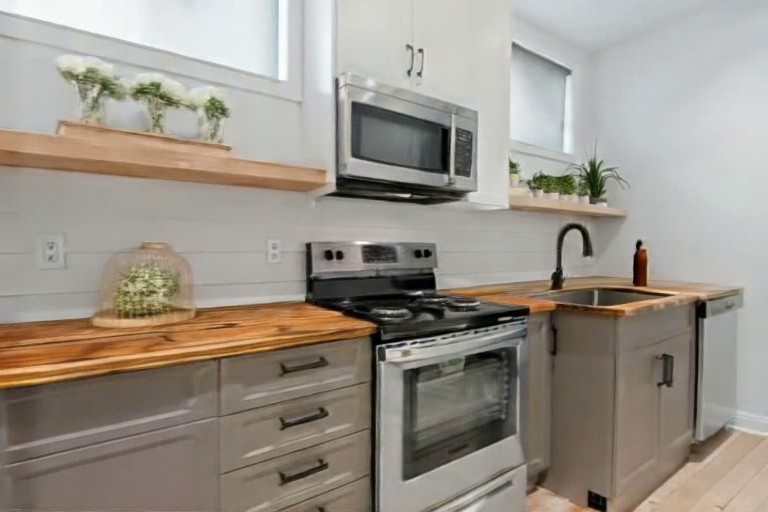} \\

        \includegraphics[width=0.19\textwidth]{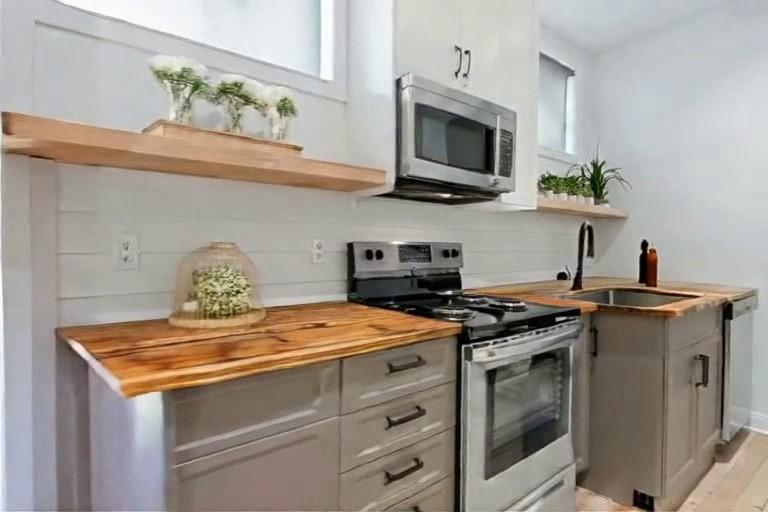}
        & \multirow{2}{*}{\rotatebox{90}{\fontsize{8}{9}\selectfont Novel Views}}
        & \includegraphics[width=0.19\textwidth]{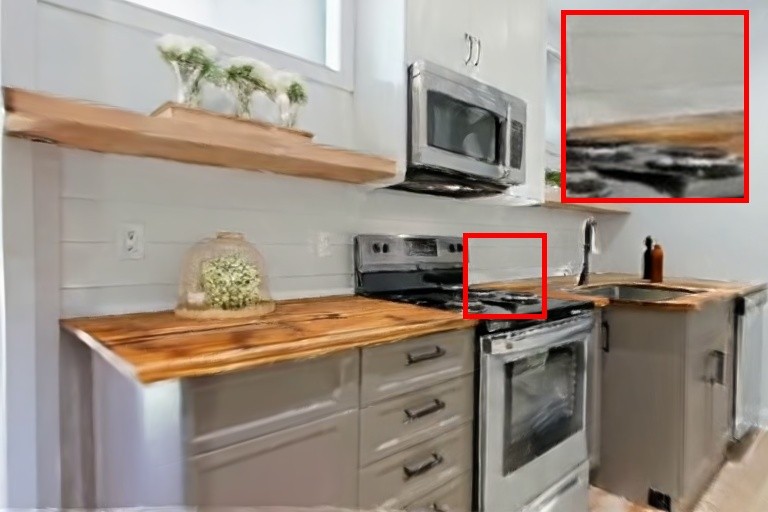}
        & \includegraphics[width=0.19\textwidth]{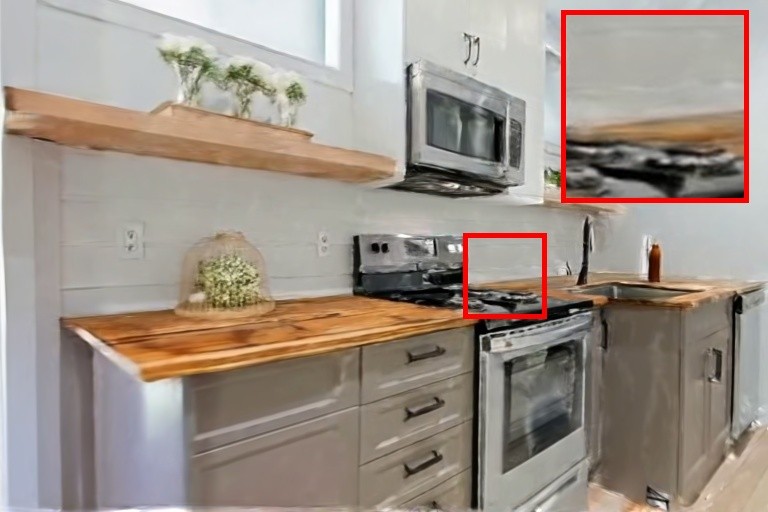}
        & \includegraphics[width=0.19\textwidth]{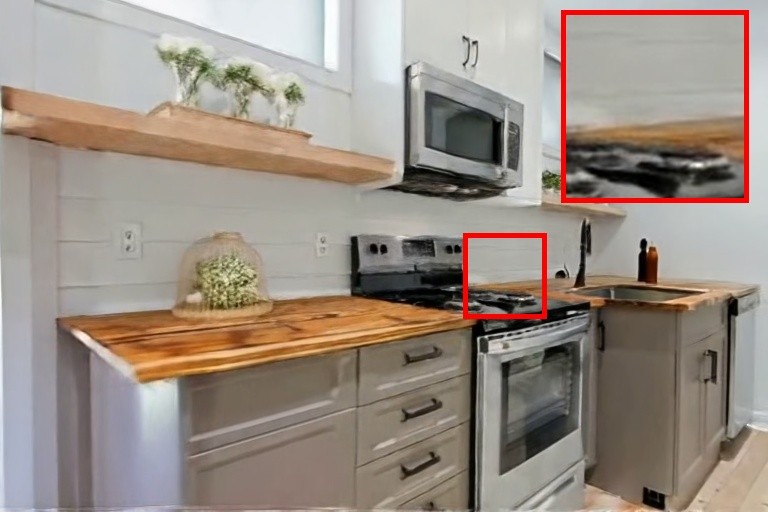}
        & \includegraphics[width=0.19\textwidth]{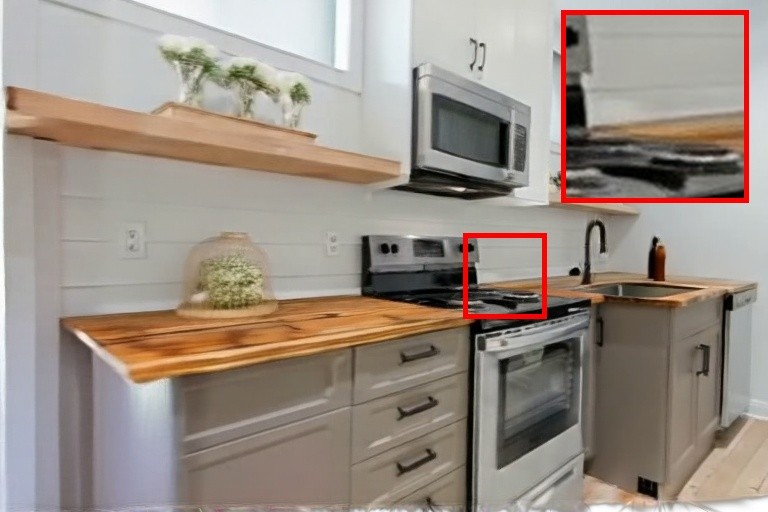} \\

        \includegraphics[width=0.19\textwidth]{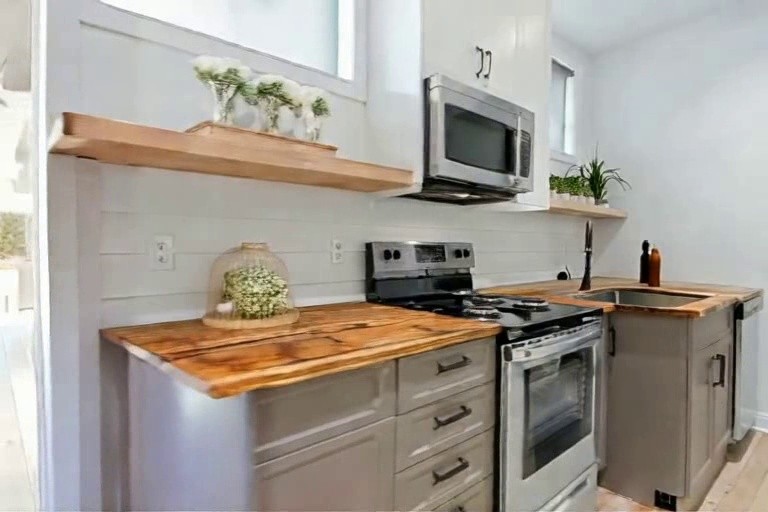}
        &
        & \includegraphics[width=0.19\textwidth]{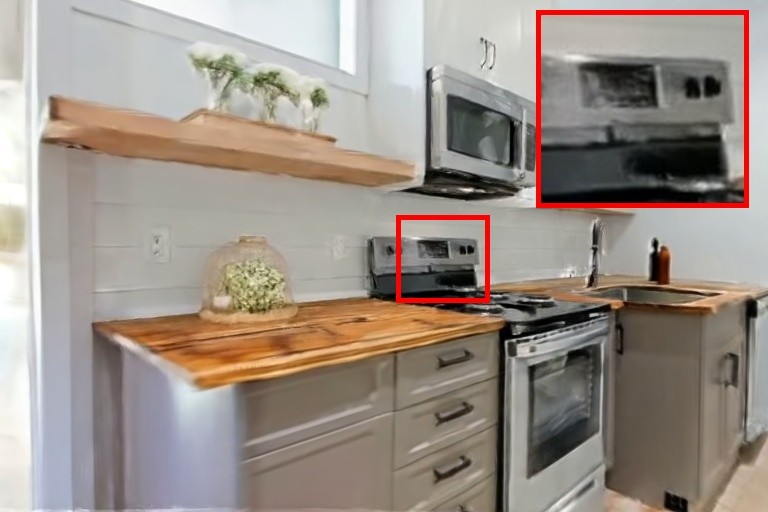}
        & \includegraphics[width=0.19\textwidth]{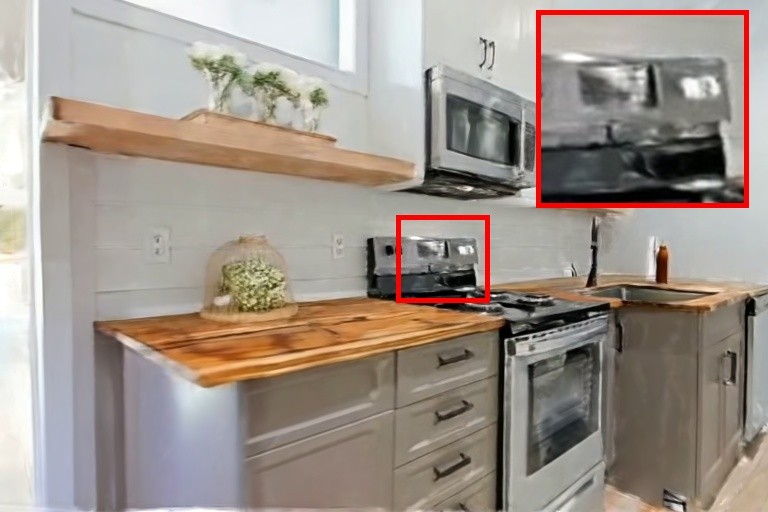}
        & \includegraphics[width=0.19\textwidth]{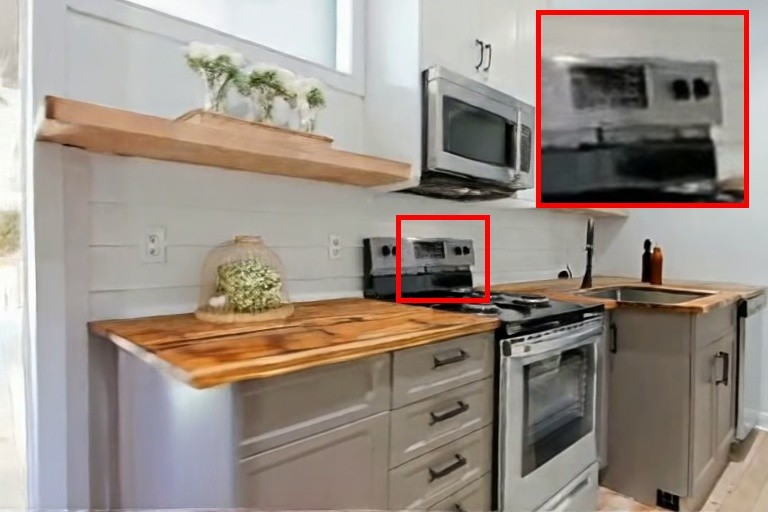}
        & \includegraphics[width=0.19\textwidth]{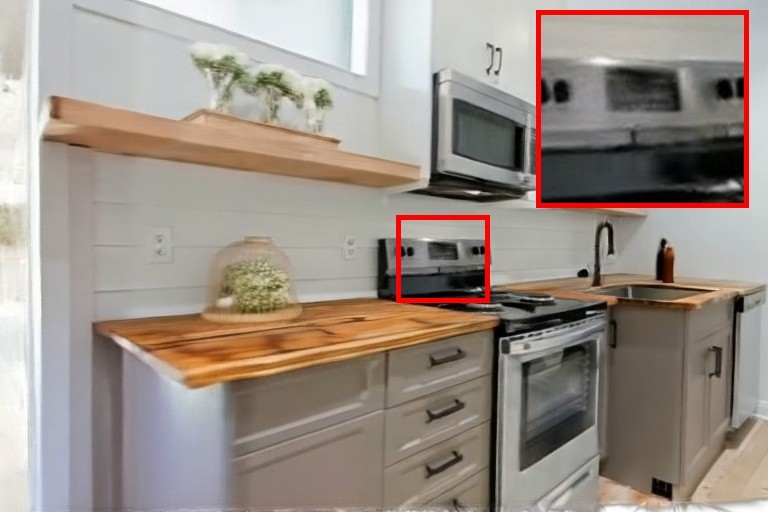} \\
        
    \end{tabular}

    \begin{tabular}{c | c c c c}

        \midrule
        \vspace{-0.5mm}

        &
        {\fontsize{8}{9}\selectfont Voyager \cite{huang2025voyager}} &
        {\fontsize{8}{9}\selectfont VGGT-X$^\dagger$ \cite{liu2025vggt}} &
        {\fontsize{8}{9}\selectfont DA3 \cite{lin2025depth}} &
        {\fontsize{8}{9}\selectfont Ours} \\

        \midrule

        \multirow{2}{*}{\rotatebox{90}{\fontsize{8}{9}\selectfont Pointcloud}}
        & \includegraphics[width=0.245\textwidth]{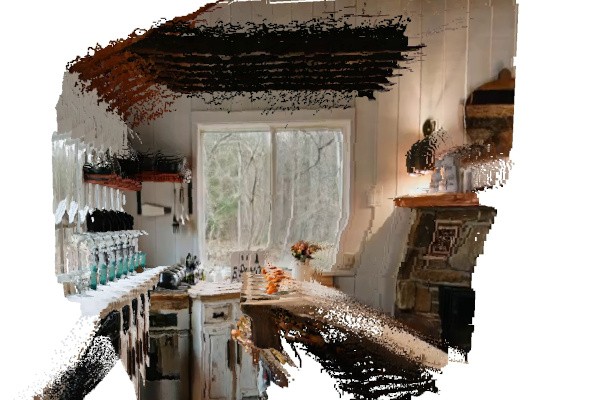}
        & \includegraphics[width=0.245\textwidth]{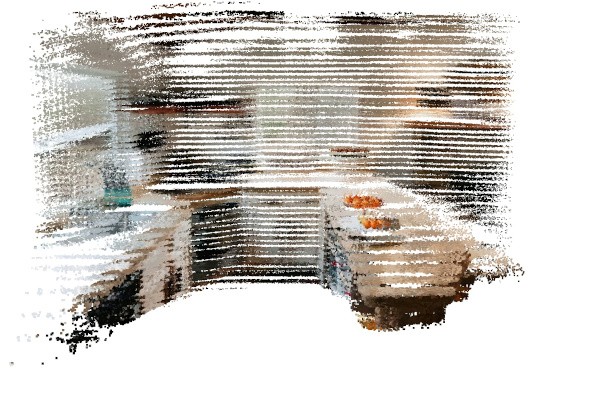}
        & \includegraphics[width=0.245\textwidth]{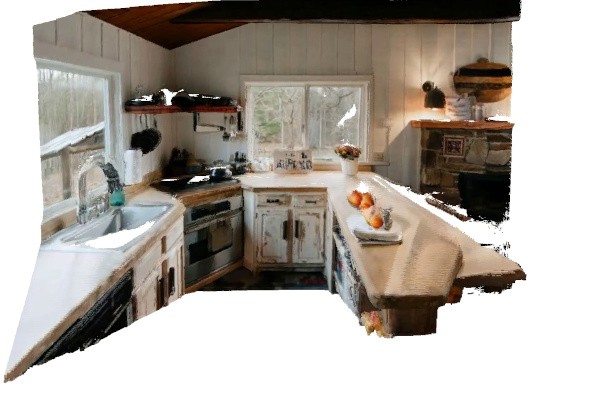}
        & \includegraphics[width=0.245\textwidth]{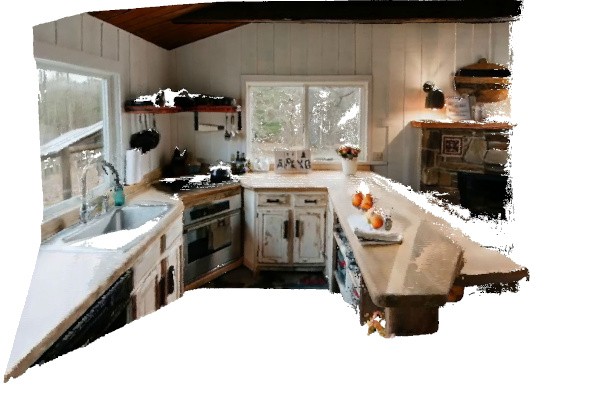} \\

        & \includegraphics[width=0.245\textwidth]{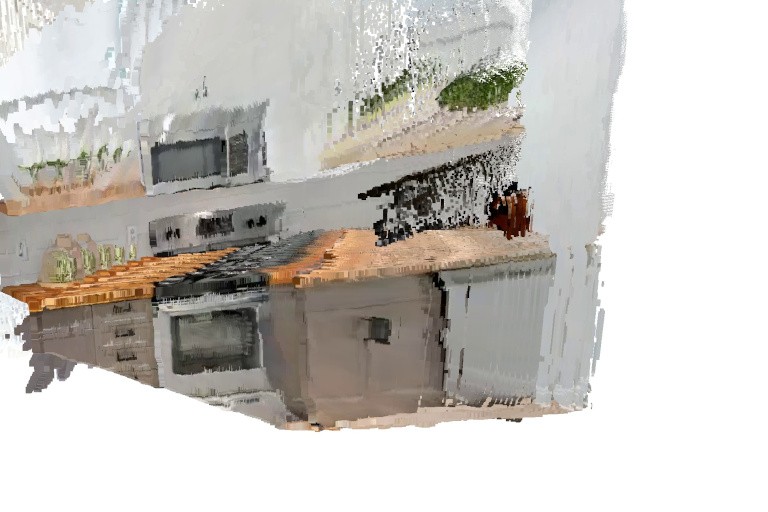}
        & \includegraphics[width=0.245\textwidth]{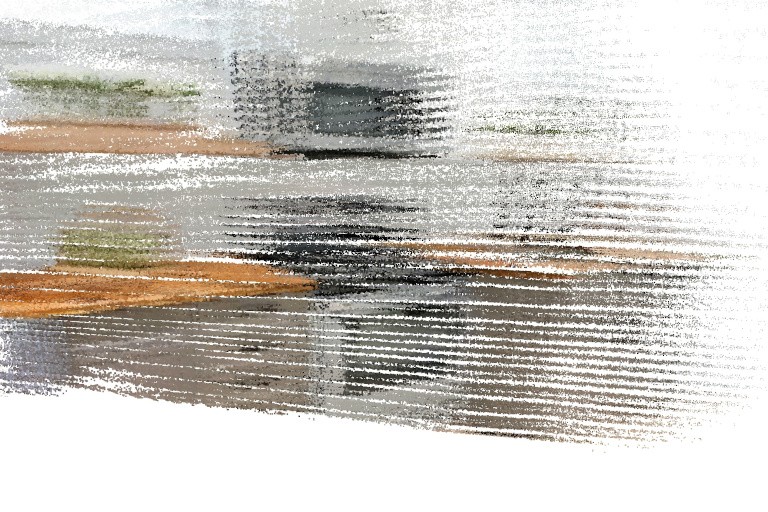}
        & \includegraphics[width=0.245\textwidth]{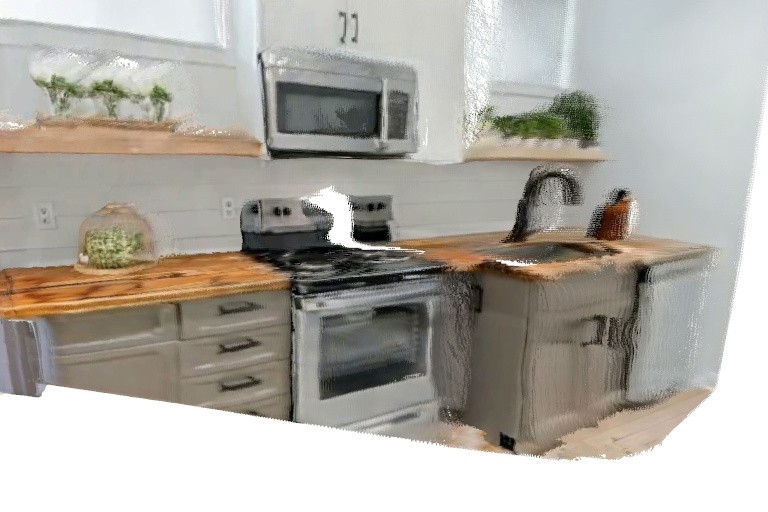}
        & \includegraphics[width=0.245\textwidth]{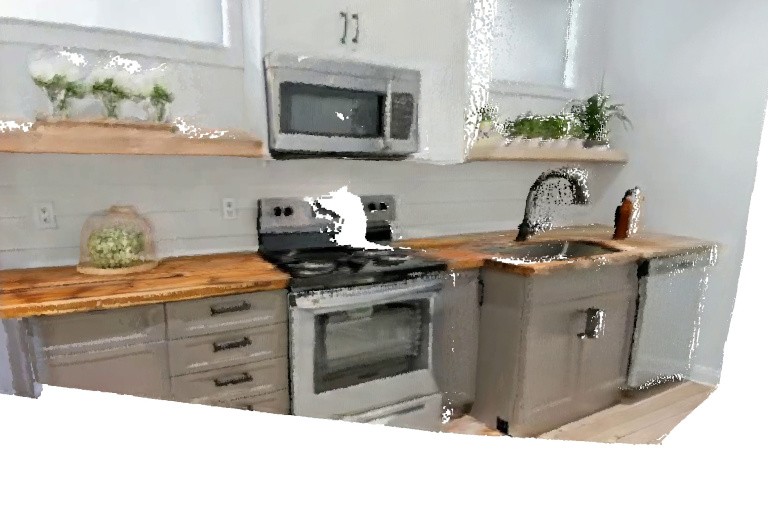} \\

    \end{tabular}

    \vspace{-2mm}
    \caption{\textbf{Single video 3D reconstruction.}     
    We generate videos with Voyager \cite{huang2025voyager} and 3D reconstruct these frames.
    Our method optimizes consistent worlds from inconsistent generated frames. 
    Compared to baselines, the renderings are of higher visual fidelity from both input and novel views.
    The pointcloud visualizations demonstrate that our method obtains aligned surfaces with compelling textures, whereas the baselines produce sparser or less aligned geometry.
    }
    \label{fig:qual_single_suppl_voyager}
\end{figure*}

We additionally compare the recent method Voyager \cite{huang2025voyager} more closely in \Cref{fig:qual_single_suppl_voyager}.
They propose a joint modeling of RGB-D sequences in the latent space of a video diffusion model, which allows for explicit 3D consistency supervision during training and thus yields improved consistency in the generated output sequence.
Nevertheless, generative drift still remains noticeable in the reconstructed and rendered scenes (\Cref{fig:qual_single_suppl_voyager} top).
Additionally, the generated RGB-D sequence still contains unaligned and overlapping surfaces with similar artifacts as the DA3~\cite{lin2025depth} predictions.
This underlines the motivation of our approach, which proposes a lightweight alignment on the reconstruction side (instead of further finetuning of a generative model).
Our pointcloud geometry contains single, aligned surfaces with compelling texture quality, which serves as strong initialization for photometric optimization with Gaussian Splatting \cite{kerbl20233d, huang20242d}.

\subsection{Large-Scale World Generation}

We show additional results on the large-scale world generation task in \Cref{fig:qual_multi_suppl1,fig:qual_multi_suppl2,fig:qual_multi_suppl3}.
Concretely, we generate and then reconstruct up to 32 video sequences with SEVA \cite{zhou2025stable} via the progressive scene expansion strategy of \cite{schneider_hoellein_2025_worldexplorer}.
The generative drift grows larger across multiple videos and thus the 3D reconstructions suffer from increased consistency problems.
While the baselines can depict complete and diverse worlds, the amount of exploration they enable is still limited (i.e., novel views far beyond the training poses suffer from extensive floating artifacts).
In contrast, our worlds remain consistent and retain a high rendering fidelity, even from extreme novel perspectives.

\newcommand{\cropppimg}[1]{%
  \adjustbox{
    width=0.245\textwidth,
    trim=0 {.05\height} 0 {.21\height},
    clip
  }{\includegraphics{#1}}%
}

\newcommand{\cropppimgg}[1]{%
  \adjustbox{
    width=0.245\textwidth,
    trim=0 {.05\height} 0 {.21\height},
    clip
  }{\includegraphics{#1}}%
}

\newcommand{\cropppimggg}[1]{%
  \adjustbox{
    width=0.33\textwidth,
    trim=0 {.1\height} 0 {.16\height},
    clip
  }{\includegraphics{#1}}%
}

\newcommand{\cropppimgggg}[1]{%
  \adjustbox{
    width=0.33\textwidth,
    trim=0 {.13\height} 0 {.13\height},
    clip
  }{\includegraphics{#1}}%
}

\begin{figure*}
    \centering
    \setlength{\tabcolsep}{1pt}
    \renewcommand{\arraystretch}{1.1}

    \resizebox{0.97\textwidth}{!}{
    \begin{minipage}{\textwidth}
    \begin{tabular}{c | c | c c c}

        &
        {\fontsize{8}{9}\selectfont Video Frames} &
        {\fontsize{8}{9}\selectfont WorldExplorer \cite{schneider_hoellein_2025_worldexplorer}} &
        {\fontsize{8}{9}\selectfont VGGT-X$^\dagger$ \cite{liu2025vggt}} &
        {\fontsize{8}{9}\selectfont Ours} \\

        \midrule

        \rotatebox{90}{\fontsize{8}{9}\selectfont Input Pose}
        & \cropppimg{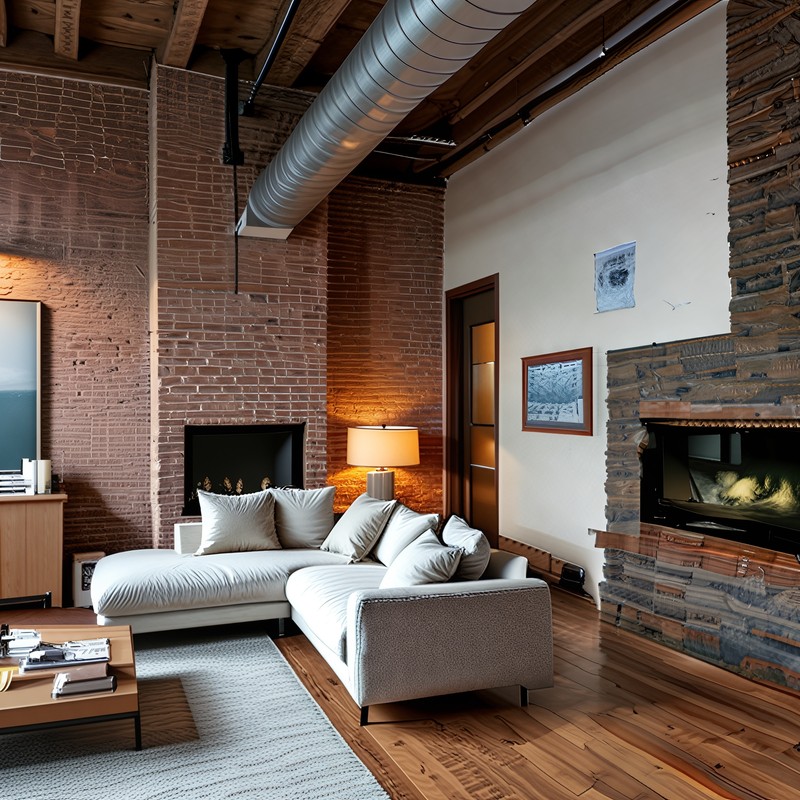}
        & \cropppimg{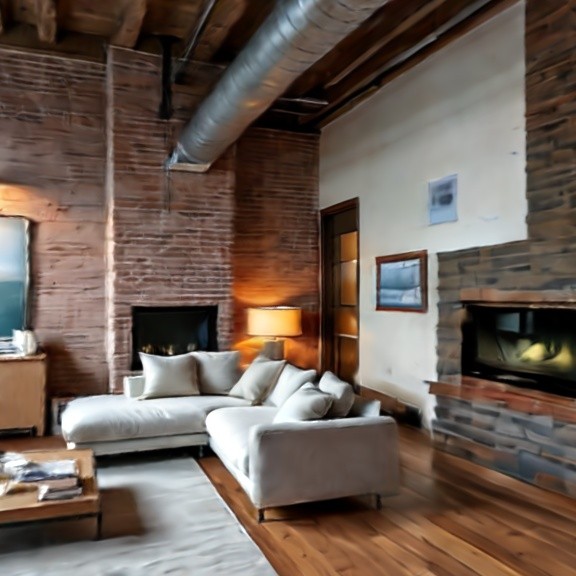}
        & \cropppimg{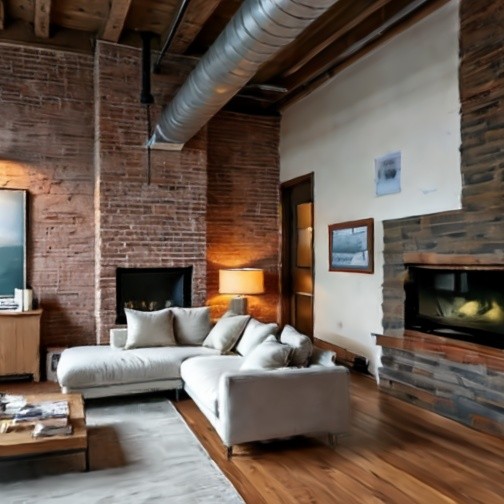}
        & \cropppimg{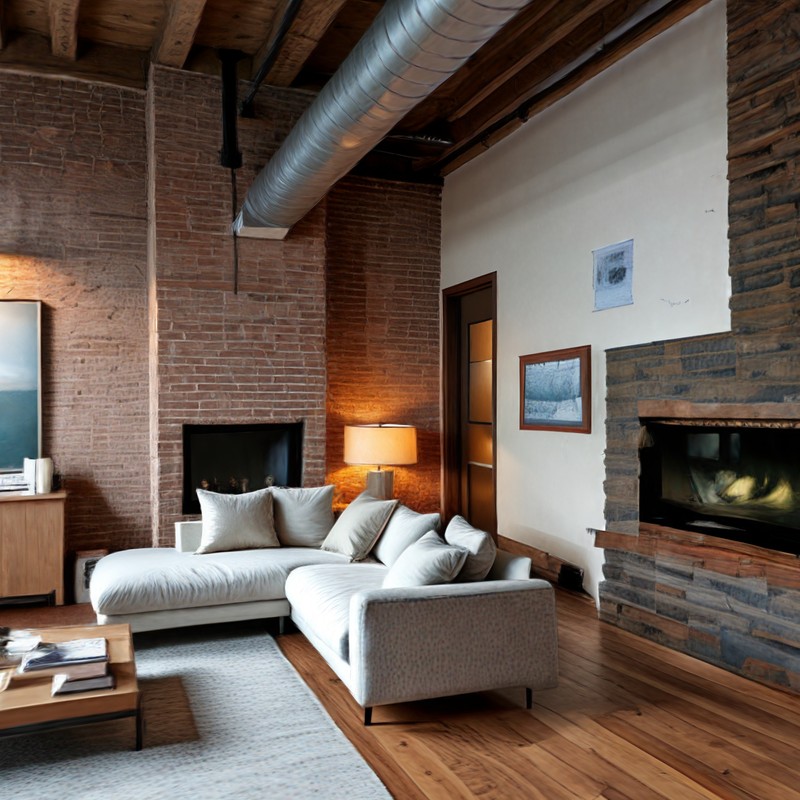} \\

        \rotatebox{90}{\fontsize{8}{9}\selectfont Input Pose}
        & \cropppimgg{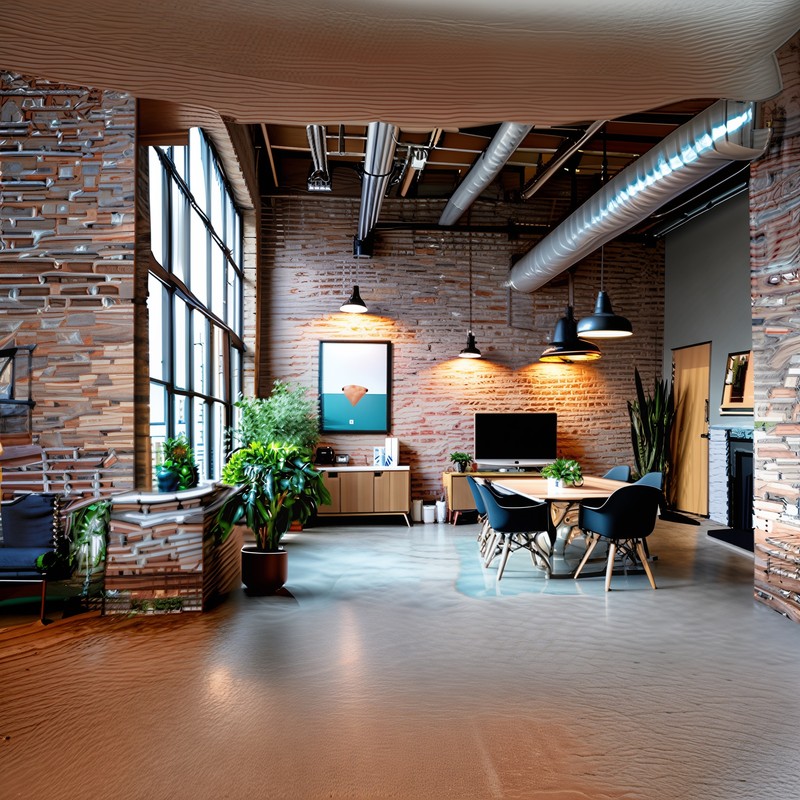}
        & \cropppimgg{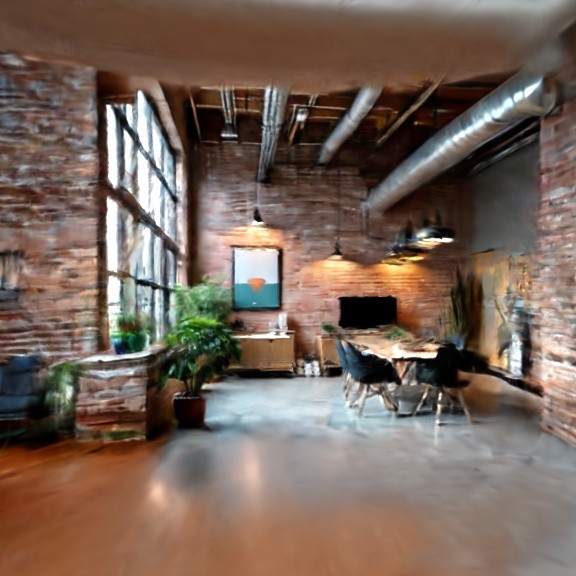}
        & \cropppimgg{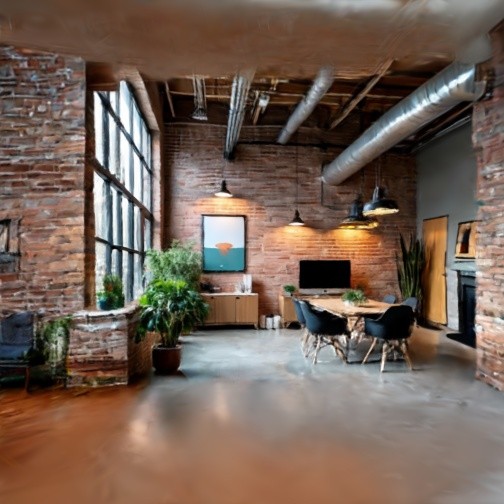}
        & \cropppimgg{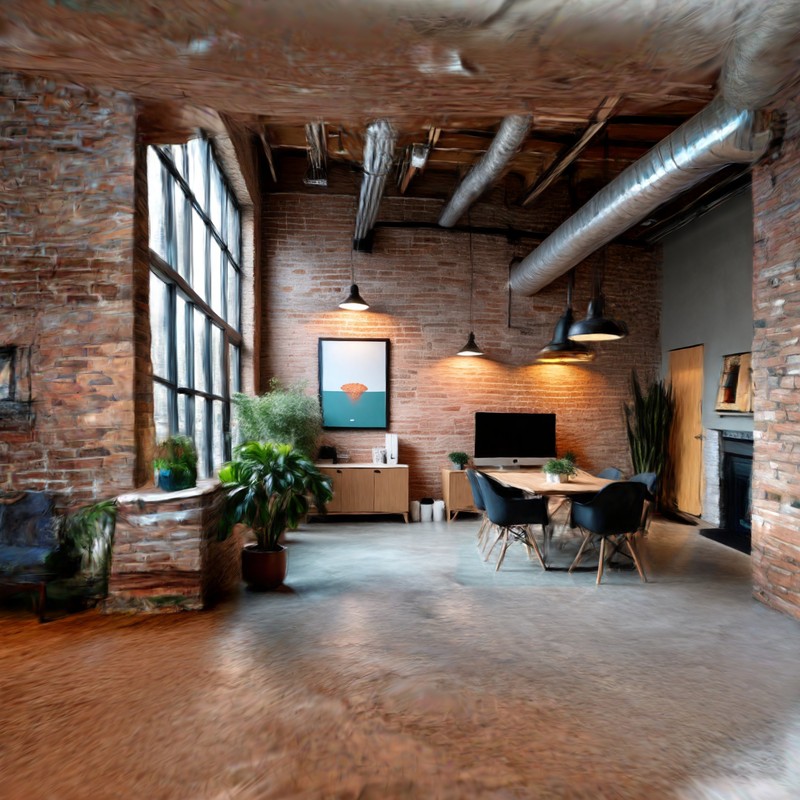} \\

    \end{tabular}

    \vspace{-1mm}

    \begin{tabular}{c | c c c}

        \midrule
        \vspace{-0.5mm}

        &
        {\fontsize{8}{9}\selectfont WorldExplorer \cite{schneider_hoellein_2025_worldexplorer}} &
        {\fontsize{8}{9}\selectfont VGGT-X$^\dagger$ \cite{liu2025vggt}} &
        {\fontsize{8}{9}\selectfont Ours} \\

        \midrule

        \rotatebox{90}{\fontsize{8}{9}\selectfont Novel Views}  
        & \cropppimggg{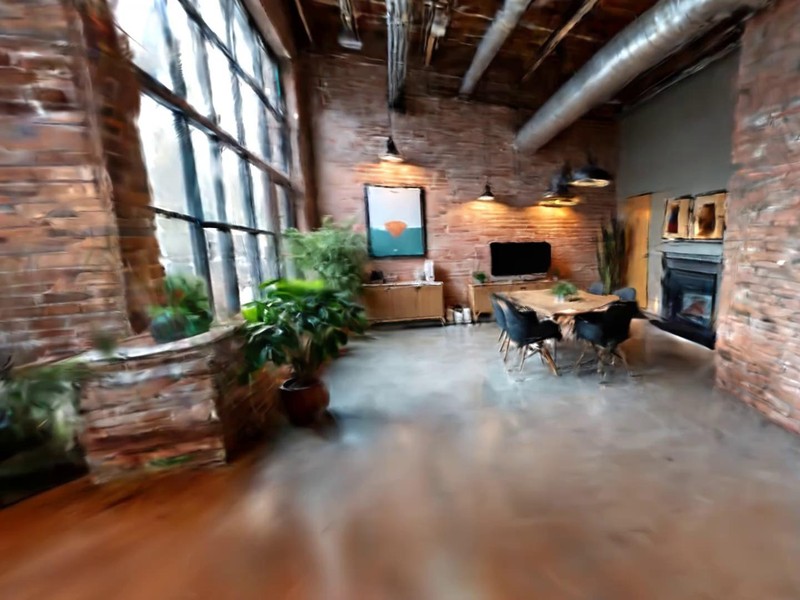}
        & \cropppimggg{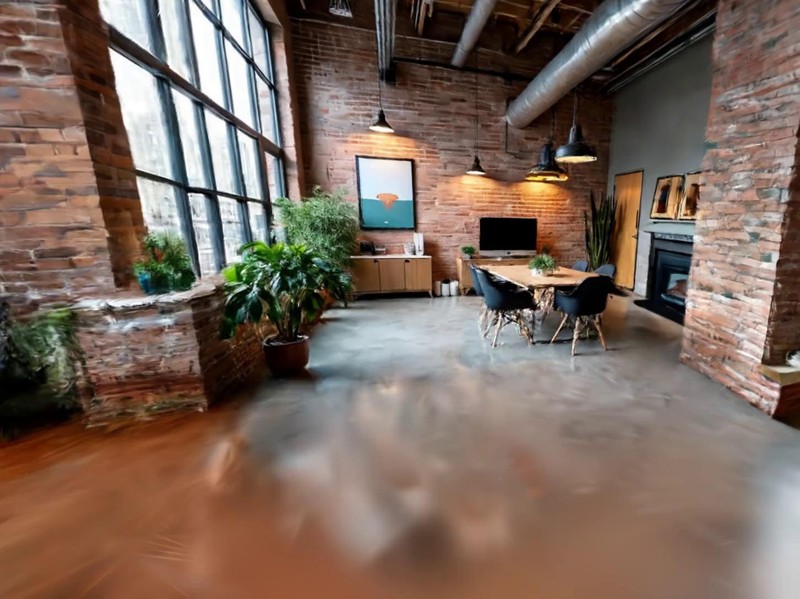}
        & \cropppimggg{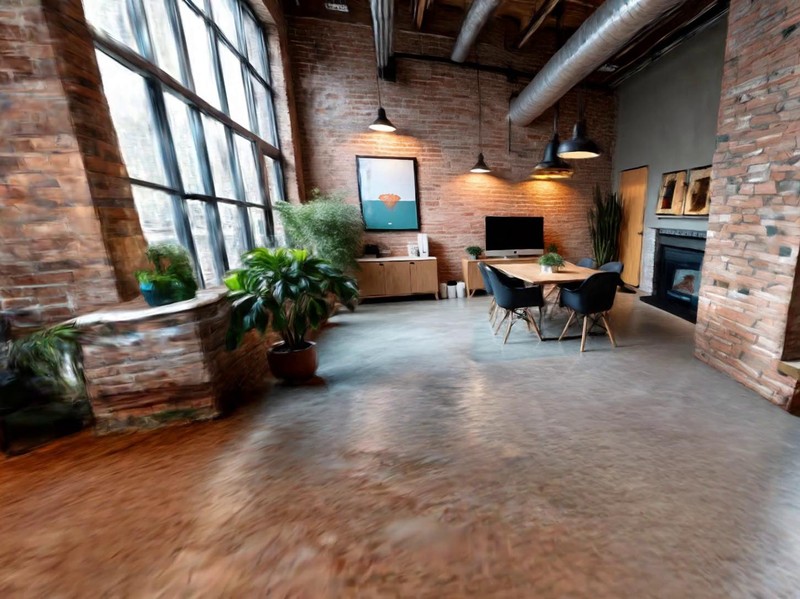} \\

        \rotatebox{90}{\fontsize{8}{9}\selectfont Novel Views}  
        & \cropppimgggg{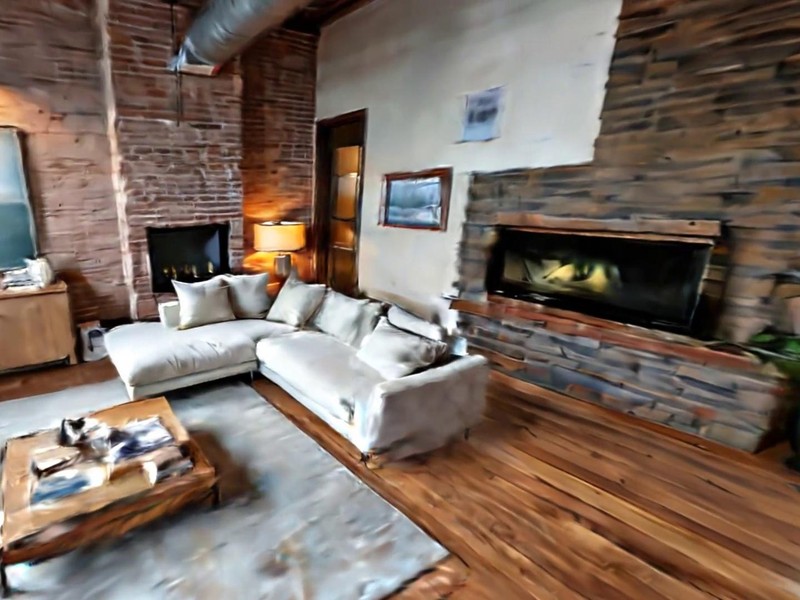}
        & \cropppimgggg{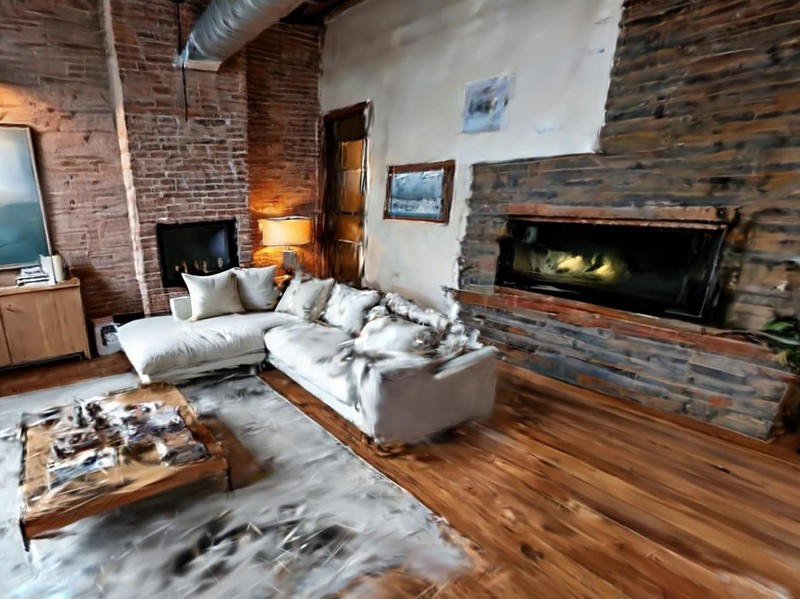}
        & \cropppimgggg{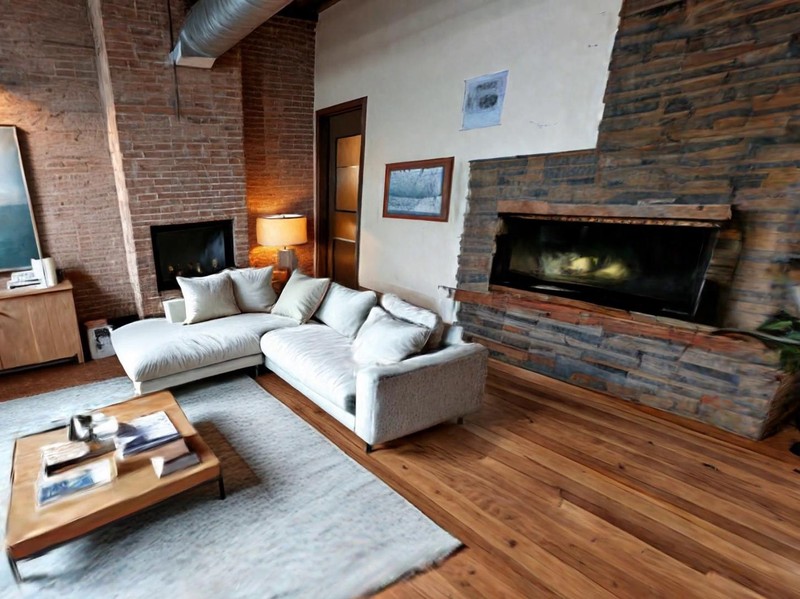} \\

    \end{tabular}
    \end{minipage}
    }

    \vspace{-4mm}
    \caption{\textbf{Large-scale 3D reconstructions.} 
    We compare rendering quality on input poses and novel views for entire 360 degree explorable scenes.
    Our method creates 3D consistent worlds with a high rendering fidelity far beyond the training views.
    Please see the supplementary video for animated flythroughs.
    }
    \label{fig:qual_multi_suppl1}
\end{figure*}

\newcommand{\croppppimg}[1]{%
  \adjustbox{
    width=0.245\textwidth,
    trim=0 {.03\height} 0 {.23\height},
    clip
  }{\includegraphics{#1}}%
}

\newcommand{\croppppimgg}[1]{%
  \adjustbox{
    width=0.245\textwidth,
    trim=0 {.05\height} 0 {.21\height},
    clip
  }{\includegraphics{#1}}%
}

\newcommand{\croppppimggg}[1]{%
  \adjustbox{
    width=0.33\textwidth,
    trim=0 {.03\height} 0 {.23\height},
    clip
  }{\includegraphics{#1}}%
}

\newcommand{\croppppimgggg}[1]{%
  \adjustbox{
    width=0.33\textwidth,
    trim=0 {.13\height} 0 {.13\height},
    clip
  }{\includegraphics{#1}}%
}

\begin{figure*}
    \centering
    \setlength{\tabcolsep}{1pt}
    \renewcommand{\arraystretch}{1.1}

    \resizebox{0.97\textwidth}{!}{
    \begin{minipage}{\textwidth}
    \begin{tabular}{c | c | c c c}

        &
        {\fontsize{8}{9}\selectfont Video Frames} &
        {\fontsize{8}{9}\selectfont WorldExplorer \cite{schneider_hoellein_2025_worldexplorer}} &
        {\fontsize{8}{9}\selectfont VGGT-X$^\dagger$ \cite{liu2025vggt}} &
        {\fontsize{8}{9}\selectfont Ours} \\

        \midrule

        \rotatebox{90}{\fontsize{8}{9}\selectfont Input Pose}
        & \croppppimg{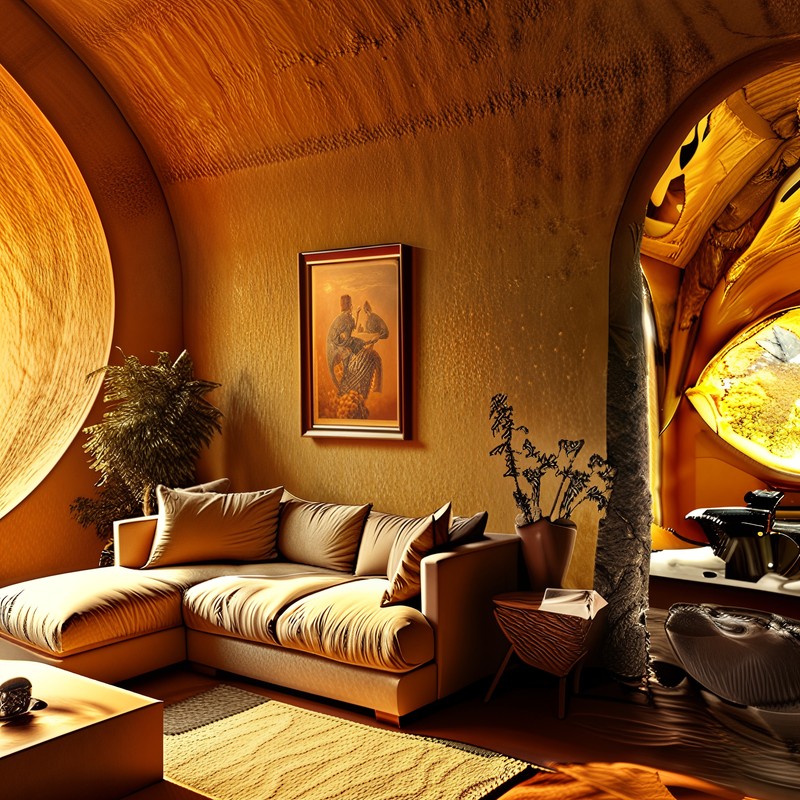}
        & \croppppimg{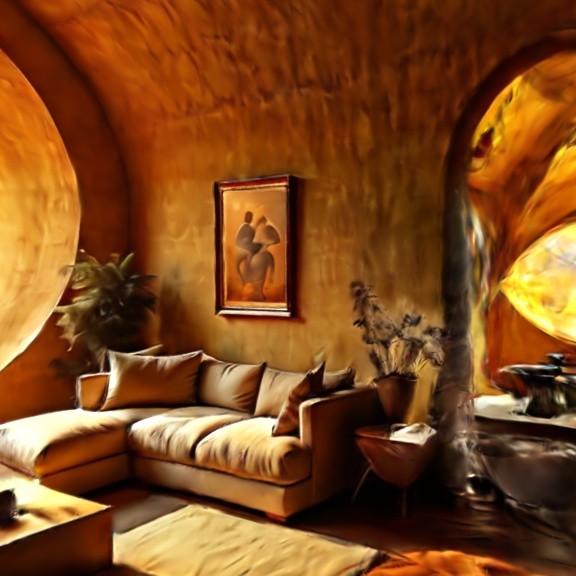}
        & \croppppimg{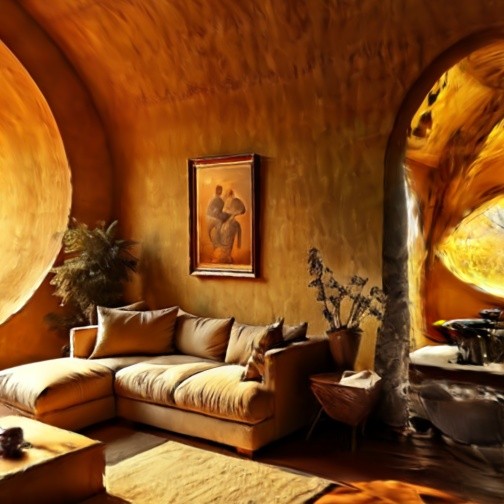}
        & \croppppimg{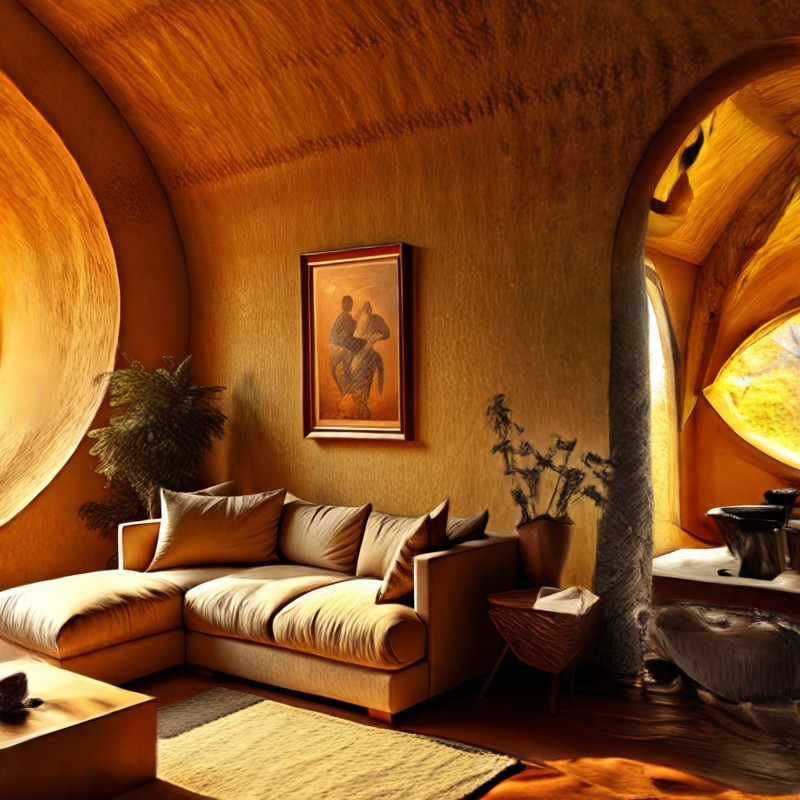} \\

        \rotatebox{90}{\fontsize{8}{9}\selectfont Input Pose}
        & \croppppimgg{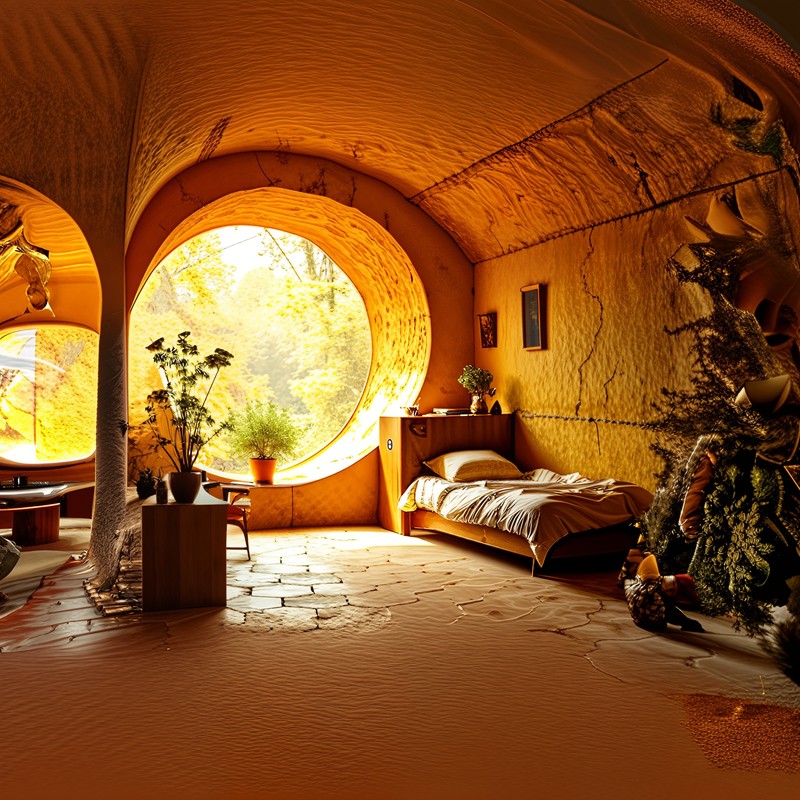}
        & \croppppimgg{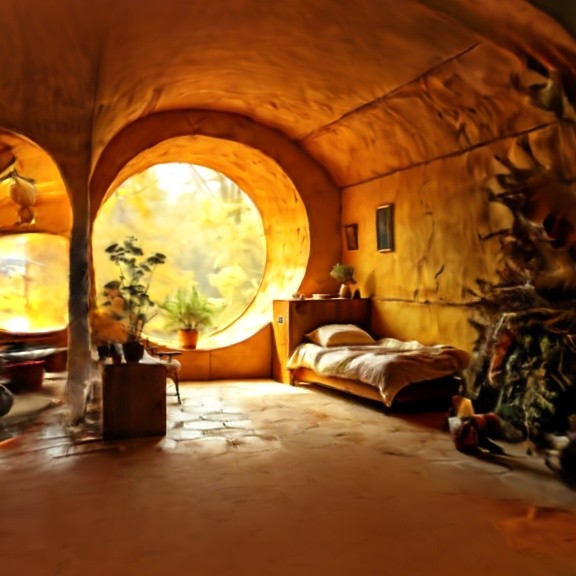}
        & \croppppimgg{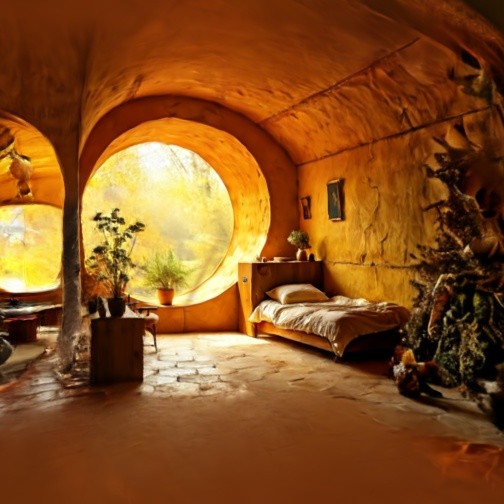}
        & \croppppimgg{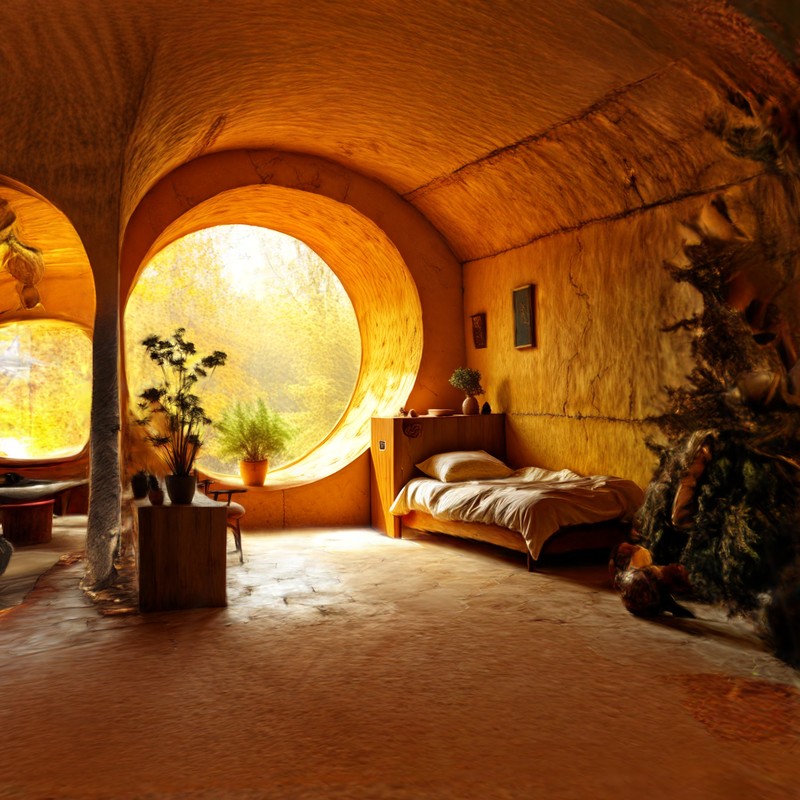} \\

    \end{tabular}

    \vspace{-1mm}

    \begin{tabular}{c | c c c}

        \midrule
        \vspace{-0.5mm}

        &
        {\fontsize{8}{9}\selectfont WorldExplorer \cite{schneider_hoellein_2025_worldexplorer}} &
        {\fontsize{8}{9}\selectfont VGGT-X$^\dagger$ \cite{liu2025vggt}} &
        {\fontsize{8}{9}\selectfont Ours} \\

        \midrule

        \rotatebox{90}{\fontsize{8}{9}\selectfont Novel Views}  
        & \croppppimggg{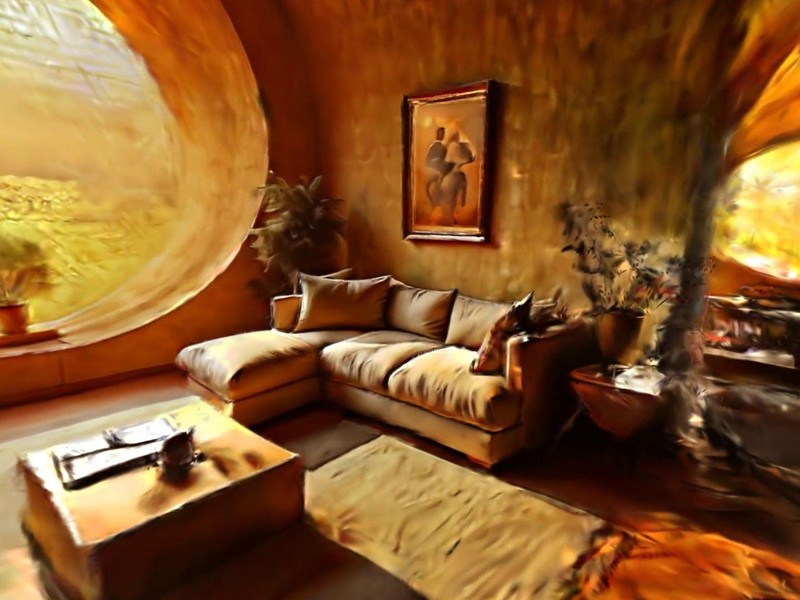}
        & \croppppimggg{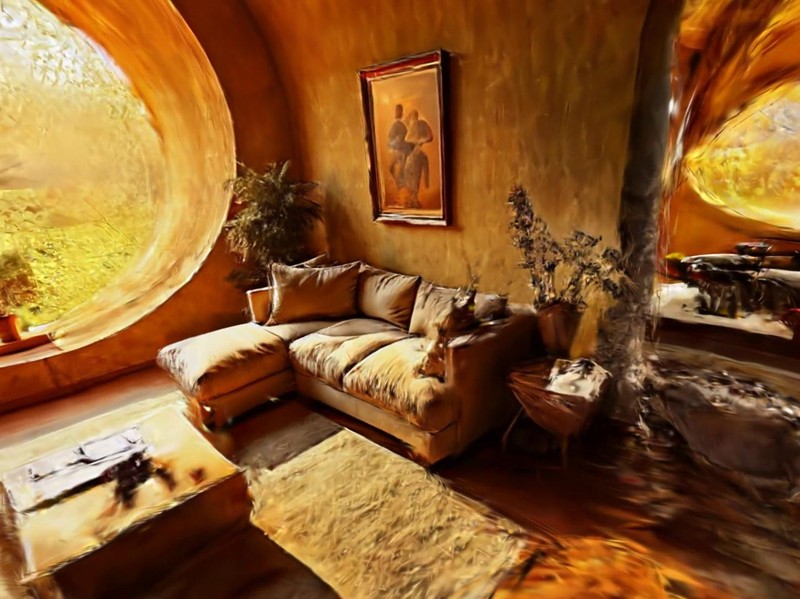}
        & \croppppimggg{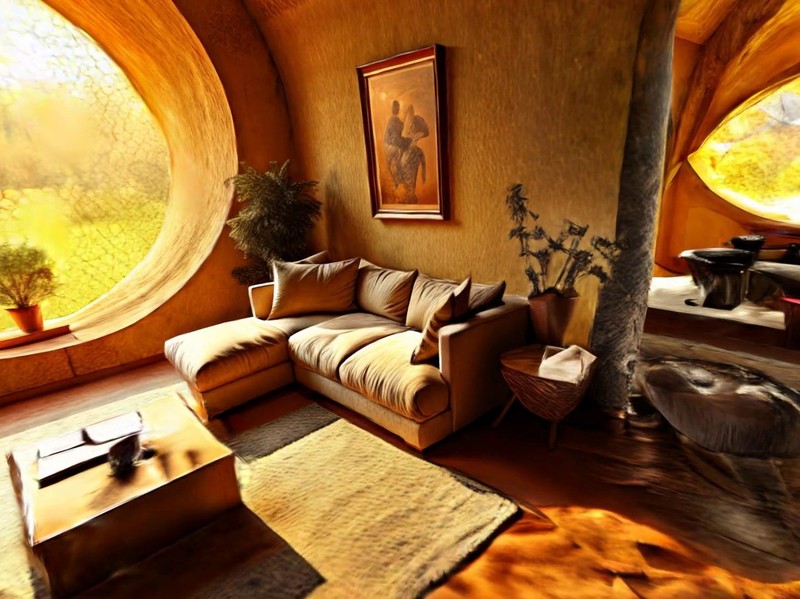} \\

        \rotatebox{90}{\fontsize{8}{9}\selectfont Novel Views}  
        & \croppppimgggg{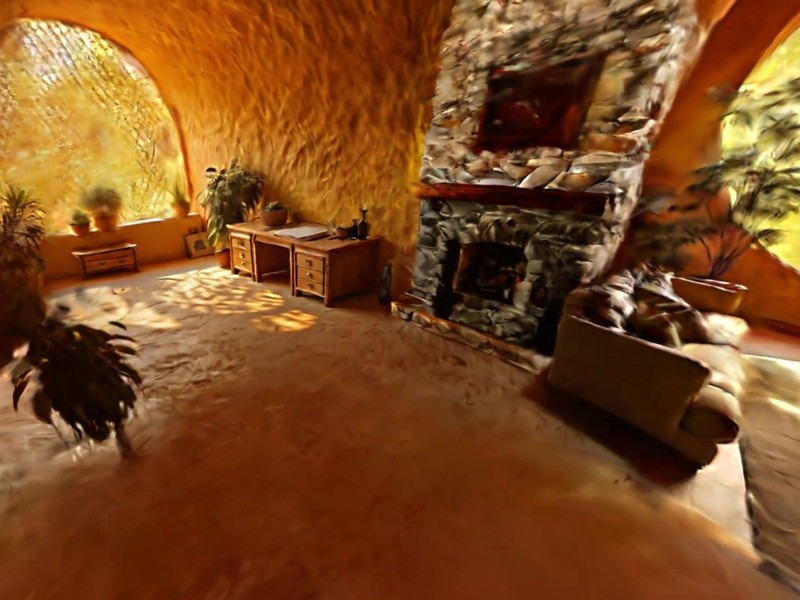}
        & \croppppimgggg{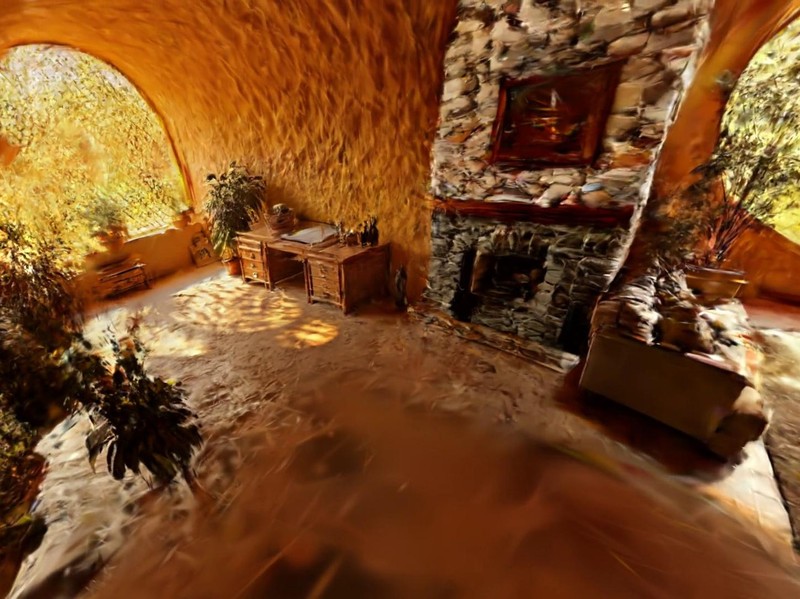}
        & \croppppimgggg{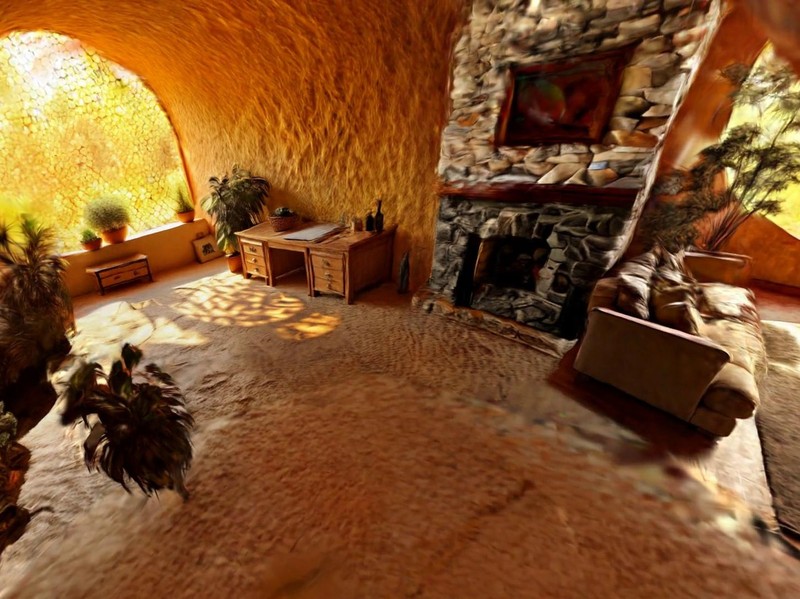} \\

    \end{tabular}
    \end{minipage}
    }

    \vspace{-4mm}
    \caption{\textbf{Large-scale 3D reconstructions.} 
    We compare rendering quality on input poses and novel views for entire 360 degree explorable scenes.
    Our method creates 3D consistent worlds with a high rendering fidelity far beyond the training views.
    Please see the supplementary video for animated flythroughs.
    }
    \label{fig:qual_multi_suppl2}
\end{figure*}

\newcommand{\cropppppimg}[1]{%
  \adjustbox{
    width=0.245\textwidth,
    trim=0 {.1\height} 0 {.16\height},
    clip
  }{\includegraphics{#1}}%
}

\newcommand{\cropppppimgg}[1]{%
  \adjustbox{
    width=0.245\textwidth,
    trim=0 {.05\height} 0 {.21\height},
    clip
  }{\includegraphics{#1}}%
}

\newcommand{\cropppppimggg}[1]{%
  \adjustbox{
    width=0.33\textwidth,
    trim=0 {.1\height} 0 {.16\height},
    clip
  }{\includegraphics{#1}}%
}

\newcommand{\cropppppimgggg}[1]{%
  \adjustbox{
    width=0.33\textwidth,
    trim=0 {.13\height} 0 {.13\height},
    clip
  }{\includegraphics{#1}}%
}

\begin{figure*}
    \centering
    \setlength{\tabcolsep}{1pt}
    \renewcommand{\arraystretch}{1.1}

    \resizebox{0.97\textwidth}{!}{
    \begin{minipage}{\textwidth}
    \begin{tabular}{c | c | c c c}

        &
        {\fontsize{8}{9}\selectfont Video Frames} &
        {\fontsize{8}{9}\selectfont WorldExplorer \cite{schneider_hoellein_2025_worldexplorer}} &
        {\fontsize{8}{9}\selectfont VGGT-X$^\dagger$ \cite{liu2025vggt}} &
        {\fontsize{8}{9}\selectfont Ours} \\

        \midrule

        \rotatebox{90}{\fontsize{8}{9}\selectfont Input Pose}
        & \cropppppimg{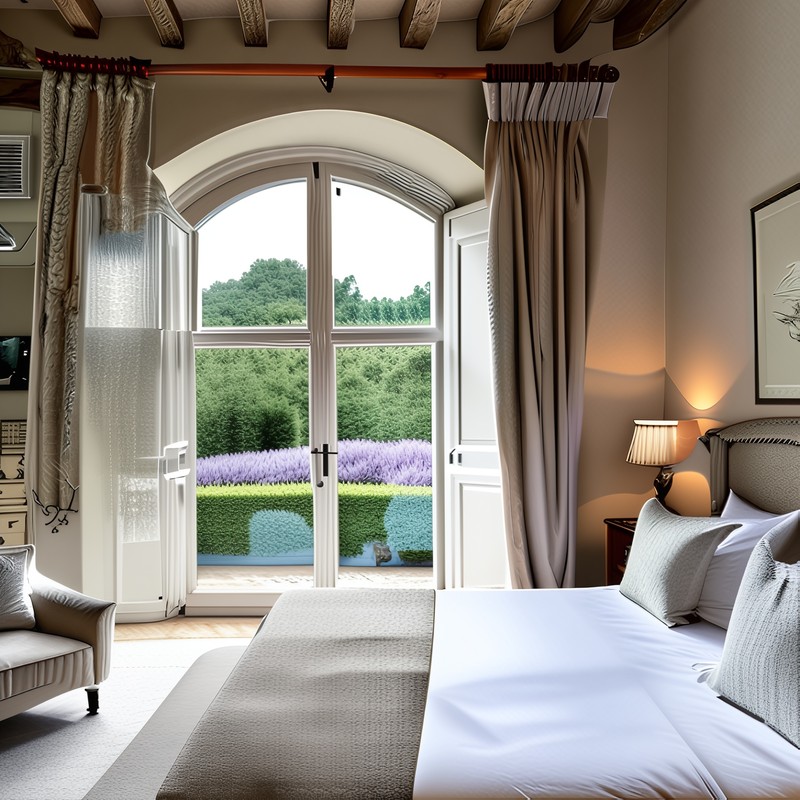}
        & \cropppppimg{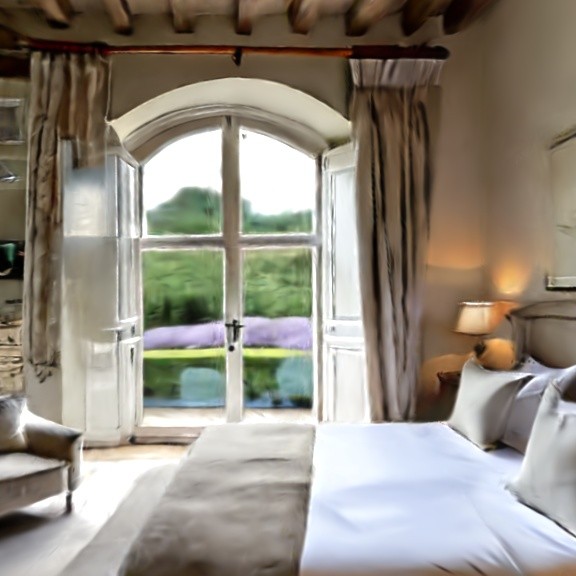}
        & \cropppppimg{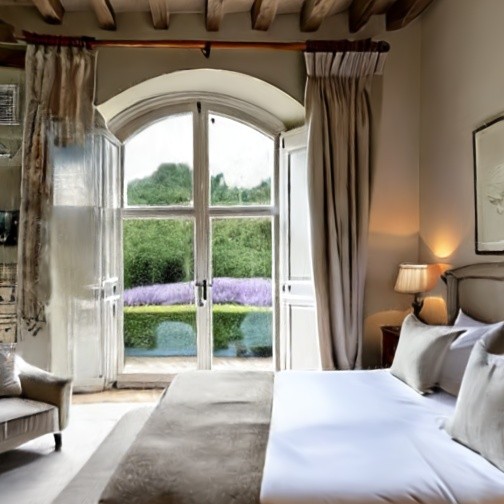}
        & \cropppppimg{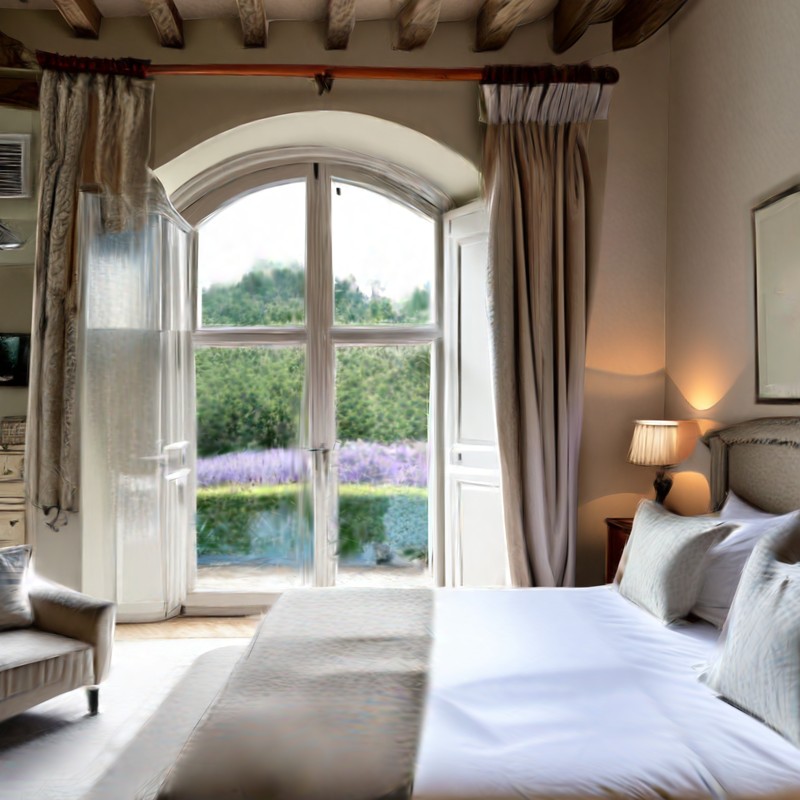} \\

        \rotatebox{90}{\fontsize{8}{9}\selectfont Input Pose}
        & \cropppppimgg{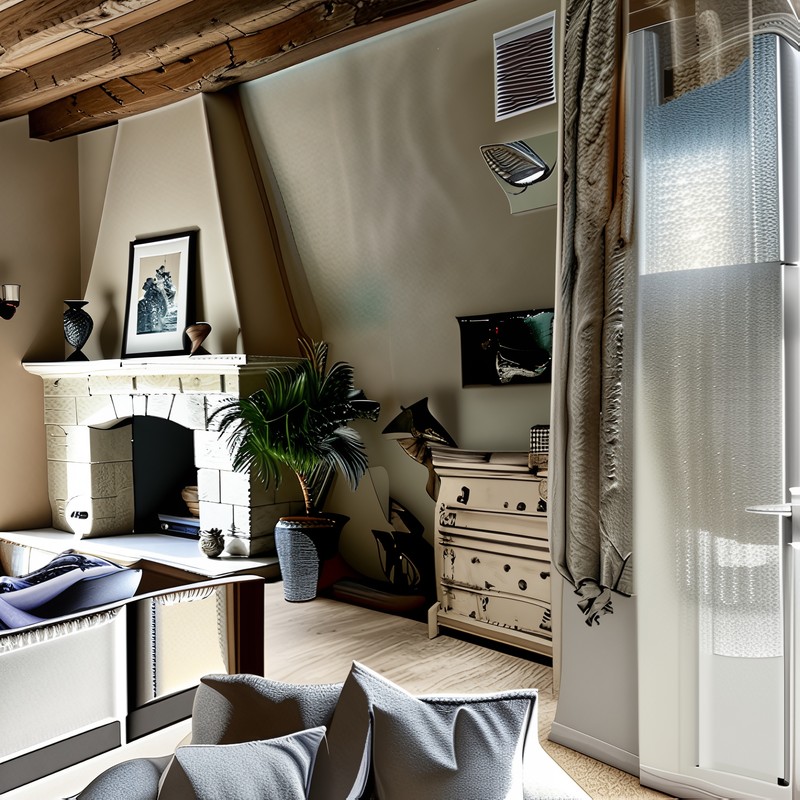}
        & \cropppppimgg{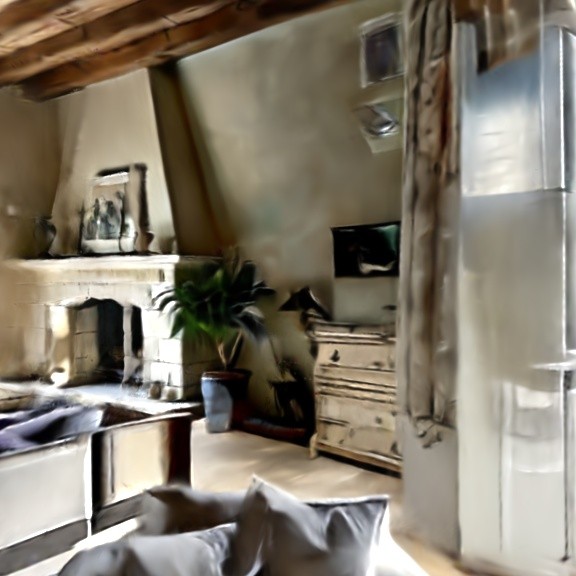}
        & \cropppppimgg{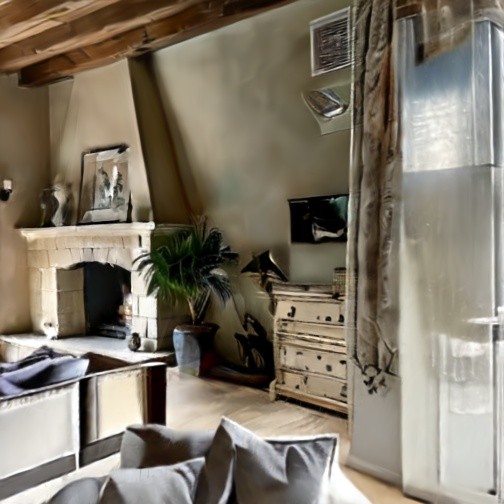}
        & \cropppppimgg{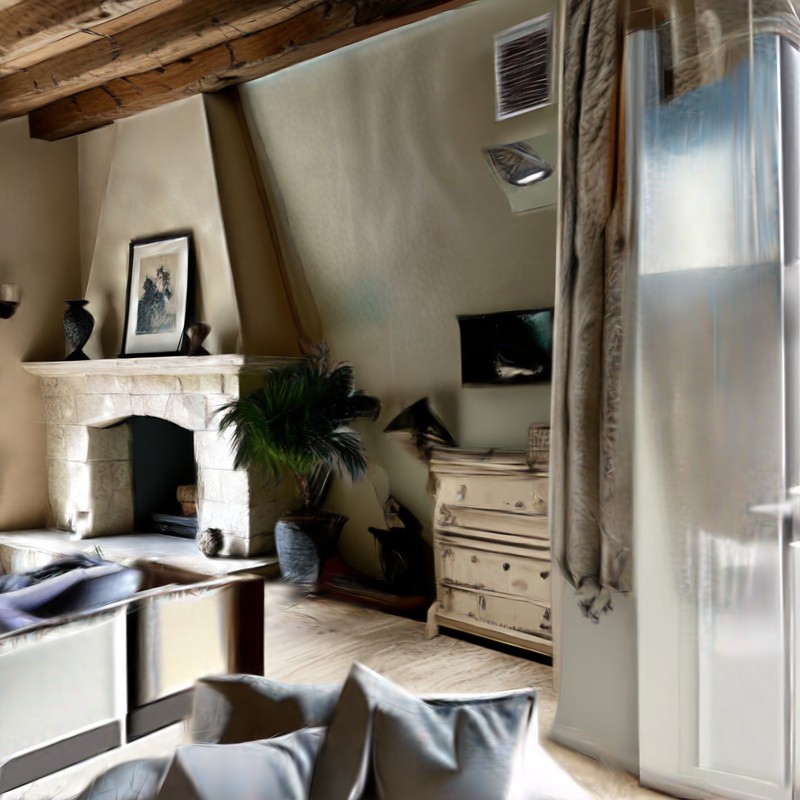} \\

    \end{tabular}

    \vspace{-1mm}

    \begin{tabular}{c | c c c}

        \midrule
        \vspace{-0.5mm}

        &
        {\fontsize{8}{9}\selectfont WorldExplorer \cite{schneider_hoellein_2025_worldexplorer}} &
        {\fontsize{8}{9}\selectfont VGGT-X$^\dagger$ \cite{liu2025vggt}} &
        {\fontsize{8}{9}\selectfont Ours} \\

        \midrule

        \rotatebox{90}{\fontsize{8}{9}\selectfont Novel Views}  
        & \cropppppimggg{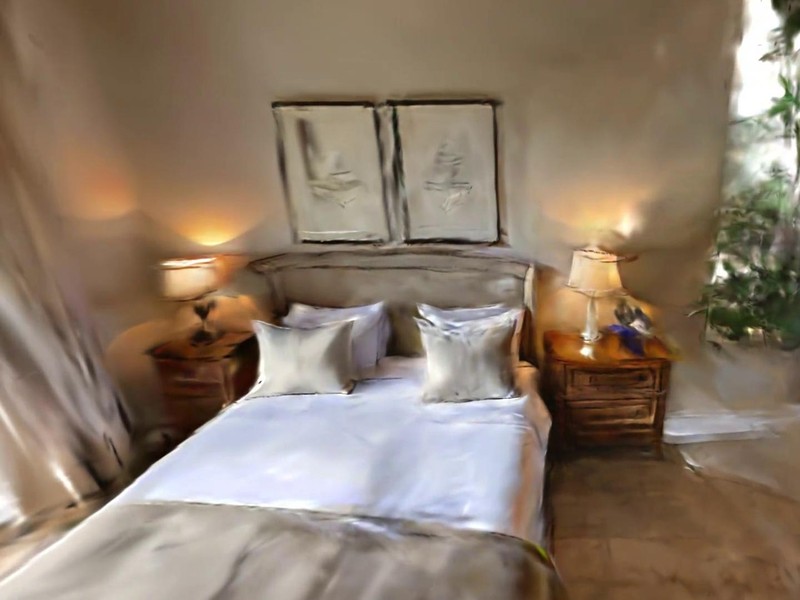}
        & \cropppppimggg{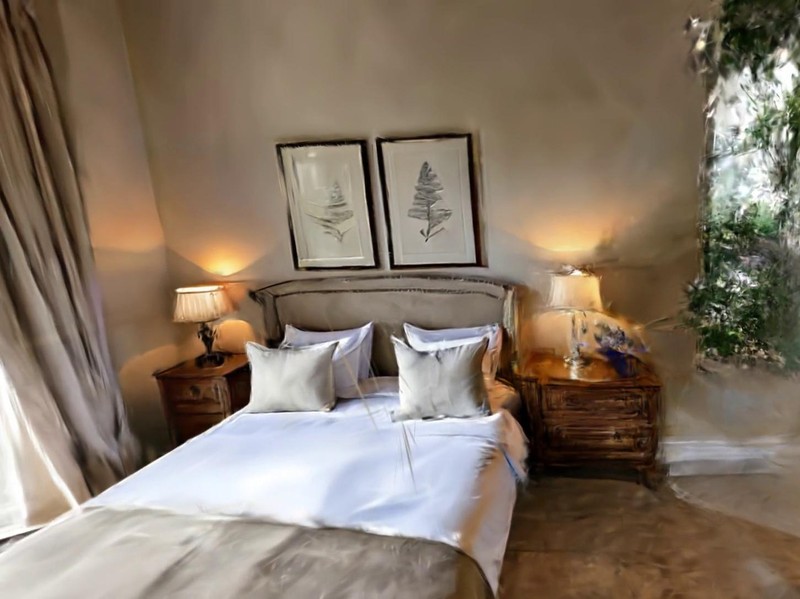}
        & \cropppppimggg{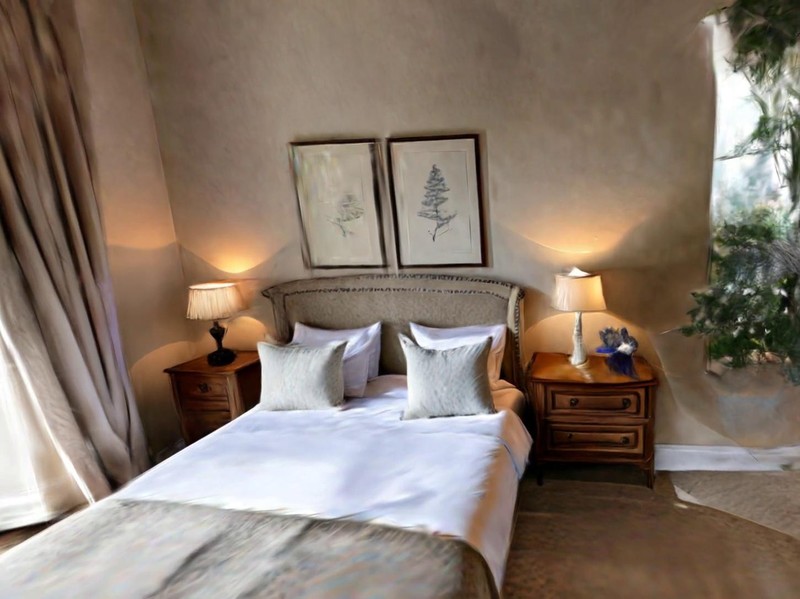} \\

        \rotatebox{90}{\fontsize{8}{9}\selectfont Novel Views}  
        & \cropppppimgggg{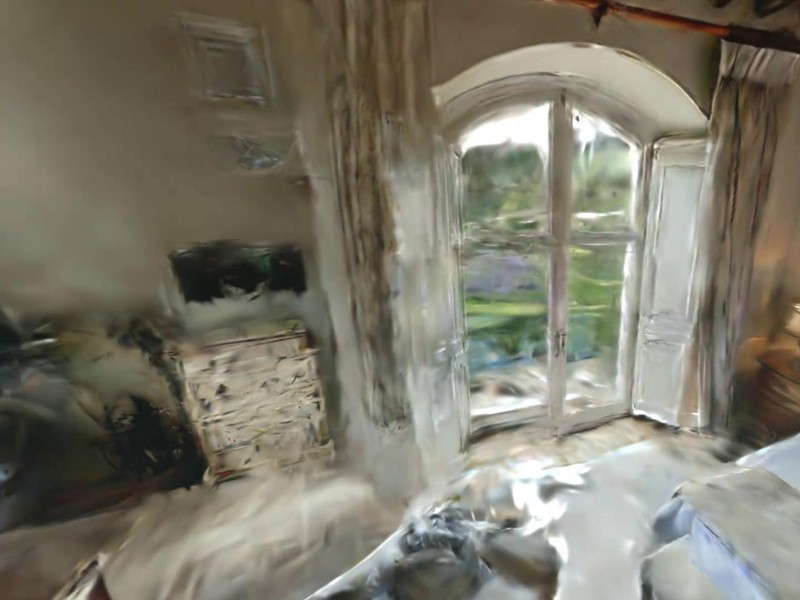}
        & \cropppppimgggg{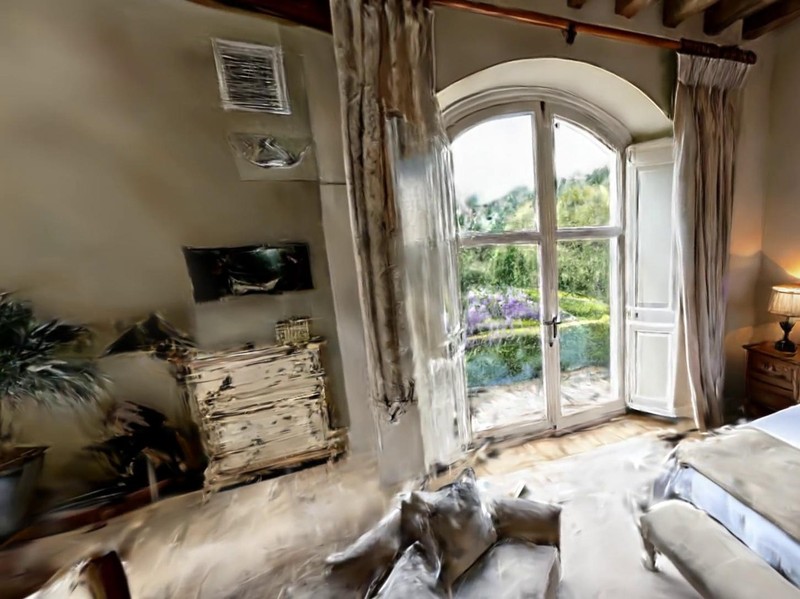}
        & \cropppppimgggg{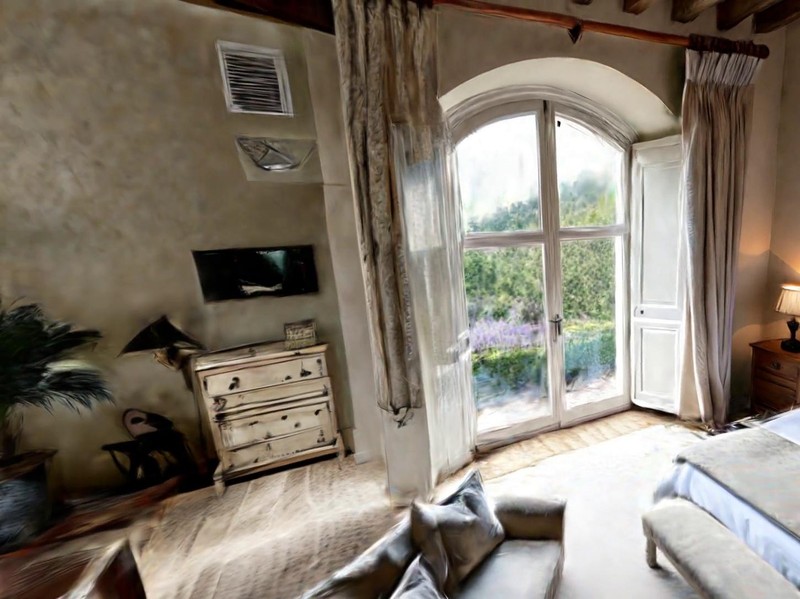} \\

    \end{tabular}
    \end{minipage}
    }

    \vspace{-4mm}
    \caption{\textbf{Large-scale 3D reconstructions.} 
    We compare rendering quality on input poses and novel views for entire 360 degree explorable scenes.
    Our method creates 3D consistent worlds with a high rendering fidelity far beyond the training views.
    Please see the supplementary video for animated flythroughs.
    }
    \label{fig:qual_multi_suppl3}
\end{figure*}

\newpage

\end{document}